 \documentclass[10pt,journal,compsoc]{IEEEtran}
\ifCLASSOPTIONcompsoc
  \usepackage[nocompress]{cite}
\else
  \usepackage{cite}
\fi
\ifCLASSINFOpdf
\else
\fi
\usepackage{amsmath}
\usepackage{amssymb}
\usepackage{mathtools}
\DeclareMathOperator*{\EV}{\mathbb{E}}
\usepackage{subcaption}
\usepackage{xcolor}

\usepackage{array}
\usepackage{dblfloatfix}
\begin{document}
%
\title{Characterizing Inter-Layer Functional Mappings of Deep Learning Models}
%
%
%
%
\author{Donald~Waagen, Katie~Rainey, Jamie~Gantert, David~Gray, Megan~King, M. Shane~Thompson, Jonathan~Barton,
Will Waldron, Samantha Livingston, and Don Hulsey
\IEEEcompsocitemizethanks{\IEEEcompsocthanksitem Donald Waagen is with the Air Force Research Laboratory, Eglin Air Force Base,
FL.\protect\\
E-mail: donald.waagen@us.af.mil
\IEEEcompsocthanksitem  Katie Rainey is with the Space and Naval Warfare Systems Center Pacific, San Diego, CA.\protect\\ 
E-mail: kate.rainey@navy.mil
\IEEEcompsocthanksitem Jamie Gantert and David Gray are with the Air Force Research Laboratory, Eglin Air Force Base,
FL.
\IEEEcompsocthanksitem Megan King and Shane Thompson are with the Army Combat Capabilities Development Command, Aviation and Missile Center, Redstone Arsenal,
AL.
\IEEEcompsocthanksitem Jonathan Barton, Will Waldron, and Don Hulsey are with Dynetics, Inc., Huntsville, AL.
\IEEEcompsocthanksitem Samantha Livingston is with Modern Technology Solutions, Inc., Alexandria, VA.

}

\thanks{Distribution A:  Approved for public release, distribution is unlimited, 96TW-2018-0343.}}

\IEEEtitleabstractindextext{%
\begin{abstract}
Deep learning architectures have demonstrated state-of-the-art performance for object classification and have become ubiquitous in commercial products.  These methods are often applied without understanding (a) the difficulty of a classification task given the input data, and (b) how a specific deep learning architecture transforms that data.  To answer (a) and (b), we illustrate the utility of a multivariate nonparametric estimator of class separation, the Henze-Penrose (HP) statistic, in the original as well as layer-induced representations.  Given an $N$-class problem, our contribution defines the $C(N,2)$ combinations of HP statistics as a sample from a distribution of class-pair separations. This allows us to characterize the distributional change to class separation induced at each layer of the model.  Fisher permutation tests are used to detect statistically significant changes within a model.  By comparing the HP statistic distributions between layers, one can statistically characterize: layer adaptation during training, the contribution of each layer to the classification task, and the presence or absence of consistency between training and validation data.  This is demonstrated for a simple deep neural network using CIFAR10 with random-labels, CIFAR10, and MNIST datasets.%
\end{abstract}

}

\maketitle

\IEEEdisplaynontitleabstractindextext

%
\IEEEpeerreviewmaketitle

\IEEEraisesectionheading{\section{Introduction}\label{sec:introduction}}
\IEEEPARstart{D}{eep} learning architectures have become the new standard practice for solving pattern recognition
and machine learning problems due to their predictive performance across a variety of datasets.  However,
the design process continues to be an ad hoc process considering these black box systems can have
millions of free parameters.  Most practitioners select a member of the standard set
of architectures that have been shown to work well on related problems and make small
adjustments or modifications to the respective architecture.

Beyond predictive performance, many questions remain. Interpretability and model explanation are
becoming important requirements as deep learning architectures are increasingly incorporated into
critical systems. We believe that understanding the nature of the data with respect to the task and the transformations induced by the model on the data
are essential steps to facilitate trust and robustness for these systems.

Understanding the nature of deep learning functional mappings has been and continues to be an exciting
area of research.  Visualization techniques abound  (e.g. activation maps~\cite{ziller13}) which
attempt to identify the spatial regions and the associated neurons of importance in the
classification of an image.  Estimates of the sensitivity of the network to
perturbations at each neuron~\cite{berishahero} can provide insight into the relative importance of each
neuron in the network to overall performance. 

Zhang et al.~\cite{zhang17} demonstrated that modern deep learning models (e.g.
AlexNet~\cite{krizhevshy12}) have the capacity to memorize random labels.  The generalization paradox
describes the phenomenon that deep learning models have generalized well despite the high capacity.
The mechanism of learning and associated generalization in the presence of high-capacity learning was
posed by Bartlett et al.~\cite{bartlett17} who demonstrate margin gap analysis on the output layer in an attempt to
address the paradox of generalization and capacity. 
The link between generalization and flat minima~\cite{hock97}
is an area of current interest.  Chaudhari et al.~\cite{cha17} exploit stochastic  gradient  descent
(SGD) to favor flat minima and maximize the generalization of the derived solution. Dziugaite and
Roy~\cite{roy17} utilize flat minima in a Probably Approximately Correct (PAC)-Bayes setting to
derive numerical bounds for the generalization error of deep neural network models.

\subsection{Our contribution}

Our research is not attempting to quantify the resultant generalization of the output. Rather, we seek to
characterize the utility of the functional mappings occurring within each layer of a deep learning
network.  Note that we focus on the functional mapping of a layer, not the individual neurons. 
This focus on layer induced mappings is also seen in earlier work by members of our team ~\cite{8010591}.

For our purposes, producing deep learning models that achieve \textit{state-of-the-art} or
\textit{super-human} performance is \textit{not} a goal for this paper.  Insight into the changes in class
sample separation as the data are transformed through a deep learning model is the goal.  Class-pair separation is estimated via the Henze-Penrose (HP) statistic. Given an $N$-class problem, we compute the $C(N,2)$ combinations of HP statistics and treat them as a collection or sample from a distribution of separations.  These statistics can also be computed for the data in the original measurement space (e.g. image), as well as at any location within a deep learning model.  This allows us to statistically characterize the distributional change induced by each layer of the model with respect to the overall classification task.  By computing and comparing these samples of separation statistics on deep models in different states, we can numerically quantify the changes to the data, as well as the changes to the data representations in the model.

The experiments are designed to compare and contrast the mappings learned under various
data-label conditions: well separated data, poorly separated data, and identically distributed
data. All experiments were performed using a single simple convolutional neural network
architecture which is outlined in Section~\ref{sub:ourBB}.

Given this simple motivational framework, we wish to investigate the following questions:
\begin{itemize}
\item[$\bullet$] How separable are the classes in the original measurement space (Section~\ref{sub:classSepOrig})?  
\item[$\bullet$] What happens to the data when passed through a model before training (Section~\ref{sub:initModel})?
\item[$\bullet$] How much adaptation has a representation produced by a layer undergone during training (Section~\ref{sub:effModelAdapt})?
\item[$\bullet$] How much is each layer contributing to the classification process (Section~\ref{sub:indivLayerContrib})?
\item[$\bullet$] Which layers in the trained model transform  the validation and training data equivalently, and conversely, differently? (Section~\ref{sub:trainVsVal})?
\end{itemize}

Answering these questions can provide a practitioner insight into model design, optimization, and
robustness. In this paper, we provide a short and very myopic discussion on
nonparametric tests and estimators of distributional separation in Section~\ref{sec:nonParamTests}.
We then discuss the datasets in Sections~\ref{sub:datasets} and~\ref{sub:classSepOrig}, a simple
convolutional neural network model in Section~\ref{sub:ourBB}, and the training process used in our experiments.  
Finally, we present the experiment results including nonparametric statistical hypothesis tests to
address the questions above in Section \ref{sub:section4}. Note that a supplemental document is provided that includes the results in greater detail.

\section{Nonparametric Two Sample Tests}\label{sec:nonParamTests}

Statisticians have been and continue to be interested in defining procedures and tests for sameness.
For example, given two random variables $X$ and $Y$ with cumulative distributions $F_X$ and $G_Y$,
define a procedure to test \({H_0:\phi(X) = \phi(Y)}\) vs. ${H_A:\phi(X) \neq \phi(Y)}$, where
$\phi(X)$ represents the distribution or a statistic (e.g. mean, variance) of the random variables
$X$.  Many tests make some strong assumptions about the distributional form of $F_X$ and $G_Y$, while
other procedures, called distribution-free or nonparametric tests, require minimal prior
information on the nature of the distributions.  For the purposes of this paper, we are interested 
in  nonparametric measures of the distributional differences between $X$ and $Y$, (i.e. under the ${H_A:F_X \neq G_Y}$ hypothesis).  

Given two univariate and independent sample observations  $X_n=(x_1,x_2,x_3,...,x_n)$ and
$Y_m=(y_1,y_2,y_3,...,y_m)$ of random variables $X$ and $Y$, Wald and Wolfowitz~\cite{wald40} defined
the runs test, a nonparametric procedure to test the hypothesis $H_0:F_X = G_Y$ vs. $H_A:F_X \not=
G_Y$.  The construction of the test consists of pooling the $N=n+m$ samples and sorting the pooled
sample, generating an ordered list.  From the ordered list, replace the values with their associated class labels $C$
and count the number of runs, i.e., the number of consecutive sequences of identical labels, or
equivalently, the number of times the $i$th and $i$th+1 class labels $c_{(i)}$ and $c_{(i+1)}$ disagree.
Let $R$ denote the number of runs and $S$ be the number of times neighboring ordered labels
disagree, giving $R=S+1$ and $\textstyle S = \sum^{N-1}_{i=1} |c_{(i)} \not= c_{(i+1)} |$.

The expected value and expected variance of $R$ under $H_0$ are
\begin{equation} \label{eq:frmean}
  \EV(R) = \frac{2mn}{m+n}+1
\end{equation}
\begin{equation}\label{eq:frvar}
  \sigma^2(R) = \frac{2mn(2mn-m-n)}{(m+n)^2(m+n-1)},
\end{equation}
respectively. The null hypothesis $H_0:F_X = G_Y$ can then be rejected or fail to be rejected using
the Wald-Wolfowitz test statistic
\begin{equation} \label{eq:wws}
  W = \frac{R-\EV(R)}{\sqrt{\sigma^2(R)}}
\end{equation}
which asymptotically converges to a standard normal distribution $N(0,1)$ as $m,n\rightarrow\infty$.
Wald and Wolfowitz point out that the runs test is a one-sided test, that is, rejection of
the null hypothesis occurs for small $W$.

In high dimensional spaces, the concept of defining a strict ordering via
sorting of samples becomes ill-defined.  This led Friedman and Rafsky~\cite{friedman79} to develop a
surrogate method for defining an ordering and generalizing Wald's runs test for multivariate data. Let
$\mathbf{X},\mathbf{Y}\in \Re^d$ denote two multivariate random variables with distributions
$F_{\mathbf{X}}$ and $G_{\mathbf{Y}}$, and let $X_n=(x_1,x_2,x_3,\dots,x_n)$ and
$Y_m=(y_1,y_2,y_3,\dots,y_m)$ denote $n$ and $m$ \textit{i.i.d.} observations of $\mathbf{X}$ and
$\mathbf{Y}$, respectively. To test $H_0:F_{\mathbf{X}} = G_{\mathbf{X}}$ vs. $H_A:F_{\mathbf{X}} \not = G_{\mathbf{X}}$, Friedman pools the
samples $Z_{n+m}={X_n\cup Y_m}$ and constructs a minimal spanning tree  $T = (E,V)= \text{MST}(Z)$ on the
pooled sample $Z$.  The sample observations from $Z$ define the vertices $V$ of the graph, and the
ordering is defined by the set of edges $E$ connecting the sample points. By construction, the
number of edges $|E|$ in $T$ is $n+m-1$.   Let $E_c\subset E$ denote the subset of edges of $E$
connecting vertices $v_i , v_j\in V$ associated with differing class labels, that is, $class(v_i)
\not= class(v_j)$.  The Friedman-Rafsky test statistics $S$ is defined as the number of edges in
$T$ which connect vertices with differing class labels, $S=|E_c|$. Closely related to $S$ is the \textit{runs}
statistic $R$ which is defined as the number of connected sets produced when the edges that connect
vertices connecting different labels are removed, with $R = S+1$.  This statistic is a member of
complexity measures identified and used for characterizing the complexity of a classification
problem~\cite{basu02}.

Friedman and Rafsky also provide the asymptotic convergence of $W$ for the multivariate case, for
which~\eqref{eq:frmean} and~\eqref{eq:wws} are shown to still hold. However, the variance
has the form
\begin{multline}\label{eq:frVarMST}
  \sigma^2(R|C) = 
  \frac{2mn}{N(N-1)} \Biggr[ \frac{2mn-N}{N} + \\
  \frac{C-N+2}{(N-2)(N-3)}[N(N-1)-4mn+2]\Biggr]
\end{multline}
where $C$ is the number of edge pairs that share a common node.  Thus the variance under $H_0$ is
conditioned on the topology of the minimal spanning tree. Fixing the topology and performing a
permutation test under $H_0$ sampling values of $R$ with randomly permuted class labels allows one
to empirically calculate $\textstyle{\sigma^2(R)\equiv\sigma^2(R|C)}$. 

\subsection{Connections to the $f$-Divergence Function Family}
The $f$-divergence functions~\cite{csiszar2004information} are a family of general measures of distributional separation between two
probability distributions. Henze and Penrose~\cite{henze99} proved that a simple function of the Friedman-Rafsky statistic $S$
asymptotically converges to a member of the $f$-divergence family.  Given samples $X_n$ and ${Y_m}$
sampled from probability distributions $f$ and $g$, they show that
\begin{equation}
\frac{S}{n+m} \rightarrow 2qp\int{\frac{f(x)g(x)}{pf(x)+qg(x)}d\mathbf{x}}=1 - \delta(f,g,p)
\end{equation}
where
\begin{equation} \label{eq:ffam1}
\delta(f,g,p) = \int{\frac{p^2f^2(x)+q^2g^2(x)}{pf(x)+qg(x)}dx}
\end{equation}
as ${n,m}\rightarrow \infty$ in a linked manner (\textit{i.e.} $\textstyle \frac {m}{m+n}
\rightarrow p$ and $q=1-p$ ), with $\delta(f,g,p)$ in~\eqref{eq:ffam1} being a member of the $f$-divergence  
family of functions.  Therefore, given two multivariate samples ${{X}_n,{Y}_m}$, the Friedman-Rafsky
test statistic $S$ can provide a sample-based estimate of the distributional separation in high
dimensions, that is,
\begin{equation} \label{eq:hpdiv1}
  \hat{\delta}(X_n,Y_m,p) = 1 - \frac{S}{n+m}.
\end{equation}
Under the null hypothesis $H_0: F=G$, $\delta(f,g,p) \equiv \delta(f,f,p)=p^2 + q^2$,
and the distribution of the sample statistic $\hat\delta$ under $H_0$ is conditioned on the
proportion of points from each class (the values of $p$ and $q$).  For example, when $p=0.5$, $\EV_{H_0}[
\hat\delta(X_n,Y_n,0.5)]=0.5$.

Extending the work of Henze and Penrose, Berisha et al.~\cite{berisha16}  define a closely
related distributional measure of separation, which they denote as ${D}_p(f,g)$, defined as
\begin{equation}\label{eq:berishaDistSep}
D_p(f,g) = \frac{1}{4pq}\left[\int\frac{(pf(x)-qg(x))^2}{pf(x)+qg(x)}dx-(p-q)^2\right].
\end{equation}
Berisha demonstrates that under $H_0:F_x=G_y$, $D_p(f,g)=D_p(f,f)=0$. Thus, the expected value of
$D_p$ under $H_0$ is zero and is independent of the prior $p$.

Similar to Henze and Penrose and under the same linked conditions previously described, Berisha
proves that a simple function of the Friedman-Rafsky test statistic $S$ converges asymptotically to%
~\eqref{eq:berishaDistSep},
\begin{equation}\label{eq:frAsymConv}
  1-S\frac{n+m}{2nm} \rightarrow {D}_p(f,g).
\end{equation}
As one can see from~\eqref{eq:frAsymConv}, a sample-based estimate of $D_p(f,g)$ can easily be
computed using the Friedman-Rafsky statistic $S$.  Given two multivariate samples $X_n$ and $Y_m$,
the sample estimate of $D_p(f,g)$, which we denote as $\mathcal{H}$, is given by
\begin{equation} \label{eq:hpdiv2}
\mathcal{H} = 1-S\frac{n+m}{2nm}.
\end{equation}
We refer to $\mathcal{H}$ in~\eqref{eq:hpdiv2} as the Henze-Penrose-Berisha-Hero divergence sample statistic, or
HP statistic for short.  When the class sample sizes are equal (i.e. $m = n$)
\begin{equation}
  \mathcal{H} = 1-S\frac{n+m}{2nm} = 1-2\frac{S}{n+m},
\end{equation}
which is equivalent to a rescaling of~\eqref{eq:hpdiv1}.

\begin{figure}[htb]
\centering
\begin{subfigure}[b]{0.3\linewidth}
  \includegraphics[width=\linewidth]{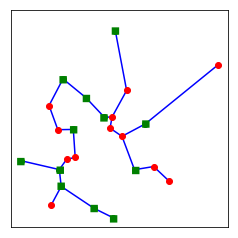}
  \caption{S =11, $\mathcal{H}$ = 0.08}
\end{subfigure}
\begin{subfigure}[b]{0.3\linewidth}
  \includegraphics[width=\linewidth]{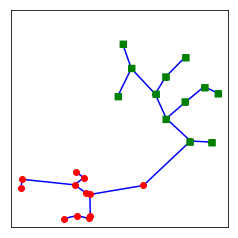}
  \caption{S=1, $\mathcal{H}$  = 0.92}
\end{subfigure}
\caption{Friedman-Rafsky ($S$) and Henze-Penrose-Berisha-Hero ($\mathcal{H}$) statistics for~(a)~mixed
         and~(b)~separated samples.}
\label{fig:frHpExample}
\end{figure}

$\mathcal{H}$ has the desirable property of
providing consistent and easily interpretable results, generating values near zero when the two
distributions are statistically indistinguishable, and near one when the distributions are well
separated. Figure~\ref{fig:frHpExample} provides an example of minimal spanning trees and
associated Friedman-Rafsky and HP statistics ($S,\mathcal{H}$)  under mixed and well separated
distributional conditions.

Given a measure of proximity, the simple nonparametric statistic such as the $\mathcal{H}$ is invariant to changes in rotation and scale.  Additionally, since it can be computed in arbitrarily finite dimensional spaces, computing and comparing class separation in differently dimensioned representations is easily performed.  One can therefore compare the distributional separation between classes before and after a functional transformation $f: \Re^d \rightarrow \Re^p$ is applied.  

A caveat to this approach is that the measure of proximity
applied to the measurements should be consistent with the algorithmic mappings under evaluation
and/or thoughtfully selected by the practitioner.  For instance, Euclidean distance may not be the best
measure for comparing images or signals without an alignment or correlation of the
data measurements. With this in mind and given an appropriate proximity/distance measure,
one can meaningfully compare the distributional separation in the original representation space as
well as the separation in the space induced by a mathematical mapping (e.g. feature transformation)%
\footnote{Note that this classification-oriented comparative analysis is not just for analysis of deep
learning models, but can be useful in evaluating the representations produced by any functional
mapping for classification, including engineered features selected by experts.}.
Distributional separation for multiple classes can be computed pairwise or as one
against all.  For our experiments, we will compute the pairwise separation between classes and
compare/contrast the sample distributions of those statistics.  Because Euclidean distance is the typically 
 proximity measure applied with $\mathcal{H}$, our baseline analysis uses it.%
\footnote{Additional analysis results are included in the supplemental material.}
An equivalent analysis was also performed with cosine distance as the measure for computing $\mathcal{H}$. 
It produced results consistent with the Euclidian distance case.%
\footnote{The results of the analysis using cosine distance are included in the supplemental material.}

For all the datasets under evaluation, this work will compare and contrast the $\mathcal{H}$ computed
between all pairs of classes.  Let $\boldsymbol{\mathcal{H}}$ denote the set or distribution
of all class pairwise $\mathcal{H}$ values computed between all classes in the population, that is,
\begin{equation}
  \boldsymbol{\mathcal{H}} = \{\mathcal{H}({x_i},{x_j}) \hspace{2mm} \forall  i, j = 1,2,...n, i < j \}
 \end{equation}
 where $\mathcal{H}({x_i},{x_j})$ is the HP divergence between samples from class $x_i$ and  $x_j$,
 and $n$ is the number of classes in the data.  And given a sample $\boldsymbol{\mathcal{H}}$, let $\bar{\mathcal{H}}$ denote its sample mean,  
 \begin{equation}\label{eq:meanh}
  \bar{\mathcal{H}} = \frac{2}{n(n-1)}\sum_i \sum_{j|i<j} {\mathcal{H}({x_i},{x_j})}
 \end{equation}
In all of our experiments, there are 10 classes defined in each dataset, and therefore 45 pairwise
$\mathcal{H}$ (\textit{i.e.} $C(10,2)$) values define the set or distribution of $\boldsymbol{\mathcal{H}}$
values.  This set quantifies the pairwise separation between all the classes in a space, which
allows us to compare and statistically evaluate $\boldsymbol{\mathcal{H}}$ for the original measurement
space, as well as the representations produced on the data by the transformations defined by the
deep learning model.

For the discussions that follow, let $\boldsymbol{\mathcal{H}}_{(L,M)}$ denote the distribution of HP
statistics for the output of layer $L$ of a model and at a prescribed model state $M$. We
limit our analysis of model states $M$ to two values, specifically $M=0$ (when initialized) and
$M=T$ (final or trained).  Let $\boldsymbol{\mathcal{H}}^{(t)}_{(L,M)}$ and
$\boldsymbol{\mathcal{H}}^{(v)}_{(L,M)}$ represent the sets of $\mathcal{H}$ values for training and
validation data, respectively.  If the superscript is not explicitly provided, the set
corresponds to the training set, that is, $\boldsymbol{\mathcal{H}}_{(L,M)} \equiv  \boldsymbol{\mathcal{H}}^{(t)}_{(L,M)}$.

\section{Methodology}

\subsection{The datasets a.k.a. the Good, the Bad, and the Ugly}\label{sub:datasets}

Our experimental design matrix is given in Table~\ref{table:experiments}. The experimental datasets we use for
illustration and analysis include the public domain
MNIST~\cite{mnist} and CIFAR10~\cite{cifar10} datasets, as well as the CIFAR10 dataset with randomly
permuted labels.  By permuting the CIFAR10 dataset
labels, each sampled pseudo-class actually consists of examples from all classes, and therefore
each class is sampling from the same underlying distribution.

\begin{table}[!t]
\renewcommand{\arraystretch}{1.3}
\caption{Model Instances}
\label{table:experiments}
\centering
\begin{tabular}{c c c c}
\hline
\bfseries Dataset & \bfseries Class Labels &  \bfseries \shortstack{Spaghetti Western\\Projection} \\
\hline \hline
MNIST & Truth & \textit{the Good} \\
CIFAR10 & Truth & \textit{the Bad} \\
CIFAR10 & Random & \textit{the Ugly} \\
\hline
\end{tabular}
\end{table}

We randomly subdivided the nominal training data for CIFAR10~(50,000 example image chips) and MNIST%
~(60,000 example image chips) into non-overlapping training and validation subsets. Our
validation subsets for CIFAR10 and MNIST each contain~1,000 examples per class, leaving~50,000 and%
~40,000 images in the training subsets, respectively. The~10,000 image test datasets for
CIFAR10 and MNIST were maintained as the source provided. For the CIFAR10 randomized label experiments, we copied the
CIFAR10 training and validation subsets with their true labels and generated
pseudo-class identifiers for the images by randomly permuting labels within each subset. The
randomized labeling processes was repeated  five times to generate  five different instances of
the CIFAR10 dataset with random labels for use in training multiple models.

For the statistical analysis of each neural network model, we randomly selected an analysis sample
from each training subset such that the sample contained~10,000 images with~1,000 images per
class to match the size and class composition of the images in the associated validation
subset. By restricting the sample size, the analysis becomes more computationally
tractable.  The statistical characterization of data passing through a neural network was performed
using the~10,000 sample subset of training data and the full~10,000 sample validation dataset.

For each experiment, we implemented and trained our reference convolutional neural network model
with Tensorflow~1.8~\cite{tf2015} using the Keras~2.1~\cite{chollet2015} interface. During training,
we selected categorical cross entropy as the loss function and simple stochastic gradient descent for optimization with
nominal hyperparameters (minibatch size of~32 and learning rate fixed at~0.01).  No regularization
or data augmentation was applied during training. In the experiments using true labels,
we trained five different network instances per experiment,
each starting with different random initializations of the layer weights (sampling from Glorot
uniform distributions). Training was  stopped when the peak accuracy was achieved on the validation
set. For the random label case, five network instances were also trained with different random
initializations of the layer weights.  However, a different instance of the randomly-permuted
image labels was used for each.  The training ended at~200 epochs. This provided a sufficient
number of epochs for the classification performance of the model to stabilize.

To test various hypotheses and evaluate the mappings of the network models with respect to the training and
validation data, we apply a permutation test~\cite{efron1994introduction}.  For each hypothesis test, the
permutation tests are estimated via Monte Carlo sampling, with~50,000 samples generated to estimate
each distribution under the null hypothesis. A critical value of $\alpha = 0.025$ is chosen for our threshold of statistical
significance for all tests.

\subsection{Our Black Box: a simple convolutional neural network}\label{sub:ourBB}

A single, simple representative convolutional neural network architecture is defined for our
experiments to facilitate communication of the concepts and interpretation of the results.  The
model architecture is outlined in Table~\ref{table:model}. We are not precluded from analyzing
deeper and/or complex architectures, but our choice of a simple model is driven by a desire to
maximize transparency in the approach and analysis.
\begin{table}[htb]
\renewcommand{\arraystretch}{1.3}
\caption{Experimental Convolutional Neural Network}
\centering
\begin{tabular}{l c c}
\hline
\bfseries \bfseries Layer & \bfseries Type & \bfseries Configuration \\
\hline \hline
0.Input & Input Space & \shortstack{Rows x Columns \\ x Channels}\\  
1.Conv & 2D Convolutional Layer & 32 3x3xChannels\\
1.ReLU & ReLU Activation & \\
2.Conv & 2D Convolutional Layer & 32 3x3 \\
2.ReLU & ReLU Activation & \\
2.MaxPool & Max-Pooling Operation & 2x2 \\
3.Conv & 2D Convolutional Layer & 64 3x3 \\
3.ReLU & ReLU Activation & \\
4.Conv & 2D Convolutional Layer & 64 3x3 \\
4.ReLU & ReLU Activation & \\
4.MaxPool & Max-Pooling Operation & 2x2 \\
5.Dense & Fully Connected Dense Layer& 512 \\
5.ReLU & ReLu Activation  \\
6.Dense & Fully Connected Dense Layer & 10 \\
6.SoftMax & Softmax Activation\\  [1ex]
\hline
\end{tabular}
\label{table:model}
\end{table}

\section{Answer me, these questions three -- I mean five}\label{sub:section4}
\subsection{Class separability in the original measurement space}\label{sub:classSepOrig}

When a practitioner is supplied a set of measurements and a task, one of the first
questions to be investigated should be in regards to how easy or difficult the task is given the measurements
as represented in the original measurement space.  For a classification task, this can be
subdivided into two subproblems:~(1)~what is the ratio of the signals of interest to the not-signals 
of interest in the measurements, and~(2)~how different are the signals of interest.  Signal
conditioning or preprocessing is generally performed to  retain as much of the differences of the
signals of interest as possible while minimizing any non-signal content in the
measurements in the process. However, this can require human intervention.  It has been a goal of
machine learning to eliminate the task of preprocessing as much as possible, allowing the system to
learn what is signal and what is not.  Therefore, we wish to understand what the intrinsic
separation of classes are in the ambient (measurement) representation \textit{before} any signal
conditioning or algorithmic transformations are performed.

\begin{figure*}[!t]
\centering
\includegraphics[width=0.8\linewidth]{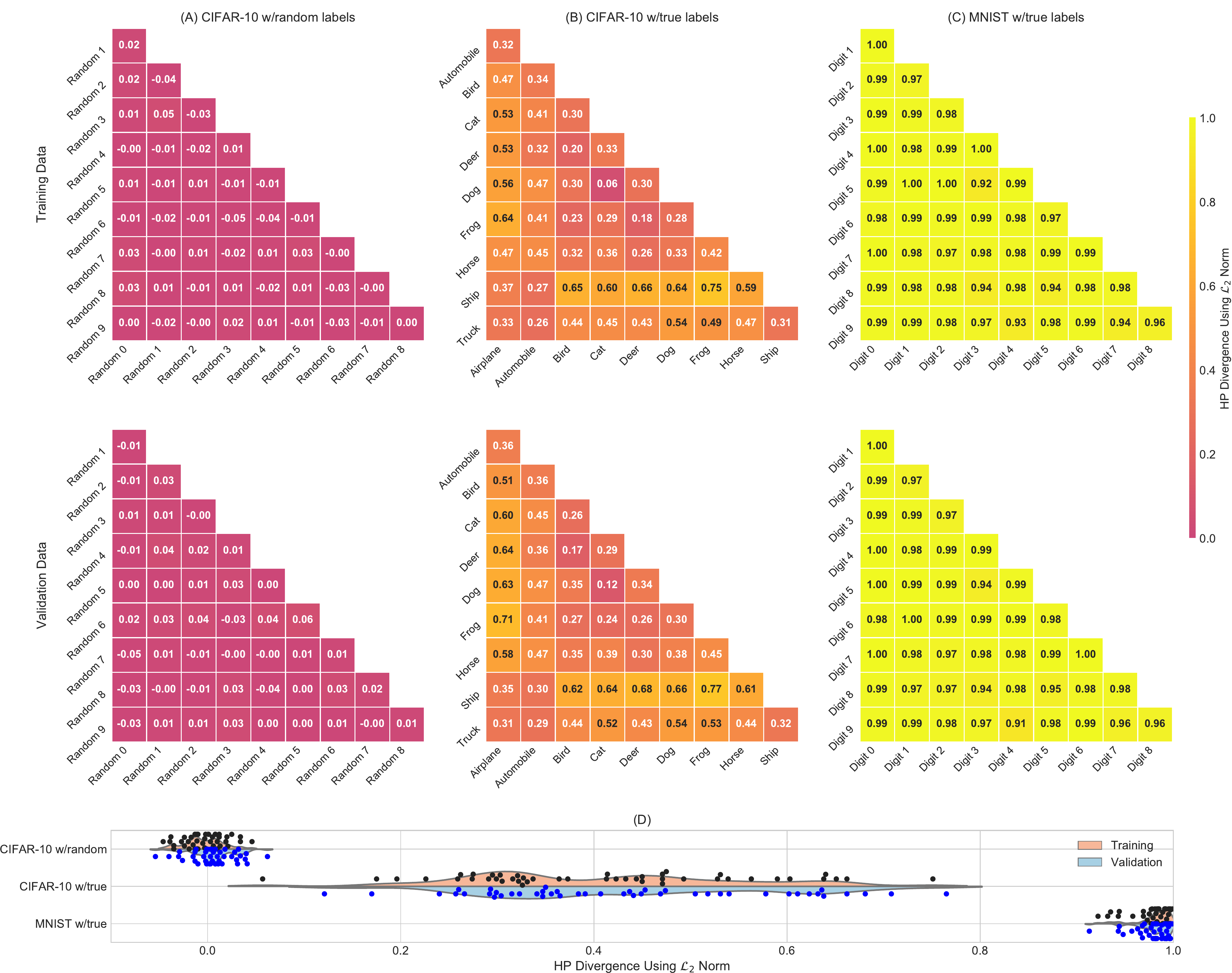}
\caption{Pairwise class HP statistics (training data above, validation data below) computed on~(A)~CIFAR10
         with random labels,~(B)~CIFAR10 with true labels, and~(C)~MNIST with true labels.~(D)~The
         $\boldsymbol{\mathcal{H}}^{(t)}$ (black dots, above),  $\boldsymbol{\mathcal{H}}^{(v)}$ (blue dots, below)
         values and respective kernel-based density functions (orange = training, blue = validation)
         for each task, which illustrate that the estimated class separation for each task in their
         respective ambient representations are quite distinct.}
\label{figure:rawdataprox}
\end{figure*}

For each dataset, we compute the set of class-pairwise statistics $\boldsymbol{\mathcal{H}}^{(t)}$ and
$\boldsymbol{\mathcal{H}}^{(v)}$ for training and validation samples, respectively. The lower-triangular
breakdown of the individual $\mathcal{H}$ values and a kernel density estimate of the
$\boldsymbol{\mathcal{H}}^{(t)}$ and $\boldsymbol{\mathcal{H}}^{(v)}$ distributions are presented in Figure%
~\ref{figure:rawdataprox}.  As can be seen, the corresponding training and validation distributions
are visually similar (and will be statistically characterized in the subsections below), while the
distributions for each task are all quite different.

As indicated in Figure~\ref{figure:rawdataprox}, the class pairwise $\mathcal{H}$ statistics for the MNIST
dataset given the true class labels
are all between~0.9 and~1.0, which indicate that measurements
from each class are quite well separated from each other \textit{in the original space}.
The lowest $\mathcal{H}$ value, corresponds to the~4 and~9 class problem (see Figure~\ref{figure:rawdataprox}C), which is
naturally the easiest pair to confuse.  For the CIFAR10 dataset, given true class labels, the
$\mathcal{H}$ values vary widely, ranging from~0.05 to~0.8.  From Figure~\ref{figure:rawdataprox}B, the
HP statistics are lowest for the Cat-Dog and Bird-Deer image class distributions, while Ship-Frog
images have the highest value.  CIFAR10 is obviously more difficult than MNIST due to the
less processed nature of data (exemplars are not necessarily centered, etc.) which is captured by
the HP statistics.

The matrix of pairwise values for the CIFAR10 with random class labels (Figure%
~\ref{figure:rawdataprox}A) provides more insight into the behavior of the HP-statistic.  In this
case, the distributional separation of the pseudo-labels should be near zero, since each pseudo-class is
sampled from the union of classes. 
This is indeed the case, as the histogram of statistics (Figure%
~\ref{figure:rawdataprox}D) is centered around 0 for these pairs, with the HP test statistics
indicating that the class labels are well mixed (which is by experimental design).  By applying the
HP statistic to a classification task, a practitioner can learn about the difficulty or impossibility of data
classification using the unaltered/initial data representations. The information about the difficulty of the
task will provide valuable guidance in signal conditioning or potential model designs to perform
the given classification task.

The utility of estimating the difficulty of a problem given a set of measurements should not be
underestimated.  As shown in Figure~\ref{figure:rawdataprox}D, the distribution of
$\mathcal{H}$ statistics for each task are linearly separable from each other.  Therefore, under the
Euclidean measure of proximity and the original representations (i.e. pixels), we have discovered a
natural and unambiguous easy-harder-hardest ordering for the classification tasks, with MNIST being
the easiest and CIFAR10 with random class labels being extremely difficult. Furthermore, we now have
baseline estimates of separation which we can compare with estimates of separation generated by
the layers of the deep learning models.  Given these distributions, we can now evaluate the
interlayer representations produced by the deep learning model. 

\subsection{Inside the Black Box: before training}\label{sub:initModel}

We wish to characterize the transformations produced by the neural network layers at model
initialization as well as after training.  Characterizing the model at initialization allows one to
define a baseline to compare with the model produced via the training process, enabling
quantification of the adaptation (or learning) that each model layer undergoes as a model is trained.

For deep learning model initialization, a lesson every practitioner is taught early in their machine learning
education is  that their model layers are to be initialized with small random values.  These values
are generally drawn from a normal distribution and appropriately scaled by a function of the number
of filters in the layer.  

This random initialization has an interesting property.  In~1984, Johnson and Lindenstrauss~\cite{johnson84} proved that any $n$ point subset of
Euclidean space can be mapped into a random subspace of $k = \mathcal{O}(\log{n/\epsilon^2})$ dimensions,
and the interpoint distances of the points projected in that subspace differ from the distances
in the original space only by $(1\pm\epsilon)$.   Increasing $k$, the number of random projections,
decreases the expected interpoint distance error. Thus, the process of randomly initializing the weights of each layer can possibly act as
a set of random projections and retain the interpoint
distances in the original space within some distortion value $\epsilon$.
That is, a deep learning layer may retain the interpoint distances
of the training and validation sets via the random initialization alone.  This is a motivational factor in
the design of \textit{extreme learning machines}~\cite{huang06}, which randomly initialize and fix
the single hidden layer input weights.  In effect, random projections can provide `data-agnostic
transfer learning' as they can retain interpoint distances independent of the input data
distributions.  This is something to note, as it provides the mechanism for the initial state of a
deep learning system to `do no harm'.   It also would allow arbitrarily large networks to
operate on relatively easy problems within the bounds of the product of the distortions induced at
each layer  $\textstyle(1\pm\epsilon) = \prod_{i=1}^{N}{(1\pm\epsilon_i)}$.

\begin{figure*}[!t]
\centering
\begin{subfigure}[b]{0.3\linewidth}
  \includegraphics[width=\linewidth]{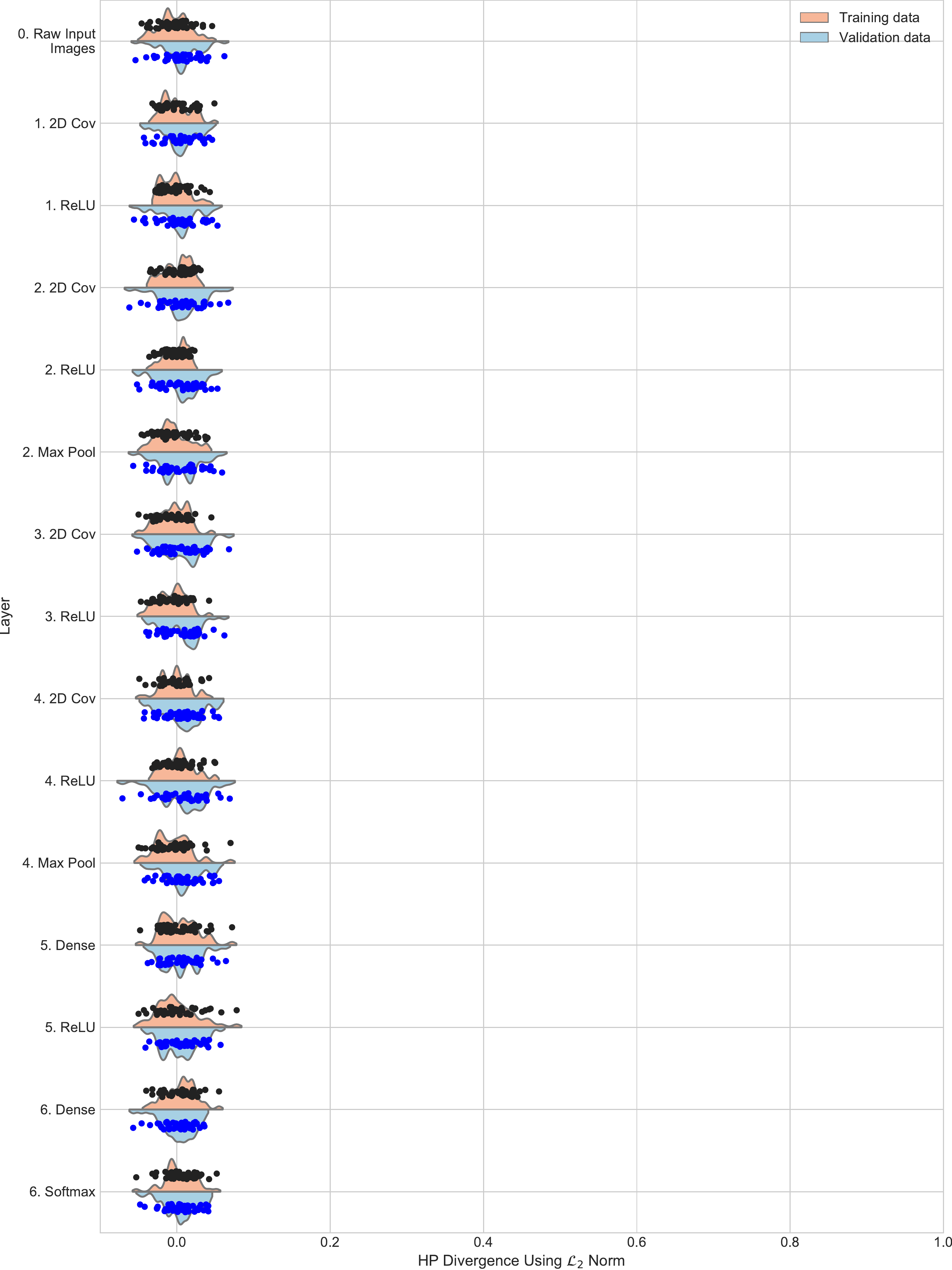}
  \caption{}
\end{subfigure}
\begin{subfigure}[b]{0.3\linewidth}
  \includegraphics[width=\linewidth]{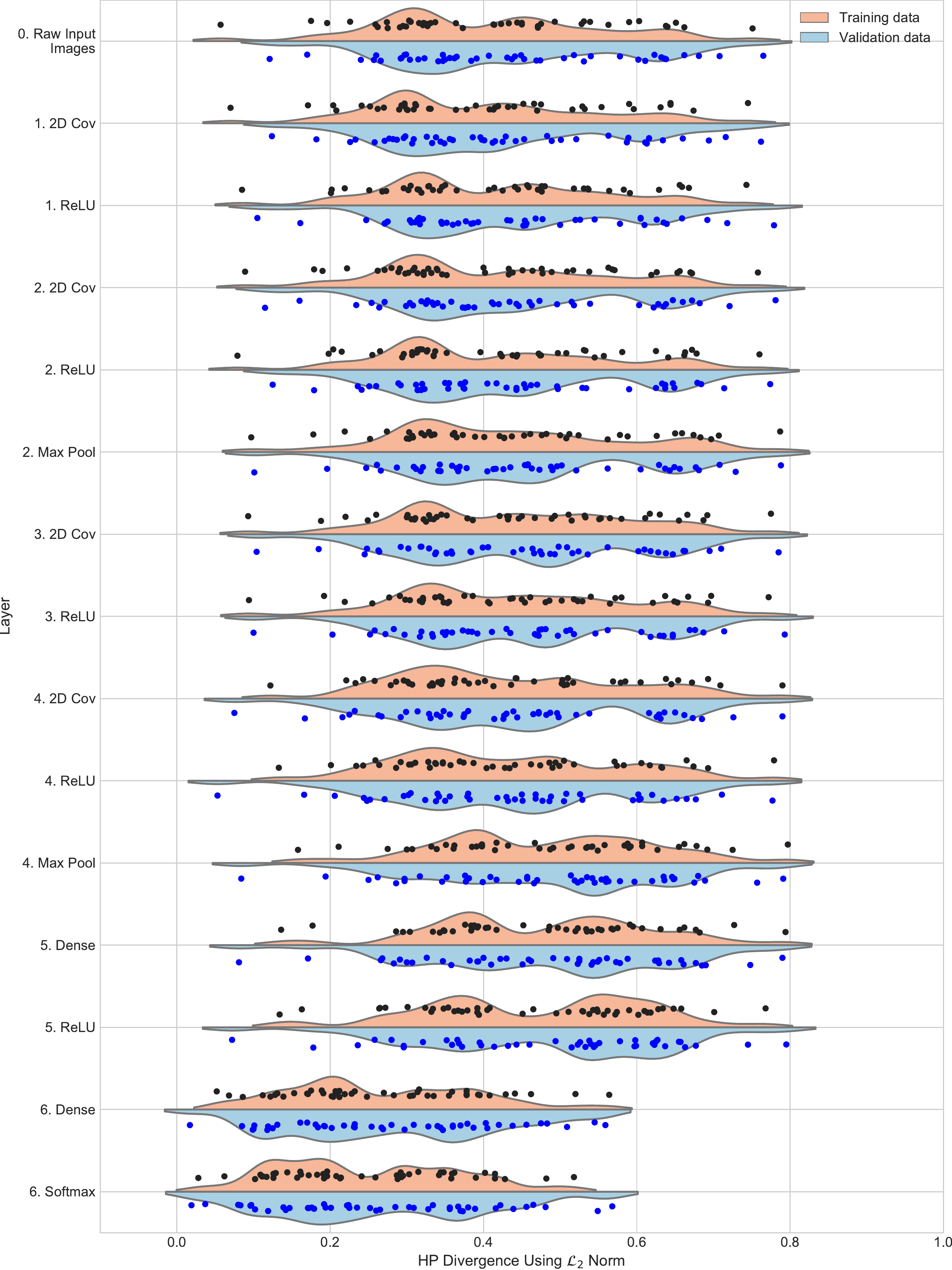}
  \caption{}
\end{subfigure}
\begin{subfigure}[b]{0.3\linewidth}
  \includegraphics[width=\linewidth]{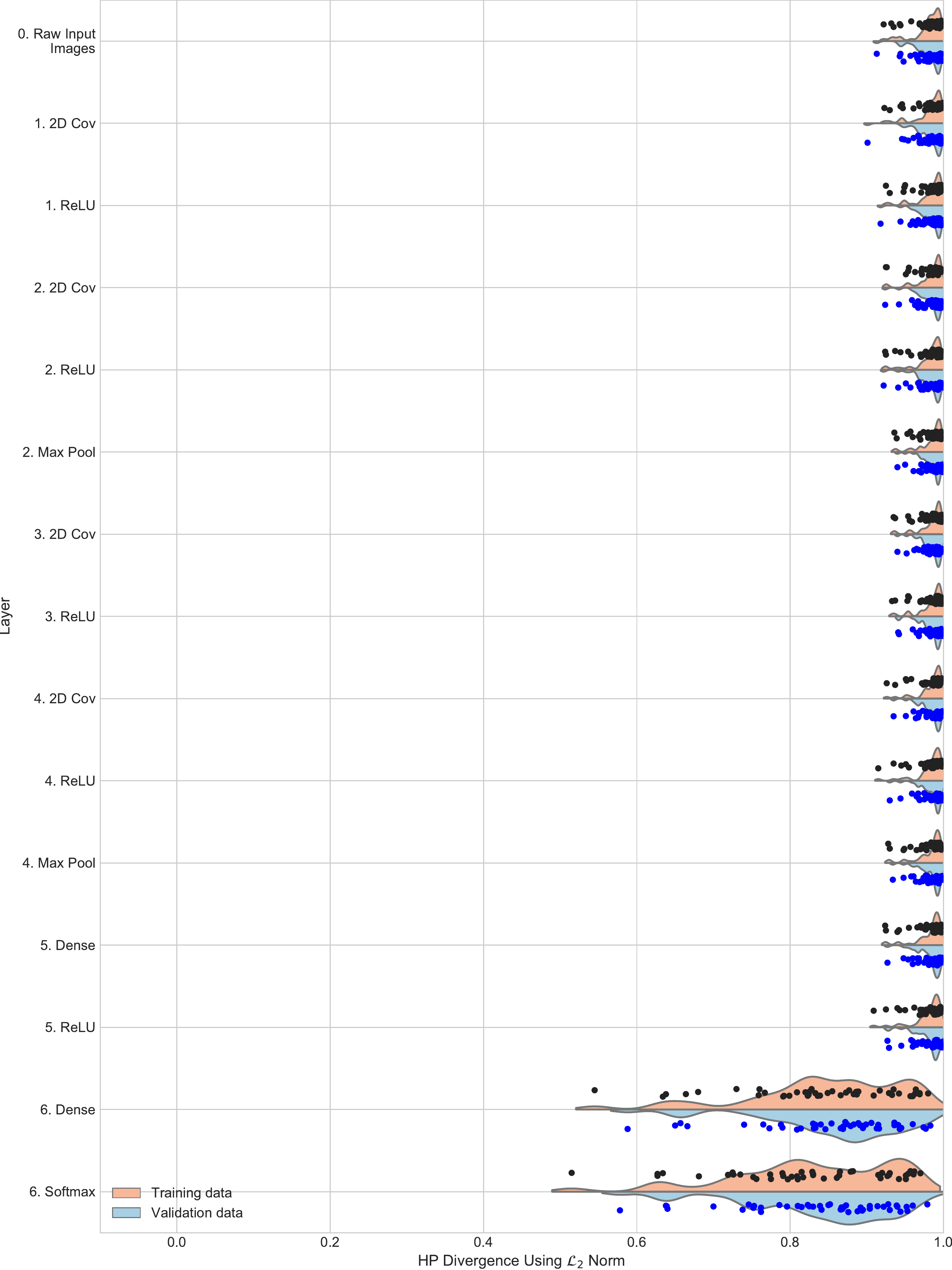}
  \caption{}
\end{subfigure}
\caption{$\mathcal{H}$ class-pair statistics at each layer for the randomly initialized model state
         (epoch~0) for~(a)~CIFAR10 with random class labels,~(b)~CIFAR10 with true class
         labels, and~(c)~MNIST with true class labels.}
\label{figure:randomModels}
\end{figure*}

For each layer of the neural network model, we compute the class-pair HP statistics for the induced
representations of the training and validation data. Figure~\ref{figure:randomModels} illustrates the
$\boldsymbol{\mathcal{H}}$ distributions at the output of each layer of randomly initialized networks.  In
each figure, the HP statistics for the training data are represented by top/black dots and an orange kernel
density at each layer of the network, and the HP statistics generated by the test data are represented
by bottom/blue dots and a corresponding blue kernel density estimate.

The motivation for this discussion is two-fold.  First, demonstrating that random initialization of
deep learning models can maintain the interpoint distances between classes.  Second, this
initialized behavior provides us a baseline to quantify the effective changes in the distribution of
class-pair separation $\boldsymbol{\mathcal{H}}$ produced in training the models. Figure%
~\ref{figure:randomModels} illustrates the distribution of $\mathcal{H}$ statistics produced by each layer
for the initialized models before any adaptation or training.  We are interested to test if the
initialized state of each layer is behaving as a set of distance-preserving random projections, or
given an appropriate number of random projections, if the average class-pair separation between input
and output spaces are statistically equivalent.
A nonparametric two-sided permutation test of the
statistical equivalence of the mean class separation $\bar{\mathcal{H}}$~\eqref{eq:meanh} between the input and output spaces of each layer is
applied. For each layer, the hypotheses under test corresponds to the following:
\begin{equation} 
  \label{eq:diffEqTest1}
  \mathbf{H_0}: {\bar{\mathcal{H}}}_{(k-1,0)} = {\bar{\mathcal{H}}}_{(k,0)} \hspace{5mm} vs.
  \hspace{5mm} \mathbf{H_A}:  {\bar{\mathcal{H}}}_{(k-1,0)} \neq {\bar{\mathcal{H}}}_{(k,0)} \hspace{5mm}
\end{equation}

For each layer, we select an $\alpha = 0.025$ level of significance for the two-sided test, and
compute the $p$-values for the model instances of each task.
These results are summarized in Table%
~\ref{table:randomFlow}.

\begin{table*}[!t]
\centering
\caption{Two-sided permutation test to detect change in $\bar{\mathcal{H}}$ between layers (before
         training) with critical value $\alpha = 0.025$. \textcolor{red}{Red font} denotes layer
         instances for which we reject $\mathbf{H_0}$, and black font denotes layers for which we fail
         to reject $\mathbf{H_0}$ \eqref{eq:diffEqTest1}. 
          (Note: $\Delta{\bar{\mathcal{H}}} = \bar{\mathcal{H}}^{(t)}_{(k,0)} - \bar{\mathcal{H}}^{(t)}_{(k-1,0)}$) }
\label{table:randomFlow}
\begin{tabular}{>{\centering\arraybackslash}m{12mm}|>{\centering\arraybackslash}m{12mm}|>{\centering\arraybackslash}m{22mm}
|>{\centering\arraybackslash}m{19mm}|>{\centering\arraybackslash}m{22mm}|>{\centering\arraybackslash}m{19mm}
|>{\centering\arraybackslash}m{22mm}|>{\centering\arraybackslash}m{19mm}}
\hline
\bfseries Input Space&\bfseries Output Space& \multicolumn{2}{c}{\bfseries CIFAR10 w Random}& \multicolumn{2}{c}{\bfseries CIFAR10 w True}&
\multicolumn{2}{c}{\bfseries MNIST w True} \\
& &$\Delta{\bar{\mathcal{H}}}$ & \textit{p}-values & $\Delta{\bar{\mathcal{H}}}$ & \textit{p}-values&
$\Delta{\bar{\mathcal{H}}}$ & \textit{p}-values \\
\hline \hline
0.Input&1.Conv&.003; .002; -.004; .003; -.003&.460; .649; .445; .556; .515&-.012; .005; -.017;
-.003; -.004&.714; .864; .593; .923; .912&.002; .001; -.001; .000; .000&.672; .792; .864; .934;
.900\\
\hline
1.Conv&1.ReLU&-.002; -.003; .004; .002; .003&.531; .523; .434; .675; .641&.016; -.022; -.003; .001;
-.000&.607; .463; .933; .969; .991&.001; .000; .001; .002; .000&.875; .906; .713; .648; .918\\
\hline
1.ReLU&2.Conv&.003; .001; -.001; -.002; -.001&.375; .801; .856; .650; .906&.003; -.006; -.001;
-.009; -.005&.924; .842; .964; .771; .878&.000; -.000; -.001; -.001; -.000&.970; .981; .878; .826;
.933\\
\hline
2.Conv&2.ReLU&-.001; -.002; .003; -.002; .001&.671; .626; .526; .667; .906&.002; -.017; .005; -.003;
.003&.959; .562; .869; .929; .912&-.001; -.000; -.001; -.001; .001&.730; .962; .803; .748; .839\\
\hline
2.ReLU&2.MaxPool&-.005; .004; -.001; -.007; -.004&.202; .390; .832; .195; .576&.019; .035; .015;
.020; .015&.563; .241; .623; .505; .624&.003; .003; .005; .004; .002&.365; .394; .202; .294; .565\\
\hline
2.MaxPool&3.Conv&.001; -.001; .004; -.006; -.003&.873; .876; .428; .286; .695&.003; -.001; .003;
-.002; -.009&.936; .978; .931; .948; .755&.000; .000; -.000; -.000; .000&.998; 1.00; .998; .956;
.973\\
\hline
3.Conv&3.ReLU&-.001; -.000; -.003; .002; .007&.747; .969; .574; .629; .246&.002; .003; .007; .004;
-.002&.960; .914; .827; .895; .957&-.001; -.001; -.001; -.001; -.001&.872; .849; .721; .768; .895\\
\hline
3.ReLU&4.Conv&.005; .004; -.000; .000; -.005&.231; .345; .932; .954; .379&-.006; -.001; -.001;
-.004; -.004&.861; .967; .975; .897; .904&.000; .001; .001; .001; .001&.998; .860; .823; .777;
.842\\
\hline
4.Conv&4.ReLU&.005; -.000; .001; .003; .001&.220; .945; .842; .551; .821&-.003; -.003; -.006; -.008;
.005&.936; .913; .851; .797; .878&-.001; .000; -.000; -.000; .000&.750; .945; .954; .893; .983\\
\hline
4.ReLU&4.MaxPool&-.009; .008; -.004; -.004; .004&.059; .092; .364; .425; .444&.049; .054; .040;
.051; .038&.113; .061; .204; .096; .212&-.000; .000; .001; .001; .001&.962; .987; .840; .778; .777\\
\hline
4.MaxPool&5.Dense&.009; .000; .000; .000; .001&.061; .976; .925; .996; .838&-.005; -.001; .002;
.003; -.003&.858; .976; .962; .920; .908&-.001; -.001; -.001; -.001; -.001&.758; .746; .633; .880;
.682\\
\hline
5.Dense&5.ReLU&-.004; .004; .005; .001; -.006&.424; .385; .283; .925; .199&-.007; -.014; -.008;
-.003; .003&.813; .596; .789; .908; .915&-.002; -.002; -.001; -.002; -.002&.696; .654; .723; .549;
.666\\
\hline
5.ReLU&6.Dense&.005; \textbf{\textcolor{red}{-.017}}; \textbf{\textcolor{red}{-.017}}; .005; .003&.330;
\textbf{\textcolor{red}{.001}}; \textbf{\textcolor{red}{.000}}; .305; .549&\textbf{\textcolor{red}{-.202}};
\textbf{\textcolor{red}{-.226}}; \textbf{\textcolor{red}{-.182}}; \textbf{\textcolor{red}{-.189}};
\textbf{\textcolor{red}{-.198}}&\textbf{\textcolor{red}{.000}}; \textbf{\textcolor{red}{.000}}; \textbf{\textcolor{red}{.000}};
\textbf{\textcolor{red}{.000}}; \textbf{\textcolor{red}{.000}}&\textbf{\textcolor{red}{-.135}}; \textbf{\textcolor{red}{-.173}};
\textbf{\textcolor{red}{-.202}}; \textbf{\textcolor{red}{-.219}}; \textbf{\textcolor{red}{-.177}}&\textbf{\textcolor{red}{.000}};
\textbf{\textcolor{red}{.000}}; \textbf{\textcolor{red}{.000}}; \textbf{\textcolor{red}{.000}}; \textbf{\textcolor{red}{.000}}\\
\hline
6.Dense&6.Softmax&-.001; -.002; -.002; -.007; -.000&.891; .665; .688; .145; .933&-.018; -.011;
-.010; -.013; -.017&.473; .457; .690; .618; .446&-.018; -.030; -.030; -.014; -.016&.419; .280; .296;
.652; .582\\
\hline
\end{tabular}
\end{table*}

Table~\ref{table:randomFlow} demonstrates that even the simple convolutional neural network model
(as defined in Table~\ref{table:model}) has sufficient number of convolutional filters such that the random
values in each layer preserve the intrinsic separation between the three datasets.  The notable
exception is the~6.Dense layer, for which the null hypothesis is rejected in~12 of the~15
experiments.  Note that the input space of the~6.Dense layer is~512, but the output space is the number
of classes, which in each case is~10 dimensional.  Hence the change in class interpoint distances is easily
explained by the expected distortion induced by randomly projecting the originally high
dimensional data into a~10 dimensional subspace.

In our view, this distance preserving property of the initialization phase has positive and negative
consequences.  On the positive side, any adequately sized initialized model has the
demonstrable ability to retain interpoint distances between classes. On the negative side, this
`do no harm' property allows a machine learning practitioner to blindly apply any deep learning
model to a task at hand without spending time understanding the data and underlying
phenomenon being operating on, treating the model as a black box.
While the ability to apply any deep learning model to multiple tasks
is not necessarily negative, we caution application without an understanding and deep analysis of
the original task versus any new task and the potential hidden affects of doing so.

\subsection{Inside the Black Box: layer adaptation}\label{sub:effModelAdapt}

\begin{figure*}[!t]
\centering
\begin{subfigure}[b]{0.3\linewidth}
  \includegraphics[width=\linewidth]{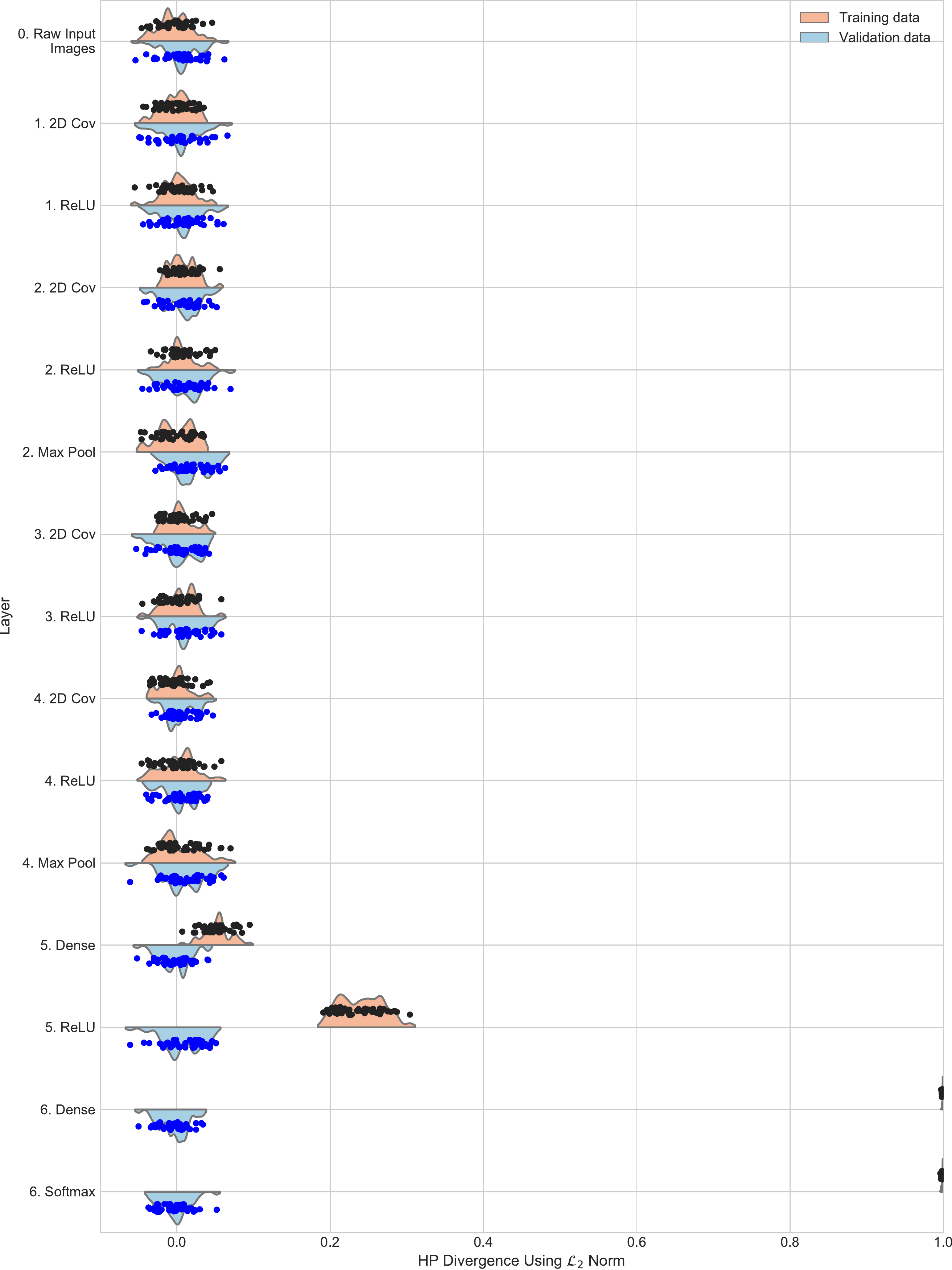}
  \caption{}
\end{subfigure}
\begin{subfigure}[b]{0.3\linewidth}
  \includegraphics[width=\linewidth]{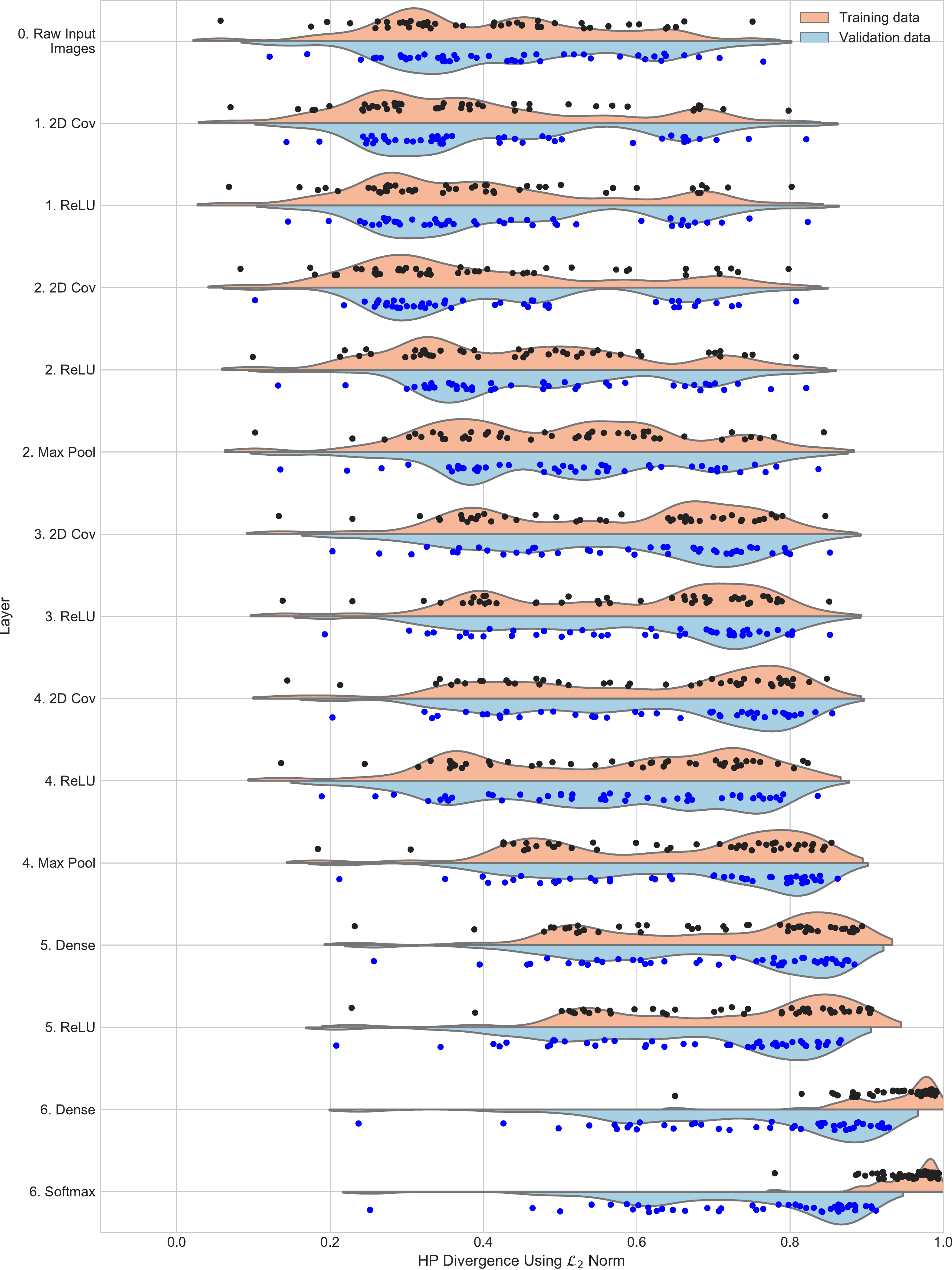}
  \caption{}
\end{subfigure}
\begin{subfigure}[b]{0.3\linewidth}
  \includegraphics[width=\linewidth]{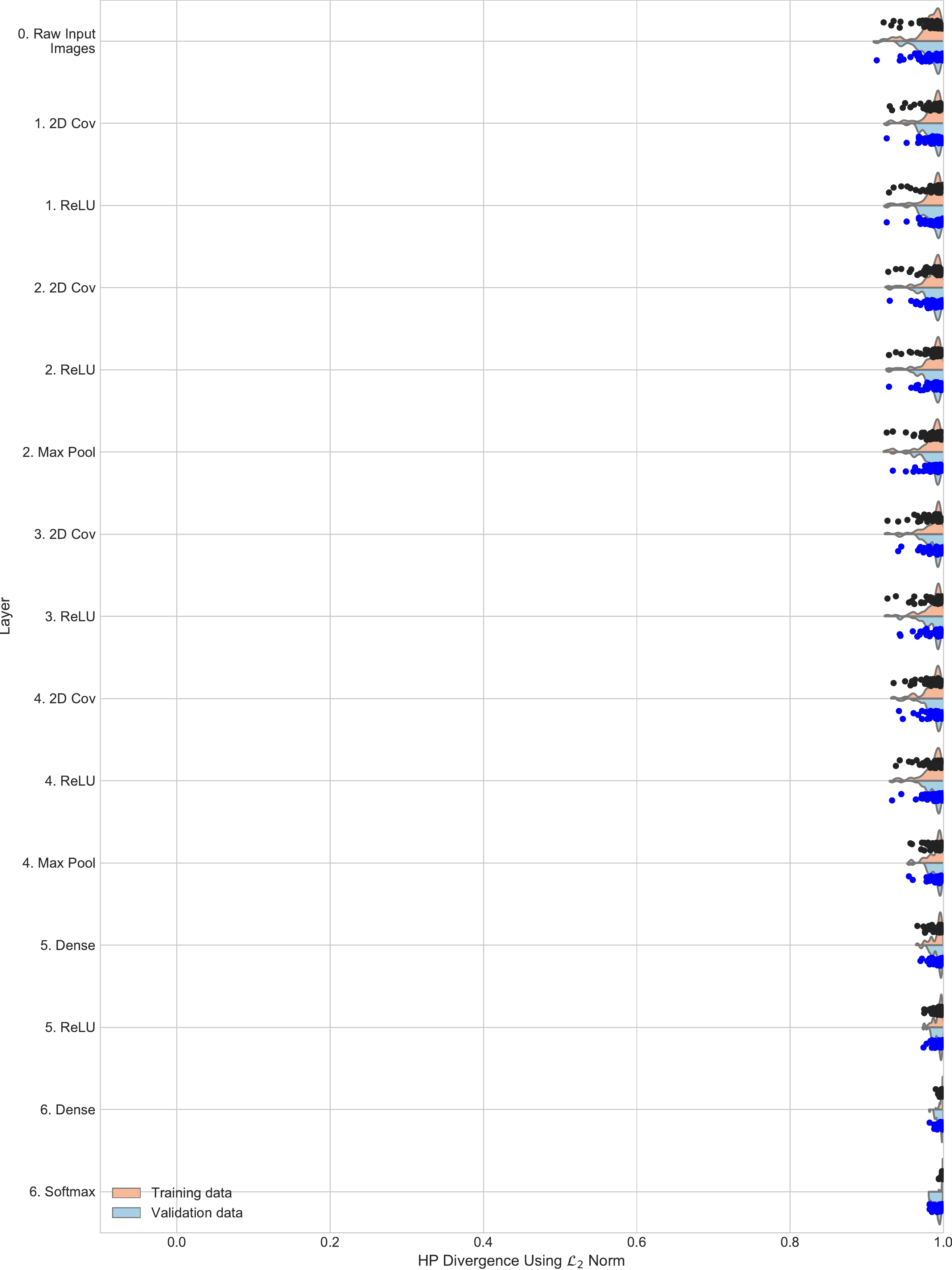}
  \caption{}
\end{subfigure}
\caption{Illustrative HP class-pair statistics of~(a)~CIFAR10 with random labels,~(b)~CIFAR10 with
         true labels, and~(c)~MNIST with true labels, computed on respective models after training
         is complete.}
\label{figure:trained}
\end{figure*}

As described in the previous section, five model instantiations are trained on each task,
where training is stopped~(a)~when training performance vs the validation performance is maximal, or~(b)~in the
random label case, after 200 epochs.  Figure~\ref{figure:trained} illustrates the training (orange,
above) and validation (blue, below) $\boldsymbol{\mathcal{H}}$ statistics at each layer of a trained model
for each dataset.

As previously discussed, one would hope that what is learned by a model via training manifests
itself as a concentration of class measure, that is, some combination of within-class distances
decreasing and/or between-class distances increasing.  The HP statistics would capture
this concentration with a corresponding higher value as a result of fewer edges in a minimal
spanning tree connecting disparate class samples. Therefore, using the training set we compare
the class separation statistics in each layer before and after training, and test if there is a
statistical significant change to mean class separation $\bar{\mathcal{H}}$~\eqref{eq:meanh}. Using a permutation test, we test the following hypotheses:
\begin{equation}\label{eq:diffT0}
  \mathbf{H}_0 : \bar{\mathcal{H}}_{(k,T)} - \bar{\mathcal{H}}_{(k,0)} \leq 0 \hspace{5mm} vs.
  \hspace{5mm} \mathbf{H}_A : \bar{\mathcal{H}}_{(k,T)} - \bar{\mathcal{H}}_{(k,0)} > 0
\end{equation}
For each model instantiation, the mean differences and associated $p$-values for~\eqref{eq:diffT0}
are given in Table~\ref{table:InitvsTrainedModels}.

\begin{table*}[!t]
\centering
\caption{Differences between the trained and initialized $\boldsymbol{\mathcal{H}}$ class-pair statistics of a layer,
         and respective $p$-values for the
         corresponding one-sided permutation test~\eqref{eq:diffT0}. \textcolor{red}{Red font}
         denotes layer instances for which we reject $\mathbf{H_0}$, and black font denotes layers for
         which we fail to reject $\mathbf{H_0}$. 
         (Note: $\Delta{\bar{\mathcal{H}}} = \bar{\mathcal{H}}^{(t)}_{(k,T)} - \bar{\mathcal{H}}^{(t)}_{(k,0)}$)}
\label{table:InitvsTrainedModels}
\begin{tabular}{>{\centering\arraybackslash}m{12mm}|>{\centering\arraybackslash}m{22mm}
|>{\centering\arraybackslash}m{22mm}|>{\centering\arraybackslash}m{22mm}|>{\centering\arraybackslash}m{22mm}
|>{\centering\arraybackslash}m{22mm}|>{\centering\arraybackslash}m{22mm}}
\hline
\bfseries Output Space& \multicolumn{2}{c|}{\bfseries CIFAR10 w Random}& \multicolumn{2}{c|}{\bfseries CIFAR10 w True}&
\multicolumn{2}{c}{\bfseries MNIST w True} \\
 &$\Delta{\bar{\mathcal{H}}}$ & \textit{p}-values & $\Delta{\bar{\mathcal{H}}}$ & \textit{p}-values&
 $\Delta{\bar{\mathcal{H}}}$ & \textit{p}-values \\
\hline \hline
1.Conv&-.000; -.002; .000; -.007; .000&.507; .645; .472; .888; .482&-.007; .005; -.008; .004;
-.004&.580; .435; .591; .453; .542&.001; .002; .004; .003; .003&.346; .277; .126; .209; .248\\
\hline
1.ReLU&.004; .004; -.004; -.009; -.001&.137; .225; .775; .955; .587&-.012; .029; .005; .015;
.005&.646; .183; .440; .320; .448&.001; .002; .003; .001; .002&.417; .316; .208; .354; .274\\
\hline
2.Conv&.007; -.009; -.011; -.014; -.004&.032; .973; .976; .995; .812&-.017; .039; -.006; .028;
.001&.690; .121; .570; .200; .491&.001; .002; .004; .003; .003&.422; .247; .161; .243; .202\\
\hline
2.ReLU&\textbf{\textcolor{red}{.008}}; -.003; -.012; -.015; -.008&\textbf{\textcolor{red}{.017}}; .706; .989; .998;
.913&.032; \textbf{\textcolor{red}{.095}}; .051; \textbf{\textcolor{red}{.066}}; .051&.174; \textbf{\textcolor{red}{.002}};
.068; \textbf{\textcolor{red}{.024}}; .068&.002; .003; .005; .004; .002&.281; .233; .105; .161; .271\\
\hline
2.MaxPool&.005; .004; -.015; -.010; -.004&.126; .161; .999; .967; .725&.065; \textbf{\textcolor{red}{.100}};
\textbf{\textcolor{red}{.094}}; \textbf{\textcolor{red}{.081}}; \textbf{\textcolor{red}{.073}}&.030; \textbf{\textcolor{red}{.002}};
\textbf{\textcolor{red}{.003}}; \textbf{\textcolor{red}{.008}}; \textbf{\textcolor{red}{.016}}&-.002; -.002; -.001; -.001;
-.001&.722; .669; .647; .655; .603\\
\hline
3.Conv&\textbf{\textcolor{red}{.009}}; .001; -.021; .000; .004&\textbf{\textcolor{red}{.011}}; .388; 1.00; .498;
.233&\textbf{\textcolor{red}{.127}}; \textbf{\textcolor{red}{.117}}; \textbf{\textcolor{red}{.168}}; \textbf{\textcolor{red}{.097}};
\textbf{\textcolor{red}{.118}}&\textbf{\textcolor{red}{.000}}; \textbf{\textcolor{red}{.000}}; \textbf{\textcolor{red}{.000}};
\textbf{\textcolor{red}{.002}}; \textbf{\textcolor{red}{.000}}&-.000; .001; .001; .001; .001&.525; .433; .328; .355;
.440\\
\hline
3.ReLU&\textbf{\textcolor{red}{.010}}; .001; -.020; -.006; -.008&\textbf{\textcolor{red}{.010}}; .365; 1.00; .896;
.926&\textbf{\textcolor{red}{.140}}; \textbf{\textcolor{red}{.135}}; \textbf{\textcolor{red}{.184}}; \textbf{\textcolor{red}{.117}};
\textbf{\textcolor{red}{.149}}&\textbf{\textcolor{red}{.000}}; \textbf{\textcolor{red}{.000}}; \textbf{\textcolor{red}{.000}};
\textbf{\textcolor{red}{.000}}; \textbf{\textcolor{red}{.000}}&-.000; .001; .002; .002; .001&.500; .431; .264; .281;
.390\\
\hline
4.Conv&-.002; .006; -.014; -.011; .008&.669; .111; .999; .989; .053&\textbf{\textcolor{red}{.168}};
\textbf{\textcolor{red}{.131}}; \textbf{\textcolor{red}{.205}}; \textbf{\textcolor{red}{.129}};
\textbf{\textcolor{red}{.148}}&\textbf{\textcolor{red}{.000}}; \textbf{\textcolor{red}{.000}}; \textbf{\textcolor{red}{.000}};
\textbf{\textcolor{red}{.000}}; \textbf{\textcolor{red}{.000}}&.000; .001; .002; .001; .001&.444; .381; .278; .330;
.358\\
\hline
4.ReLU&-.001; .007; -.006; -.007; -.003&.610; .084; .919; .909; .704&\textbf{\textcolor{red}{.128}};
\textbf{\textcolor{red}{.098}}; \textbf{\textcolor{red}{.190}}; .011; \textbf{\textcolor{red}{.106}}&\textbf{\textcolor{red}{.000}};
\textbf{\textcolor{red}{.002}}; \textbf{\textcolor{red}{.000}}; .366; \textbf{\textcolor{red}{.001}}&.001; .001; .002; .003;
.002&.336; .397; .228; .210; .314\\
\hline
4.MaxPool&.009; -.007; -.016; .006; -.011&.045; .928; .999; .127; .988&\textbf{\textcolor{red}{.161}};
\textbf{\textcolor{red}{.163}}; \textbf{\textcolor{red}{.207}}; \textbf{\textcolor{red}{.125}};
\textbf{\textcolor{red}{.181}}&\textbf{\textcolor{red}{.000}}; \textbf{\textcolor{red}{.000}}; \textbf{\textcolor{red}{.000}};
\textbf{\textcolor{red}{.000}}; \textbf{\textcolor{red}{.000}}&\textbf{\textcolor{red}{.006}}; .005; \textbf{\textcolor{red}{.006}};
.005; .004&\textbf{\textcolor{red}{.023}}; .049; \textbf{\textcolor{red}{.015}}; .038; .068\\
\hline
5.Dense&\textbf{\textcolor{red}{.049}}; \textbf{\textcolor{red}{.024}}; \textbf{\textcolor{red}{.036}}; \textbf{\textcolor{red}{.035}};
\textbf{\textcolor{red}{.029}}&\textbf{\textcolor{red}{.000}}; \textbf{\textcolor{red}{.000}}; \textbf{\textcolor{red}{.000}};
\textbf{\textcolor{red}{.000}}; \textbf{\textcolor{red}{.000}}&\textbf{\textcolor{red}{.227}}; \textbf{\textcolor{red}{.231}};
\textbf{\textcolor{red}{.257}}; \textbf{\textcolor{red}{.225}}; \textbf{\textcolor{red}{.258}}&\textbf{\textcolor{red}{.000}};
\textbf{\textcolor{red}{.000}}; \textbf{\textcolor{red}{.000}}; \textbf{\textcolor{red}{.000}};
\textbf{\textcolor{red}{.000}}&\textbf{\textcolor{red}{.009}}; \textbf{\textcolor{red}{.008}}; \textbf{\textcolor{red}{.009}};
\textbf{\textcolor{red}{.007}}; \textbf{\textcolor{red}{.007}}&\textbf{\textcolor{red}{.001}}; \textbf{\textcolor{red}{.001}};
\textbf{\textcolor{red}{.000}}; \textbf{\textcolor{red}{.001}}; \textbf{\textcolor{red}{.002}}\\
\hline
5.ReLU&\textbf{\textcolor{red}{.241}}; \textbf{\textcolor{red}{.234}}; \textbf{\textcolor{red}{.233}}; \textbf{\textcolor{red}{.238}};
\textbf{\textcolor{red}{.246}}&\textbf{\textcolor{red}{.000}}; \textbf{\textcolor{red}{.000}}; \textbf{\textcolor{red}{.000}};
\textbf{\textcolor{red}{.000}}; \textbf{\textcolor{red}{.000}}&\textbf{\textcolor{red}{.248}}; \textbf{\textcolor{red}{.256}};
\textbf{\textcolor{red}{.279}}; \textbf{\textcolor{red}{.237}}; \textbf{\textcolor{red}{.270}}&\textbf{\textcolor{red}{.000}};
\textbf{\textcolor{red}{.000}}; \textbf{\textcolor{red}{.000}}; \textbf{\textcolor{red}{.000}};
\textbf{\textcolor{red}{.000}}&\textbf{\textcolor{red}{.012}}; \textbf{\textcolor{red}{.011}}; \textbf{\textcolor{red}{.012}};
\textbf{\textcolor{red}{.010}}; \textbf{\textcolor{red}{.011}}&\textbf{\textcolor{red}{.000}}; \textbf{\textcolor{red}{.000}};
\textbf{\textcolor{red}{.000}}; \textbf{\textcolor{red}{.000}}; \textbf{\textcolor{red}{.000}}\\
\hline
6.Dense&\textbf{\textcolor{red}{.994}}; \textbf{\textcolor{red}{1.002}}; \textbf{\textcolor{red}{1.00}};
\textbf{\textcolor{red}{.989}}; \textbf{\textcolor{red}{.996}}&\textbf{\textcolor{red}{.000}}; \textbf{\textcolor{red}{.000}};
\textbf{\textcolor{red}{.000}}; \textbf{\textcolor{red}{.000}}; \textbf{\textcolor{red}{.000}}&\textbf{\textcolor{red}{.669}};
\textbf{\textcolor{red}{.681}}; \textbf{\textcolor{red}{.610}}; \textbf{\textcolor{red}{.718}};
\textbf{\textcolor{red}{.668}}&\textbf{\textcolor{red}{.000}}; \textbf{\textcolor{red}{.000}}; \textbf{\textcolor{red}{.000}};
\textbf{\textcolor{red}{.000}}; \textbf{\textcolor{red}{.000}}&\textbf{\textcolor{red}{.152}}; \textbf{\textcolor{red}{.188}};
\textbf{\textcolor{red}{.218}}; \textbf{\textcolor{red}{.234}}; \textbf{\textcolor{red}{.192}}&\textbf{\textcolor{red}{.000}};
\textbf{\textcolor{red}{.000}}; \textbf{\textcolor{red}{.000}}; \textbf{\textcolor{red}{.000}}; \textbf{\textcolor{red}{.000}}\\
\hline
6.Softmax&\textbf{\textcolor{red}{.995}}; \textbf{\textcolor{red}{1.004}}; \textbf{\textcolor{red}{1.002}};
\textbf{\textcolor{red}{.996}}; \textbf{\textcolor{red}{.996}}&\textbf{\textcolor{red}{.000}}; \textbf{\textcolor{red}{.000}};
\textbf{\textcolor{red}{.000}}; \textbf{\textcolor{red}{.000}}; \textbf{\textcolor{red}{.000}}&\textbf{\textcolor{red}{.706}};
\textbf{\textcolor{red}{.711}}; \textbf{\textcolor{red}{.634}}; \textbf{\textcolor{red}{.741}};
\textbf{\textcolor{red}{.706}}&\textbf{\textcolor{red}{.000}}; \textbf{\textcolor{red}{.000}}; \textbf{\textcolor{red}{.000}};
\textbf{\textcolor{red}{.000}}; \textbf{\textcolor{red}{.000}}&\textbf{\textcolor{red}{.171}}; \textbf{\textcolor{red}{.219}};
\textbf{\textcolor{red}{.250}}; \textbf{\textcolor{red}{.249}}; \textbf{\textcolor{red}{.209}}&\textbf{\textcolor{red}{.000}};
\textbf{\textcolor{red}{.000}}; \textbf{\textcolor{red}{.000}}; \textbf{\textcolor{red}{.000}}; \textbf{\textcolor{red}{.000}}\\
\hline
\end{tabular}
\end{table*}

In reviewing the mean difference statistics and associated $p$-values in Table%
~\ref{table:InitvsTrainedModels}, one will note that for all datasets the mean separation as
estimated by $\bar{\mathcal{H}}$ for the first convolutional and ReLU layers have not significantly
improved between initialized and trained states.  Moreover, in four out of five models for the
CIFAR10 dataset trained with
random labels and all five MNIST trained models, we fail to reject that all the convolutional layers of the
trained network are producing statistically equivalent or less class separation statistics.  These
two cases are extreme, one being quite easy, and the other impossible. In neither case are the
models required or able to improve class separation in the convolutional layers to optimize its
performance.  Conversely, the models trained with true CIFAR10 labels demonstrate statistically significant
changes in separation statistics in the third and fourth convolutional layers.  In all cases,
statistically significant differences in mean class separation occur by the time the representations
are mapped in the dense layers, indicating that the dense layers have adapted and the corresponding
learned mappings significantly improve ${\mathcal{H}}$ class statistics for the training data.

\subsection{Inside the Black Box: individual layer contribution}\label{sub:indivLayerContrib}

To quantitatively estimate the improvement in class separation for each layer in the trained models,
we again compute the difference between the class-pair statistics $\boldsymbol{\mathcal{H}}$, but now
between the input and output representations of each layer after training.  That is,
\begin{equation}\label{eq:diffTk}
\bar{\mathcal{H}}_{(k,T)} - \bar{\mathcal{H}}_{(k-1,T)}.
\end{equation} The results for each dataset and their five trained models are given in Table%
~\ref{table:InterLayerContribution} for  one-sided, permutation-based null hypothesis tests of the form
\begin{eqnarray}  \label{eq:diffTkNHT}
\mathbf{H_0}&:&\bar{\mathcal{H}}_{(k,T)} - \bar{\mathcal{H}}_{(k-1,T)} \leq 0 \hspace{5mm} vs.\nonumber\\
\mathbf{H_A}&:&\bar{\mathcal{H}}_{(k,T)} - \bar{\mathcal{H}}_{(k-1,T)} > 0.
\end{eqnarray}

Unlike Table~\ref{table:InitvsTrainedModels}, the results in Table%
~\ref{table:InterLayerContribution} are not characterizing the cumulative changes due to training,
but rather looking at the individual change in separation statistics given the input and output
representations of a layer.  We should caveat that failure to reject the null hypothesis does not
mean that a layer is not mapping the data into an improved (more separated) representation, but that
the differences are not statistically significant enough.  The behavior of the second
dense layer (6.Dense) is unique, where in all 15 cases we reject the null hypothesis with very
small $p$-values.  Significant changes to the data representations are occurring on that layer.
Nowhere is that more true than for the CIFAR10 dataset with random labels, where both dense layers effectively
translate the training data from a totally mixed state with the $\mathcal{H}$ values centered at zero, 
to completely separated with $\mathcal{H}$ statistics extremely close to one.  Since
this separation is meaningless because the classes are sampled from the same distribution, the
learning in this case is just overfitting or memorization of the training set.  In this example,
the statistics have clearly identified the layers in this model responsible for memorization.

\begin{table*}[!t]
\centering
\caption{Training data difference between input and output of each layer $\boldsymbol{\mathcal{H}}$ class-pair statistics
         for the trained models, and respective
         $p$-values for the one sided permutation test~\eqref{eq:diffTkNHT}.  \textcolor{red}{Red font}
         denotes layer instances for which we reject $\mathbf{H_0}$, and black font denotes layers for
         which we fail to reject $\mathbf{H_0}$. 
         (Note: $\Delta{\bar{\mathcal{H}}} = \bar{\mathcal{H}}^{(t)}_{(k,T)} - \bar{\mathcal{H}}^{(t)}_{(k-1,T)}$)}

\label{table:InterLayerContribution}
\begin{tabular}{>{\centering\arraybackslash}m{12mm}|>{\centering\arraybackslash}m{12mm}
|>{\centering\arraybackslash}m{22mm}|>{\centering\arraybackslash}m{19mm}
|>{\centering\arraybackslash}m{22mm}|>{\centering\arraybackslash}m{19mm}
|>{\centering\arraybackslash}m{22mm}|>{\centering\arraybackslash}m{19mm}}
\hline
\bfseries Input Space &\bfseries Output Space& \multicolumn{2}{c|}{\bfseries CIFAR10 w Random}& 
\multicolumn{2}{c|}{\bfseries CIFAR10 w True}& \multicolumn{2}{c}{\bfseries MNIST w True} \\
 & &$\Delta{\bar{\mathcal{H}}}$ & \textit{p}-values & $\Delta{\bar{\mathcal{H}}}$ & \textit{p}-values&
 $\Delta{\bar{\mathcal{H}}}$ & \textit{p}-values \\
\hline
\hline
0.Input&1.Conv&.003; .000; -.004; -.003; -.003&.243; .474; .741; .729; .721&-.019; .011; -.025;
.001; -.007&.708; .378; .767; .492; .582&.003; .003; .004; .003; .003&.200; .199; .169; .181; .208\\
\hline
1.Conv&1.ReLU&.002; .002; -.000; -.001; .001&.312; .303; .531; .542; .421&.010; .002; .010; .013;
.008&.386; .481; .389; .356; .410&-.000; .000; .000; .000; .000&.506; .492; .481; .480; .493\\
\hline
1.ReLU&2.Conv&.006; -.012; -.008; -.007; -.003&.070; .990; .911; .896; .756&-.002; .004; -.013;
.004; -.009&.525; .455; .635; .453; .591&.000; .001; .000; .000; .001&.494; .425; .502; .468; .433\\
\hline
2.Conv&2.ReLU&-.001; .004; .002; -.003; -.003&.573; .210; .330; .740; .752&.051; .039; .062; .035;
.053&.080; .142; .052; .164; .076&.000; .000; .000; .000; .000&.481; .496; .482; .500; .494\\
\hline
2.ReLU&2.MaxPool&-.007; \textbf{\textcolor{red}{.010}}; -.005; -.002; .000&.955; \textbf{\textcolor{red}{.015}}; .827;
.682; .471&.051; .040; .058; .035; .037&.071; .130; .055; .164; .152&-.001; -.001; -.001; -.001;
-.001&.599; .609; .629; .632; .608\\
\hline
2.MaxPool&3.Conv&.005; -.003; -.001; .004; .005&.127; .775; .603; .193; .146&.065; .017;
\textbf{\textcolor{red}{.077}}; .015; .036&.035; .318; \textbf{\textcolor{red}{.020}}; .334; .158&.002; .002; .003;
.003; .002&.301; .274; .210; .235; .323\\
\hline
3.Conv&3.ReLU&-.001; -.000; -.002; -.004; -.005&.610; .503; .616; .776; .840&.014; .021; .023; .024;
.029&.346; .285; .270; .251; .210&-.000; -.001; -.001; -.000; -.000&.534; .581; .568; .543; .513\\
\hline
3.ReLU&4.Conv&-.007; .009; .005; -.005; \textbf{\textcolor{red}{.011}}&.947; .039; .132; .865;
\textbf{\textcolor{red}{.010}}&.023; -.005; .020; .008; -.004&.276; .550; .307; .412; .545&.000; .001; .001;
.000; .001&.444; .371; .425; .440; .374\\
\hline
4.Conv&4.ReLU&.006; .000; \textbf{\textcolor{red}{.009}}; .007; -.009&.107; .462; \textbf{\textcolor{red}{.007}};
.058; .975&-.043; -.036; -.020; -.126; -.038&.866; .839; .701; 1.00; .845&-.000; .000; .000; .001;
.001&.522; .476; .458; .404; .428\\
\hline
4.ReLU&4.MaxPool&.001; -.005; -.015; .009; -.004&.397; .869; .999; .041; .806&\textbf{\textcolor{red}{.081}};
\textbf{\textcolor{red}{.119}}; .057; \textbf{\textcolor{red}{.165}}; \textbf{\textcolor{red}{.113}}&\textbf{\textcolor{red}{.015}};
\textbf{\textcolor{red}{.001}}; .068; \textbf{\textcolor{red}{.000}}; \textbf{\textcolor{red}{.002}}&.004; .004; .004; .003;
.004&.051; .068; .059; .089; .085\\
\hline
4.MaxPool&5.Dense&\textbf{\textcolor{red}{.049}}; \textbf{\textcolor{red}{.031}}; \textbf{\textcolor{red}{.053}};
\textbf{\textcolor{red}{.029}}; \textbf{\textcolor{red}{.040}}&\textbf{\textcolor{red}{.000}}; \textbf{\textcolor{red}{.000}};
\textbf{\textcolor{red}{.000}}; \textbf{\textcolor{red}{.000}}; \textbf{\textcolor{red}{.000}}&.061; .067; .051;
\textbf{\textcolor{red}{.103}}; \textbf{\textcolor{red}{.074}}&.042; .028; .080; \textbf{\textcolor{red}{.002}};
\textbf{\textcolor{red}{.020}}&.002; .002; .002; .002; .002&.111; .136; .192; .165; .195\\
\hline
5.Dense&5.ReLU&\textbf{\textcolor{red}{.188}}; \textbf{\textcolor{red}{.214}}; \textbf{\textcolor{red}{.202}};
\textbf{\textcolor{red}{.203}}; \textbf{\textcolor{red}{.212}}&\textbf{\textcolor{red}{.000}}; \textbf{\textcolor{red}{.000}};
\textbf{\textcolor{red}{.000}}; \textbf{\textcolor{red}{.000}}; \textbf{\textcolor{red}{.000}}&.013; .010; .015; .009;
.015&.351; .380; .338; .396; .332&.001; .001; .002; .001; .002&.176; .190; .126; .292; .091\\
\hline
5.ReLU&6.Dense&\textbf{\textcolor{red}{.758}}; \textbf{\textcolor{red}{.752}}; \textbf{\textcolor{red}{.750}};
\textbf{\textcolor{red}{.756}}; \textbf{\textcolor{red}{.753}}&\textbf{\textcolor{red}{.000}}; \textbf{\textcolor{red}{.000}};
\textbf{\textcolor{red}{.000}}; \textbf{\textcolor{red}{.000}}; \textbf{\textcolor{red}{.000}}&\textbf{\textcolor{red}{.220}};
\textbf{\textcolor{red}{.199}}; \textbf{\textcolor{red}{.148}}; \textbf{\textcolor{red}{.292}};
\textbf{\textcolor{red}{.201}}&\textbf{\textcolor{red}{.000}}; \textbf{\textcolor{red}{.000}}; \textbf{\textcolor{red}{.000}};
\textbf{\textcolor{red}{.000}}; \textbf{\textcolor{red}{.000}}&\textbf{\textcolor{red}{.005}}; \textbf{\textcolor{red}{.004}};
\textbf{\textcolor{red}{.004}}; \textbf{\textcolor{red}{.005}}; \textbf{\textcolor{red}{.004}}&\textbf{\textcolor{red}{.000}};
\textbf{\textcolor{red}{.000}}; \textbf{\textcolor{red}{.000}}; \textbf{\textcolor{red}{.000}}; \textbf{\textcolor{red}{.000}}\\
\hline
6.Dense&6.Softmax&-.000; -.000; .000; .000; -.000&.985; .581; .250; .145; .858&.019; .019; .015;
\textbf{\textcolor{red}{.010}}; .020&.049; .124; .221; \textbf{\textcolor{red}{.000}}; .070&\textbf{\textcolor{red}{.001}};
\textbf{\textcolor{red}{.001}}; \textbf{\textcolor{red}{.001}}; .001; \textbf{\textcolor{red}{.001}}&\textbf{\textcolor{red}{.001}};
\textbf{\textcolor{red}{.000}}; \textbf{\textcolor{red}{.000}}; .028; \textbf{\textcolor{red}{.000}}\\

\hline
\end{tabular}
\end{table*}

\subsection{Inside the Black Box: comparing model behavior on training vs. validation data}
\label{sub:trainVsVal}

Finally, we wish to investigate the layer-wise change in class separation induced on validation
data, that is, data that has not been explicitly used for model training, but in theory have the
same labeling distribution as the training data.  To remind the reader, Figure~\ref{figure:trained}
illustrates $\boldsymbol{\mathcal{H}}^{(t)}$  training (above, orange) and $\boldsymbol{\mathcal{H}}^{(v)}$ validation
(below, blue) class-pair distributions that have been computed at each layer for each trained model.
Additionally, we will look at detecting whether the change in statistics for validation and training $\mathcal{H}$  
distributions are significantly different. For the randomized CIFAR10 dataset, any difference in
layer training and validation behavior are by design due to the model overfitting and outright
memorization. This has allowed us to identify the layers responsible for memorization. 
For the datasets with non-randomized class labels, we will not attempt to tease out the
root causes for differences in behavior between training and validation data. Differences can be
due not only to overfitting by the deep learning model, but also from potential domain-shifts
between training and validation data. See~\cite{berishahero, ben-david10} for more information on
domain-shifted datasets and derived relationships between error bounds. For the true-labeled data,
identifying contributions due to domain-shifts and overfitting will be the focus for a future
paper or an exercise for an intrepid reader. To restate, our purpose is to demonstrate techniques that evoke
and enable the reader to gain insights into the behavior of individual layers of an optimized/trained deep learning model.  For the discussion that follows, we explicitly
use $\boldsymbol{\mathcal{H}}^{(t)}$ and $\boldsymbol{\mathcal{H}}^{(v)}$ to denote the distribution of class-pair
statistics computed on the training and validation data, respectively. 

We will now attempt to address the following questions:
\begin{itemize}
\item[Q1:]  Is there a statistically significant change in $\boldsymbol{\mathcal{H}}^{(v)}$ between
the input and output spaces of each layer?
\item[Q2:] Is the induced change in distributional separation of training and validation data
statistically equivalent?
\end{itemize}

To investigate Q1, we compute the difference in mean
distributional separation and test if this difference is statistically significant via a one-sided
permutation test, that is,
\begin{eqnarray}  \label{eq:PTvk}
\mathbf{H_0}&:&\bar{\mathcal{H}}^{(v)}_{(k,T)} - \bar{\mathcal{H}}^{(v)}_{(k-1,T)} \leq 0 \hspace{5mm} vs.\nonumber\\
\mathbf{H_A}&:&\bar{\mathcal{H}}^{(v)}_{(k,T)} - \bar{\mathcal{H}}^{(v)}_{(k-1,T)} > 0.
\end{eqnarray}
We are using a one-sided permutation test of means with our arbitrary critical value $\alpha =
0.025$.  The results for each dataset and their five trained models are given in Table%
~\ref{table:InterLayerVChange}.


\begin{table*}[!t]
\centering
\caption{Validation data differences in mean class-pair separation statistics between the
         input and output representations of a layer, and respective one-sided
         permutation test~$p$-values.  \textcolor{red}{Red font} denotes layer
         instances for which we reject $\mathbf{H_0}$, and black font denotes layers for which we fail
         to reject $\mathbf{H_0}$ ~\eqref{eq:PTvk}. 
          (Note: $\Delta{\bar{\mathcal{H}}} = \bar{\mathcal{H}}^{(v)}_{(k,T)} - \bar{\mathcal{H}}^{(v)}_{(k-1,T)}$)}
\label{table:InterLayerVChange}
\begin{tabular}{>{\centering\arraybackslash}m{12mm}|>{\centering\arraybackslash}m{12mm}
|>{\centering\arraybackslash}m{22mm}|>{\centering\arraybackslash}m{19mm}
|>{\centering\arraybackslash}m{22mm}|>{\centering\arraybackslash}m{19mm}
|>{\centering\arraybackslash}m{22mm}|>{\centering\arraybackslash}m{19mm}}
\hline
\bfseries Input Space &\bfseries Output Space& \multicolumn{2}{c|}{\bfseries CIFAR10 w Random}
& \multicolumn{2}{c|}{\bfseries CIFAR10 w True}& \multicolumn{2}{c}{\bfseries MNIST w True} \\
 & &$\Delta{\bar{\mathcal{H}}}$ & \textit{p}-values & $\Delta{\bar{\mathcal{H}}}$ & \textit{p}-values&
 $\Delta{\bar{\mathcal{H}}}$ & \textit{p}-values \\
 \hline \hline

0.Input&1.Conv&-.002; .006; -.003; -.001; -.004&.639; .035; .728; .543; .822&-.014; .017; -.021;
.005; -.003&.662; .308; .729; .441; .533&.004; .004; .004; .004; .004&.094; .113; .137; .119; .103\\
\hline
1.Conv&1.ReLU&.002; .000; -.002; .001; .001&.348; .488; .676; .418; .412&.009; .001; .010; .012;
.008&.402; .490; .395; .372; .415&.000; .000; .000; .000; -.000&.482; .488; .481; .466; .509\\
\hline
1.ReLU&2.Conv&.000; -.003; -.001; -.004; -.003&.499; .729; .583; .752; .789&-.006; -.003; -.019;
-.004; -.006&.569; .531; .700; .544; .563&.000; .001; .001; .000; .001&.470; .429; .412; .449;
.424\\
\hline
2.Conv&2.ReLU&.002; .005; .000; .004; -.001&.314; .138; .479; .251; .585&.054; .038; .064; .038;
.046&.065; .136; .043; .138; .095&-.000; .000; .000; -.000; .000&.516; .498; .504; .510; .496\\
\hline
2.ReLU&2.MaxPool&.010; -.006; -.012; .002; -.001&.030; .919; .994; .323; .608&.041; .029; .053;
.030; .031&.113; .198; .068; .195; .186&-.000; -.000; -.000; -.000; -.001&.517; .516; .536; .533;
.568\\
\hline
2.MaxPool&3.Conv&-.014; .008; \textbf{\textcolor{red}{.012}}; -.007; .004&.996; .042; \textbf{\textcolor{red}{.016}};
.947; .150&.067; .027; \textbf{\textcolor{red}{.073}}; .019; .040&.028; .219; \textbf{\textcolor{red}{.024}}; .290;
.122&.001; .001; .002; .001; .002&.332; .353; .284; .307; .281\\
\hline
3.Conv&3.ReLU&.005; -.002; .003; .006; .000&.157; .622; .301; .093; .495&.014; .021; .021; .022;
.022&.346; .277; .286; .266; .260&-.000; -.000; -.001; -.000; -.001&.570; .542; .582; .551; .581\\
\hline
3.ReLU&4.Conv&-.005; -.005; -.009; -.002; -.002&.888; .858; .970; .689; .723&.022; -.006; .019;
.002; -.002&.274; .570; .305; .476; .522&.001; .001; .001; .001; .001&.400; .409; .408; .374; .390\\
\hline
4.Conv&4.ReLU&.001; -.003; \textbf{\textcolor{red}{.008}}; .004; -.002&.405; .705; \textbf{\textcolor{red}{.023}};
.180; .640&-.057; -.045; -.026; -.120; -.053&.941; .906; .750; 1.00; .941&-.000; .000; .000; .000;
.000&.523; .480; .455; .476; .434\\
\hline
4.ReLU&4.MaxPool&.007; .007; .005; .002; \textbf{\textcolor{red}{.013}}&.077; .050; .138; .348;
\textbf{\textcolor{red}{.002}}&\textbf{\textcolor{red}{.089}}; \textbf{\textcolor{red}{.115}}; .061; \textbf{\textcolor{red}{.154}};
\textbf{\textcolor{red}{.119}}&\textbf{\textcolor{red}{.007}}; \textbf{\textcolor{red}{.000}}; .050; \textbf{\textcolor{red}{.000}};
\textbf{\textcolor{red}{.000}}&.003; .004; .004; .003; .003&.086; .042; .050; .077; .084\\
\hline
4.MaxPool&5.Dense&-.015; .003; -.009; -.010; -.008&.999; .237; .952; .973; .977&.049; .054; .035;
\textbf{\textcolor{red}{.084}}; .058&.076; .056; .162; \textbf{\textcolor{red}{.008}}; .046&.002; .001; .001; .001;
.002&.163; .225; .234; .217; .171\\
\hline
5.Dense&5.ReLU&\textbf{\textcolor{red}{.010}}; .000; -.002; .005; -.007&\textbf{\textcolor{red}{.018}}; .456; .693;
.134; .944&-.027; -.016; .000; -.054; -.014&.782; .677; .500; .949; .662&.001; .001; .001; .001;
.001&.288; .274; .168; .251; .147\\
\hline
5.ReLU&6.Dense&-.010; -.016; .006; -.011; .007&.987; 1.00; .071; .976; .089&\textbf{\textcolor{red}{.086}};
\textbf{\textcolor{red}{.084}}; .058; \textbf{\textcolor{red}{.118}}; \textbf{\textcolor{red}{.076}}&\textbf{\textcolor{red}{.006}};
\textbf{\textcolor{red}{.008}}; .043; \textbf{\textcolor{red}{.000}}; \textbf{\textcolor{red}{.012}}&\textbf{\textcolor{red}{.003}};
.002; .002; \textbf{\textcolor{red}{.002}}; .001&\textbf{\textcolor{red}{.008}}; .041; .081; \textbf{\textcolor{red}{.018}};
.103\\
\hline
6.Dense&6.Softmax&.001; \textbf{\textcolor{red}{.017}}; \textbf{\textcolor{red}{.010}}; .007; -.006&.407;
\textbf{\textcolor{red}{.000}}; \textbf{\textcolor{red}{.011}}; .096; .880&-.009; .000; -.004; -.026; -.016&.611;
.496; .557; .783; .692&-.003; -.003; -.004; -.005; -.005&.999; .999; 1.00; 1.00; 1.00\\
\hline
\end{tabular}
\end{table*}

In reviewing Table~\ref{table:InterLayerVChange} on the CIFAR10 dataset with random labels,
no layers are consistently significant in increasing the average interclass separation across the
five models, which is in contrast to the behavior of the training data in the dense layers (Table%
~\ref{table:InterLayerContribution}) .   This across the board discrepancy is a clear indicator of
model overfitting.  On the other hand, two layers of the models trained  on CIFAR10 with true labels
demonstrate statistically significant change in mean separation on the validation data, namely
the 4.MaxPooling and 6.Dense layers.  This indicates that the mappings learned for
these layers have provided a statistically significant improvement to the interpoint separation on
unseen data.

The final question and statistical test is whether or not the change in $\boldsymbol{\mathcal{H}}$ on each layer
is statistically equivalent for training and validation data (Q2, above). In other words, is the measured change
in the mean separation produced by the layer on the training data statistically equivalent to the
change induced on the validation data, as quantified by $\bar{\mathcal{H}}^{(t)}$ and
$\bar{\mathcal{H}}^{(v)}$?

Define $\boldsymbol{\Delta{\mathcal{H}}}$ to be the set of paired differences
between two class-paired separability statistics from different layers, that is, 
\begin{equation}\label{eq:pairdiff}
  \boldsymbol{\Delta{\mathcal{H}}}^{(D)}_{(k_1,k_2)} = \{ \mathcal{H}(y_i^{(k_1)}, y_j^{(k_1)})- \mathcal{H}(y_i^{(k_2)},y_j^{(k_2)})
  \hspace{1mm}\forall(i,j), \hspace{1mm} i < j \}
\end{equation}
where $D$ indicates the data set (training data, $t$, or the validation data, $v$) being used, and
$y_m^{(k)}$ is the set of feature vectors associated with class $m$ of data set $D$ produced by the model at layers $k_1$ and $k_2$.
Thus,  $\boldsymbol{\Delta{\mathcal{H}}}^{(t)}_{(k,k-1)}$ and $\boldsymbol{\Delta{\mathcal{H}}}^{(v)}_{(k,k-1)}$ are the distributions of 
changes produced by the $k$th layer
on the training and validation datasets, respectively.  

We attempt to quantify whether the means of the
differences or differences are statistically equivalent or not
\begin{eqnarray} 
\label{eq:deltaTvsV}
\mathbf{H_0}&: &\overline{\Delta{\mathcal{H}}}^{\:(t)}_{(k, k-1)} -  \overline{\Delta{\mathcal{H}}}^{\:(v)}_{(k,k-1)} = 0 \hspace{5mm} vs.\nonumber\\
\mathbf{H_A}&:&\overline{\Delta{\mathcal{H}}}^{\:(t)}_{(k, k-1)} -  \overline{\Delta{\mathcal{H}}}^{\:(v)}_{(k,k-1)} \neq 0.
\end{eqnarray}  

We test the hypothesis on the instances of our models via the random permutation test of means.  The results for
each model/dataset are given in Table~\ref{table:InterLayerTvsV}. The results of this hypothesis test provide an indication 
of which individual layers are
disproportionately improving class separation for the training data with respect to the improvement
to the validation data.  In all three cases, a portion of the back-end dense layers are
showing a statistically significant bias in training data separation improvement vs. the validation
data.  These results provide the practitioner actionable information with respect to targeting these
layers via regularization or other approaches to mitigate the training bias.

\begin{table*}[!t]
\centering
\caption{Two-sided permutation test~\eqref{eq:deltaTvsV} comparing the differences in the mean change induced on the
         training and validation statistics ( $\boldsymbol{\Delta{\mathcal{H}}}^{(t)}_{(k,k-1)}$ and $\boldsymbol{\Delta{\mathcal{H}}}^{(v)}_{(k,k-1)}$  ). \textcolor{red}{Red font}
         denotes layer instances for which we reject $\mathbf{H_0}$, and black font denotes layers for
         which we fail to reject $\mathbf{H_0}$. (Note: $\Delta\mu = \overline{\Delta{\mathcal{H}}}^{\:(t)}_{(k, k-1)} -  \overline{\Delta{\mathcal{H}}}^{\:(v)}_{(k,k-1)} $)}
\label{table:InterLayerTvsV}
\begin{tabular}{>{\centering\arraybackslash}m{12mm}|>{\centering\arraybackslash}m{12mm}
|>{\centering\arraybackslash}m{22mm}|>{\centering\arraybackslash}m{19mm}
|>{\centering\arraybackslash}m{22mm}|>{\centering\arraybackslash}m{19mm}
|>{\centering\arraybackslash}m{22mm}|>{\centering\arraybackslash}m{19mm}}
\hline
\bfseries Input Space &\bfseries Output Space& \multicolumn{2}{c|}{\bfseries CIFAR10 w Random}
& \multicolumn{2}{c|}{\bfseries CIFAR10 w True}& \multicolumn{2}{c}{\bfseries MNIST w True} \\
  & &$\Delta\mu$ & \textit{p}-values &
  $\Delta\mu$ & \textit{p}-values&
  $\Delta\mu$ & \textit{p}-values \\
\hline \hline
0.Input&1.Conv&.005; -.006; -.001; -.003; .001&.156; .052; .769; .332; .811&-.004; -.006; -.004;
-.004; -.005&.682; .225; .691; .457; .547&-.001; -.001; -.000; -.001; -.001&.212; .307; .852; .497;
.242\\
\hline
1.Conv&1.ReLU&-.000; .002; .002; -.002; .000&.966; .204; .406; .376; .891&.001; .001; .001; .001;
.000&.549; .569; .780; .598; .963&-.000; .000; .000; -.000; .000&.308; 1.00; 1.00; .459; .347\\
\hline
1.ReLU&2.Conv&.006; \textbf{\textcolor{red}{-.009}}; -.007; -.003; -.000&.250; \textbf{\textcolor{red}{.020}}; .192;
.467; .980&.004; .007; .006; .008; -.003&.390; .172; .184; .146; .541&-.000; .000; -.001; -.000;
.000&.735; .825; .079; .923; .936\\
\hline
2.Conv&2.ReLU&-.003; -.001; .002; \textbf{\textcolor{red}{-.007}}; -.002&.362; .809; .509;
\textbf{\textcolor{red}{.011}}; .592&-.003; .001; -.001; -.003; .007&.719; .892; .861; .575;
.248&\textbf{\textcolor{red}{.000}}; .000; .000; .000; .000&\textbf{\textcolor{red}{.007}}; .832; .350; .672; .858\\
\hline
2.ReLU&2.MaxPool&\textbf{\textcolor{red}{-.017}}; \textbf{\textcolor{red}{.016}}; .008; -.004;
.002&\textbf{\textcolor{red}{.000}}; \textbf{\textcolor{red}{.000}}; .088; .258; .701&.010; .011; .005; .004;
.007&.130; .031; .434; .350; .154&-.001; -.001; -.001; -.001; -.001&.217; .188; .219; .227; .473\\
\hline
2.MaxPool&3.Conv&\textbf{\textcolor{red}{.018}}; -.012; \textbf{\textcolor{red}{-.013}}; \textbf{\textcolor{red}{.011}};
.000&\textbf{\textcolor{red}{.000}}; .027; \textbf{\textcolor{red}{.009}}; \textbf{\textcolor{red}{.014}}; .929&-.002; -.010;
.004; -.004; -.004&.848; .049; .791; .321; .513&.001; .001; .001; .001; -.000&.421; .258; .332;
.323; .974\\
\hline
3.Conv&3.ReLU&-.006; .002; -.005; \textbf{\textcolor{red}{-.010}}; -.005&.095; .616; .219;
\textbf{\textcolor{red}{.011}}; .251&.000; .000; .002; .002; .007&.886; .970; .470; .614; .108&.000; -.000;
-.000; .000; .000&.630; .335; 1.00; 1.00; .150\\
\hline
3.ReLU&4.Conv&-.001; \textbf{\textcolor{red}{.014}}; \textbf{\textcolor{red}{.015}}; -.003;
\textbf{\textcolor{red}{.013}}&.805; \textbf{\textcolor{red}{.005}}; \textbf{\textcolor{red}{.005}}; .598;
\textbf{\textcolor{red}{.007}}&.001; .001; .000; .006; -.002&.861; .772; .951; .281; .623&-.000; .000; -.000;
-.000; .000&.672; .621; .948; .483; .717\\
\hline
4.Conv&4.ReLU&.005; .003; .001; .003; -.008&.406; .589; .783; .634; .177&.015; .009; .005; -.006;
.016&.088; .267; .403; .695; .048&.000; .000; .000; .001; .000&1.00; .985; .969; .275; .890\\
\hline
4.ReLU&4.MaxPool&-.006; \textbf{\textcolor{red}{-.012}}; \textbf{\textcolor{red}{-.020}}; .007;
\textbf{\textcolor{red}{-.017}}&.300; \textbf{\textcolor{red}{.012}}; \textbf{\textcolor{red}{.000}}; .247;
\textbf{\textcolor{red}{.000}}&-.008; .004; -.004; .011; -.006&.332; .710; .516; .453; .581&.001; .000; .000;
.000; .000&.348; .975; .789; .892; .731\\
\hline
4.MaxPool&5.Dense&\textbf{\textcolor{red}{.064}}; \textbf{\textcolor{red}{.028}}; \textbf{\textcolor{red}{.062}};
\textbf{\textcolor{red}{.039}}; \textbf{\textcolor{red}{.049}}&\textbf{\textcolor{red}{.000}}; \textbf{\textcolor{red}{.000}};
\textbf{\textcolor{red}{.000}}; \textbf{\textcolor{red}{.000}}; \textbf{\textcolor{red}{.000}}&\textbf{\textcolor{red}{.012}};
\textbf{\textcolor{red}{.013}}; \textbf{\textcolor{red}{.016}}; .020; \textbf{\textcolor{red}{.016}}&\textbf{\textcolor{red}{.013}};
\textbf{\textcolor{red}{.020}}; \textbf{\textcolor{red}{.001}}; .030; \textbf{\textcolor{red}{.006}}&.001; .001; .000; .001;
.000&.431; .210; .387; .392; .938\\
\hline
5.Dense&5.ReLU&\textbf{\textcolor{red}{.179}}; \textbf{\textcolor{red}{.213}}; \textbf{\textcolor{red}{.204}};
\textbf{\textcolor{red}{.198}}; \textbf{\textcolor{red}{.218}}&\textbf{\textcolor{red}{.000}}; \textbf{\textcolor{red}{.000}};
\textbf{\textcolor{red}{.000}}; \textbf{\textcolor{red}{.000}}; \textbf{\textcolor{red}{.000}}&\textbf{\textcolor{red}{.040}};
\textbf{\textcolor{red}{.026}}; \textbf{\textcolor{red}{.015}}; \textbf{\textcolor{red}{.063}};
\textbf{\textcolor{red}{.029}}&\textbf{\textcolor{red}{.000}}; \textbf{\textcolor{red}{.000}}; \textbf{\textcolor{red}{.000}};
\textbf{\textcolor{red}{.000}}; \textbf{\textcolor{red}{.000}}&.001; .000; .000; -.000; .001&.217; .472; .511; .856;
.239\\
\hline
5.ReLU&6.Dense&\textbf{\textcolor{red}{.768}}; \textbf{\textcolor{red}{.768}}; \textbf{\textcolor{red}{.744}};
\textbf{\textcolor{red}{.767}}; \textbf{\textcolor{red}{.746}}&\textbf{\textcolor{red}{.000}}; \textbf{\textcolor{red}{.000}};
\textbf{\textcolor{red}{.000}}; \textbf{\textcolor{red}{.000}}; \textbf{\textcolor{red}{.000}}&\textbf{\textcolor{red}{.134}};
\textbf{\textcolor{red}{.115}}; \textbf{\textcolor{red}{.090}}; \textbf{\textcolor{red}{.174}};
\textbf{\textcolor{red}{.124}}&\textbf{\textcolor{red}{.000}}; \textbf{\textcolor{red}{.000}}; \textbf{\textcolor{red}{.000}};
\textbf{\textcolor{red}{.000}}; \textbf{\textcolor{red}{.000}}&\textbf{\textcolor{red}{.002}}; \textbf{\textcolor{red}{.002}};
\textbf{\textcolor{red}{.003}}; \textbf{\textcolor{red}{.003}}; \textbf{\textcolor{red}{.003}}&\textbf{\textcolor{red}{.014}};
\textbf{\textcolor{red}{.015}}; \textbf{\textcolor{red}{.001}}; \textbf{\textcolor{red}{.003}}; \textbf{\textcolor{red}{.000}}\\
\hline
6.Dense&6.Softmax&-.001; \textbf{\textcolor{red}{-.017}}; \textbf{\textcolor{red}{-.010}}; -.007; .006&.750;
\textbf{\textcolor{red}{.000}}; \textbf{\textcolor{red}{.025}}; .223; .266&\textbf{\textcolor{red}{.028}};
\textbf{\textcolor{red}{.019}}; \textbf{\textcolor{red}{.020}}; \textbf{\textcolor{red}{.036}};
\textbf{\textcolor{red}{.036}}&\textbf{\textcolor{red}{.000}}; \textbf{\textcolor{red}{.004}}; \textbf{\textcolor{red}{.000}};
\textbf{\textcolor{red}{.000}}; \textbf{\textcolor{red}{.000}}&\textbf{\textcolor{red}{.004}}; \textbf{\textcolor{red}{.005}};
\textbf{\textcolor{red}{.005}}; \textbf{\textcolor{red}{.006}}; \textbf{\textcolor{red}{.006}}&\textbf{\textcolor{red}{.000}};
\textbf{\textcolor{red}{.000}}; \textbf{\textcolor{red}{.000}}; \textbf{\textcolor{red}{.000}}; \textbf{\textcolor{red}{.000}}\\

\hline
\end{tabular}
\end{table*}

\section{Conclusions}

Advancements in hierarchical approaches can be accelerated given the statistical approach discussed.  
Using the Henze-Penrose-Berisha-Hero statistic, we have demonstrated the statistical characterization of the functional mappings of deep learning
models trained on disparate datasets.  These characterizations include the identification of layers that (a) are memorizing, (b) acting as distance-preserving random projections, or (c) inducing a domain mismatch between training and validation sets.  Actionable insights include optimization of network topology, identification of layers which require additional regularization,  and testing of data domain consistency or mismatch.

It has been demonstrated that estimating the distributional characteristics of the data with respect to a vector space can provide insight into the difficulty of the task as well as the mechanisms and
utility of each functional mapping applied.    The comparative analysis between datasets illustrates that the
required complexity of the model depends on the measurements and the class relationships therein
and is not intrinsic to the architecture.

\bibliographystyle{ieeetr}

\begin{thebibliography}{10}

\bibitem{ziller13}
M.~D. Zeiler and R.~Fergus, ``Visualizing and understanding convolutional
  networks,'' {\em CoRR}, vol.~abs/1311.2901, 2013.

\bibitem{berishahero}
V.~Berisha and A.~O. Hero, ``Empirical non-parametric estimation of the fisher
  information,'' {\em IEEE Signal Processing Letters}, vol.~22, pp.~988--992,
  July 2015.

\bibitem{zhang17}
C.~Zhang, S.~Bengio, M.~Hardt, B.~Recht, and O.~Vinyals, ``Understanding deep
  learning requires rethinking generalization,'' {\em CoRR},
  vol.~abs/1611.03530, 2016.

\bibitem{krizhevshy12}
A.~Krizhevsky, I.~Sutskever, and G.~E. Hinton, ``Imagenet classification with
  deep convolutional neural networks,'' in {\em Proceedings of the 25th
  International Conference on Neural Information Processing Systems - Volume
  1}, NIPS'12, (USA), pp.~1097--1105, Curran Associates Inc., 2012.

\bibitem{bartlett17}
P.~L. Bartlett, D.~J. Foster, and M.~Telgarsky, ``Spectrally-normalized margin
  bounds for neural networks,'' {\em CoRR}, vol.~abs/1706.08498, 2017.

\bibitem{hock97}
S.~Hochreiter and J.~Schmidhuber, ``Long short-term memory,'' {\em Neural
  Computation}, vol.~9, no.~8, pp.~1735--1780, 1997.

\bibitem{cha17}
P.~Chaudhari, A.~Choromanska, S.~Soatto, Y.~LeCun, C.~Baldassi, C.~Borgs, J.~T.
  Chayes, L.~Sagun, and R.~Zecchina, ``Entropy-sgd: Biasing gradient descent
  into wide valleys,'' {\em CoRR}, vol.~abs/1611.01838, 2016.

\bibitem{roy17}
G.~{Karolina Dziugaite} and D.~M. {Roy}, ``{Computing Nonvacuous Generalization
  Bounds for Deep (Stochastic) Neural Networks with Many More Parameters than
  Training Data},'' {\em ArXiv e-prints}, Mar. 2017.

\bibitem{8010591}
M.~Dotter, K.~Rainey, and D.~Waagen, ``Visualization of high dimensional image
  features for classification,'' in {\em 2016 IEEE Applied Imagery Pattern
  Recognition Workshop (AIPR)}, pp.~1--6, Oct 2016.

\bibitem{wald40}
A.~Wald and J.~Wolfowitz, ``On a test whether two samples are from the same
  population,'' {\em Ann. Math. Statist.}, vol.~11, pp.~147--162, 06 1940.

\bibitem{friedman79}
J.~H. Friedman and L.~C. Rafsky, ``Multivariate generalizations of the
  wald-wolfowitz and smirnov two-sample tests,'' {\em Ann. Statist.}, vol.~7,
  pp.~697--717, 07 1979.

\bibitem{basu02}
T.~K. Ho and M.~Basu, ``Complexity measures of supervised classification
  problems,'' {\em IEEE Transactions on Pattern Analysis and Machine
  Intelligence}, vol.~24, pp.~289--300, March 2002.

\bibitem{csiszar2004information}
I.~Csisz{\'a}r, P.~C. Shields, {\em et~al.}, ``Information theory and
  statistics: A tutorial,'' {\em Foundations and Trends{\textregistered} in
  Communications and Information Theory}, vol.~1, no.~4, pp.~417--528, 2004.

\bibitem{henze99}
N.~Henze and M.~D. Penrose, ``On the multivariate runs test,'' {\em Ann.
  Statist.}, vol.~27, pp.~290--298, 03 1999.

\bibitem{berisha16}
V.~Berisha, A.~Wisler, A.~O. Hero, and A.~Spanias, ``Empirically estimable
  classification bounds based on a nonparametric divergence measure,'' {\em
  IEEE Transactions on Signal Processing}, vol.~64, pp.~580--591, Feb 2016.

\bibitem{mnist}
Y.~Lecun, L.~Bottou, Y.~Bengio, and P.~Haffner, ``Gradient-based learning
  applied to document recognition,'' {\em Proceedings of the IEEE}, vol.~86,
  pp.~2278--2324, Nov 1998.

\bibitem{cifar10}
A.~Krizhevsky, V.~Nair, and G.~Hinton, ``Cifar-10,''
\newblock http://www.cs.toronto.edu/{\textasciitilde{}}kriz/cifar.html.

\bibitem{tf2015}
M.~Abadi, A.~Agarwal, P.~Barham, E.~Brevdo, Z.~Chen, C.~Citro, G.~S. Corrado,
  A.~Davis, J.~Dean, M.~Devin, S.~Ghemawat, I.~Goodfellow, A.~Harp, G.~Irving,
  M.~Isard, Y.~Jia, R.~Jozefowicz, L.~Kaiser, M.~Kudlur, J.~Levenberg,
  D.~Man\'{e}, R.~Monga, S.~Moore, D.~Murray, C.~Olah, M.~Schuster, J.~Shlens,
  B.~Steiner, I.~Sutskever, K.~Talwar, P.~Tucker, V.~Vanhoucke, V.~Vasudevan,
  F.~Vi\'{e}gas, O.~Vinyals, P.~Warden, M.~Wattenberg, M.~Wicke, Y.~Yu, and
  X.~Zheng, ``{TensorFlow}: Large-scale machine learning on heterogeneous
  systems,'' 2015.
\newblock Software available from tensorflow.org, https://www.tensorflow.org/.

\bibitem{chollet2015}
F.~Chollet {\em et~al.}, ``Keras,'' 2015.
\newblock https://github.com/fchollet/keras.

\bibitem{efron1994introduction}
B.~Efron and R.~J. Tibshirani, {\em An introduction to the bootstrap}.
\newblock CRC press, 1994.

\bibitem{johnson84}
W.~B. Johnson, J.~Lindenstrauss, and G.~Schechtman, ``Extensions of lipschitz
  maps into banach spaces,'' {\em Israel Journal of Mathematics}, vol.~54,
  pp.~129--138, Jun 1986.

\bibitem{huang06}
G.-B. Huang, Q.-Y. Zhu, and C.-K. Siew, ``Extreme learning machine: Theory and
  applications,'' {\em Neurocomputing}, vol.~70, no.~1, pp.~489 -- 501, 2006.
\newblock Neural Networks.

\bibitem{ben-david10}
S.~Ben-David, J.~Blitzer, K.~Crammer, A.~Kulesza, F.~Pereira, and J.~W.
  Vaughan, ``A theory of learning from different domains,'' {\em Machine
  Learning}, vol.~79, pp.~151--175, May 2010.

\end{thebibliography}

\end{document}


\maketitle

\section{Statistical analysis using Euclidean distance as measure of proximity}\label{app:euclidean}

This supplemental material  presents more detailed results from the three experiments (CIFAR10 with random labels,
CIFAR10 with true labels, and MNIST with true labels) using the Euclidean distance metric for the
proximity measure of the HP statistics.

\begin{figure}[p!]
\centering
\includegraphics[width=1.0\linewidth]{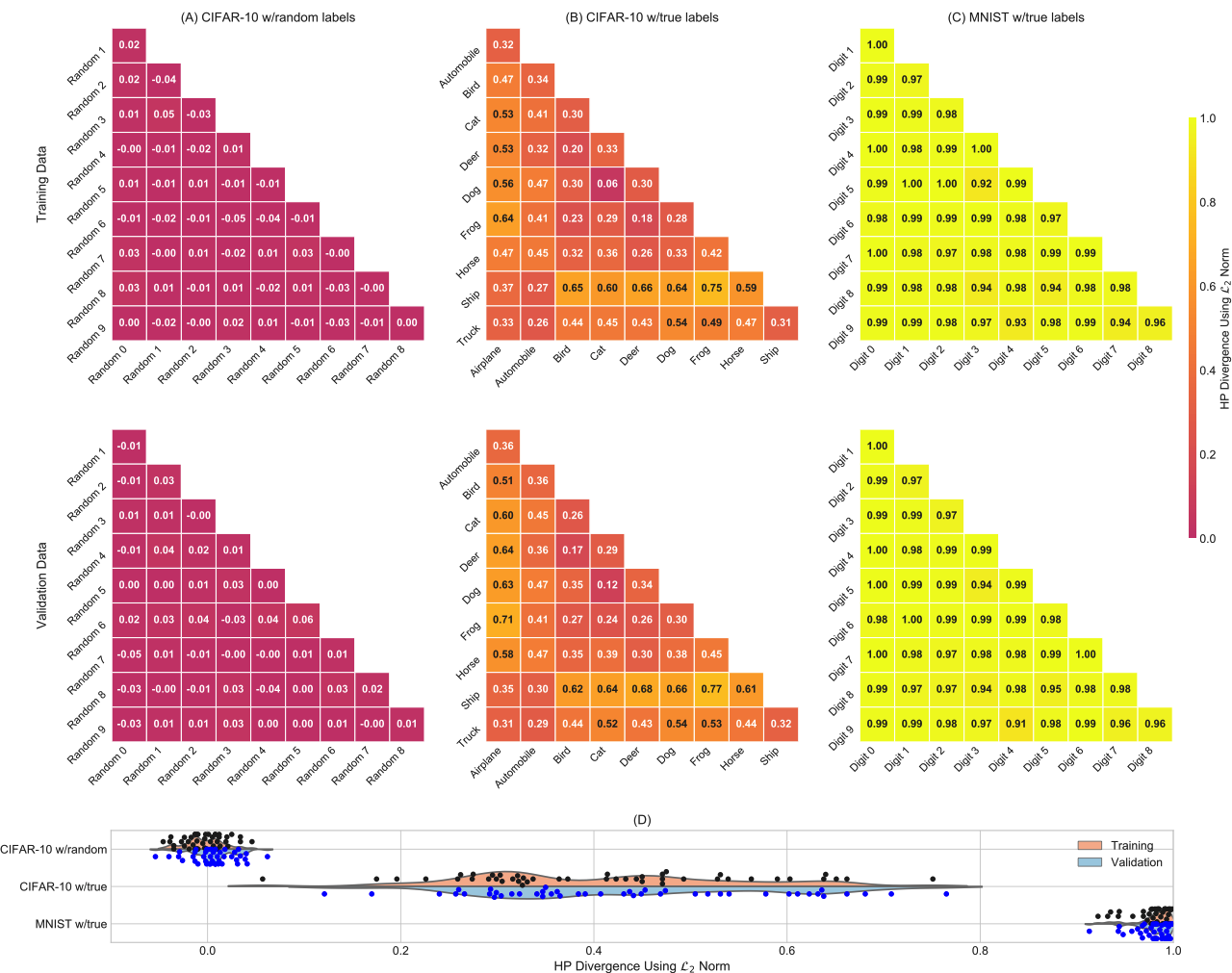}
\caption{Pairwise class HP statistics using Euclidean distance (training data above, validation data
below) computed on (A) CIFAR10 with random labels, (B) CIFAR10 with true labels, and (C) MNIST with
true labels. (D) The $\boldsymbol{\mathcal{H}}^{(t)}$ (black dots above) and $\boldsymbol{\mathcal{H}}^{(v)}$(blue
dots, below) values and respective kernel-based density functions (orange = training, blue =
validation) for each task which illustrate that the estimated class separation for each task in
their respective ambient representations are quite distinct.}
\label{figure:A1}
\end{figure}

Figure~\ref{figure:A1} shows the between-class HP statistics of the raw images, or class
separability in the original measurement space. The comparisons were made using the analysis subset
of the training data~(1000 images per class) and the validation data~(1000 images per class). As    
described in Section~3 of the paper, for the case using CIFAR10 with random labels there are
five different versions of the randomly permuted labels; one per instance of the training network.
The results for only one of these versions is tabulated and plotted in Figures~\ref{figure:A1} (A)
and (D), respectively.

\begin{figure}[p!]
\centering
\begin{subfigure}[(A)]{0.49\linewidth}
  \includegraphics[width=\linewidth]{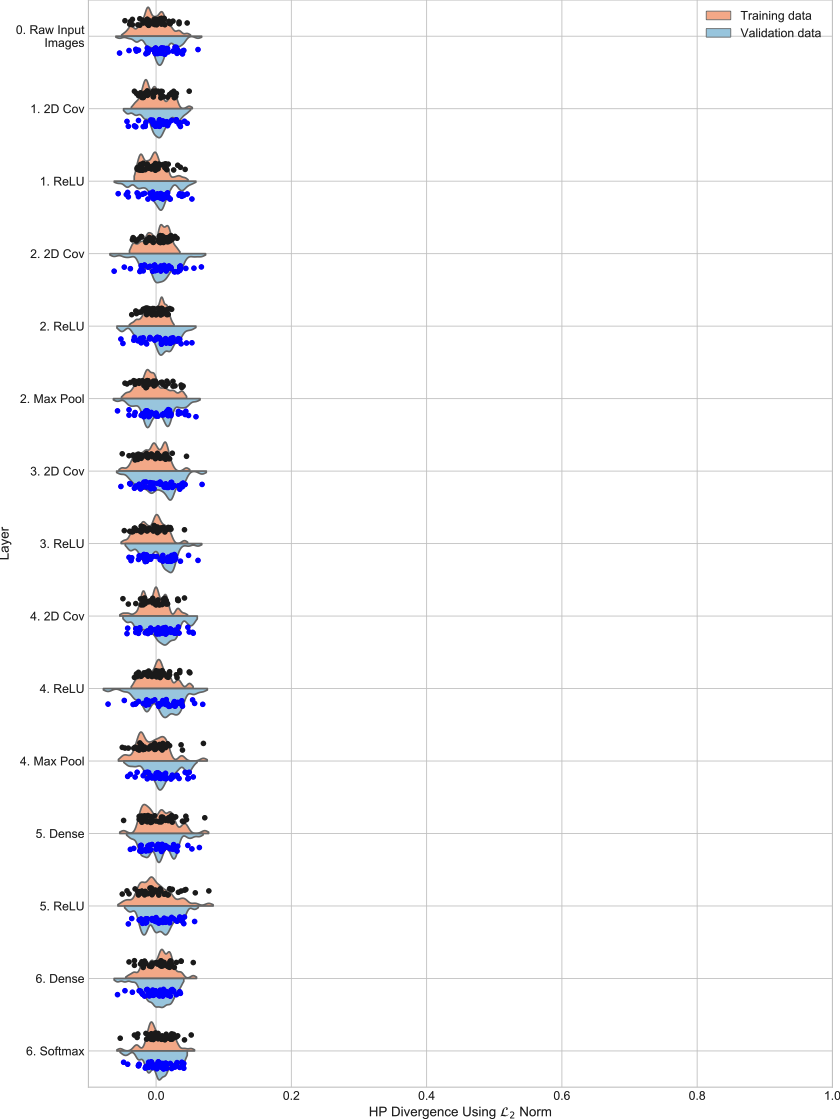}
  \caption{Untrained model}
\end{subfigure}
\begin{subfigure}[(B)]{0.49\linewidth}
  \includegraphics[width=\linewidth]{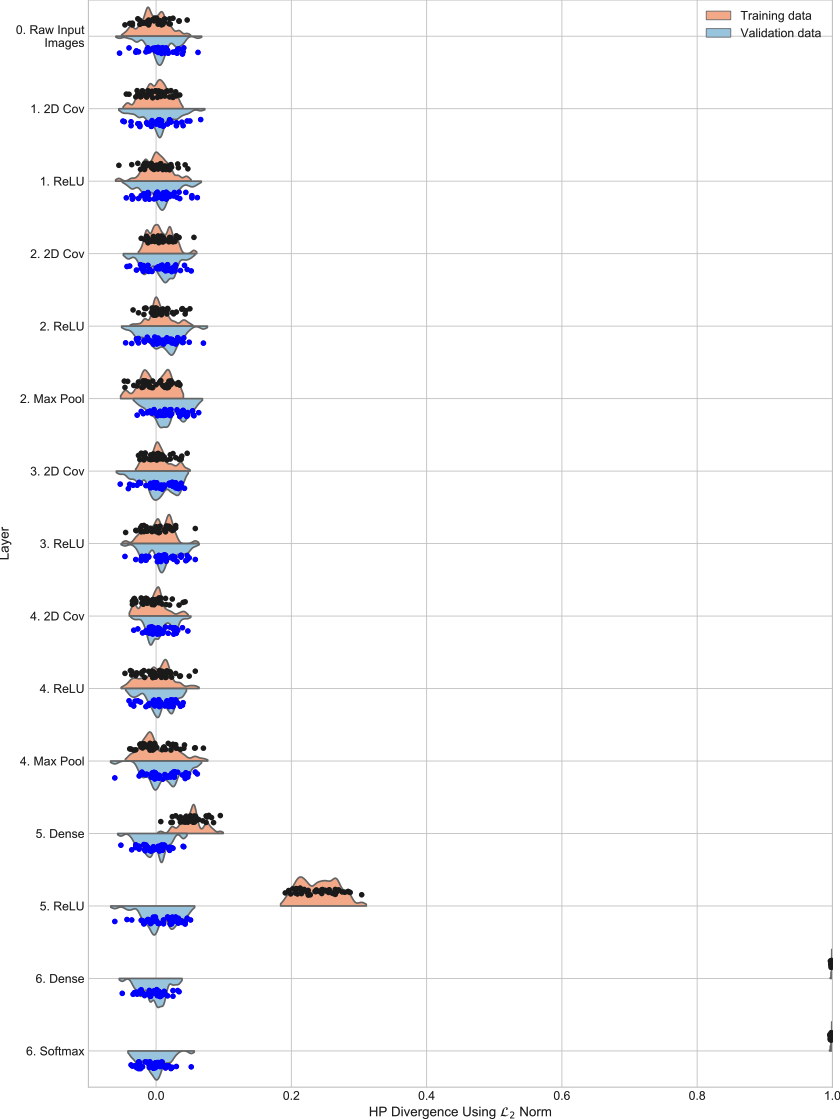}
  \caption{Trained model}
\end{subfigure}
\caption{$\mathcal{H}$ class-pair statistics at each layer for instance 1 of the model for CIFAR10 with
         random class labels. (a) shows results for the data for passing through the randomly
         initialized model (epoch~0 state). (b) shows the results for the data  passing through the
         fully trained  model (epoch~200 state). (Note:~Euclidean distance is used as the proximity
         measure.)}
\label{figure:A2}
\end{figure}

\begin{figure}[p!]
\centering
\begin{subfigure}[(A)]{0.49\linewidth}
  \includegraphics[width=\linewidth]{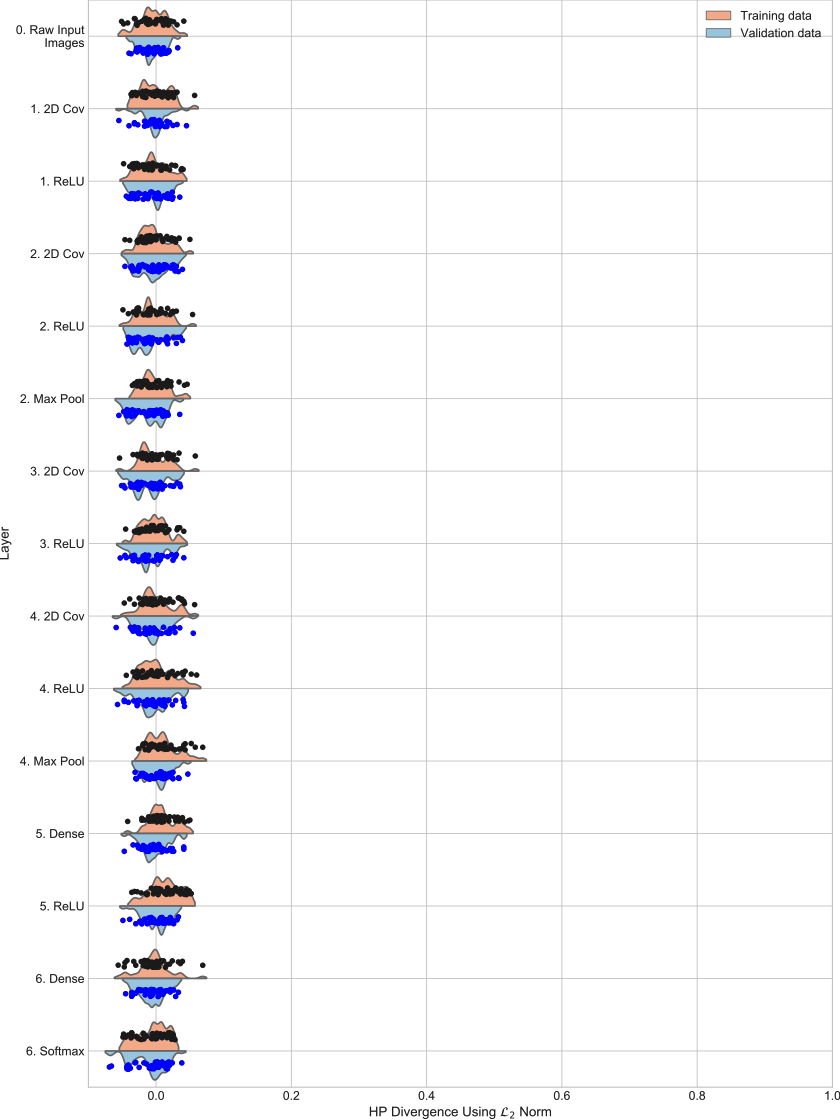}
  \caption{Untrained model}
\end{subfigure}
\begin{subfigure}[(B)]{0.49\linewidth}
  \includegraphics[width=\linewidth]{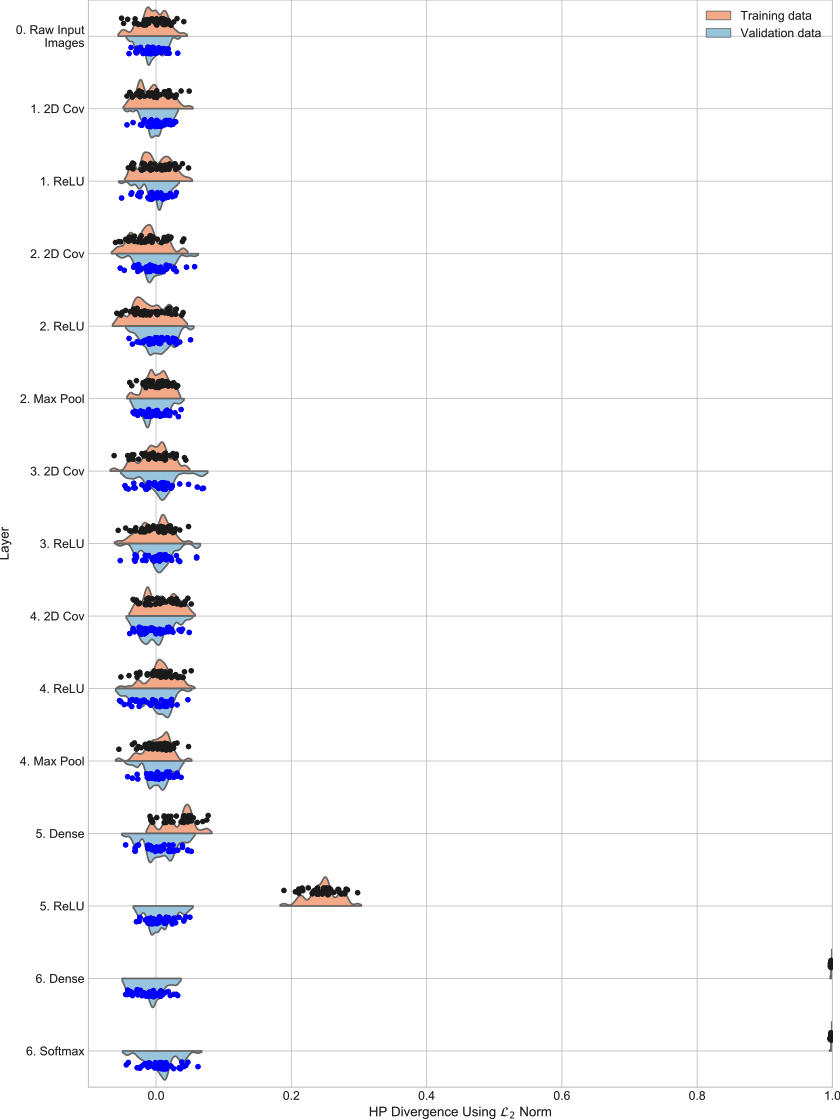}
  \caption{Trained model}
\end{subfigure}
\caption{$\mathcal{H}$ class-pair statistics at each layer for instance 2 of the model for CIFAR10 with
         random class labels. (a) shows results for the data for passing through the randomly
         initialized model  (epoch~0 state). (b) shows the results for the data  passing through the
         fully trained  model (epoch~200 state). (Note:~Euclidean distance is used as the proximity
         measure.)}
\label{figure:A3}
\end{figure}

\begin{figure}[p!]
\centering
\begin{subfigure}[(A)]{0.49\linewidth}
  \includegraphics[width=\linewidth]{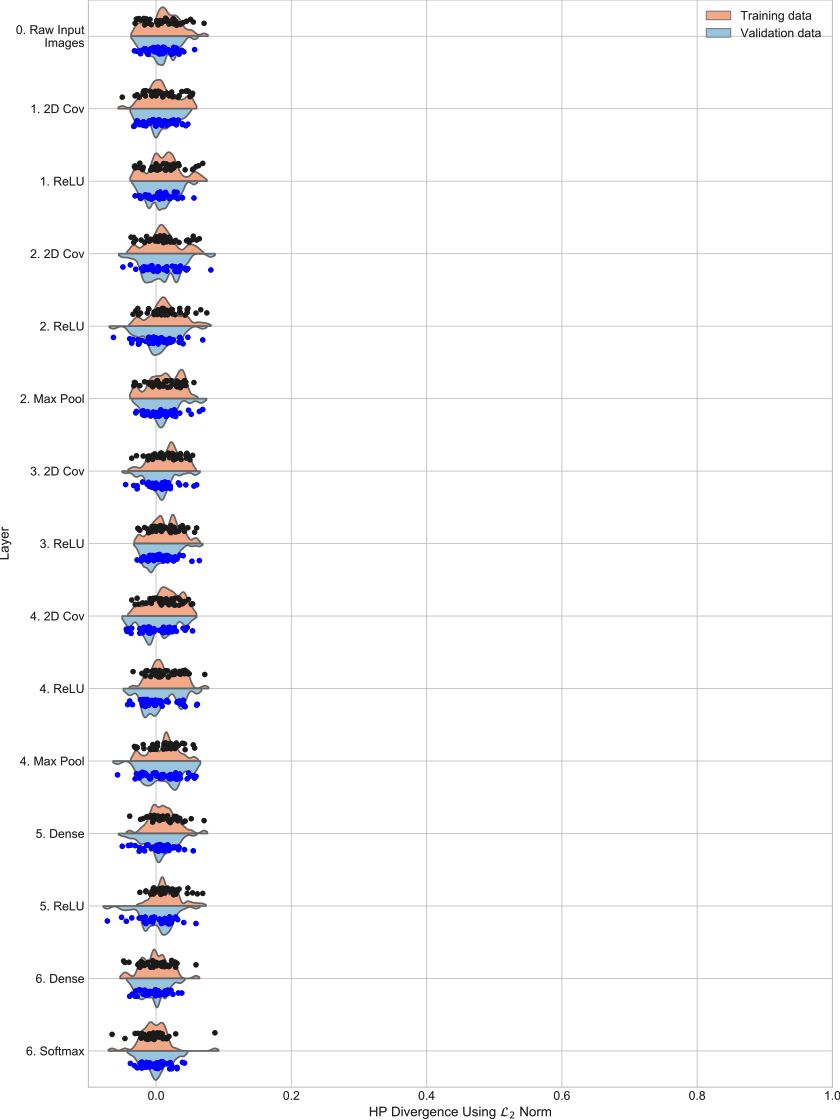}
  \caption{Untrained model}
\end{subfigure}
\begin{subfigure}[(B)]{0.49\linewidth}
  \includegraphics[width=\linewidth]{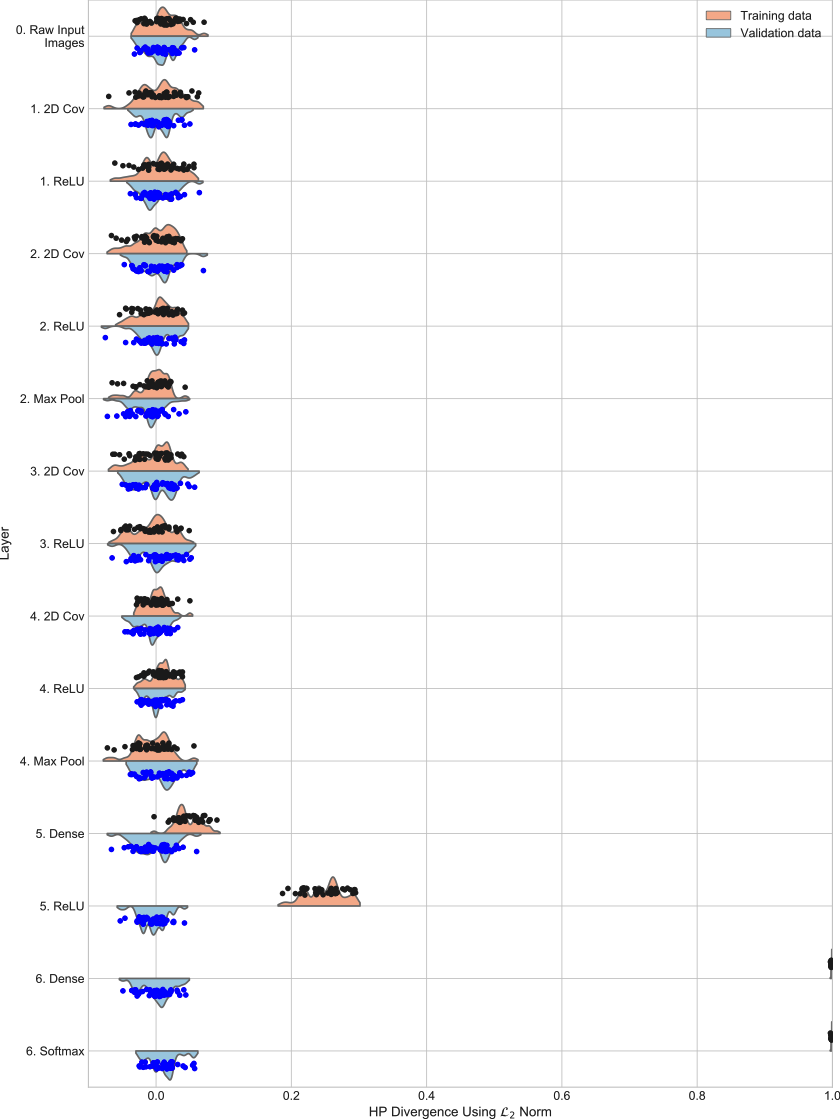}
  \caption{Trained model}
\end{subfigure}
\caption{$\mathcal{H}$ class-pair statistics at each layer for instance 3 of the model for CIFAR10 with
         random class labels. (a) shows results for the data for passing through the randomly
         initialized model  (epoch~0 state). (b) shows the results for the data  passing through the
         fully trained  model (epoch~200 state). (Note:~Euclidean distance is used as the proximity
         measure.)}
\label{figure:A4}
\end{figure}

\begin{figure}[p!]
\centering
\begin{subfigure}[(A)]{0.49\linewidth}
  \includegraphics[width=\linewidth]{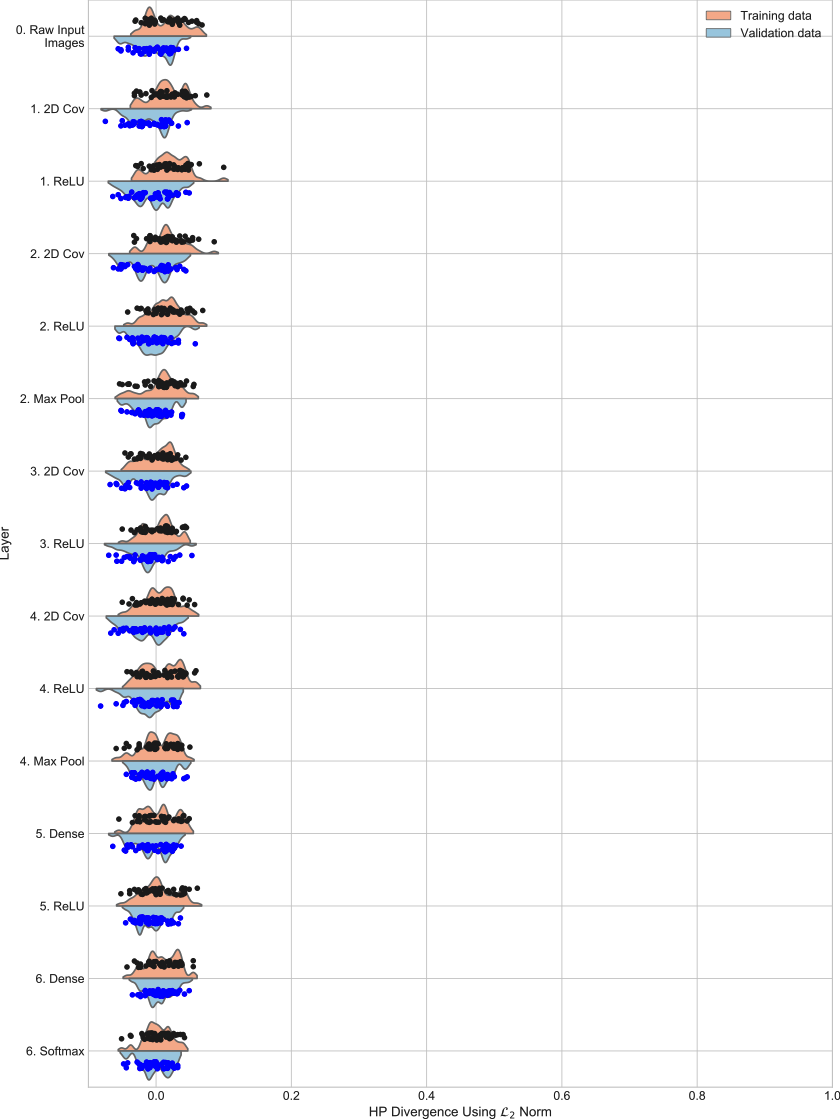}
  \caption{Untrained model}
\end{subfigure}
\begin{subfigure}[(B)]{0.49\linewidth}
  \includegraphics[width=\linewidth]{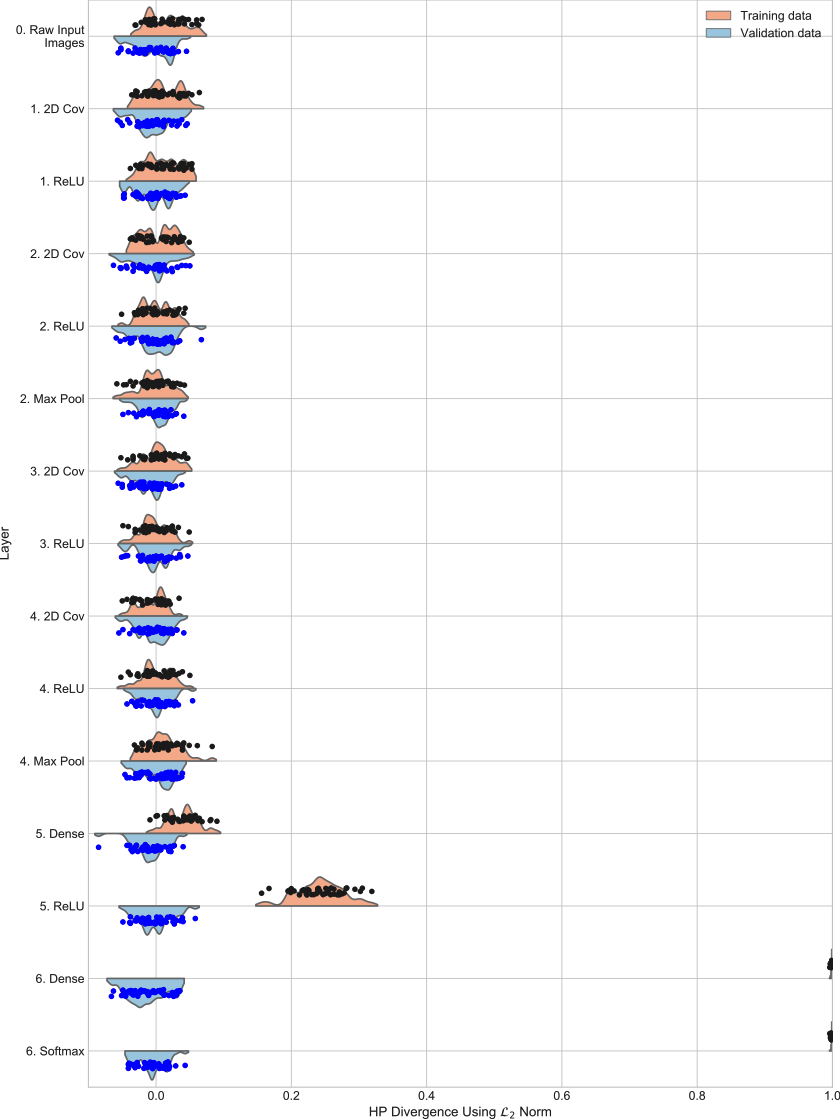}
  \caption{Trained model}
\end{subfigure}
\caption{$\mathcal{H}$ class-pair statistics at each layer for instance 4 of the model for CIFAR10 with
         random class labels. (a) shows results for the data for passing through the randomly
         initialized model  (epoch~0 state). (b) shows the results for the data  passing through the
         fully trained  model (epoch~200 state). (Note:~Euclidean distance is used as the proximity
         measure.)}
\label{figure:A5}
\end{figure}

\begin{figure}[p!]
\centering
\begin{subfigure}[(A)]{0.49\linewidth}
  \includegraphics[width=\linewidth]{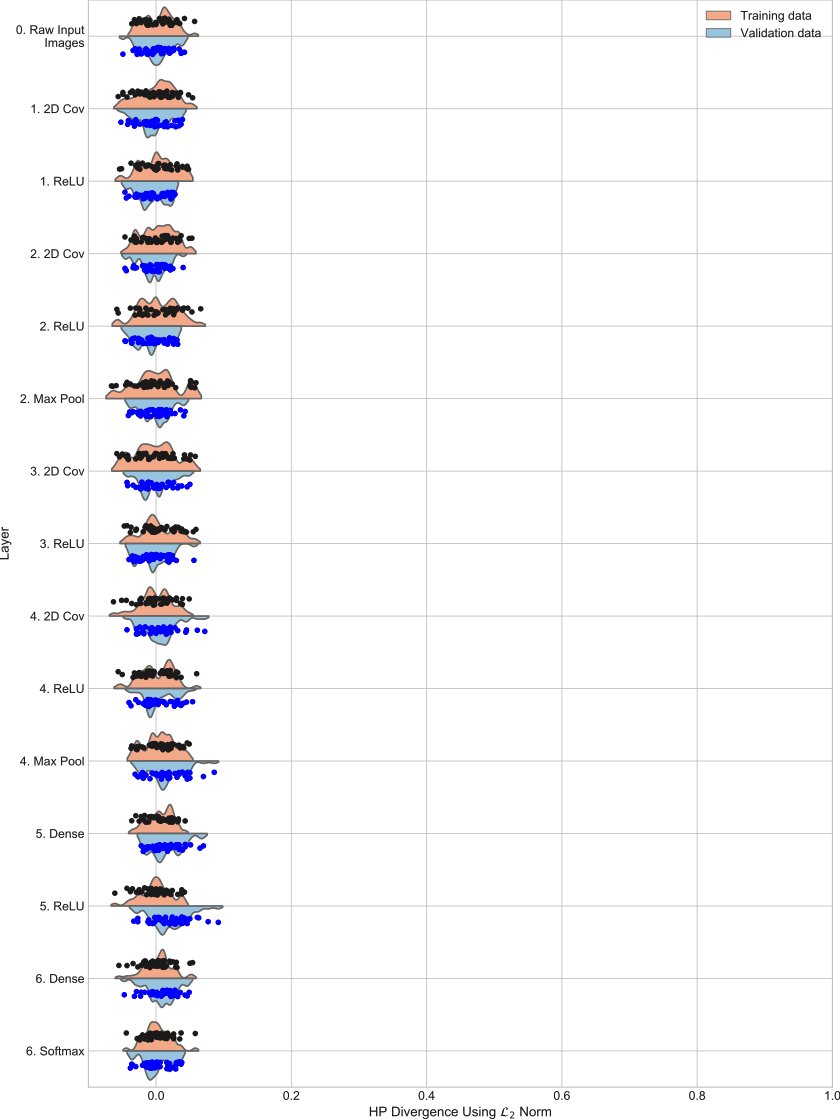}
  \caption{Untrained model}
\end{subfigure}
\begin{subfigure}[(B)]{0.49\linewidth}
  \includegraphics[width=\linewidth]{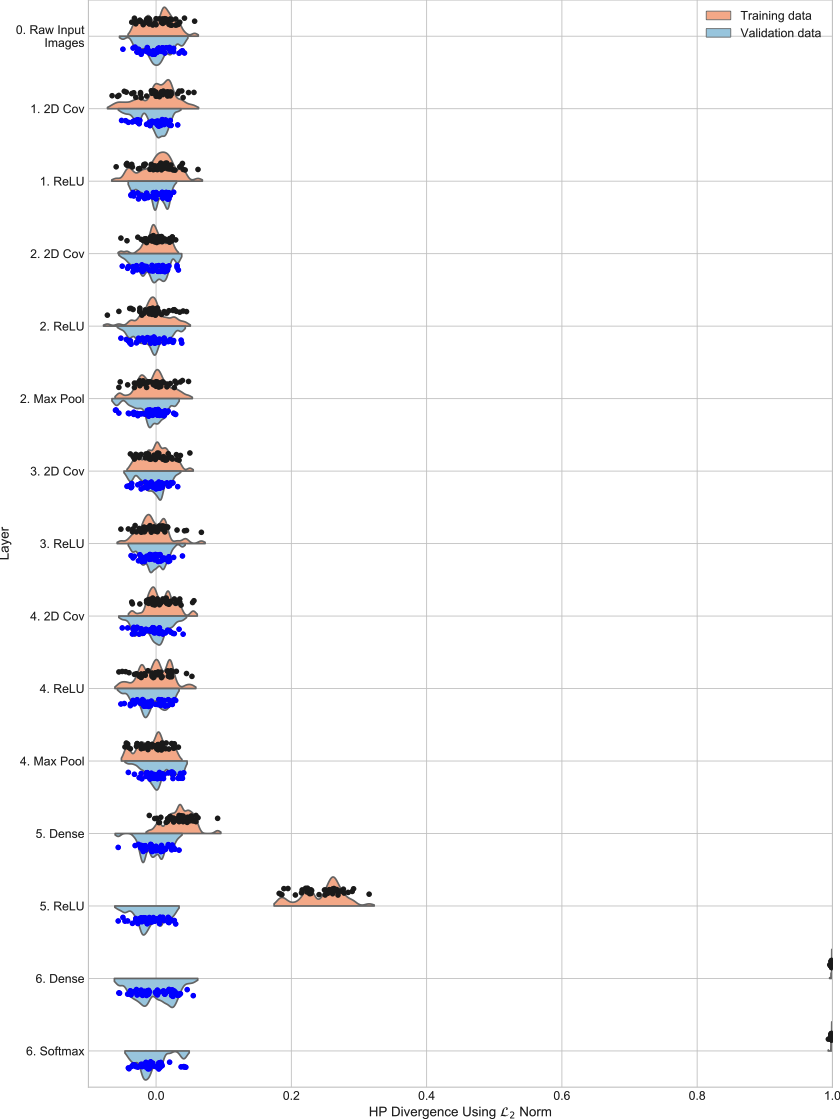}
  \caption{Trained model}
\end{subfigure}
\caption{$\mathcal{H}$ class-pair statistics at each layer for instance 5 of the model for CIFAR10 with
         random class labels. (a) shows results for the data for passing through the randomly
         initialized model  (epoch~0 state). (b) shows the results for the data  passing through the
         fully trained  model (epoch~200 state). (Note:~Euclidean distance is used as the proximity
         measure.)}
\label{figure:A6}
\end{figure}

\begin{figure}[p!]
\centering
\begin{subfigure}[(A)]{0.49\linewidth}
  \includegraphics[width=\linewidth]{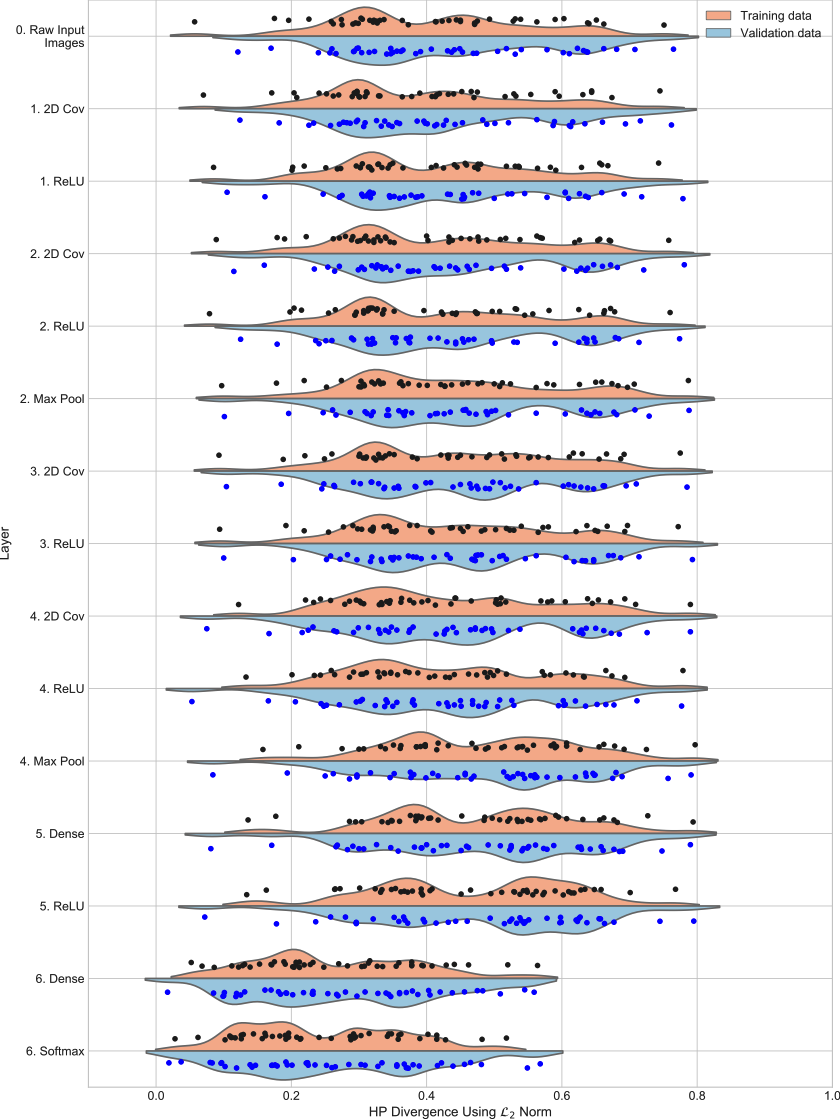}
  \caption{Untrained model}
\end{subfigure}
\begin{subfigure}[(B)]{0.49\linewidth}
  \includegraphics[width=\linewidth]{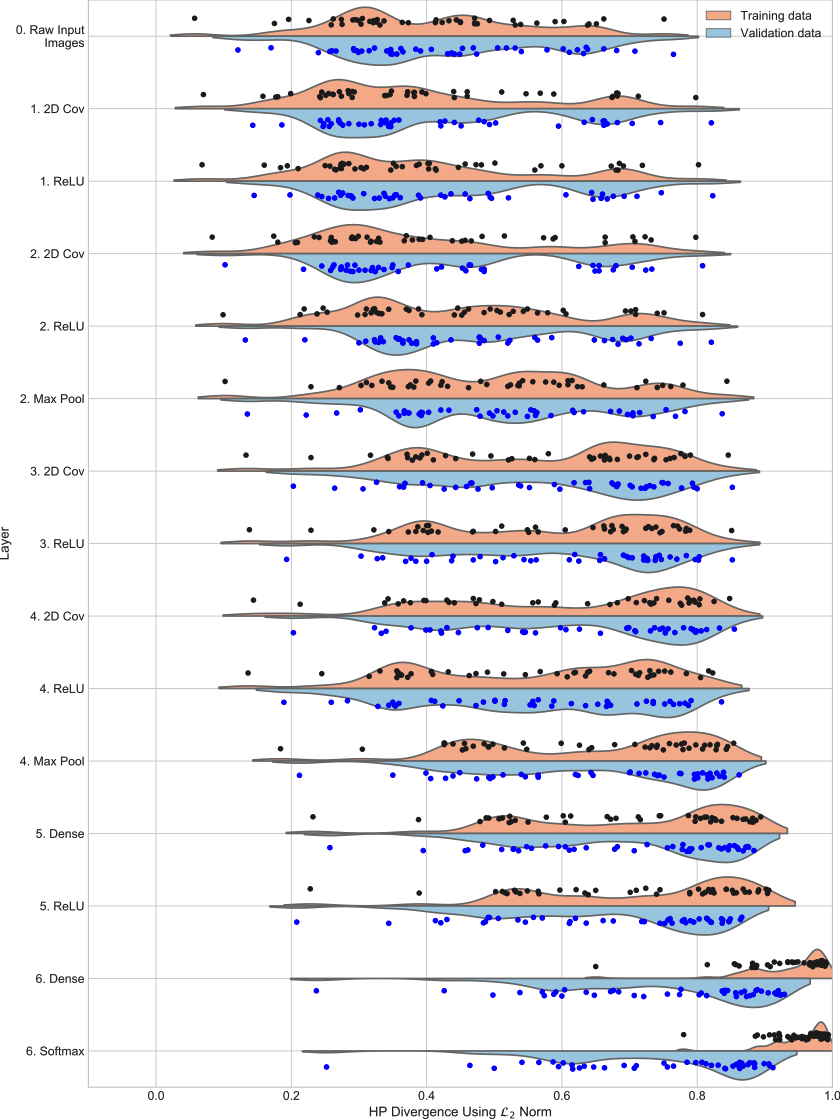}
  \caption{Trained model}
\end{subfigure}
\caption{$\mathcal{H}$ class-pair statistics at each layer for instance 1 of the model for CIFAR10 with
         true class labels. (a) shows results for the data for passing through the randomly
         initialized model (epoch~0 state). (b) shows the results for the data  passing through the
         fully trained  model (stopping at peak validation set accuracy). (Note:~Euclidean distance
         is used as the proximity measure.)}
\label{figure:A7}
\end{figure}

\begin{figure}[p!]
\centering
\begin{subfigure}[(A)]{0.49\linewidth}
  \includegraphics[width=\linewidth]{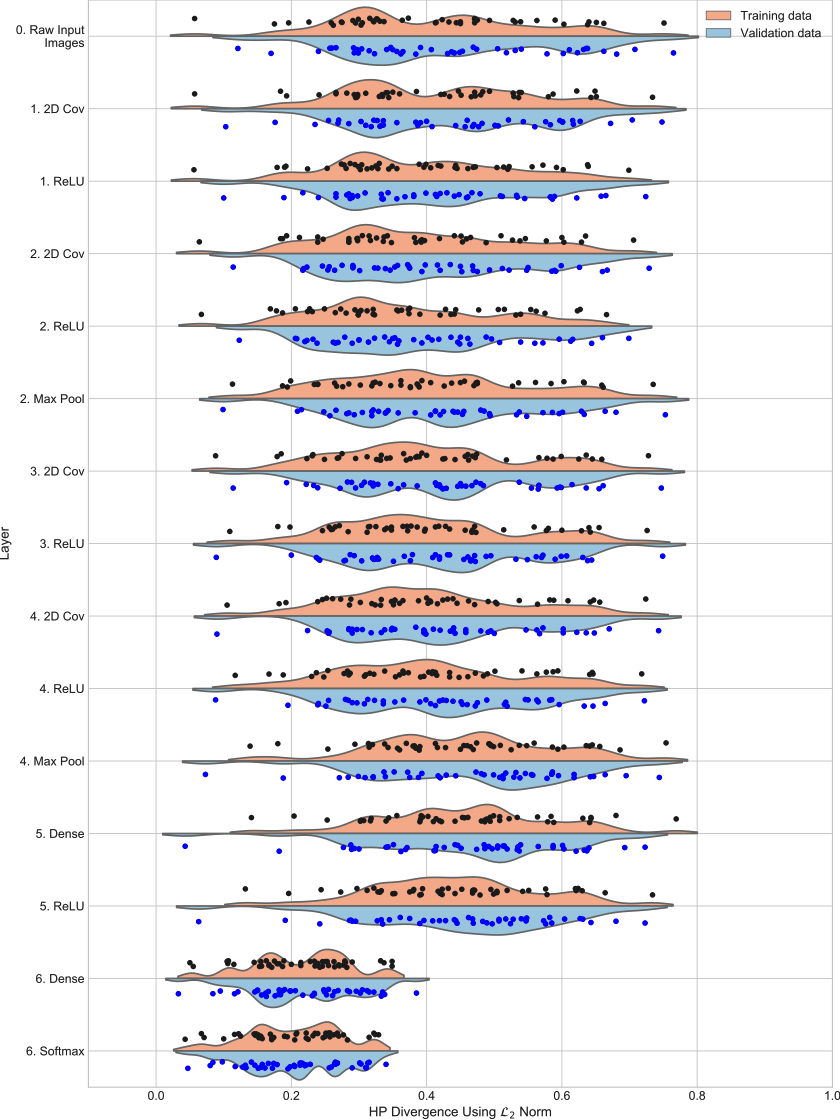}
  \caption{Untrained model}
\end{subfigure}
\begin{subfigure}[(B)]{0.49\linewidth}
  \includegraphics[width=\linewidth]{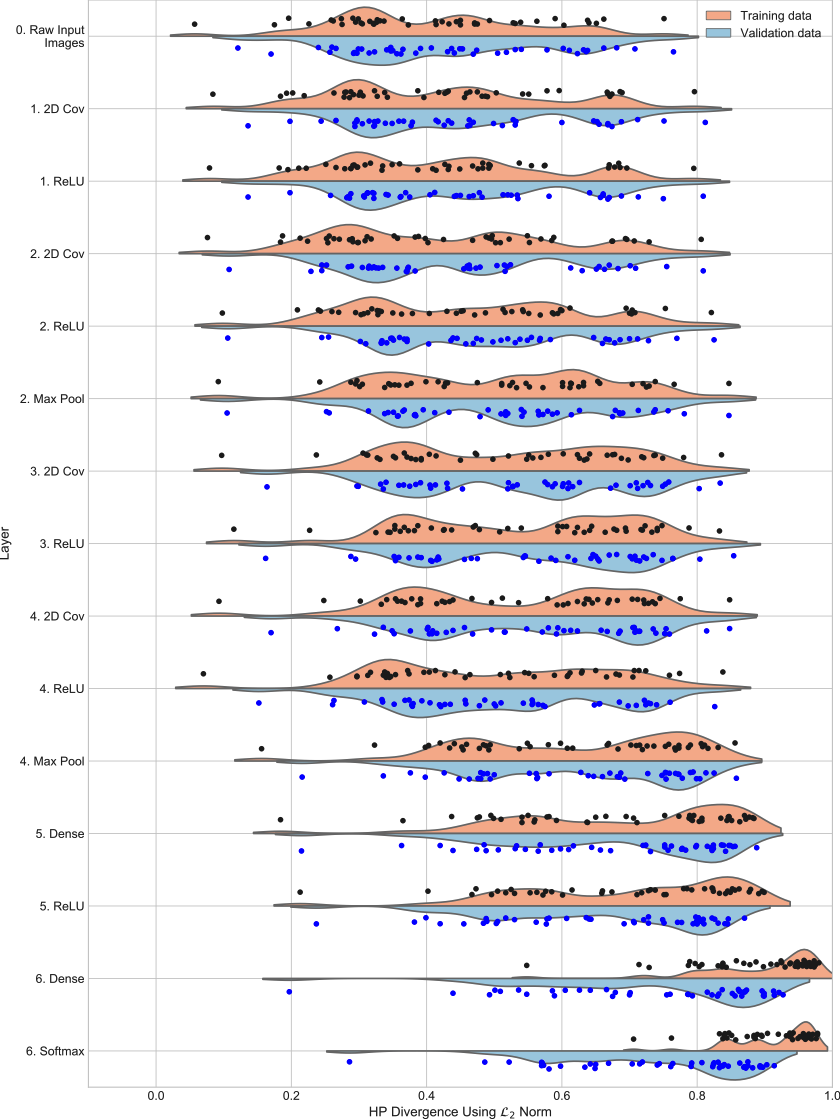}
  \caption{Trained model}
\end{subfigure}
\caption{$\mathcal{H}$ class-pair statistics at each layer for instance 2 of the model for CIFAR10 with
         true class labels. (a) shows results for the data for passing through the randomly
         initialized model  (epoch~0 state).  (b) shows the results for the data  passing through
         the fully trained  model  (stopping at peak validation set accuracy). (Note:~Euclidean
         distance is used as the proximity measure.)}
\label{figure:A8}
\end{figure}

\begin{figure}[p!]
\centering
\begin{subfigure}[(A)]{0.49\linewidth}
  \includegraphics[width=\linewidth]{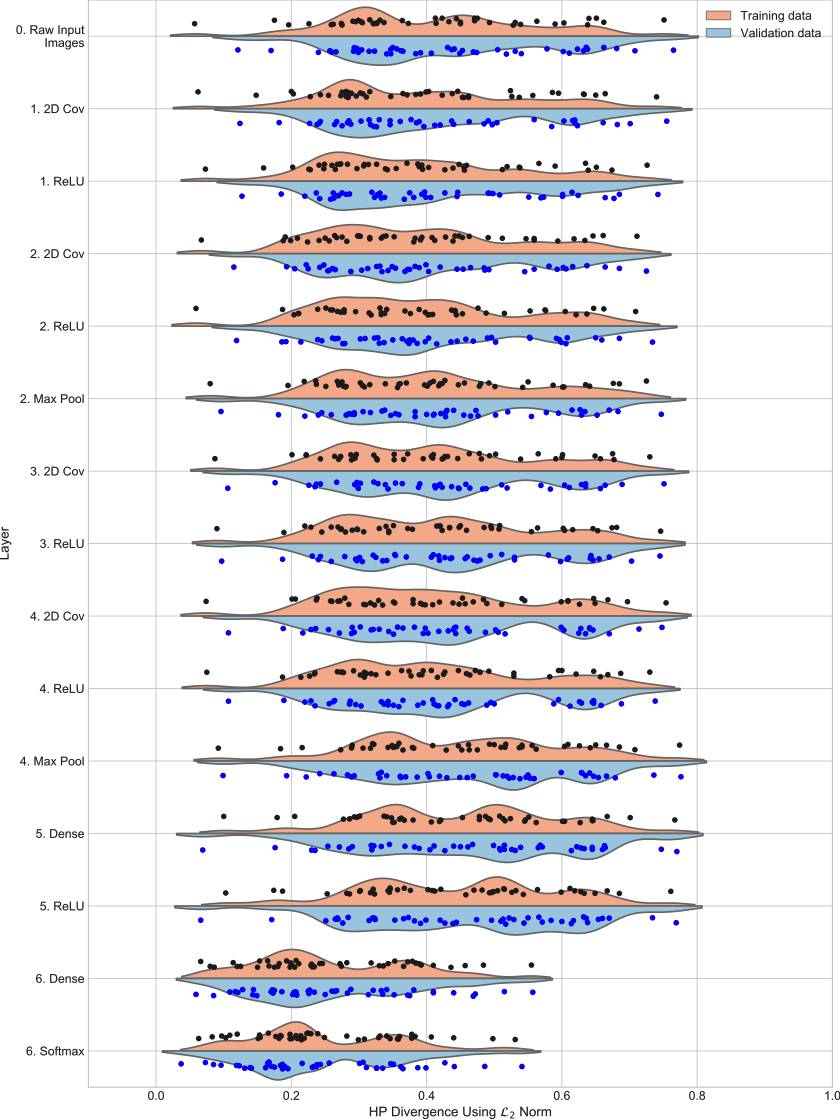}
  \caption{Untrained model}
\end{subfigure}
\begin{subfigure}[(B)]{0.49\linewidth}
  \includegraphics[width=\linewidth]{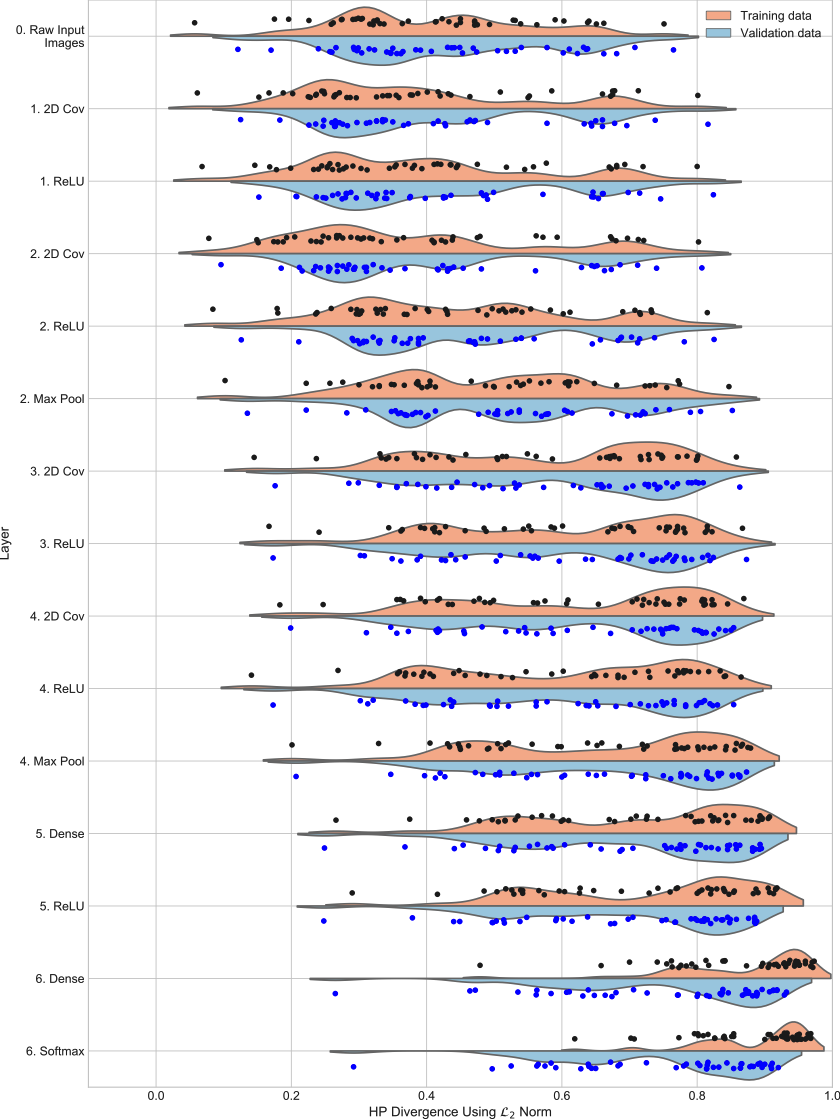}
  \caption{Trained model}
\end{subfigure}
\caption{$\mathcal{H}$ class-pair statistics at each layer for instance 3 of the model for CIFAR10 with
         true class labels. (a) shows results for the data for passing through the randomly
         initialized model  (epoch~0 state).  (b) shows the results for the data  passing through
         the fully trained  model  (stopping at peak validation set accuracy). (Note:~Euclidean
         distance is used as the proximity measure.)}
\label{figure:A9}
\end{figure}

\begin{figure}[p!]
\centering
\begin{subfigure}[(A)]{0.49\linewidth}
  \includegraphics[width=\linewidth]{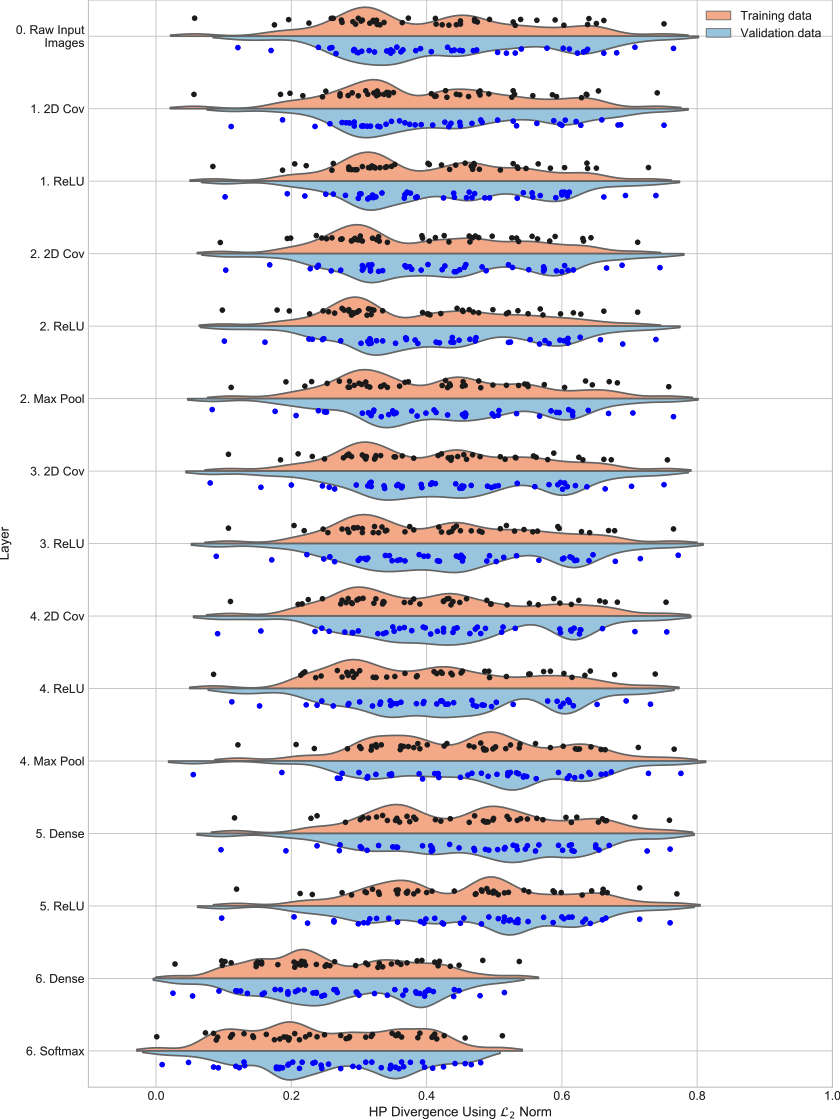}
  \caption{Untrained model}
\end{subfigure}
\begin{subfigure}[(B)]{0.49\linewidth}
  \includegraphics[width=\linewidth]{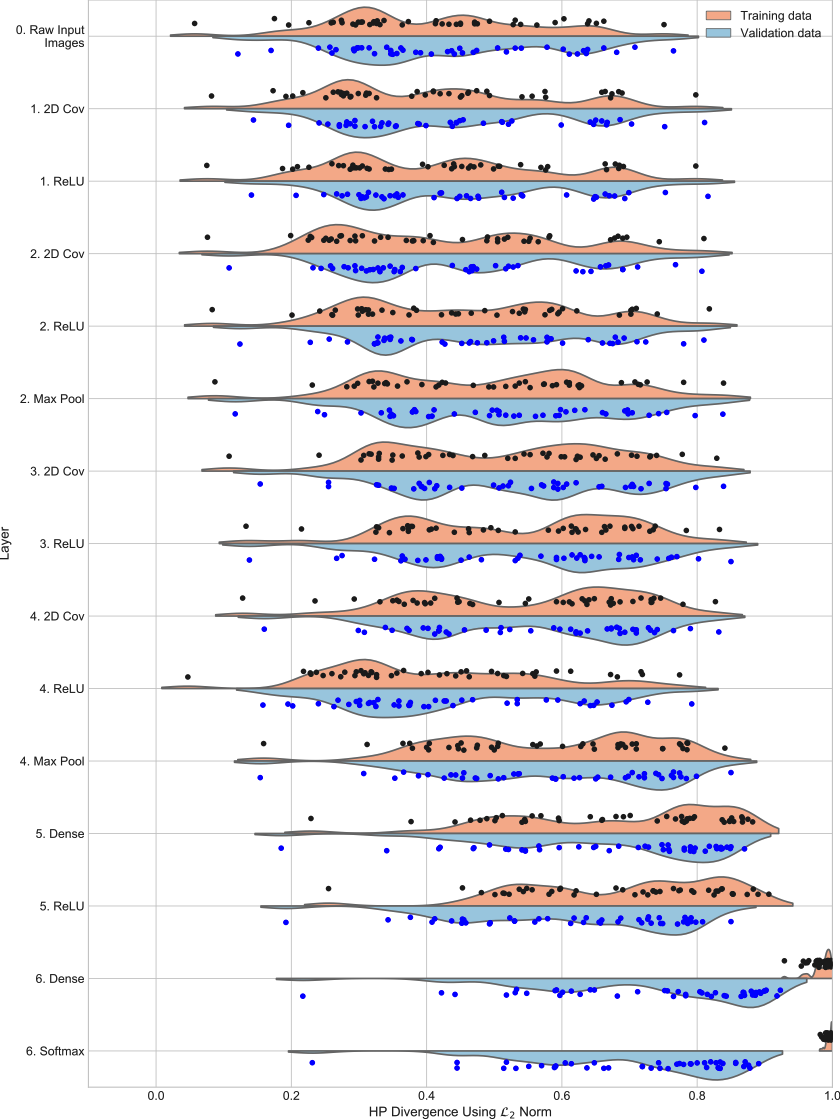}
  \caption{Trained model}
\end{subfigure}
\caption{$\mathcal{H}$ class-pair statistics at each layer for instance 4 of the model for CIFAR10 with
         true class labels. (a) shows results for the data for passing through the randomly
         initialized model  (epoch~0 state).  (b) shows the results for the data  passing through
         the fully trained  model  (stopping at peak validation set accuracy). (Note:~Euclidean
         distance is used as the proximity measure.)}
\label{figure:A10}
\end{figure}

\begin{figure}[p!]
\centering
\begin{subfigure}[(A)]{0.49\linewidth}
  \includegraphics[width=\linewidth]{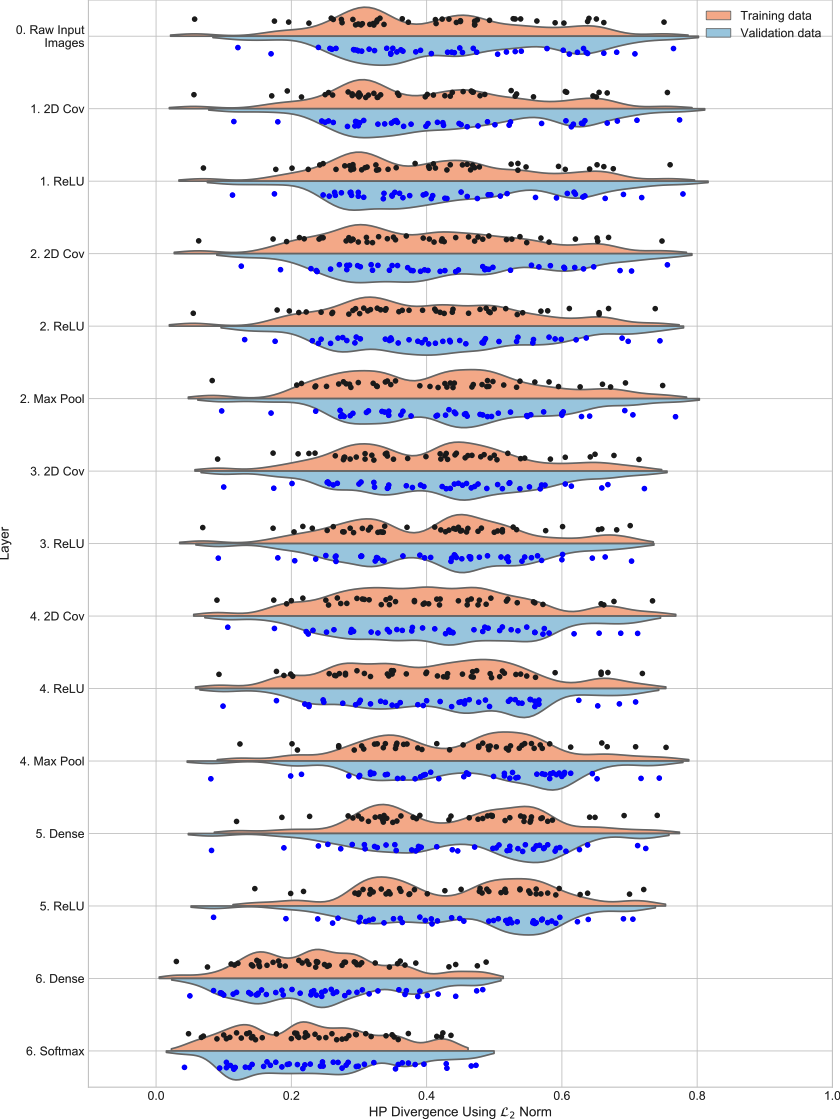}
  \caption{Untrained model}
\end{subfigure}
\begin{subfigure}[(B)]{0.49\linewidth}
  \includegraphics[width=\linewidth]{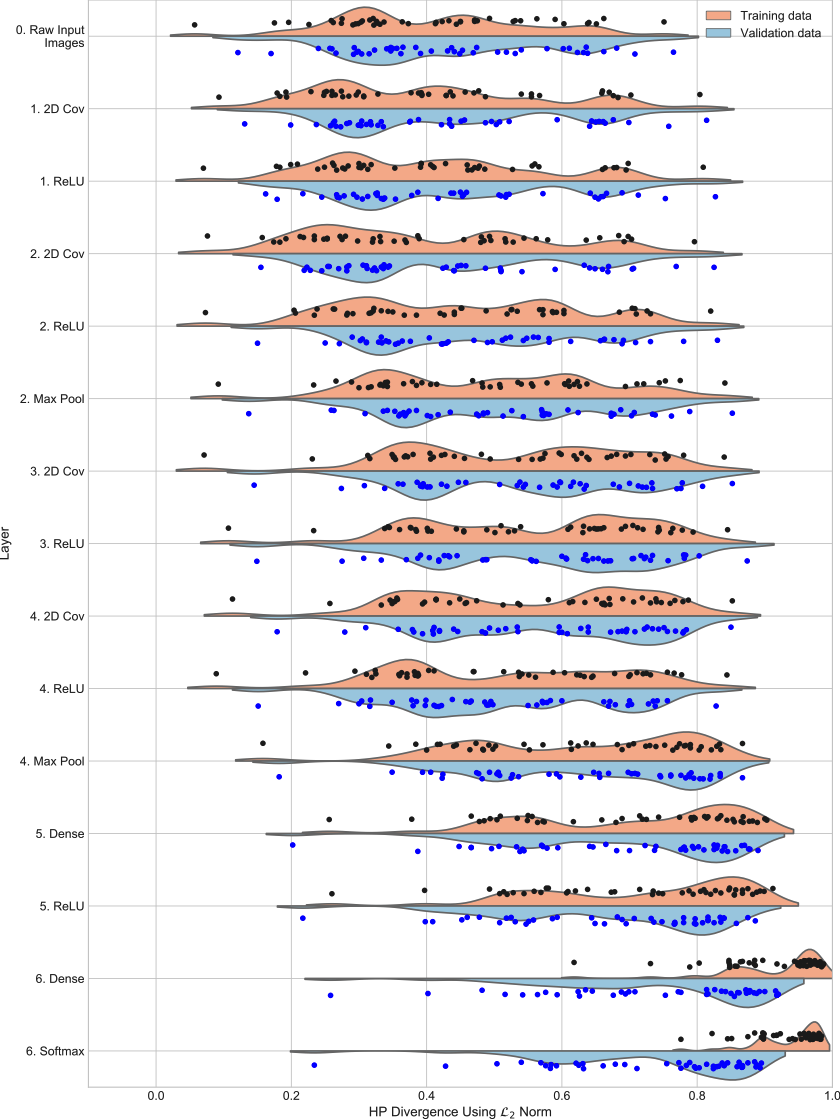}
  \caption{Trained model}
\end{subfigure}
\caption{$\mathcal{H}$ class-pair statistics at each layer for instance 5 of the model for CIFAR10 with
         true class labels. (a) shows results for the data for passing through the randomly
         initialized model  (epoch~0 state).  (b) shows the results for the data  passing through
         the fully trained  model  (stopping at peak validation set accuracy). (Note:~Euclidean
         distance is used as the proximity measure.)}
\label{figure:A11}
\end{figure}

\begin{figure}[p!]
\centering
\begin{subfigure}[(A)]{0.49\linewidth}
  \includegraphics[width=\linewidth]{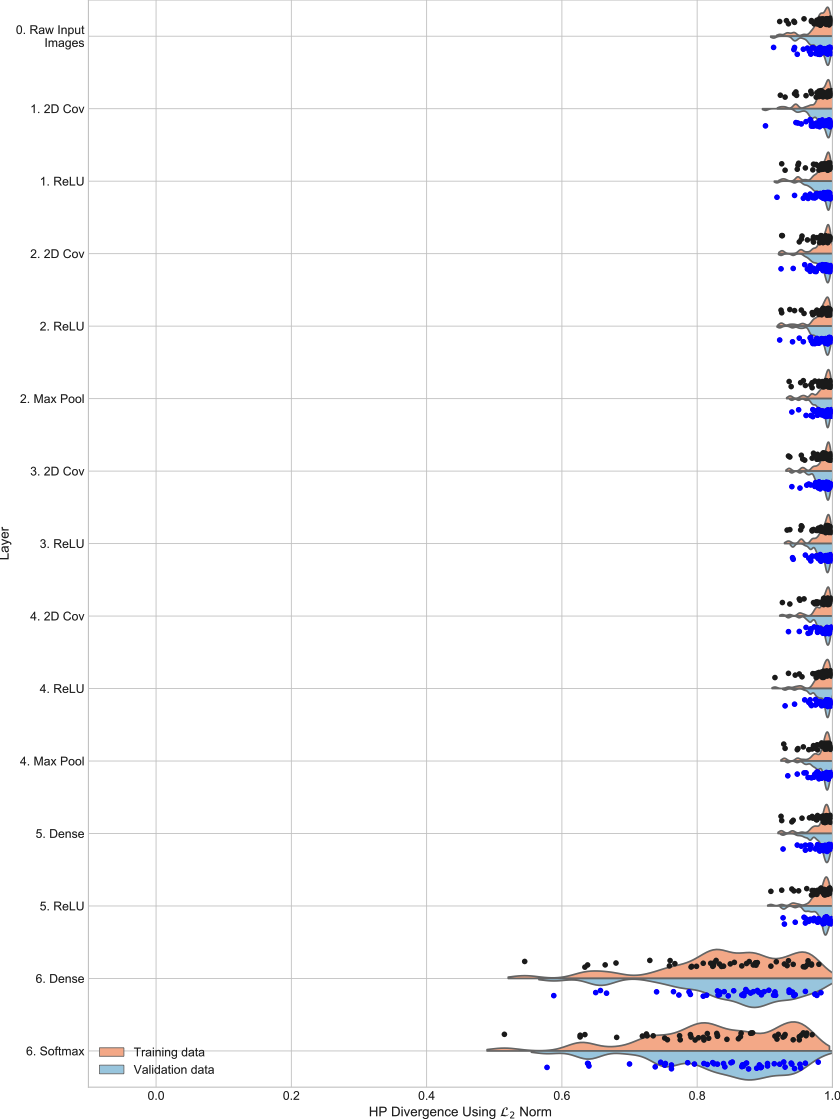}
  \caption{Untrained model}
\end{subfigure}
\begin{subfigure}[(B)]{0.49\linewidth}
  \includegraphics[width=\linewidth]{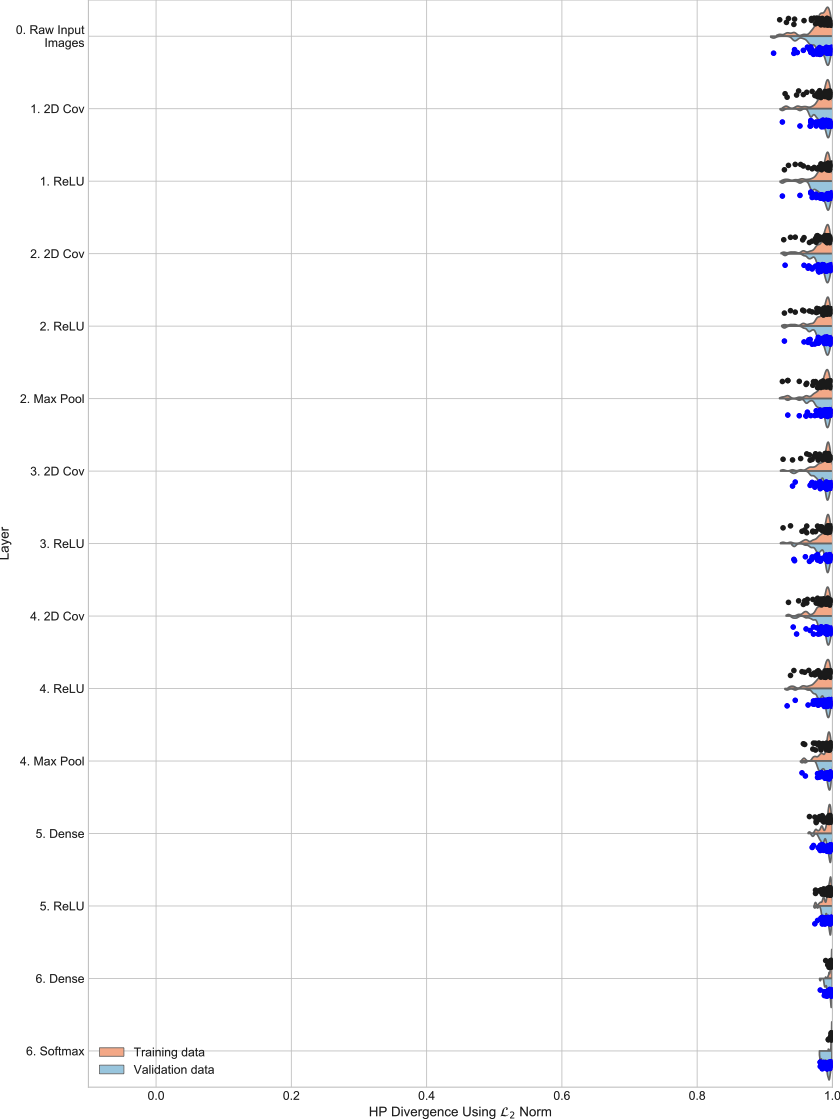}
  \caption{Trained model}
\end{subfigure}
\caption{$\mathcal{H}$ class-pair statistics at each layer for instance 1 of the model for MNIST with
         true class labels. (a) shows results for the data for passing through the randomly
         initialized model  (epoch~0 state).  (b) shows the results for the data  passing through
         the fully trained  model  (stopping at peak validation set accuracy). (Note:~Euclidean
         distance is used as the proximity measure.)}
\label{figure:A12}
\end{figure}

\begin{figure}[p!]
\centering
\begin{subfigure}[(A)]{0.49\linewidth}
  \includegraphics[width=\linewidth]{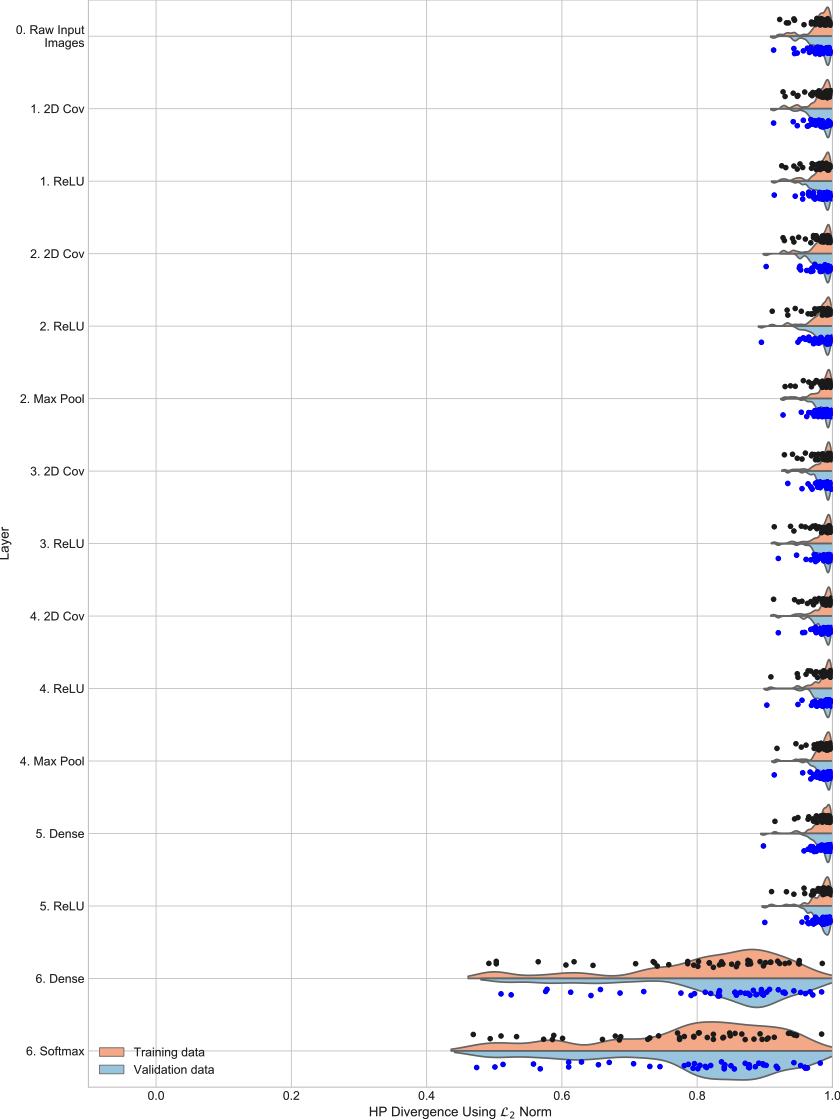}
  \caption{Untrained model}
\end{subfigure}
\begin{subfigure}[(B)]{0.49\linewidth}
  \includegraphics[width=\linewidth]{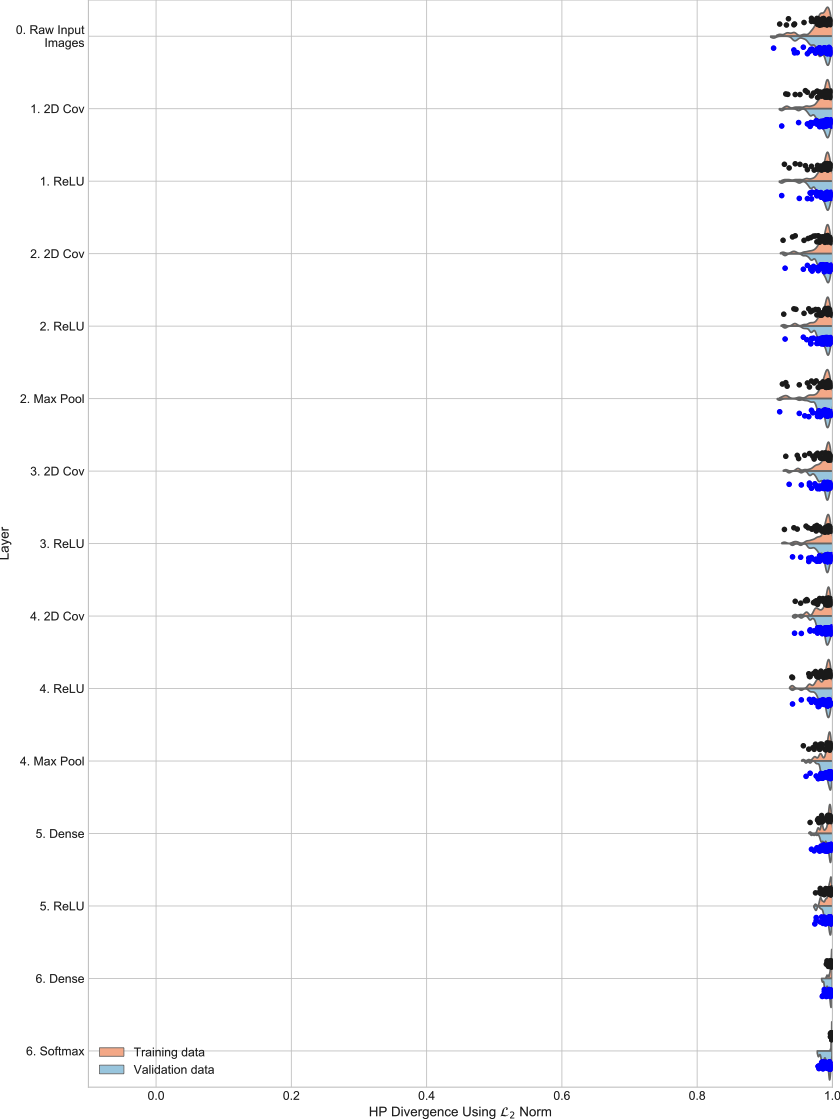}
  \caption{Trained model}
\end{subfigure}
\caption{$\mathcal{H}$ class-pair statistics at each layer for instance 2 of the model for MNIST with
         true class labels. (a) shows results for the data for passing through the randomly
         initialized model  (epoch~0 state).  (b) shows the results for the data  passing through
         the fully trained  model  (stopping at peak validation set accuracy). (Note:~Euclidean
         distance is used as the proximity measure.)}
\label{figure:A13}
\end{figure}

\begin{figure}[p!]
\centering
\begin{subfigure}[(A)]{0.49\linewidth}
  \includegraphics[width=\linewidth]{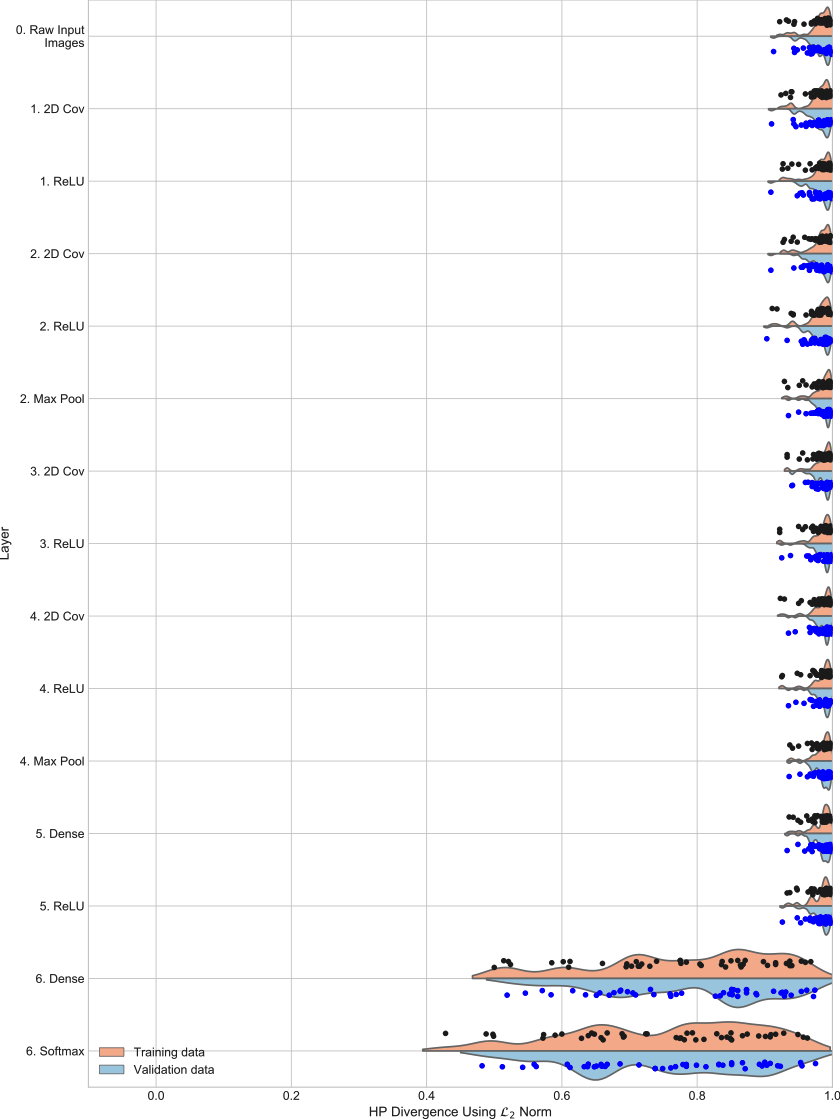}
  \caption{Untrained model}
\end{subfigure}
\begin{subfigure}[(B)]{0.49\linewidth}
  \includegraphics[width=\linewidth]{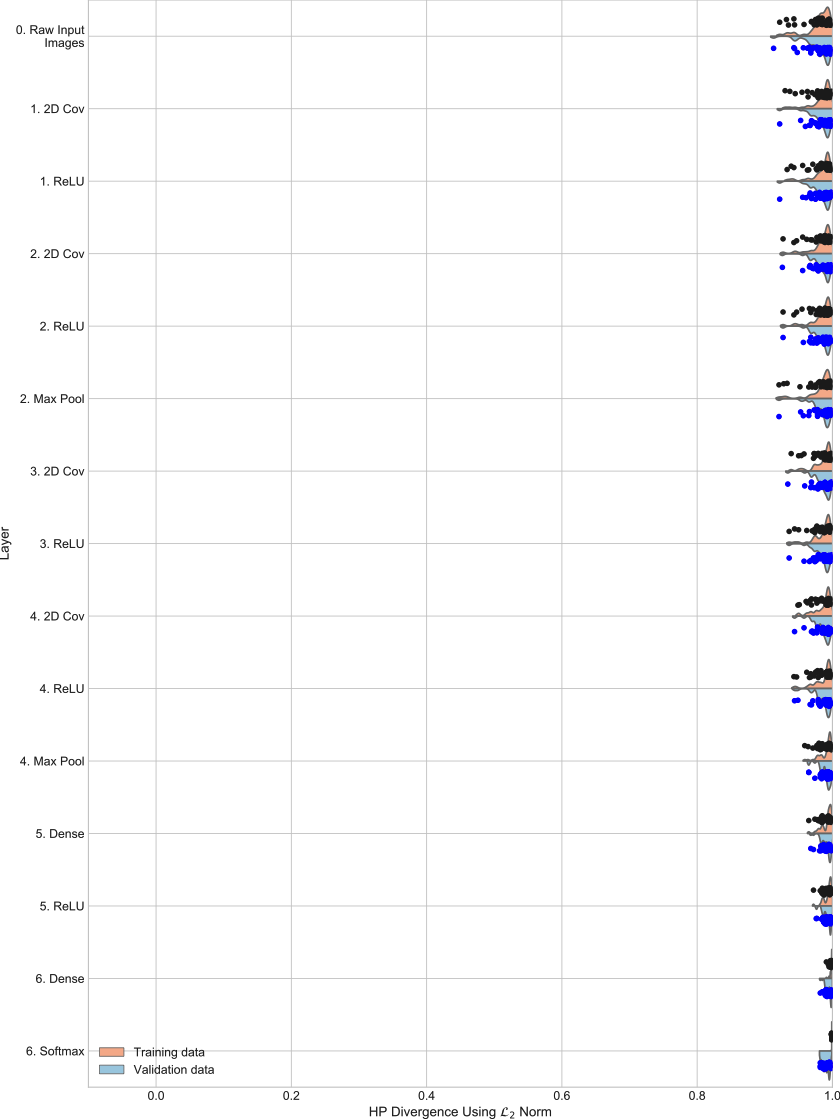}
  \caption{Trained model}
\end{subfigure}
\caption{$\mathcal{H}$ class-pair statistics at each layer for instance 3 of the model for MNIST with
         true class labels. (a) shows results for the data for passing through the randomly
         initialized model  (epoch~0 state).  (b) shows the results for the data  passing through
         the fully trained  model  (stopping at peak validation set accuracy). (Note:~Euclidean
         distance is used as the proximity measure.)}
\label{figure:A14}
\end{figure}

\begin{figure}[p!]
\centering
\begin{subfigure}[(A)]{0.49\linewidth}
  \includegraphics[width=\linewidth]{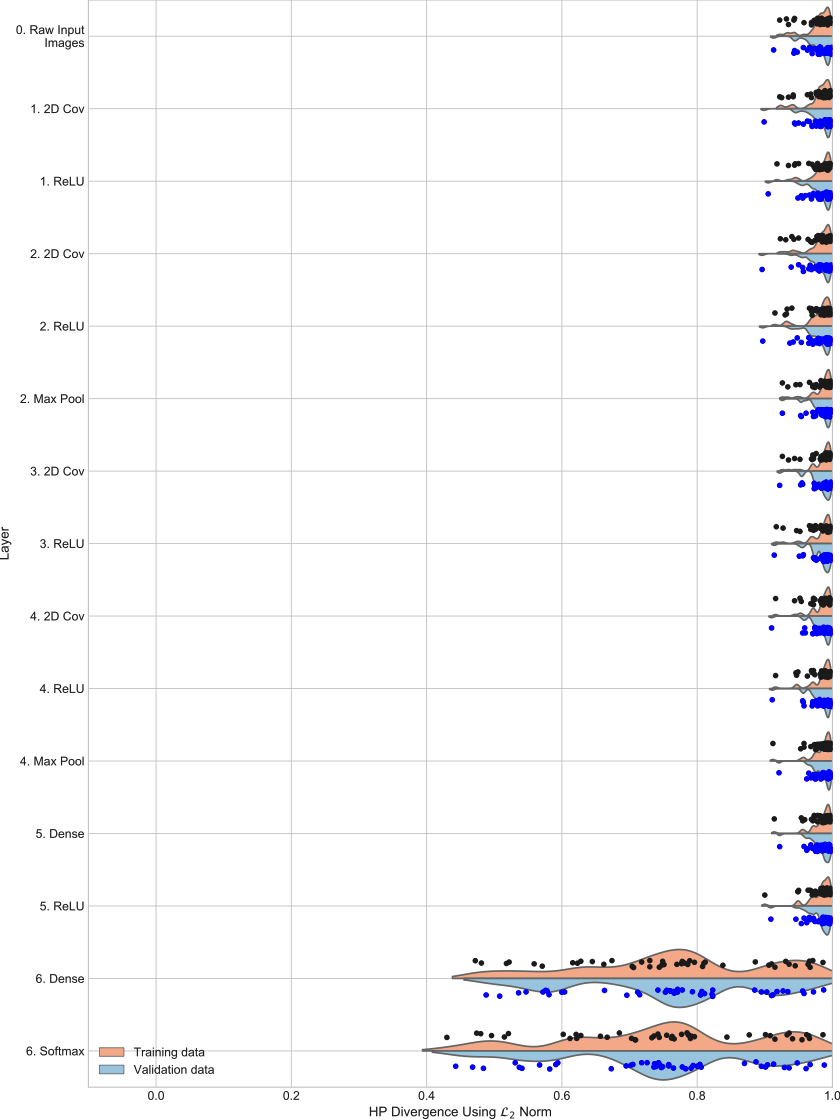}
  \caption{Untrained model}
\end{subfigure}
\begin{subfigure}[(B)]{0.49\linewidth}
  \includegraphics[width=\linewidth]{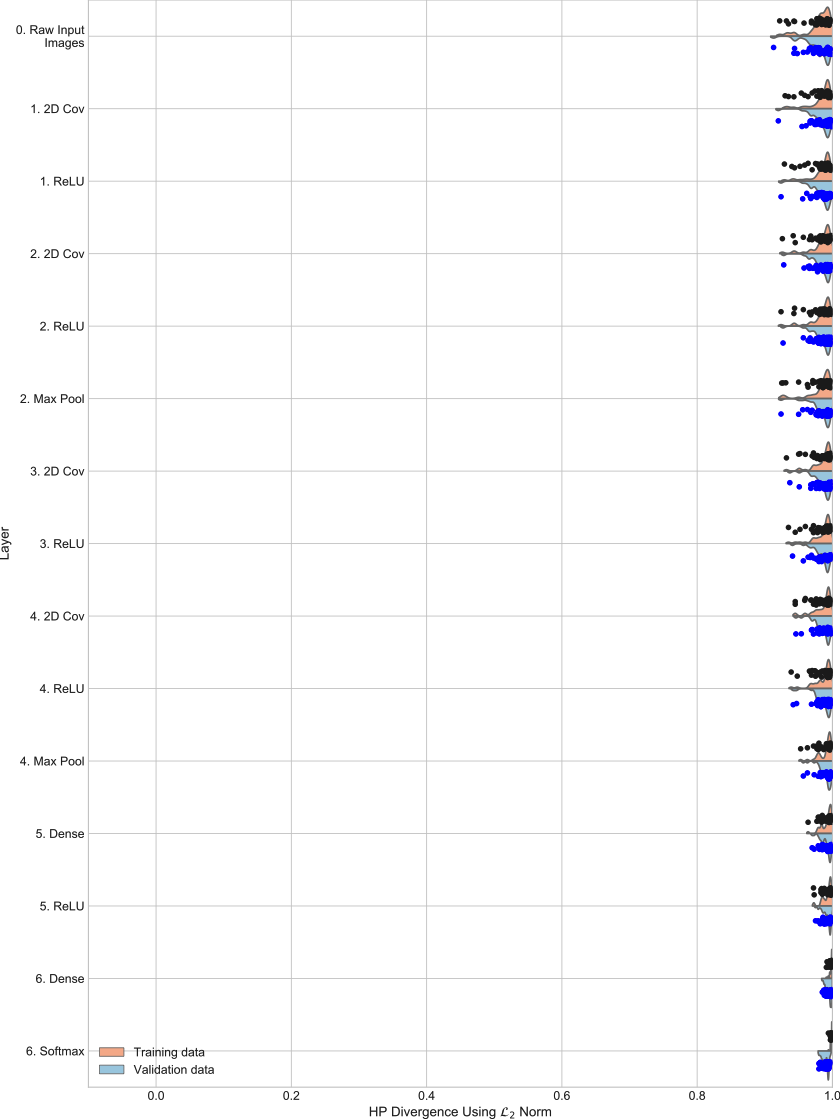}
  \caption{Trained model}
\end{subfigure}
\caption{$\mathcal{H}$ class-pair statistics at each layer for instance 4 of the model for MNIST with
         true class labels. (a) shows results for the data for passing through the randomly
         initialized model  (epoch~0 state).  (b) shows the results for the data  passing through
         the fully trained  model  (stopping at peak validation set accuracy). (Note:~Euclidean
         distance is used as the proximity measure.)}
\label{figure:A15}
\end{figure}

\begin{figure}[p!]
\centering
\begin{subfigure}[(A)]{0.49\linewidth}
  \includegraphics[width=\linewidth]{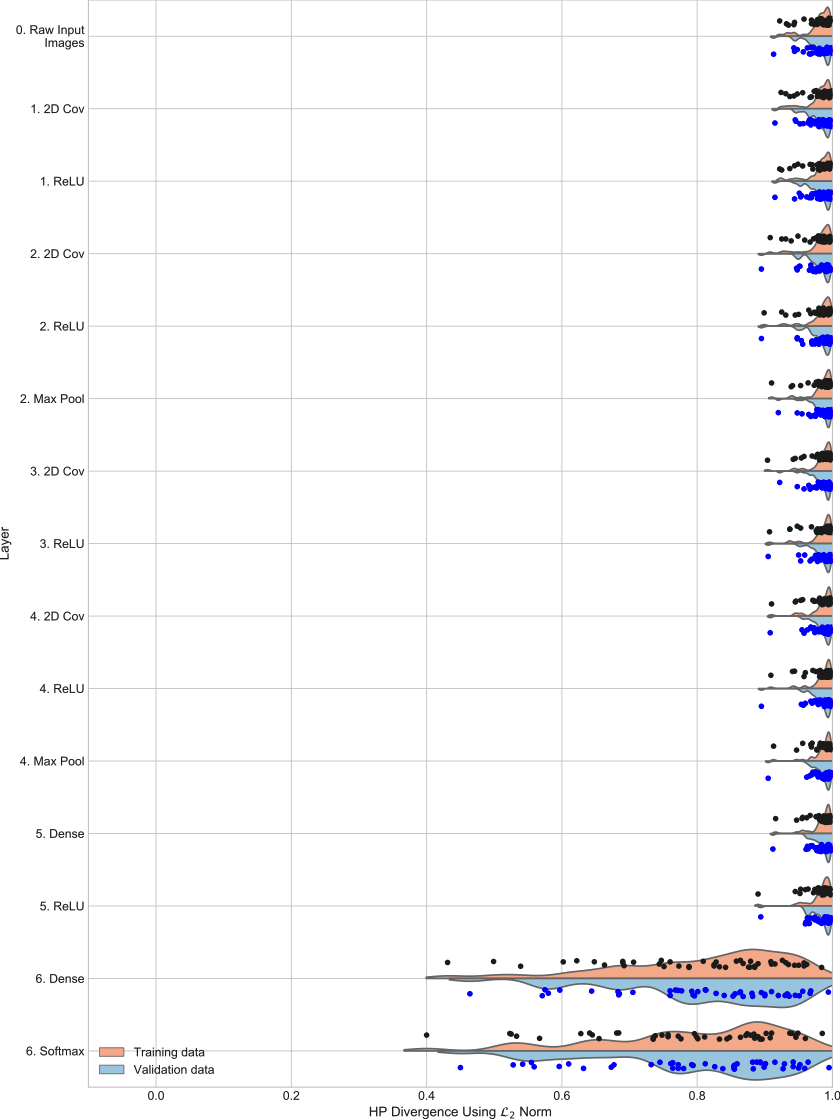}
  \caption{Untrained model}
\end{subfigure}
\begin{subfigure}[(B)]{0.49\linewidth}
  \includegraphics[width=\linewidth]{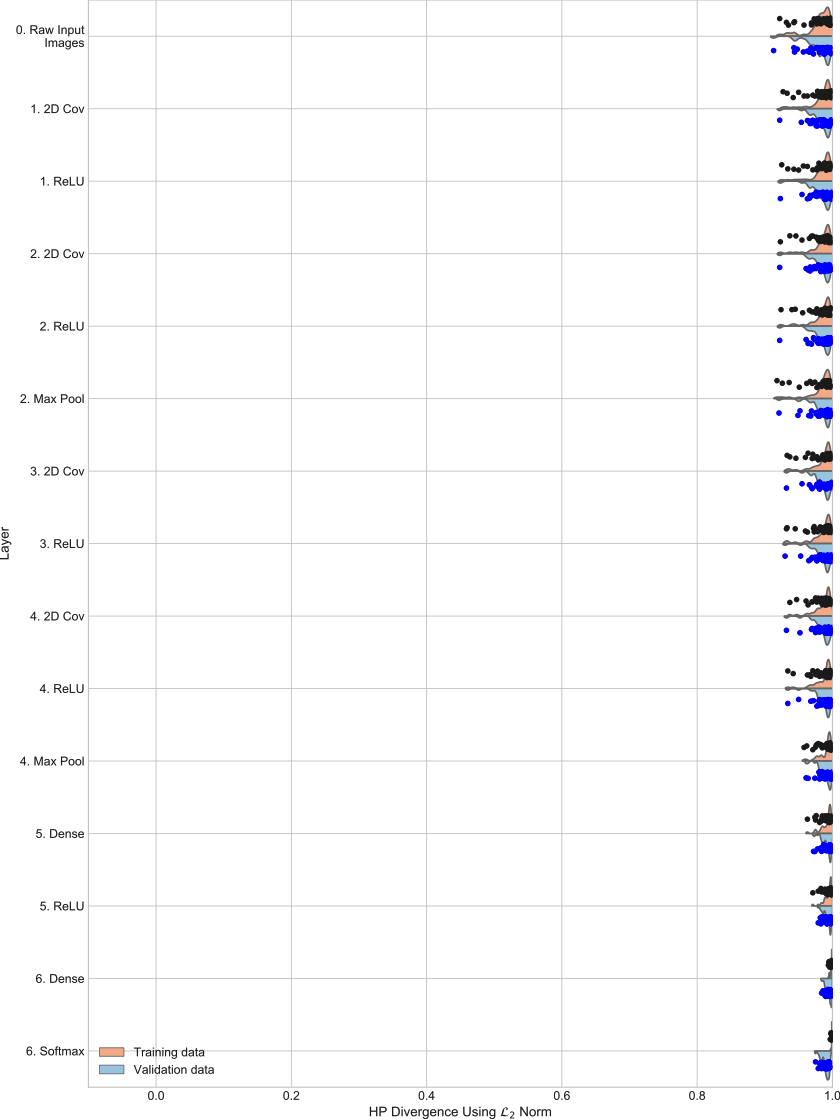}
  \caption{Trained model}
\end{subfigure}
\caption{$\mathcal{H}$ class-pair statistics at each layer for instance 5 of the model for MNIST with
         true class labels. (a) shows results for the data for passing through the randomly
         initialized model  (epoch~0 state).  (b) shows the results for the data  passing through
         the fully trained  model  (stopping at peak validation set accuracy). (Note:~Euclidean
         distance is used as the proximity measure.)}
\label{figure:A16}
\end{figure}

Figures~\ref{figure:A2} through~\ref{figure:A6} present the class-wise HP statistics of the training
and validation  samples of the CIFAR10 data with random labels as they pass through an associated
model. Each figure plots the results for one of the five training instances discussed in
Section~3 of the paper.  The~(a)~subfigures show the data passing through the untrained models
and the~(b)~subfigures show the data passing through the trained version of the models. Similarly,
Figures~\ref{figure:A7} through~\ref{figure:A11} present plots for the~5 model instances trained on
the CIFAR10 data with true labels, and Figures~\ref{figure:A12} through~\ref{figure:A16} show the
plots for the~5 MNIST-trained models.

Tables~\ref{table:A1Table} through~\ref{table:A9Table} present a superset of the results of
two-sample null hypothesis tests of means presented in the main body of the paper.  In particular
Tables~\ref{table:A4Table},~\ref{table:A5Table},~\ref{table:A7Table}, and~\ref{table:A8Table} show
the test performed between multi-layer components of the networks. The tables flag cases where the
estimated $p$-values are $< 0.025$. Note that the tests were performed using the random permutation
algorithm with~50,000 Monte Carlo trials.
\begin{table}[p!]
\tiny
\centering
\caption{Two-sided permutation test of training data to detect change in  $\bar{\mathcal{H}}$ between
         layers (before training) with critical value $\alpha = 0.025$. \textcolor{red}{Red font}
         denote layer instances for which we reject $\mathbf{H_0}$, black font denotes layers for
         which we fail to reject $\mathbf{H_0}$.  (Eq. 14 in paper)
         (Note:~Euclidean distance is used as the proximity
         measure. 50,000 Monte Carlo trials were used to estimate the $p$-values. $\Delta{\bar{\mathcal{H}}} = \bar{\mathcal{H}}^{(t)}_{(k,0)} - \bar{\mathcal{H}}^{(t)}_{(k-1,0)}$))}
\label{table:A1Table}
\begin{tabular}{p{12mm}|p{12mm}|K{14mm}|K{14mm}|K{14mm}|K{14mm}|K{14mm}|K{14mm}}
\toprule
Input Space&Output Space& \multicolumn{2}{c}{CIFAR10 w Random}& \multicolumn{2}{c}{CIFAR10 w True}&
\multicolumn{2}{c}{MNIST w True} \\
& &$\Delta{\bar{\mathcal{H}}}$ & \textit{p}-values & $\Delta{\bar{\mathcal{H}}}$ & \textit{p}-values&
$\Delta{\bar{\mathcal{H}}}$ & \textit{p}-values \\
\midrule
0.Input&1.Conv&.003; .002; -.004; .003; -.003&.460; .649; .445; .556; .515&-.012; .005; -.017;
-.003; -.004&.714; .864; .593; .923; .912&.002; .001; -.001; .000; .000&.672; .792; .864; .934;
.900\\
\hline
1.Conv&1.ReLU&-.002; -.003; .004; .002; .003&.531; .523; .434; .675; .641&.016; -.022; -.003; .001;
-.000&.607; .463; .933; .969; .991&.001; .000; .001; .002; .000&.875; .906; .713; .648; .918\\
\hline
1.ReLU&2.Conv&.003; .001; -.001; -.002; -.001&.375; .801; .856; .650; .906&.003; -.006; -.001;
-.009; -.005&.924; .842; .964; .771; .878&.000; -.000; -.001; -.001; -.000&.970; .981; .878; .826;
.933\\
\hline
2.Conv&2.ReLU&-.001; -.002; .003; -.002; .001&.671; .626; .526; .667; .906&.002; -.017; .005; -.003;
.003&.959; .562; .869; .929; .912&-.001; -.000; -.001; -.001; .001&.730; .962; .803; .748; .839\\
\hline
2.ReLU&2.MaxPool&-.005; .004; -.001; -.007; -.004&.202; .390; .832; .195; .576&.019; .035; .015;
.020; .015&.563; .241; .623; .505; .624&.003; .003; .005; .004; .002&.365; .394; .202; .294; .565\\
\hline
2.MaxPool&3.Conv&.001; -.001; .004; -.006; -.003&.873; .876; .428; .286; .695&.003; -.001; .003;
-.002; -.009&.936; .978; .931; .948; .755&.000; .000; -.000; -.000; .000&.998; 1.000; .998; .956;
.973\\
\hline
3.Conv&3.ReLU&-.001; -.000; -.003; .002; .007&.747; .969; .574; .629; .246&.002; .003; .007; .004;
-.002&.960; .914; .827; .895; .957&-.001; -.001; -.001; -.001; -.001&.872; .849; .721; .768; .895\\
\hline
3.ReLU&4.Conv&.005; .004; -.000; .000; -.005&.231; .345; .932; .954; .379&-.006; -.001; -.001;
-.004; -.004&.861; .967; .975; .897; .904&.000; .001; .001; .001; .001&.998; .860; .823; .777;
.842\\
\hline
4.Conv&4.ReLU&.005; -.000; .001; .003; .001&.220; .945; .842; .551; .821&-.003; -.003; -.006; -.008;
.005&.936; .913; .851; .797; .878&-.001; .000; -.000; -.000; .000&.750; .945; .954; .893; .983\\
\hline
4.ReLU&4.MaxPool&-.009; .008; -.004; -.004; .004&.059; .092; .364; .425; .444&.049; .054; .040;
.051; .038&.113; .061; .204; .096; .212&-.000; .000; .001; .001; .001&.962; .987; .840; .778; .777\\
\hline
4.MaxPool&5.Dense&.009; .000; .000; .000; .001&.061; .976; .925; .996; .838&-.005; -.001; .002;
.003; -.003&.858; .976; .962; .920; .908&-.001; -.001; -.001; -.001; -.001&.758; .746; .633; .880;
.682\\
\hline
5.Dense&5.ReLU&-.004; .004; .005; .001; -.006&.424; .385; .283; .925; .199&-.007; -.014; -.008;
-.003; .003&.813; .596; .789; .908; .915&-.002; -.002; -.001; -.002; -.002&.696; .654; .723; .549;
.666\\
\hline
5.ReLU&6.Dense&.005; \textcolor{red}{-.017}; \textcolor{red}{-.017}; .005; .003&.330;
\textcolor{red}{.001}; \textcolor{red}{.000}; .305; .549&\textcolor{red}{-.202};
\textcolor{red}{-.226}; \textcolor{red}{-.182}; \textcolor{red}{-.189};
\textcolor{red}{-.198}&\textcolor{red}{.000}; \textcolor{red}{.000}; \textcolor{red}{.000};
\textcolor{red}{.000}; \textcolor{red}{.000}&\textcolor{red}{-.135}; \textcolor{red}{-.173};
\textcolor{red}{-.202}; \textcolor{red}{-.219}; \textcolor{red}{-.177}&\textcolor{red}{.000};
\textcolor{red}{.000}; \textcolor{red}{.000}; \textcolor{red}{.000}; \textcolor{red}{.000}\\
\hline
6.Dense&6.Softmax&-.001; -.002; -.002; -.007; -.000&.891; .665; .688; .145; .933&-.018; -.011;
-.010; -.013; -.017&.473; .457; .690; .618; .446&-.018; -.030; -.030; -.014; -.016&.419; .280; .296;
.652; .582\\
\bottomrule
\end{tabular}
\end{table}
\begin{table}[p!]
\tiny
\centering
\caption{Differences between the trained and initialized $\boldsymbol{\mathcal{H}}$ class-pair statistics of a layer,
         and respective $p$-values for the
         corresponding one-sided permutation test~(Eq. 15 in paper). \textcolor{red}{Red font}
         denotes layer instances for which we reject $\mathbf{H_0}$, and black font denotes layers for
         which we fail to reject $\mathbf{H_0}$. 
         (Note: ~Euclidean distance is used as the proximity measure. 50,000 Monte Carlo trials to estimate the $p$-values.
         $\Delta{\bar{\mathcal{H}}} = \bar{\mathcal{H}}^{(t)}_{(k,T)} - \bar{\mathcal{H}}^{(t)}_{(k,0)}$)}

\label{table:A2Table}
\begin{tabular}{p{12mm}|K{16mm}|K{16mm}|K{16mm}|K{16mm}|K{16mm}|K{16mm}}
\toprule
Output Space& \multicolumn{2}{c|}{CIFAR10 w Random}& \multicolumn{2}{c|}{CIFAR10 w True}&
\multicolumn{2}{c}{MNIST w True} \\
 &$\Delta{\bar{\mathcal{H}}}$ & \textit{p}-values & $\Delta{\bar{\mathcal{H}}}$ & \textit{p}-values&
 $\Delta{\bar{\mathcal{H}}}$ & \textit{p}-values \\
\midrule
1.Conv&-.000; -.002; .000; -.007; .000&.507; .645; .472; .888; .482&-.007; .005; -.008; .004;
-.004&.580; .435; .591; .453; .542&.001; .002; .004; .003; .003&.346; .277; .126; .209; .248\\
\hline
1.ReLU&.004; .004; -.004; -.009; -.001&.137; .225; .775; .955; .587&-.012; .029; .005; .015;
.005&.646; .183; .440; .320; .448&.001; .002; .003; .001; .002&.417; .316; .208; .354; .274\\
\hline
2.Conv&.007; -.009; -.011; -.014; -.004&.032; .973; .976; .995; .812&-.017; .039; -.006; .028;
.001&.690; .121; .570; .200; .491&.001; .002; .004; .003; .003&.422; .247; .161; .243; .202\\
\hline
2.ReLU&\textcolor{red}{.008}; -.003; -.012; -.015; -.008&\textcolor{red}{.017}; .706; .989; .998;
.913&.032; \textcolor{red}{.095}; .051; \textcolor{red}{.066}; .051&.174; \textcolor{red}{.002};
.068; \textcolor{red}{.024}; .068&.002; .003; .005; .004; .002&.281; .233; .105; .161; .271\\
\hline
2.MaxPool&.005; .004; -.015; -.010; -.004&.126; .161; .999; .967; .725&.065; \textcolor{red}{.100};
\textcolor{red}{.094}; \textcolor{red}{.081}; \textcolor{red}{.073}&.030; \textcolor{red}{.002};
\textcolor{red}{.003}; \textcolor{red}{.008}; \textcolor{red}{.016}&-.002; -.002; -.001; -.001;
-.001&.722; .669; .647; .655; .603\\
\hline
3.Conv&\textcolor{red}{.009}; .001; -.021; .000; .004&\textcolor{red}{.011}; .388; 1.000; .498;
.233&\textcolor{red}{.127}; \textcolor{red}{.117}; \textcolor{red}{.168}; \textcolor{red}{.097};
\textcolor{red}{.118}&\textcolor{red}{.000}; \textcolor{red}{.000}; \textcolor{red}{.000};
\textcolor{red}{.002}; \textcolor{red}{.000}&-.000; .001; .001; .001; .001&.525; .433; .328; .355;
.440\\
\hline
3.ReLU&\textcolor{red}{.010}; .001; -.020; -.006; -.008&\textcolor{red}{.010}; .365; 1.000; .896;
.926&\textcolor{red}{.140}; \textcolor{red}{.135}; \textcolor{red}{.184}; \textcolor{red}{.117};
\textcolor{red}{.149}&\textcolor{red}{.000}; \textcolor{red}{.000}; \textcolor{red}{.000};
\textcolor{red}{.000}; \textcolor{red}{.000}&-.000; .001; .002; .002; .001&.500; .431; .264; .281;
.390\\
\hline
4.Conv&-.002; .006; -.014; -.011; .008&.669; .111; .999; .989; .053&\textcolor{red}{.168};
\textcolor{red}{.131}; \textcolor{red}{.205}; \textcolor{red}{.129};
\textcolor{red}{.148}&\textcolor{red}{.000}; \textcolor{red}{.000}; \textcolor{red}{.000};
\textcolor{red}{.000}; \textcolor{red}{.000}&.000; .001; .002; .001; .001&.444; .381; .278; .330;
.358\\
\hline
4.ReLU&-.001; .007; -.006; -.007; -.003&.610; .084; .919; .909; .704&\textcolor{red}{.128};
\textcolor{red}{.098}; \textcolor{red}{.190}; .011; \textcolor{red}{.106}&\textcolor{red}{.000};
\textcolor{red}{.002}; \textcolor{red}{.000}; .366; \textcolor{red}{.001}&.001; .001; .002; .003;
.002&.336; .397; .228; .210; .314\\
\hline
4.MaxPool&.009; -.007; -.016; .006; -.011&.045; .928; .999; .127; .988&\textcolor{red}{.161};
\textcolor{red}{.163}; \textcolor{red}{.207}; \textcolor{red}{.125};
\textcolor{red}{.181}&\textcolor{red}{.000}; \textcolor{red}{.000}; \textcolor{red}{.000};
\textcolor{red}{.000}; \textcolor{red}{.000}&\textcolor{red}{.006}; .005; \textcolor{red}{.006};
.005; .004&\textcolor{red}{.023}; .049; \textcolor{red}{.015}; .038; .068\\
\hline
5.Dense&\textcolor{red}{.049}; \textcolor{red}{.024}; \textcolor{red}{.036}; \textcolor{red}{.035};
\textcolor{red}{.029}&\textcolor{red}{.000}; \textcolor{red}{.000}; \textcolor{red}{.000};
\textcolor{red}{.000}; \textcolor{red}{.000}&\textcolor{red}{.227}; \textcolor{red}{.231};
\textcolor{red}{.257}; \textcolor{red}{.225}; \textcolor{red}{.258}&\textcolor{red}{.000};
\textcolor{red}{.000}; \textcolor{red}{.000}; \textcolor{red}{.000};
\textcolor{red}{.000}&\textcolor{red}{.009}; \textcolor{red}{.008}; \textcolor{red}{.009};
\textcolor{red}{.007}; \textcolor{red}{.007}&\textcolor{red}{.001}; \textcolor{red}{.001};
\textcolor{red}{.000}; \textcolor{red}{.001}; \textcolor{red}{.002}\\
\hline
5.ReLU&\textcolor{red}{.241}; \textcolor{red}{.234}; \textcolor{red}{.233}; \textcolor{red}{.238};
\textcolor{red}{.246}&\textcolor{red}{.000}; \textcolor{red}{.000}; \textcolor{red}{.000};
\textcolor{red}{.000}; \textcolor{red}{.000}&\textcolor{red}{.248}; \textcolor{red}{.256};
\textcolor{red}{.279}; \textcolor{red}{.237}; \textcolor{red}{.270}&\textcolor{red}{.000};
\textcolor{red}{.000}; \textcolor{red}{.000}; \textcolor{red}{.000};
\textcolor{red}{.000}&\textcolor{red}{.012}; \textcolor{red}{.011}; \textcolor{red}{.012};
\textcolor{red}{.010}; \textcolor{red}{.011}&\textcolor{red}{.000}; \textcolor{red}{.000};
\textcolor{red}{.000}; \textcolor{red}{.000}; \textcolor{red}{.000}\\
\hline
6.Dense&\textcolor{red}{.994}; \textcolor{red}{1.002}; \textcolor{red}{1.000};
\textcolor{red}{.989}; \textcolor{red}{.996}&\textcolor{red}{.000}; \textcolor{red}{.000};
\textcolor{red}{.000}; \textcolor{red}{.000}; \textcolor{red}{.000}&\textcolor{red}{.669};
\textcolor{red}{.681}; \textcolor{red}{.610}; \textcolor{red}{.718};
\textcolor{red}{.668}&\textcolor{red}{.000}; \textcolor{red}{.000}; \textcolor{red}{.000};
\textcolor{red}{.000}; \textcolor{red}{.000}&\textcolor{red}{.152}; \textcolor{red}{.188};
\textcolor{red}{.218}; \textcolor{red}{.234}; \textcolor{red}{.192}&\textcolor{red}{.000};
\textcolor{red}{.000}; \textcolor{red}{.000}; \textcolor{red}{.000}; \textcolor{red}{.000}\\
\hline
6.Softmax&\textcolor{red}{.995}; \textcolor{red}{1.004}; \textcolor{red}{1.002};
\textcolor{red}{.996}; \textcolor{red}{.996}&\textcolor{red}{.000}; \textcolor{red}{.000};
\textcolor{red}{.000}; \textcolor{red}{.000}; \textcolor{red}{.000}&\textcolor{red}{.706};
\textcolor{red}{.711}; \textcolor{red}{.634}; \textcolor{red}{.741};
\textcolor{red}{.706}&\textcolor{red}{.000}; \textcolor{red}{.000}; \textcolor{red}{.000};
\textcolor{red}{.000}; \textcolor{red}{.000}&\textcolor{red}{.171}; \textcolor{red}{.219};
\textcolor{red}{.250}; \textcolor{red}{.249}; \textcolor{red}{.209}&\textcolor{red}{.000};
\textcolor{red}{.000}; \textcolor{red}{.000}; \textcolor{red}{.000}; \textcolor{red}{.000}\\
\bottomrule
\end{tabular}
\end{table}
%
%
%
\begin{table}[p!]
\tiny
\centering
\caption{Training data difference between input and output of each layer's $\boldsymbol{\mathcal{H}}$ class-pair statistics
         for the trained models, and respective
         $p$-values for the one sided permutation test~(Eq. 17 in paper).  \textcolor{red}{Red font}
         denotes layer instances for which we reject $\mathbf{H_0}$, and black font denotes layers for
         which we fail to reject $\mathbf{H_0}$. 
         (Note: ~Euclidean distance is used as the proximity measure.  50,000
         Monte Carlo trials used to estimate the $p$-values. $\Delta{\bar{\mathcal{H}}} = \bar{\mathcal{H}}^{(t)}_{(k,T)} - \bar{\mathcal{H}}^{(t)}_{(k-1,T)}$)}

\label{table:A3Table}
\begin{tabular}{p{10mm}p{12mm}|K{16mm}|K{16mm}|K{16mm}|K{16mm}|K{16mm}|K{16mm}}
\toprule
Input Space &Output Space& \multicolumn{2}{c|}{CIFAR10 w Random}& \multicolumn{2}{c|}{CIFAR10 w
True}& \multicolumn{2}{c}{MNIST w True} \\
 & &$\Delta{\bar{\mathcal{H}}}$ & \textit{p}-values & $\Delta{\bar{\mathcal{H}}}$ & \textit{p}-values&
 $\Delta{\bar{\mathcal{H}}}$ & \textit{p}-values \\
\midrule
0.Input&1.Conv&.003; .000; -.004; -.003; -.003&.243; .474; .741; .729; .721&-.019; .011; -.025;
.001; -.007&.708; .378; .767; .492; .582&.003; .003; .004; .003; .003&.200; .199; .169; .181; .208\\
\hline
1.Conv&1.ReLU&.002; .002; -.000; -.001; .001&.312; .303; .531; .542; .421&.010; .002; .010; .013;
.008&.386; .481; .389; .356; .410&-.000; .000; .000; .000; .000&.506; .492; .481; .480; .493\\
\hline
1.ReLU&2.Conv&.006; -.012; -.008; -.007; -.003&.070; .990; .911; .896; .756&-.002; .004; -.013;
.004; -.009&.525; .455; .635; .453; .591&.000; .001; .000; .000; .001&.494; .425; .502; .468; .433\\
\hline
2.Conv&2.ReLU&-.001; .004; .002; -.003; -.003&.573; .210; .330; .740; .752&.051; .039; .062; .035;
.053&.080; .142; .052; .164; .076&.000; .000; .000; .000; .000&.481; .496; .482; .500; .494\\
\hline
2.ReLU&2.MaxPool&-.007; \textcolor{red}{.010}; -.005; -.002; .000&.955; \textcolor{red}{.015}; .827;
.682; .471&.051; .040; .058; .035; .037&.071; .130; .055; .164; .152&-.001; -.001; -.001; -.001;
-.001&.599; .609; .629; .632; .608\\
\hline
2.MaxPool&3.Conv&.005; -.003; -.001; .004; .005&.127; .775; .603; .193; .146&.065; .017;
\textcolor{red}{.077}; .015; .036&.035; .318; \textcolor{red}{.020}; .334; .158&.002; .002; .003;
.003; .002&.301; .274; .210; .235; .323\\
\hline
3.Conv&3.ReLU&-.001; -.000; -.002; -.004; -.005&.610; .503; .616; .776; .840&.014; .021; .023; .024;
.029&.346; .285; .270; .251; .210&-.000; -.001; -.001; -.000; -.000&.534; .581; .568; .543; .513\\
\hline
3.ReLU&4.Conv&-.007; .009; .005; -.005; \textcolor{red}{.011}&.947; .039; .132; .865;
\textcolor{red}{.010}&.023; -.005; .020; .008; -.004&.276; .550; .307; .412; .545&.000; .001; .001;
.000; .001&.444; .371; .425; .440; .374\\
\hline
4.Conv&4.ReLU&.006; .000; \textcolor{red}{.009}; .007; -.009&.107; .462; \textcolor{red}{.007};
.058; .975&-.043; -.036; -.020; -.126; -.038&.866; .839; .701; 1.000; .845&-.000; .000; .000; .001;
.001&.522; .476; .458; .404; .428\\
\hline
4.ReLU&4.MaxPool&.001; -.005; -.015; .009; -.004&.397; .869; .999; .041; .806&\textcolor{red}{.081};
\textcolor{red}{.119}; .057; \textcolor{red}{.165}; \textcolor{red}{.113}&\textcolor{red}{.015};
\textcolor{red}{.001}; .068; \textcolor{red}{.000}; \textcolor{red}{.002}&.004; .004; .004; .003;
.004&.051; .068; .059; .089; .085\\
\hline
4.MaxPool&5.Dense&\textcolor{red}{.049}; \textcolor{red}{.031}; \textcolor{red}{.053};
\textcolor{red}{.029}; \textcolor{red}{.040}&\textcolor{red}{.000}; \textcolor{red}{.000};
\textcolor{red}{.000}; \textcolor{red}{.000}; \textcolor{red}{.000}&.061; .067; .051;
\textcolor{red}{.103}; \textcolor{red}{.074}&.042; .028; .080; \textcolor{red}{.002};
\textcolor{red}{.020}&.002; .002; .002; .002; .002&.111; .136; .192; .165; .195\\
\hline
5.Dense&5.ReLU&\textcolor{red}{.188}; \textcolor{red}{.214}; \textcolor{red}{.202};
\textcolor{red}{.203}; \textcolor{red}{.212}&\textcolor{red}{.000}; \textcolor{red}{.000};
\textcolor{red}{.000}; \textcolor{red}{.000}; \textcolor{red}{.000}&.013; .010; .015; .009;
.015&.351; .380; .338; .396; .332&.001; .001; .002; .001; .002&.176; .190; .126; .292; .091\\
\hline
5.ReLU&6.Dense&\textcolor{red}{.758}; \textcolor{red}{.752}; \textcolor{red}{.750};
\textcolor{red}{.756}; \textcolor{red}{.753}&\textcolor{red}{.000}; \textcolor{red}{.000};
\textcolor{red}{.000}; \textcolor{red}{.000}; \textcolor{red}{.000}&\textcolor{red}{.220};
\textcolor{red}{.199}; \textcolor{red}{.148}; \textcolor{red}{.292};
\textcolor{red}{.201}&\textcolor{red}{.000}; \textcolor{red}{.000}; \textcolor{red}{.000};
\textcolor{red}{.000}; \textcolor{red}{.000}&\textcolor{red}{.005}; \textcolor{red}{.004};
\textcolor{red}{.004}; \textcolor{red}{.005}; \textcolor{red}{.004}&\textcolor{red}{.000};
\textcolor{red}{.000}; \textcolor{red}{.000}; \textcolor{red}{.000}; \textcolor{red}{.000}\\
\hline
6.Dense&6.Softmax&-.000; -.000; .000; .000; -.000&.985; .581; .250; .145; .858&.019; .019; .015;
\textcolor{red}{.010}; .020&.049; .124; .221; \textcolor{red}{.000}; .070&\textcolor{red}{.001};
\textcolor{red}{.001}; \textcolor{red}{.001}; .001; \textcolor{red}{.001}&\textcolor{red}{.001};
\textcolor{red}{.000}; \textcolor{red}{.000}; .028; \textcolor{red}{.000}\\
\bottomrule
\end{tabular}
\end{table}
%
\begin{table}[p!]
\tiny
\centering
\caption{Training data difference between between multi-layer component input and output $\boldsymbol{\mathcal{H}}$ class-pair statistics
         for the trained models, and respective
         $p$-values for the one sided permutation test~(Eq. 17 in paper).  \textcolor{red}{Red font}
         denotes layer instances for which we reject $\mathbf{H_0}$, and black font denotes layers for
         which we fail to reject $\mathbf{H_0}$. 
         (Note: ~Euclidean distance is used as the proximity measure.  50,000
         Monte Carlo trials used to estimate the $p$-values. $\Delta{\bar{\mathcal{H}}} = \bar{\mathcal{H}}^{(t)}_{(k_2,T)} - \bar{\mathcal{H}}^{(t)}_{(k_1,T)}$)}

\label{table:A4Table}
\begin{tabular}{p{10mm}p{12mm}|K{16mm}|K{16mm}|K{16mm}|K{16mm}|K{16mm}|K{16mm}}
\toprule
Input Space &Output Space& \multicolumn{2}{c|}{CIFAR10 w Random}& \multicolumn{2}{c|}{CIFAR10 w
True}& \multicolumn{2}{c}{MNIST w True} \\
 & &$\Delta{\bar{\mathcal{H}}}$ & \textit{p}-values & $\Delta{\bar{\mathcal{H}}}$ & \textit{p}-values&
 $\Delta{\bar{\mathcal{H}}}$ & \textit{p}-values \\
\midrule
0.Input&1.ReLU&.005; .003; -.004; -.004; -.002&.120; .283; .765; .757; .649&-.008; .013; -.015;
.013; .001&.598; .353; .668; .341; .492&.003; .003; .004; .004; .003&.203; .195; .151; .171; .201\\
\hline
1.ReLU&2.ReLU&.005; -.007; -.005; -.010; -.006&.101; .920; .833; .973; .880&.049; .043; .049; .040;
.045&.083; .114; .090; .132; .110&.000; .001; .000; .000; .001&.476; .424; .474; .467; .421\\
\hline
2.ReLU&2.MaxPool&-.007; \textcolor{red}{.010}; -.005; -.002; .000&.956; \textcolor{red}{.014}; .828;
.688; .471&.051; .040; .058; .035; .037&.072; .131; .057; .167; .152&-.001; -.001; -.001; -.001;
-.001&.606; .611; .631; .633; .608\\
\hline
2.MaxPool&3.ReLU&.004; -.003; -.003; .001; .000&.204; .781; .715; .443; .473&\textcolor{red}{.079};
.038; \textcolor{red}{.100}; .039; .065&\textcolor{red}{.015}; .142; \textcolor{red}{.005}; .133;
.037&.002; .001; .002; .002; .002&.330; .344; .265; .262; .338\\
\hline
3.ReLU&4.ReLU&-.001; .009; \textcolor{red}{.014}; .002; .001&.585; .027; \textcolor{red}{.002};
.308; .389&-.020; -.041; -.001; -.118; -.042&.703; .871; .506; 1.000; .876&.000; .001; .001; .001;
.002&.461; .358; .384; .347; .315\\
\hline
4.ReLU&4.MaxPool&.001; -.005; -.015; .009; -.004&.396; .872; .999; .041; .806&\textcolor{red}{.081};
\textcolor{red}{.119}; .057; \textcolor{red}{.165}; \textcolor{red}{.113}&\textcolor{red}{.015};
\textcolor{red}{.001}; .068; \textcolor{red}{.000}; \textcolor{red}{.001}&.004; .004; .004; .003;
.004&.051; .067; .058; .089; .087\\
\hline
4.MaxPool&5.ReLU&\textcolor{red}{.238}; \textcolor{red}{.245}; \textcolor{red}{.255};
\textcolor{red}{.232}; \textcolor{red}{.252}&\textcolor{red}{.000}; \textcolor{red}{.000};
\textcolor{red}{.000}; \textcolor{red}{.000}; \textcolor{red}{.000}&\textcolor{red}{.075};
\textcolor{red}{.078}; .065; \textcolor{red}{.112}; \textcolor{red}{.089}&\textcolor{red}{.018};
\textcolor{red}{.014}; .035; \textcolor{red}{.000}; \textcolor{red}{.005}&\textcolor{red}{.004};
.003; \textcolor{red}{.003}; .003; \textcolor{red}{.004}&\textcolor{red}{.020}; .030;
\textcolor{red}{.024}; .067; \textcolor{red}{.017}\\
\hline
5.ReLU&6.Softmax&\textcolor{red}{.758}; \textcolor{red}{.752}; \textcolor{red}{.750};
\textcolor{red}{.757}; \textcolor{red}{.753}&\textcolor{red}{.000}; \textcolor{red}{.000};
\textcolor{red}{.000}; \textcolor{red}{.000}; \textcolor{red}{.000}&\textcolor{red}{.238};
\textcolor{red}{.218}; \textcolor{red}{.163}; \textcolor{red}{.303};
\textcolor{red}{.221}&\textcolor{red}{.000}; \textcolor{red}{.000}; \textcolor{red}{.000};
\textcolor{red}{.000}; \textcolor{red}{.000}&\textcolor{red}{.006}; \textcolor{red}{.006};
\textcolor{red}{.005}; \textcolor{red}{.006}; \textcolor{red}{.005}&\textcolor{red}{.000};
\textcolor{red}{.000}; \textcolor{red}{.000}; \textcolor{red}{.000}; \textcolor{red}{.000}\\
\bottomrule
\end{tabular}
\end{table}
%
\begin{table}[p!]
\tiny
\centering
\caption{Training data difference between between multi-layer component input and output $\boldsymbol{\mathcal{H}}$ class-pair statistics
         for the trained models, and respective
         $p$-values for the one sided permutation test~(Eq.~17 in paper).  \textcolor{red}{Red font}
         denotes layer instances for which we reject $\mathbf{H_0}$, and black font denotes layers for
         which we fail to reject $\mathbf{H_0}$. 
         (Note: ~Euclidean distance is used as the proximity measure.  50,000
         Monte Carlo trials used to estimate the $p$-values. $\Delta{\bar{\mathcal{H}}} = \bar{\mathcal{H}}^{(t)}_{(k_2,T)} - \bar{\mathcal{H}}^{(t)}_{(k_1,T)}$)}
\label{table:A5Table}
\begin{tabular}{p{10mm}p{12mm}|K{16mm}|K{16mm}|K{16mm}|K{16mm}|K{16mm}|K{16mm}}
\toprule
Input Space &Output Space& \multicolumn{2}{c|}{CIFAR10 w Random}& \multicolumn{2}{c|}{CIFAR10 w
True}& \multicolumn{2}{c}{MNIST w True} \\
 & &$\Delta{\bar{\mathcal{H}}}$ & \textit{p}-values & $\Delta{\bar{\mathcal{H}}}$ & \textit{p}-values&
 $\Delta{\bar{\mathcal{H}}}$ & \textit{p}-values \\
\midrule
0.Input&2.MaxPool&.003; .006; -.014; -.016; -.008&.275; .078; .998; .998;
.949&\textcolor{red}{.093}; \textcolor{red}{.095}; \textcolor{red}{.093}; \textcolor{red}{.087};
\textcolor{red}{.083}&\textcolor{red}{.003}; \textcolor{red}{.003}; \textcolor{red}{.003};
\textcolor{red}{.006}; \textcolor{red}{.008}&.002; .003; .003; .003; .003&.271; .227; .240; .252;
.231\\
\hline
2.MaxPool&4.MaxPool&.004; .001; -.003; \textcolor{red}{.012}; -.002&.213; .402; .760;
\textcolor{red}{.011}; .699&\textcolor{red}{.141}; \textcolor{red}{.116}; \textcolor{red}{.157};
\textcolor{red}{.085}; \textcolor{red}{.136}&\textcolor{red}{.000}; \textcolor{red}{.000};
\textcolor{red}{.000}; \textcolor{red}{.007}; \textcolor{red}{.000}&\textcolor{red}{.006};
\textcolor{red}{.006}; \textcolor{red}{.007}; \textcolor{red}{.007};
\textcolor{red}{.007}&\textcolor{red}{.021}; \textcolor{red}{.016}; \textcolor{red}{.011};
\textcolor{red}{.014}; \textcolor{red}{.017}\\
\hline
4.MaxPool&5.ReLU&\textcolor{red}{.238}; \textcolor{red}{.245}; \textcolor{red}{.255};
\textcolor{red}{.232}; \textcolor{red}{.252}&\textcolor{red}{.000}; \textcolor{red}{.000};
\textcolor{red}{.000}; \textcolor{red}{.000}; \textcolor{red}{.000}&\textcolor{red}{.075};
\textcolor{red}{.078}; .065; \textcolor{red}{.112}; \textcolor{red}{.089}&\textcolor{red}{.018};
\textcolor{red}{.013}; .037; \textcolor{red}{.000}; \textcolor{red}{.005}&\textcolor{red}{.004};
.003; \textcolor{red}{.003}; .003; \textcolor{red}{.004}&\textcolor{red}{.020}; .029;
\textcolor{red}{.025}; .066; \textcolor{red}{.018}\\
\hline
5.ReLU&6.Softmax&\textcolor{red}{.758}; \textcolor{red}{.752}; \textcolor{red}{.750};
\textcolor{red}{.757}; \textcolor{red}{.753}&\textcolor{red}{.000}; \textcolor{red}{.000};
\textcolor{red}{.000}; \textcolor{red}{.000}; \textcolor{red}{.000}&\textcolor{red}{.238};
\textcolor{red}{.218}; \textcolor{red}{.163}; \textcolor{red}{.303};
\textcolor{red}{.221}&\textcolor{red}{.000}; \textcolor{red}{.000}; \textcolor{red}{.000};
\textcolor{red}{.000}; \textcolor{red}{.000}&\textcolor{red}{.006}; \textcolor{red}{.006};
\textcolor{red}{.005}; \textcolor{red}{.006}; \textcolor{red}{.005}&\textcolor{red}{.000};
\textcolor{red}{.000}; \textcolor{red}{.000}; \textcolor{red}{.000}; \textcolor{red}{.000}\\
\bottomrule
\end{tabular}
\end{table}
\begin{table}[p!]
\tiny
\centering
\caption{Validation data differences in mean of $\boldsymbol{\mathcal{H}}$ class-pair  statistics between the
         input and output representations of a layer, and respective one-sided
         permutation test~$p$-values.  \textcolor{red}{Red font} denotes layer
         instances for which we reject $\mathbf{H_0}$, and black font denotes layers for which we fail
         to reject $\mathbf{H_0}$ ~(Eq.~18 in paper). 
          (Note: Note: ~Euclidean distance is used as the proximity measure.  50,000
         Monte Carlo trials used to estimate the $p$-values. 
         $\Delta{\bar{\mathcal{H}}} = \bar{\mathcal{H}}^{(v)}_{(k,T)} - \bar{\mathcal{H}}^{(v)}_{(k-1,T)}$)}
\label{table:A6Table}
\begin{tabular}{p{10mm}p{12mm}|K{16mm}|K{16mm}|K{16mm}|K{16mm}|K{16mm}|K{16mm}}
\toprule
Input Space &Output Space& \multicolumn{2}{c|}{CIFAR10 w Random}& \multicolumn{2}{c|}{CIFAR10 w
True}& \multicolumn{2}{c}{MNIST w True} \\
 & &$\Delta{\bar{\mathcal{H}}}$ & \textit{p}-values & $\Delta{\bar{\mathcal{H}}}$ & \textit{p}-values&
 $\Delta{\bar{\mathcal{H}}}$ & \textit{p}-values \\
\midrule
0.Input&1.Conv&-.002; .006; -.003; -.001; -.004&.639; .035; .728; .543; .822&-.014; .017; -.021;
.005; -.003&.662; .308; .729; .441; .533&.004; .004; .004; .004; .004&.094; .113; .137; .119; .103\\
\hline
1.Conv&1.ReLU&.002; .000; -.002; .001; .001&.348; .488; .676; .418; .412&.009; .001; .010; .012;
.008&.402; .490; .395; .372; .415&.000; .000; .000; .000; -.000&.482; .488; .481; .466; .509\\
\hline
1.ReLU&2.Conv&.000; -.003; -.001; -.004; -.003&.499; .729; .583; .752; .789&-.006; -.003; -.019;
-.004; -.006&.569; .531; .700; .544; .563&.000; .001; .001; .000; .001&.470; .429; .412; .449;
.424\\
\hline
2.Conv&2.ReLU&.002; .005; .000; .004; -.001&.314; .138; .479; .251; .585&.054; .038; .064; .038;
.046&.065; .136; .043; .138; .095&-.000; .000; .000; -.000; .000&.516; .498; .504; .510; .496\\
\hline
2.ReLU&2.MaxPool&.010; -.006; -.012; .002; -.001&.030; .919; .994; .323; .608&.041; .029; .053;
.030; .031&.113; .198; .068; .195; .186&-.000; -.000; -.000; -.000; -.001&.517; .516; .536; .533;
.568\\
\hline
2.MaxPool&3.Conv&-.014; .008; \textcolor{red}{.012}; -.007; .004&.996; .042; \textcolor{red}{.016};
.947; .150&.067; .027; \textcolor{red}{.073}; .019; .040&.028; .219; \textcolor{red}{.024}; .290;
.122&.001; .001; .002; .001; .002&.332; .353; .284; .307; .281\\
\hline
3.Conv&3.ReLU&.005; -.002; .003; .006; .000&.157; .622; .301; .093; .495&.014; .021; .021; .022;
.022&.346; .277; .286; .266; .260&-.000; -.000; -.001; -.000; -.001&.570; .542; .582; .551; .581\\
\hline
3.ReLU&4.Conv&-.005; -.005; -.009; -.002; -.002&.888; .858; .970; .689; .723&.022; -.006; .019;
.002; -.002&.274; .570; .305; .476; .522&.001; .001; .001; .001; .001&.400; .409; .408; .374; .390\\
\hline
4.Conv&4.ReLU&.001; -.003; \textcolor{red}{.008}; .004; -.002&.405; .705; \textcolor{red}{.023};
.180; .640&-.057; -.045; -.026; -.120; -.053&.941; .906; .750; 1.000; .941&-.000; .000; .000; .000;
.000&.523; .480; .455; .476; .434\\
\hline
4.ReLU&4.MaxPool&.007; .007; .005; .002; \textcolor{red}{.013}&.077; .050; .138; .348;
\textcolor{red}{.002}&\textcolor{red}{.089}; \textcolor{red}{.115}; .061; \textcolor{red}{.154};
\textcolor{red}{.119}&\textcolor{red}{.007}; \textcolor{red}{.000}; .050; \textcolor{red}{.000};
\textcolor{red}{.000}&.003; .004; .004; .003; .003&.086; .042; .050; .077; .084\\
\hline
4.MaxPool&5.Dense&-.015; .003; -.009; -.010; -.008&.999; .237; .952; .973; .977&.049; .054; .035;
\textcolor{red}{.084}; .058&.076; .056; .162; \textcolor{red}{.008}; .046&.002; .001; .001; .001;
.002&.163; .225; .234; .217; .171\\
\hline
5.Dense&5.ReLU&\textcolor{red}{.010}; .000; -.002; .005; -.007&\textcolor{red}{.018}; .456; .693;
.134; .944&-.027; -.016; .000; -.054; -.014&.782; .677; .500; .949; .662&.001; .001; .001; .001;
.001&.288; .274; .168; .251; .147\\
\hline
5.ReLU&6.Dense&-.010; -.016; .006; -.011; .007&.987; 1.000; .071; .976; .089&\textcolor{red}{.086};
\textcolor{red}{.084}; .058; \textcolor{red}{.118}; \textcolor{red}{.076}&\textcolor{red}{.006};
\textcolor{red}{.008}; .043; \textcolor{red}{.000}; \textcolor{red}{.012}&\textcolor{red}{.003};
.002; .002; \textcolor{red}{.002}; .001&\textcolor{red}{.008}; .041; .081; \textcolor{red}{.018};
.103\\
\hline
6.Dense&6.Softmax&.001; \textcolor{red}{.017}; \textcolor{red}{.010}; .007; -.006&.407;
\textcolor{red}{.000}; \textcolor{red}{.011}; .096; .880&-.009; .000; -.004; -.026; -.016&.611;
.496; .557; .783; .692&-.003; -.003; -.004; -.005; -.005&.999; .999; 1.000; 1.000; 1.000\\
\bottomrule
\end{tabular}
\end{table}
\begin{table}[p!]
\tiny
\centering
\caption{Validation data differences in mean of $\boldsymbol{\mathcal{H}}$ class-pair  statistics between the
         input and output representations of multilayer layer compnents, and respective one-sided
         permutation test~$p$-values.  \textcolor{red}{Red font} denotes layer
         instances for which we reject $\mathbf{H_0}$, and black font denotes layers for which we fail
         to reject $\mathbf{H_0}$ ~(Eq.~18 in paper). 
          (Note: Note: ~Euclidean distance is used as the proximity measure.  50,000
         Monte Carlo trials used to estimate the $p$-values. 
         $\Delta{\bar{\mathcal{H}}} = \bar{\mathcal{H}}^{(v)}_{(k_2,T)} - \bar{\mathcal{H}}^{(v)}_{(k_1,T)}$)}
\label{table:A7Table}
\begin{tabular}{p{10mm}p{12mm}|K{16mm}|K{16mm}|K{16mm}|K{16mm}|K{16mm}|K{16mm}}
\toprule
Input Space &Output Space& \multicolumn{2}{c|}{CIFAR10 w Random}& \multicolumn{2}{c|}{CIFAR10 w
True}& \multicolumn{2}{c}{MNIST w True} \\
 & &$\Delta{\bar{\mathcal{H}}}$ & \textit{p}-values & $\Delta{\bar{\mathcal{H}}}$ & \textit{p}-values&
 $\Delta{\bar{\mathcal{H}}}$ & \textit{p}-values \\
\midrule
0.Input&1.ReLU&.000; .006; -.005; .000; -.003&.482; .041; .850; .466; .776&-.005; .017; -.011; .017;
.005&.561; .297; .631; .307; .440&.005; .004; .004; .004; .004&.086; .108; .121; .098; .104\\
\hline
1.ReLU&2.ReLU&.002; .002; -.001; .000; -.004&.316; .274; .558; .501; .846&.048; .035; .044; .035;
.040&.086; .150; .110; .159; .126&.000; .001; .001; .000; .001&.484; .426; .405; .458; .420\\
\hline
2.ReLU&2.MaxPool&.010; -.006; -.012; .002; -.001&.030; .917; .992; .325; .607&.041; .029; .053;
.030; .031&.114; .200; .070; .193; .186&-.000; -.000; -.000; -.000; -.001&.515; .525; .533; .531;
.567\\
\hline
2.MaxPool&3.ReLU&-.009; .007; \textcolor{red}{.015}; -.001; .005&.960; .063; \textcolor{red}{.003};
.583; .135&\textcolor{red}{.081}; .047; \textcolor{red}{.094}; .041; .062&\textcolor{red}{.011};
.087; \textcolor{red}{.006}; .118; .037&.001; .001; .001; .001; .001&.403; .386; .349; .346; .351\\
\hline
3.ReLU&4.ReLU&-.004; -.008; -.002; .002; -.004&.827; .938; .654; .345; .835&-.036; -.051; -.006;
-.118; -.055&.840; .932; .567; 1.000; .946&.001; .001; .001; .001; .001&.422; .393; .365; .354;
.335\\
\hline
4.ReLU&4.MaxPool&.007; .007; .005; .002; \textcolor{red}{.013}&.078; .051; .138; .350;
\textcolor{red}{.002}&\textcolor{red}{.089}; \textcolor{red}{.115}; .061; \textcolor{red}{.154};
\textcolor{red}{.119}&\textcolor{red}{.007}; \textcolor{red}{.001}; .049; \textcolor{red}{.000};
\textcolor{red}{.000}&.003; .004; .004; .003; .003&.089; .044; .049; .076; .084\\
\hline
4.MaxPool&5.ReLU&-.005; .004; -.012; -.004; -.015&.831; .175; .992; .806; 1.000&.022; .038; .035;
.029; .044&.261; .127; .163; .188; .094&.003; .002; .002; .002; .003&.066; .087; .050; .083; .028\\
\hline
5.ReLU&6.Softmax&-.009; .001; \textcolor{red}{.017}; -.004; .001&.979; .425; \textcolor{red}{.000};
.820; .415&\textcolor{red}{.077}; \textcolor{red}{.084}; .054; \textcolor{red}{.093};
.060&\textcolor{red}{.011}; \textcolor{red}{.005}; .048; \textcolor{red}{.002}; .029&-.000; -.001;
-.002; -.003; -.003&.662; .860; .972; .987; .999\\
\bottomrule
\end{tabular}
\end{table}
\begin{table}[p!]
\tiny
\centering
\caption{Validation data differences in mean of $\boldsymbol{\mathcal{H}}$ class-pair  statistics between the
         input and output representations of multilayer layer compnents, and respective one-sided
         permutation test~$p$-values.  \textcolor{red}{Red font} denotes layer
         instances for which we reject $\mathbf{H_0}$, and black font denotes layers for which we fail
         to reject $\mathbf{H_0}$ ~(Eq.~18 in paper). 
          (Note: ~Euclidean distance is used as the proximity measure.  50,000
         Monte Carlo trials used to estimate the $p$-values. 
         $\Delta{\bar{\mathcal{H}}} = \bar{\mathcal{H}}^{(v)}_{(k_2,T)} - \bar{\mathcal{H}}^{(v)}_{(k_1,T)}$)}
\label{table:A8Table}
\begin{tabular}{p{10mm}p{12mm}|K{16mm}|K{16mm}|K{16mm}|K{16mm}|K{16mm}|K{16mm}}
\toprule
Input Space &Output Space& \multicolumn{2}{c|}{CIFAR10 w Random}& \multicolumn{2}{c|}{CIFAR10 w
True}& \multicolumn{2}{c}{MNIST w True} \\
 & &$\Delta{\bar{\mathcal{H}}}$ & \textit{p}-values & $\Delta{\bar{\mathcal{H}}}$ & \textit{p}-values&
 $\Delta{\bar{\mathcal{H}}}$ & \textit{p}-values \\
\midrule
0.Input&2.MaxPool&\textcolor{red}{.012}; .003; -.018; .003; -.009&\textcolor{red}{.007}; .203;
1.000; .284; .969&\textcolor{red}{.084}; \textcolor{red}{.082}; \textcolor{red}{.086};
\textcolor{red}{.082}; \textcolor{red}{.076}&\textcolor{red}{.007}; \textcolor{red}{.009};
\textcolor{red}{.007}; \textcolor{red}{.009}; \textcolor{red}{.012}&.005; .005; .004; .004;
.004&.087; .085; .100; .098; .102\\
\hline
2.MaxPool&4.MaxPool&-.006; .006; \textcolor{red}{.018}; .003; \textcolor{red}{.014}&.879; .045;
\textcolor{red}{.000}; .288; \textcolor{red}{.001}&\textcolor{red}{.135}; \textcolor{red}{.111};
\textcolor{red}{.149}; \textcolor{red}{.077}; \textcolor{red}{.126}&\textcolor{red}{.000};
\textcolor{red}{.001}; \textcolor{red}{.000}; \textcolor{red}{.013}; \textcolor{red}{.000}&.005;
\textcolor{red}{.005}; \textcolor{red}{.006}; \textcolor{red}{.005}; \textcolor{red}{.006}&.032;
\textcolor{red}{.016}; \textcolor{red}{.011}; \textcolor{red}{.018}; \textcolor{red}{.014}\\
\hline
4.MaxPool&5.ReLU&-.005; .004; -.012; -.004; -.015&.829; .172; .992; .806; 1.000&.022; .038; .035;
.029; .044&.260; .122; .163; .187; .094&.003; .002; .002; .002; .003&.063; .087; .050; .081; .029\\
\hline
5.ReLU&6.Softmax&-.009; .001; \textcolor{red}{.017}; -.004; .001&.979; .426; \textcolor{red}{.000};
.823; .413&\textcolor{red}{.077}; \textcolor{red}{.084}; .054; \textcolor{red}{.093};
.060&\textcolor{red}{.011}; \textcolor{red}{.004}; .049; \textcolor{red}{.002}; .031&-.000; -.001;
-.002; -.003; -.003&.667; .861; .972; .988; .999\\
\bottomrule
\end{tabular}
\end{table}
\begin{table}[p!]
\tiny
\centering
\caption{Two-sided permutation test~(Eq.~20 in paper) comparing the differences in the mean change induced on the
         training and validation statistics ( $\boldsymbol{\Delta{\mathcal{H}}}^{(t)}_{(k,k-1)}$ and $\boldsymbol{\Delta{\mathcal{H}}}^{(v)}_{(k,k-1)}$  ). \textcolor{red}{Red font}
         denotes layer instances for which we reject $\mathbf{H_0}$, and black font denotes layers for
         which we fail to reject $\mathbf{H_0}$. 
         (Note: ~Euclidean distance is used as the proximity measure.  50,000
         Monte Carlo trials used to estimate the $p$-values. 
         $\Delta\mu = \overline{\Delta{\mathcal{H}}}^{\:(t)}_{(k, k-1)} -  \overline{\Delta{\mathcal{H}}}^{\:(v)}_{(k,k-1)} $)}
\label{table:A9Table}
\begin{tabular}{p{10mm}p{12mm}|K{16mm}|K{16mm}|K{16mm}|K{16mm}|K{16mm}|K{16mm}}
\toprule
Input Space &Output Space& \multicolumn{2}{c|}{CIFAR10 w Random}& \multicolumn{2}{c|}{CIFAR10 w
True}& \multicolumn{2}{c}{MNIST w True} \\
  & &$\Delta\mu$ & \textit{p}-values &
  $\Delta\mu$ & \textit{p}-values&
  $\Delta\mu$ & \textit{p}-values \\
\midrule
0.Input&1.Conv&.005; -.006; -.001; -.003; .001&.156; .052; .769; .332; .811&-.004; -.006; -.004;
-.004; -.005&.682; .225; .691; .457; .547&-.001; -.001; -.000; -.001; -.001&.212; .307; .852; .497;
.242\\
\hline
1.Conv&1.ReLU&-.000; .002; .002; -.002; .000&.966; .204; .406; .376; .891&.001; .001; .001; .001;
.000&.549; .569; .780; .598; .963&-.000; .000; .000; -.000; .000&.308; 1.000; 1.000; .459; .347\\
\hline
1.ReLU&2.Conv&.006; \textcolor{red}{-.009}; -.007; -.003; -.000&.250; \textcolor{red}{.020}; .192;
.467; .980&.004; .007; .006; .008; -.003&.390; .172; .184; .146; .541&-.000; .000; -.001; -.000;
.000&.735; .825; .079; .923; .936\\
\hline
2.Conv&2.ReLU&-.003; -.001; .002; \textcolor{red}{-.007}; -.002&.362; .809; .509;
\textcolor{red}{.011}; .592&-.003; .001; -.001; -.003; .007&.719; .892; .861; .575;
.248&\textcolor{red}{.000}; .000; .000; .000; .000&\textcolor{red}{.007}; .832; .350; .672; .858\\
\hline
2.ReLU&2.MaxPool&\textcolor{red}{-.017}; \textcolor{red}{.016}; .008; -.004;
.002&\textcolor{red}{.000}; \textcolor{red}{.000}; .088; .258; .701&.010; .011; .005; .004;
.007&.130; .031; .434; .350; .154&-.001; -.001; -.001; -.001; -.001&.217; .188; .219; .227; .473\\
\hline
2.MaxPool&3.Conv&\textcolor{red}{.018}; -.012; \textcolor{red}{-.013}; \textcolor{red}{.011};
.000&\textcolor{red}{.000}; .027; \textcolor{red}{.009}; \textcolor{red}{.014}; .929&-.002; -.010;
.004; -.004; -.004&.848; .049; .791; .321; .513&.001; .001; .001; .001; -.000&.421; .258; .332;
.323; .974\\
\hline
3.Conv&3.ReLU&-.006; .002; -.005; \textcolor{red}{-.010}; -.005&.095; .616; .219;
\textcolor{red}{.011}; .251&.000; .000; .002; .002; .007&.886; .970; .470; .614; .108&.000; -.000;
-.000; .000; .000&.630; .335; 1.000; 1.000; .150\\
\hline
3.ReLU&4.Conv&-.001; \textcolor{red}{.014}; \textcolor{red}{.015}; -.003;
\textcolor{red}{.013}&.805; \textcolor{red}{.005}; \textcolor{red}{.005}; .598;
\textcolor{red}{.007}&.001; .001; .000; .006; -.002&.861; .772; .951; .281; .623&-.000; .000; -.000;
-.000; .000&.672; .621; .948; .483; .717\\
\hline
4.Conv&4.ReLU&.005; .003; .001; .003; -.008&.406; .589; .783; .634; .177&.015; .009; .005; -.006;
.016&.088; .267; .403; .695; .048&.000; .000; .000; .001; .000&1.000; .985; .969; .275; .890\\
\hline
4.ReLU&4.MaxPool&-.006; \textcolor{red}{-.012}; \textcolor{red}{-.020}; .007;
\textcolor{red}{-.017}&.300; \textcolor{red}{.012}; \textcolor{red}{.000}; .247;
\textcolor{red}{.000}&-.008; .004; -.004; .011; -.006&.332; .710; .516; .453; .581&.001; .000; .000;
.000; .000&.348; .975; .789; .892; .731\\
\hline
4.MaxPool&5.Dense&\textcolor{red}{.064}; \textcolor{red}{.028}; \textcolor{red}{.062};
\textcolor{red}{.039}; \textcolor{red}{.049}&\textcolor{red}{.000}; \textcolor{red}{.000};
\textcolor{red}{.000}; \textcolor{red}{.000}; \textcolor{red}{.000}&\textcolor{red}{.012};
\textcolor{red}{.013}; \textcolor{red}{.016}; .020; \textcolor{red}{.016}&\textcolor{red}{.013};
\textcolor{red}{.020}; \textcolor{red}{.001}; .030; \textcolor{red}{.006}&.001; .001; .000; .001;
.000&.431; .210; .387; .392; .938\\
\hline
5.Dense&5.ReLU&\textcolor{red}{.179}; \textcolor{red}{.213}; \textcolor{red}{.204};
\textcolor{red}{.198}; \textcolor{red}{.218}&\textcolor{red}{.000}; \textcolor{red}{.000};
\textcolor{red}{.000}; \textcolor{red}{.000}; \textcolor{red}{.000}&\textcolor{red}{.040};
\textcolor{red}{.026}; \textcolor{red}{.015}; \textcolor{red}{.063};
\textcolor{red}{.029}&\textcolor{red}{.000}; \textcolor{red}{.000}; \textcolor{red}{.000};
\textcolor{red}{.000}; \textcolor{red}{.000}&.001; .000; .000; -.000; .001&.217; .472; .511; .856;
.239\\
\hline
5.ReLU&6.Dense&\textcolor{red}{.768}; \textcolor{red}{.768}; \textcolor{red}{.744};
\textcolor{red}{.767}; \textcolor{red}{.746}&\textcolor{red}{.000}; \textcolor{red}{.000};
\textcolor{red}{.000}; \textcolor{red}{.000}; \textcolor{red}{.000}&\textcolor{red}{.134};
\textcolor{red}{.115}; \textcolor{red}{.090}; \textcolor{red}{.174};
\textcolor{red}{.124}&\textcolor{red}{.000}; \textcolor{red}{.000}; \textcolor{red}{.000};
\textcolor{red}{.000}; \textcolor{red}{.000}&\textcolor{red}{.002}; \textcolor{red}{.002};
\textcolor{red}{.003}; \textcolor{red}{.003}; \textcolor{red}{.003}&\textcolor{red}{.014};
\textcolor{red}{.015}; \textcolor{red}{.001}; \textcolor{red}{.003}; \textcolor{red}{.000}\\
\hline
6.Dense&6.Softmax&-.001; \textcolor{red}{-.017}; \textcolor{red}{-.010}; -.007; .006&.750;
\textcolor{red}{.000}; \textcolor{red}{.025}; .223; .266&\textcolor{red}{.028};
\textcolor{red}{.019}; \textcolor{red}{.020}; \textcolor{red}{.036};
\textcolor{red}{.036}&\textcolor{red}{.000}; \textcolor{red}{.004}; \textcolor{red}{.000};
\textcolor{red}{.000}; \textcolor{red}{.000}&\textcolor{red}{.004}; \textcolor{red}{.005};
\textcolor{red}{.005}; \textcolor{red}{.006}; \textcolor{red}{.006}&\textcolor{red}{.000};
\textcolor{red}{.000}; \textcolor{red}{.000}; \textcolor{red}{.000}; \textcolor{red}{.000}\\
\bottomrule
\end{tabular}
\end{table}

\newpage
\section{Statistical analysis using cosine distance as measure of proximity}\label{app:cosine}

This supplemental material presents results for the three experiments (CIFAR10 with random labels, CIFAR10 with
true labels, and MNIST with true labels) using the cosine distance  for the proximity measure of the
HP statistics.  Note that the cosine distance  between two vectors x and y is defined as
\begin{equation}\label{eq:CosDist}
R_{cosine}(x,y) = 1 - \frac{x \cdot y}{\left\lVert x \right\rVert  \left\lVert y \right\rVert}
\end{equation}

\begin{figure}[p!]
\centering
\includegraphics[width=1.0\linewidth]{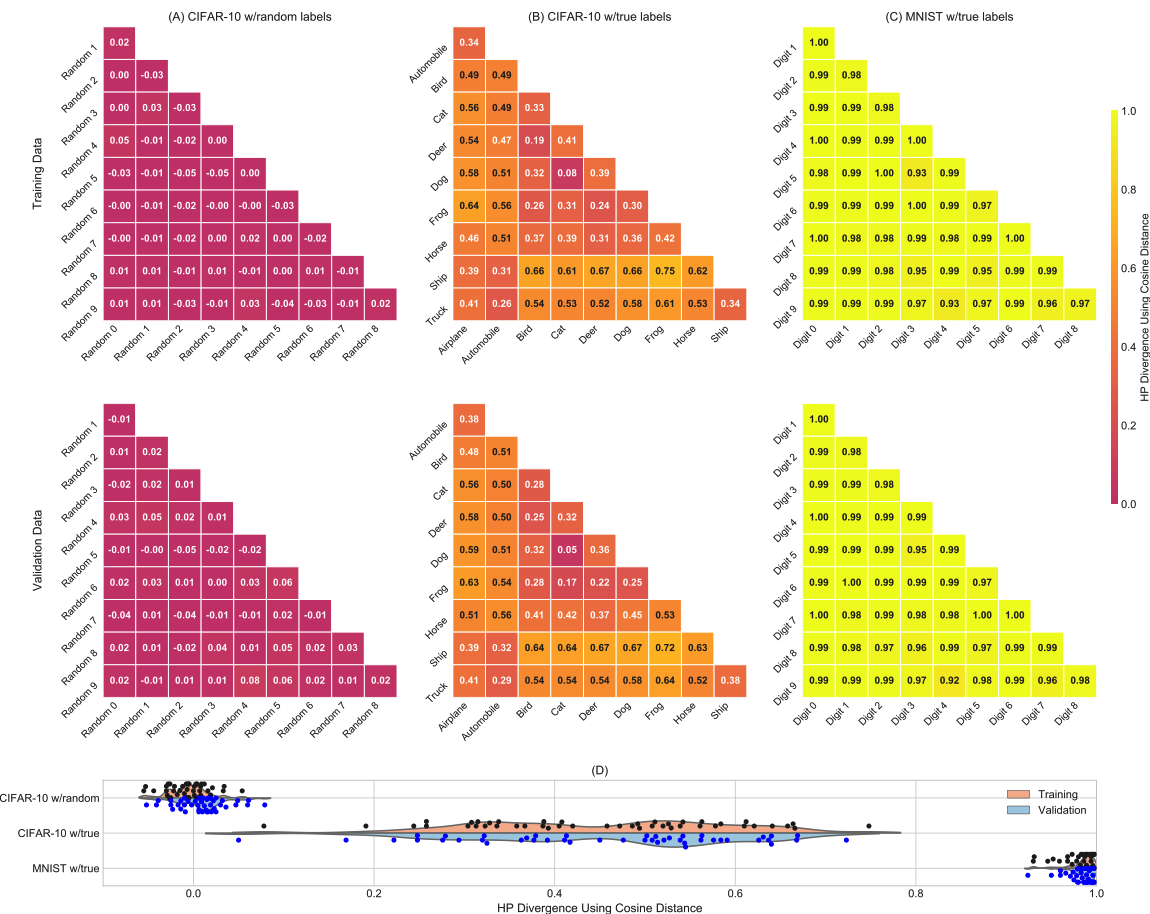}
\caption{Pairwise class HP statistics using cosine distance (training data above, validation data
below) computed on (A) CIFAR10 with random labels, (B) CIFAR10 with true labels, and (C) MNIST with
true labels. (D)
 The $\boldsymbol{\mathcal{H}}^{(t)}$ (black dots above) and $\boldsymbol{\mathcal{H}}^{(v)}$ (blue
dots, below) values and respective kernel-based density functions (orange = training, blue =
validation) for each task which illustrate that the estimated class separation for each task in
their respective ambient representations are quite distinct.}
\label{figure:B1}
\end{figure}

Figure~\ref{figure:B1} shows the between-class HP statistics of the raw images, or class
separability in the original measurement space using the cosine distance. The comparisons were made
using the analysis subset of the training data (1000 images per class) and the validation data (1000
images per class). As previously noted, there are five different versions of the randomly permuted
labels; one per instance of the training network model. The results for only one of these versions
is tabulated and plotted in Figures~\ref{figure:B1} (A) and (D), respectively.

\begin{figure}[p!]
\centering
\begin{subfigure}[(A)]{0.49\linewidth}
  \includegraphics[width=\linewidth]{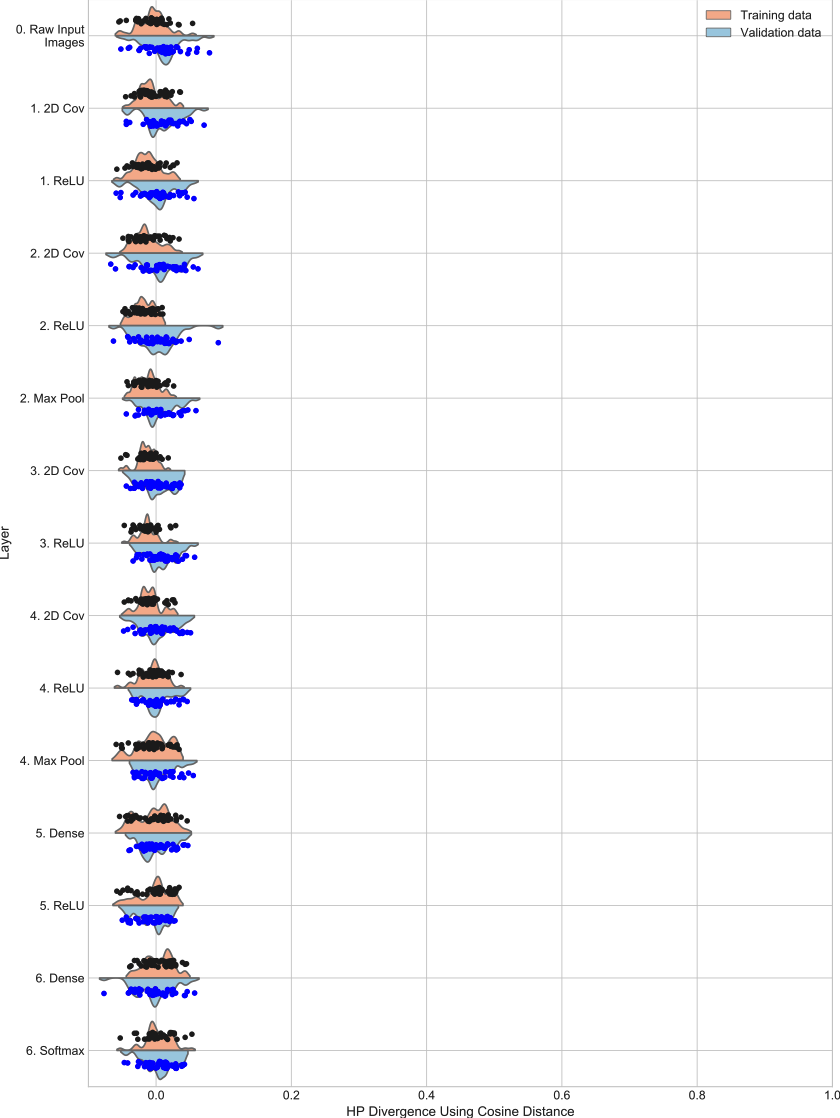}
  \caption{Untrained model}
\end{subfigure}
\begin{subfigure}[(B)]{0.49\linewidth}
  \includegraphics[width=\linewidth]{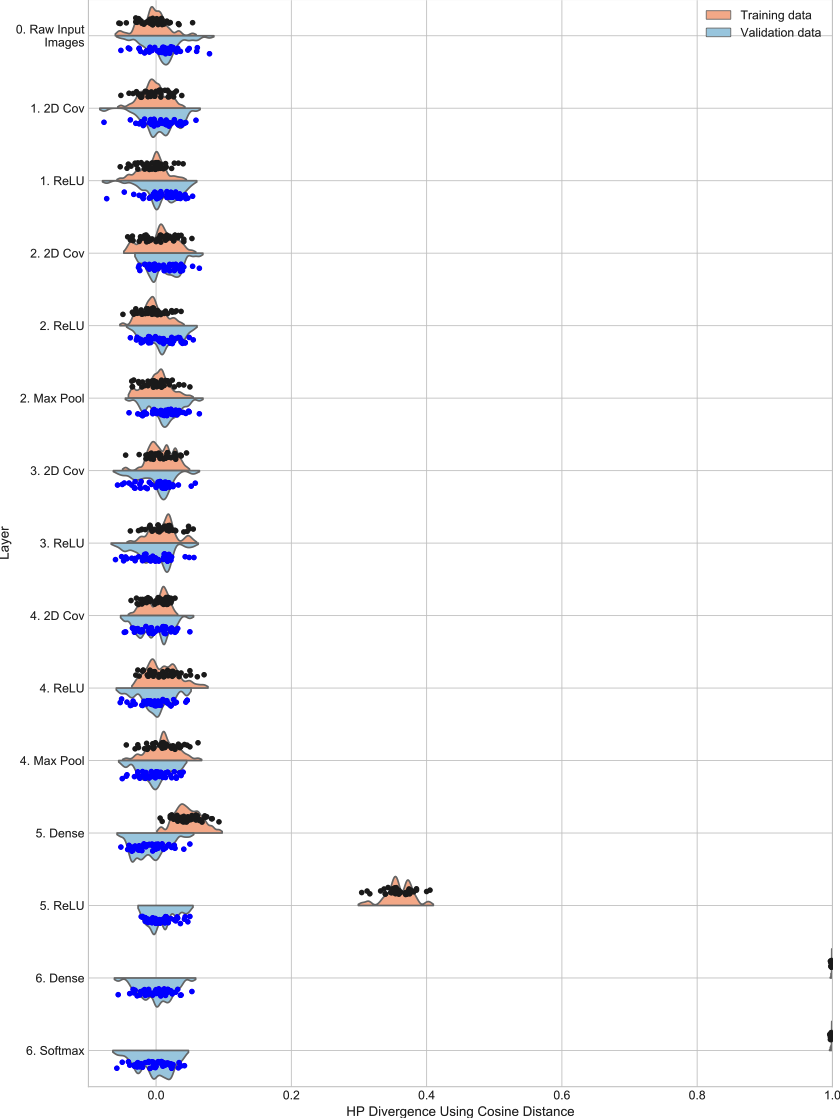}
  \caption{Trained model}
\end{subfigure}
\caption{$\mathcal{H}$ class-pair statistics at each layer for instance 1 of the model for CIFAR10 with
random class labels. (a) shows results for the data for passing through the randomly initialized
model  (epoch~0 state). (b) shows the results for the data  passing through the fully trained  model
(epoch~200 state). (Note:~Cosine distance is used as the proximity measure.)}
\label{figure:B2}
\end{figure}

\begin{figure}[p!]
\centering
\begin{subfigure}[(A)]{0.49\linewidth}
  \includegraphics[width=\linewidth]{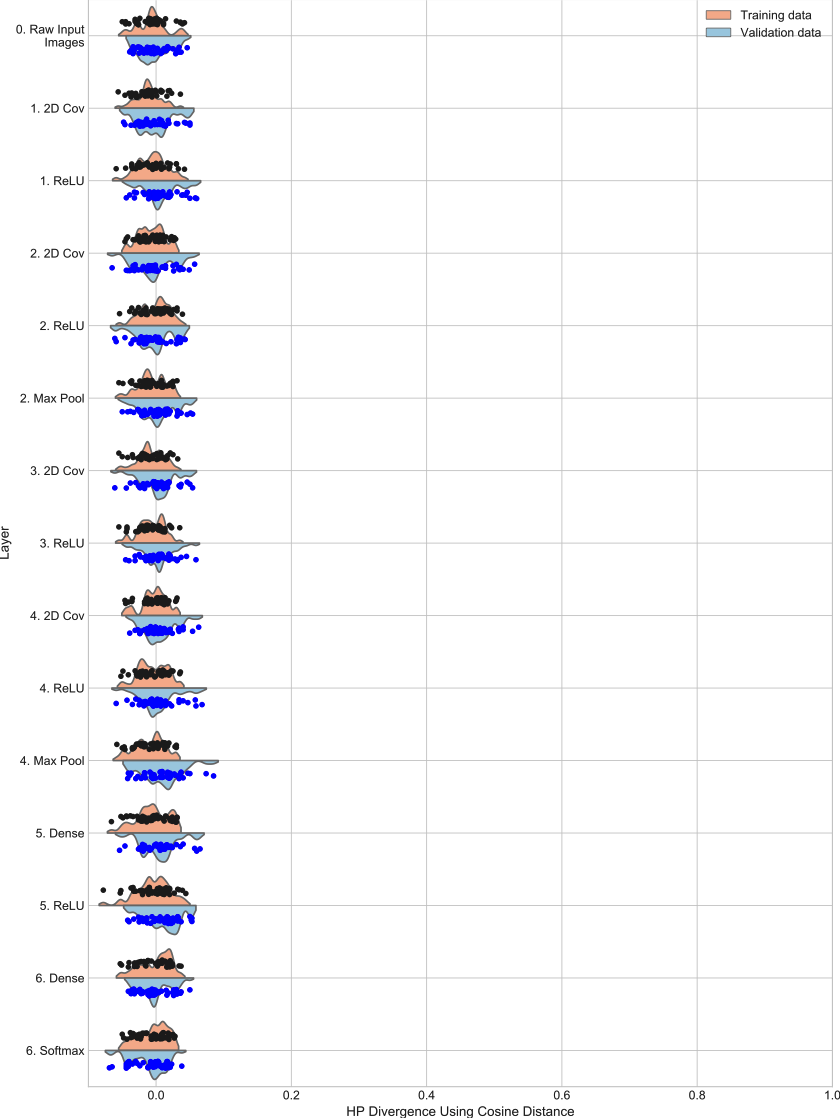}
  \caption{Untrained model}
\end{subfigure}
\begin{subfigure}[(B)]{0.49\linewidth}
  \includegraphics[width=\linewidth]{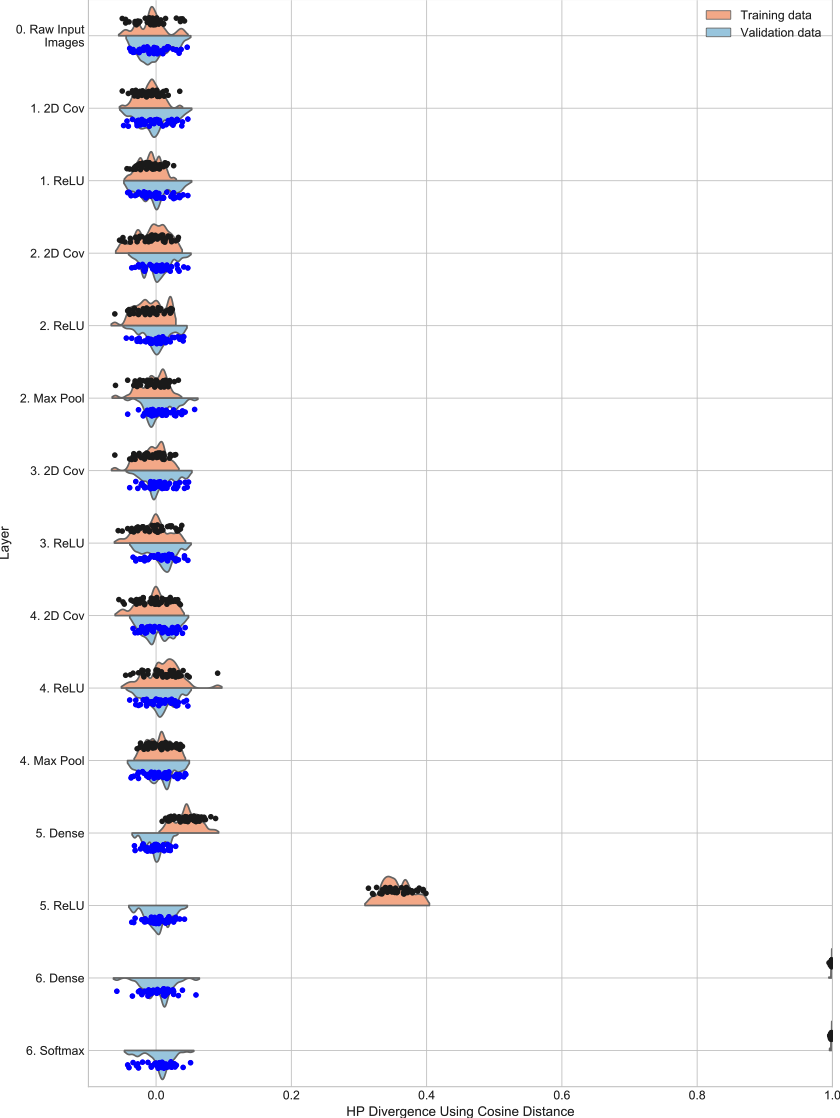}
  \caption{Trained model}
\end{subfigure}
\caption{$\mathcal{H}$ class-pair statistics at each layer for instance 2 of the model for CIFAR10 with
         random class labels. (a) shows results for the data for passing through the randomly
         initialized model  (epoch~0 state).  (b) shows the results for the data  passing through
         the fully trained  model  (epoch~200 state). (Note:~Cosine distance is used as the
         proximity measure.)}
\label{figure:B3}
\end{figure}

\begin{figure}[p!]
\centering
\begin{subfigure}[(A)]{0.49\linewidth}
  \includegraphics[width=\linewidth]{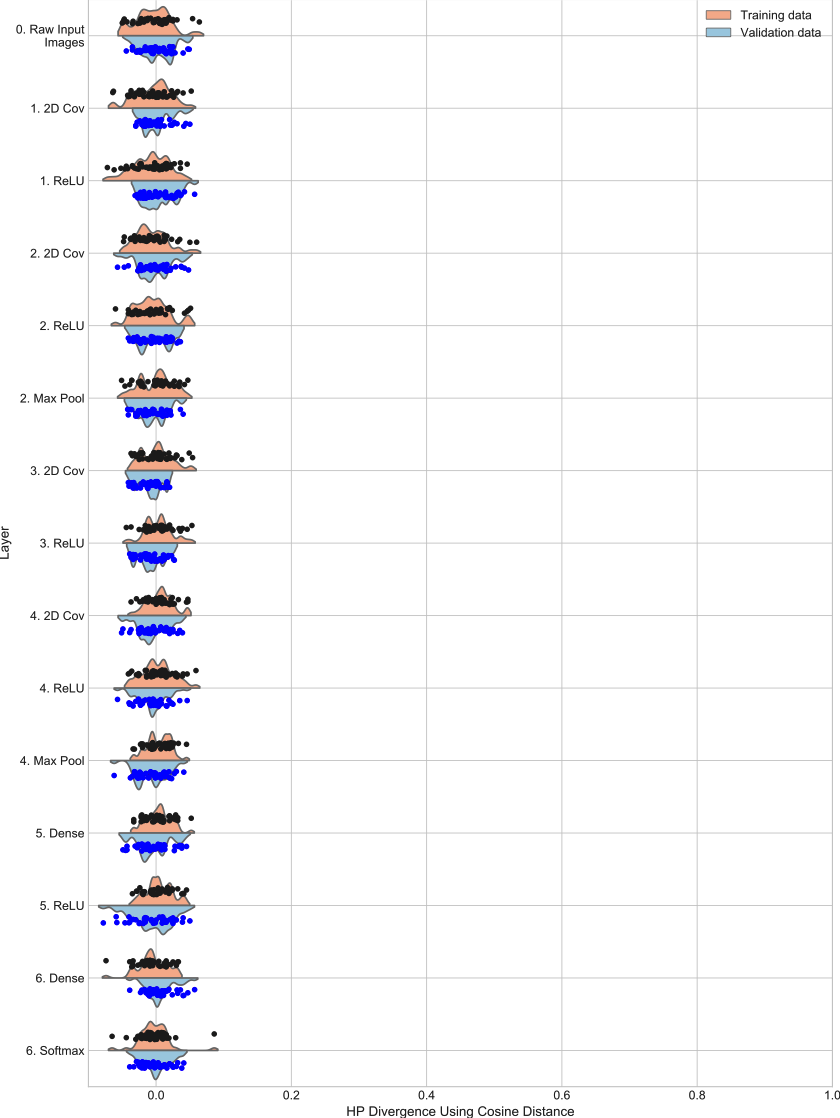}
  \caption{Untrained model}
\end{subfigure}
\begin{subfigure}[(B)]{0.49\linewidth}
  \includegraphics[width=\linewidth]{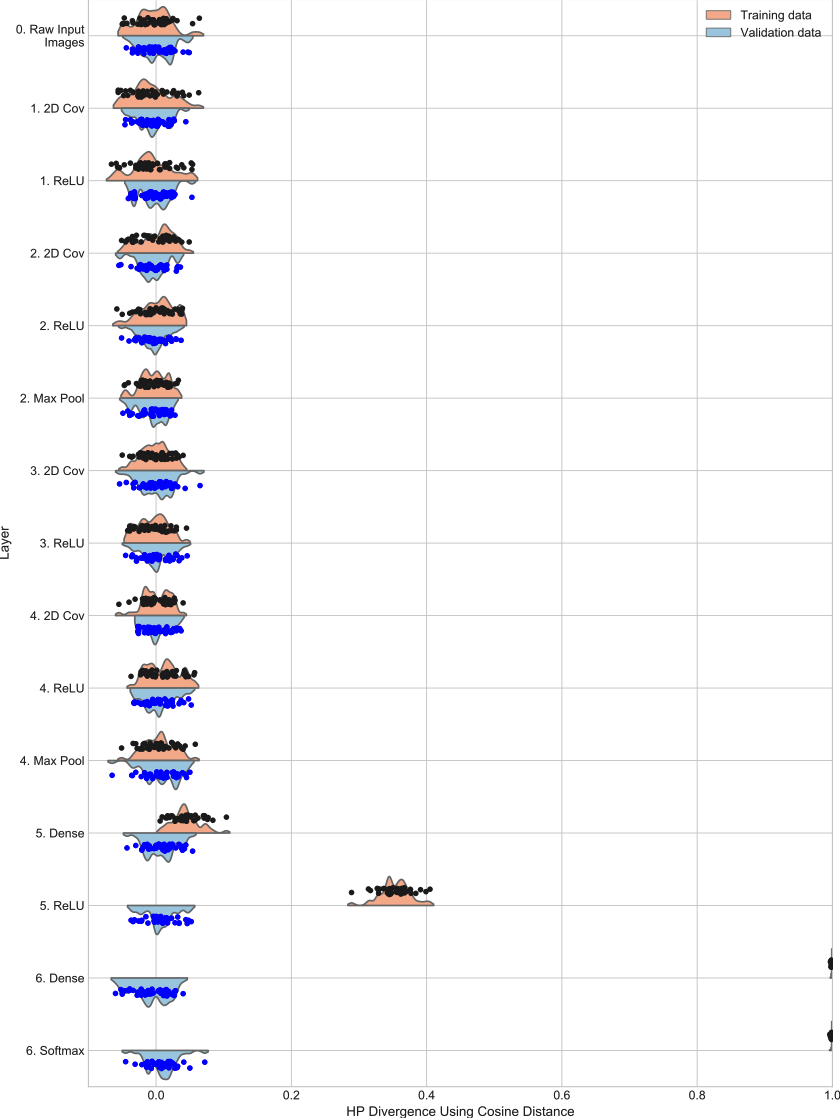}
  \caption{Trained model}
\end{subfigure}
\caption{$\mathcal{H}$ class-pair statistics at each layer for instance 3 of the model for CIFAR10 with
         random class labels. (a) shows results for the data for passing through the randomly
         initialized model  (epoch~0 state).  (b) shows the results for the data  passing through
         the fully trained  model  (epoch~200 state). (Note:~Cosine distance is used as the
         proximity measure.)}
\label{figure:B4}
\end{figure}

\begin{figure}[p!]
\centering
\begin{subfigure}[(A)]{0.49\linewidth}
  \includegraphics[width=\linewidth]{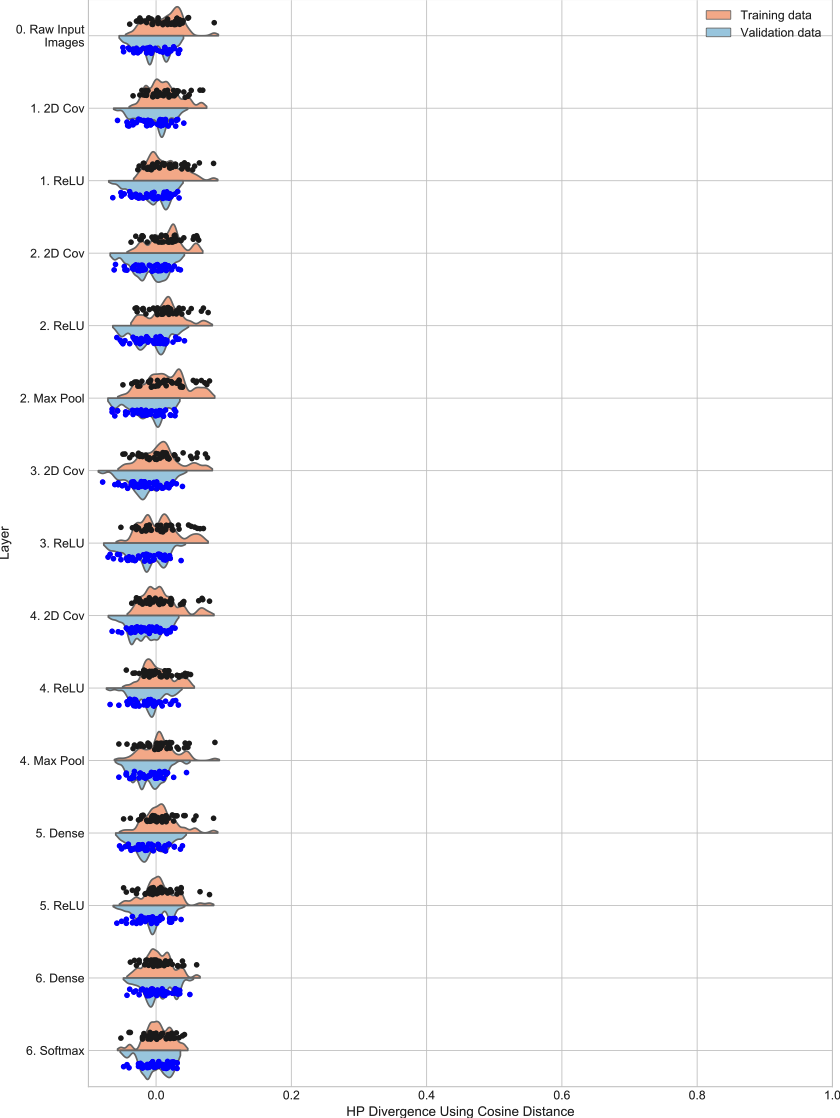}
  \caption{Untrained model}
\end{subfigure}
\begin{subfigure}[(B)]{0.49\linewidth}
  \includegraphics[width=\linewidth]{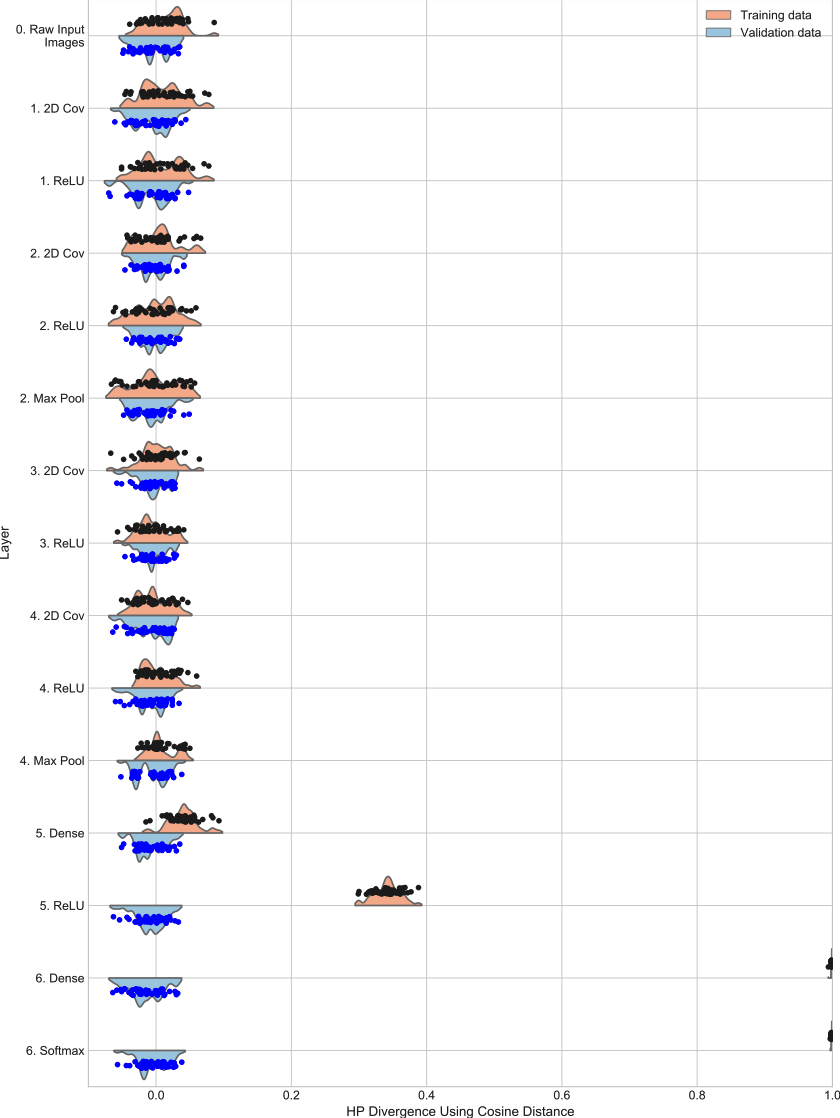}
  \caption{Trained model}
\end{subfigure}
\caption{$\mathcal{H}$ class-pair statistics at each layer for instance 4 of the model for CIFAR10 with
         random class labels. (a) shows results for the data for passing through the randomly
         initialized model  (epoch~0 state).  (b) shows the results for the data  passing through
         the fully trained  model  (epoch~200 state). (Note:~Cosine distance is used as the
         proximity measure.)}
\label{figure:B5}
\end{figure}

\begin{figure}[p!]
\centering
\begin{subfigure}[(A)]{0.49\linewidth}
  \includegraphics[width=\linewidth]{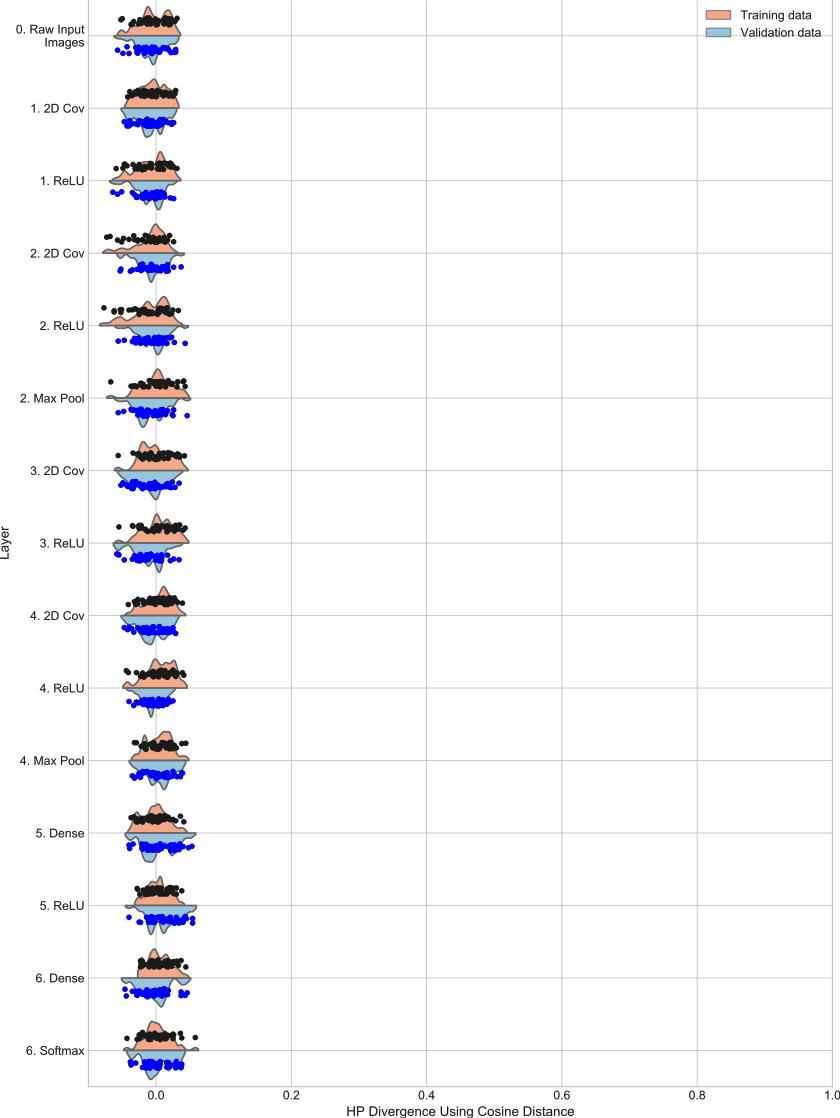}
  \caption{Untrained model}
\end{subfigure}
\begin{subfigure}[(B)]{0.49\linewidth}
  \includegraphics[width=\linewidth]{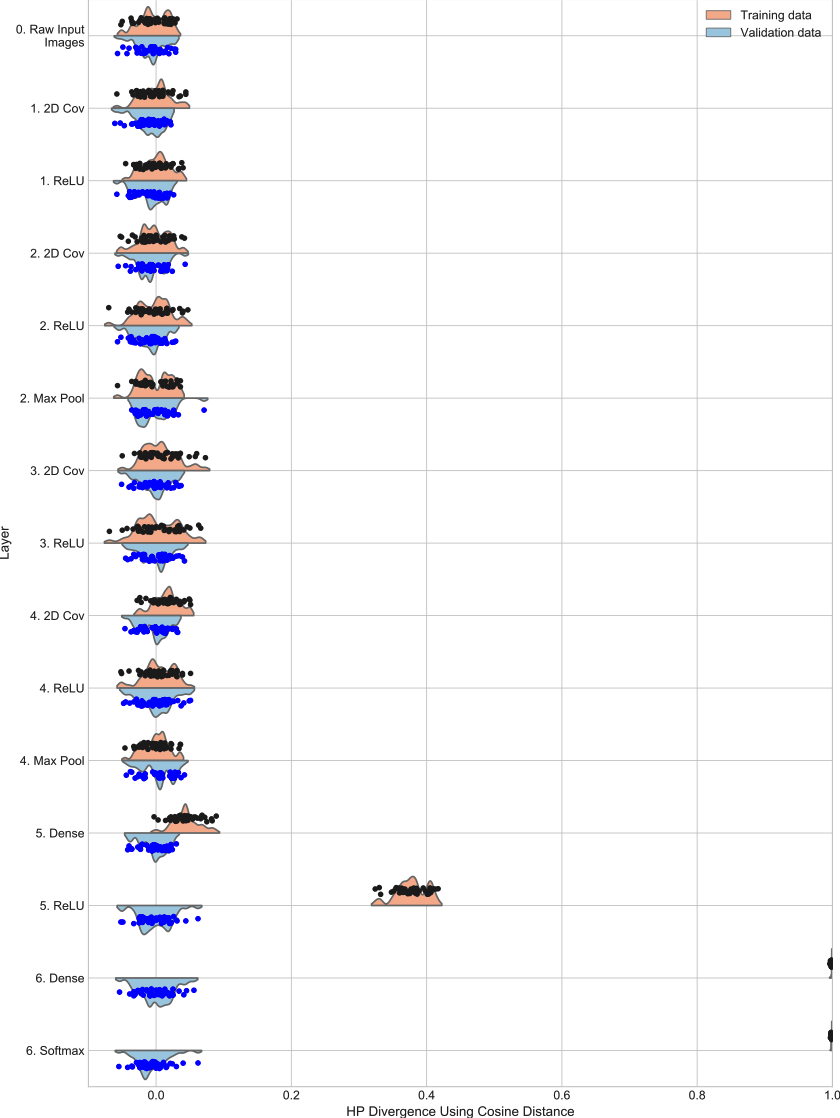}
  \caption{Trained model}
\end{subfigure}
\caption{$\mathcal{H}$ class-pair statistics at each layer for instance 5 of the model for CIFAR10 with
         random class labels. (a) shows results for the data for passing through the randomly
         initialized model  (epoch~0 state).  (b) shows the results for the data  passing through
         the fully trained  model  (epoch~200 state). (Note:~Cosine distance is used as the
         proximity measure.)}
\label{figure:B6}
\end{figure}

\begin{figure}[p!]
\centering
\begin{subfigure}[(A)]{0.49\linewidth}
  \includegraphics[width=\linewidth]{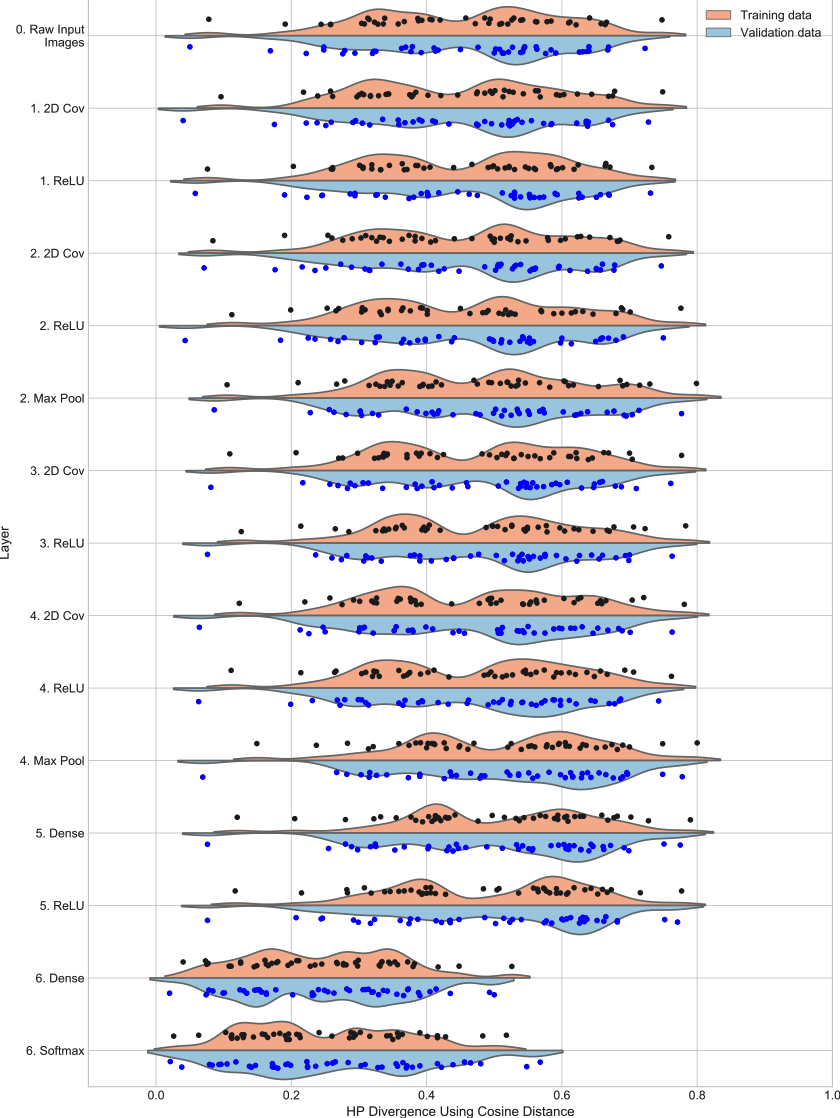}
  \caption{Untrained model}
\end{subfigure}
\begin{subfigure}[(B)]{0.49\linewidth}
  \includegraphics[width=\linewidth]{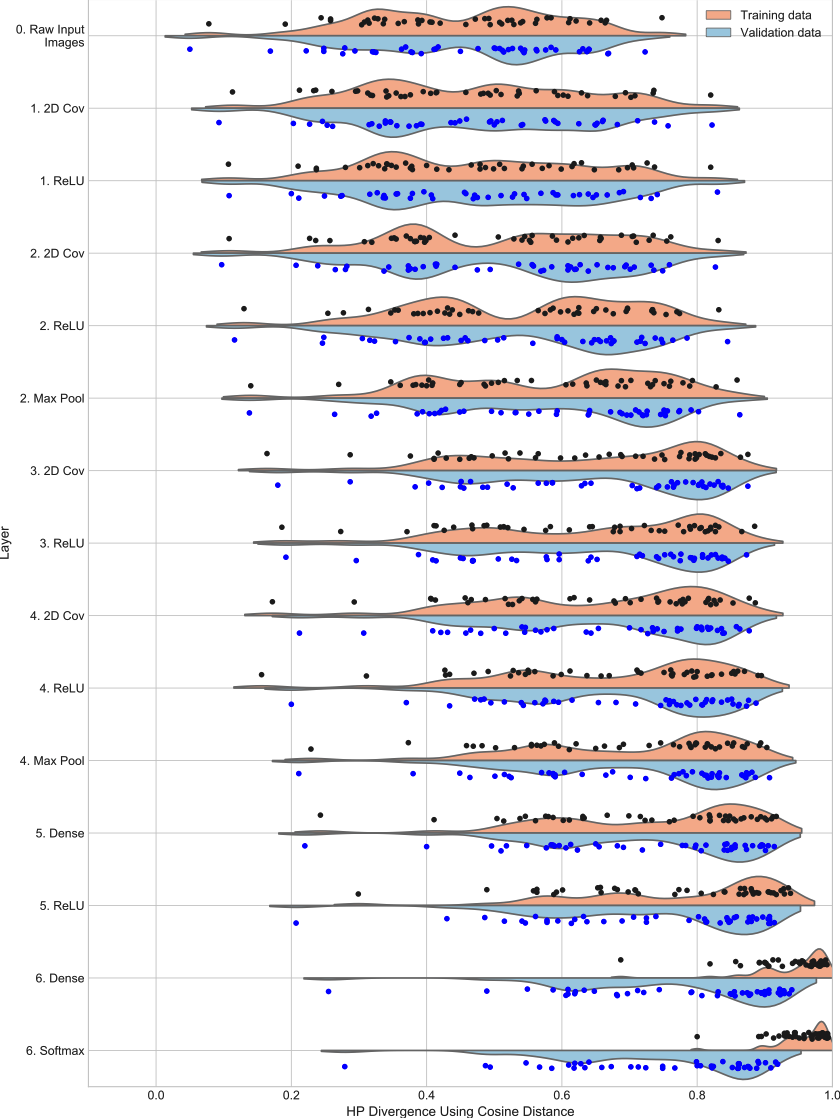}
  \caption{Trained model}
\end{subfigure}
\caption{$\mathcal{H}$ class-pair statistics at each layer for instance 1 of the model for CIFAR10 with
         true class labels. (a) shows results for the data for passing through the randomly
         initialized model  (epoch~0 state).  (b) shows the results for the data  passing through
         the fully trained  model  (stopping at peak validation set accuracy). (Note:~Cosine
         distance is used as the proximity measure.)}
\label{figure:B7}
\end{figure}

\begin{figure}[p!]
\centering
\begin{subfigure}[(A)]{0.49\linewidth}
  \includegraphics[width=\linewidth]{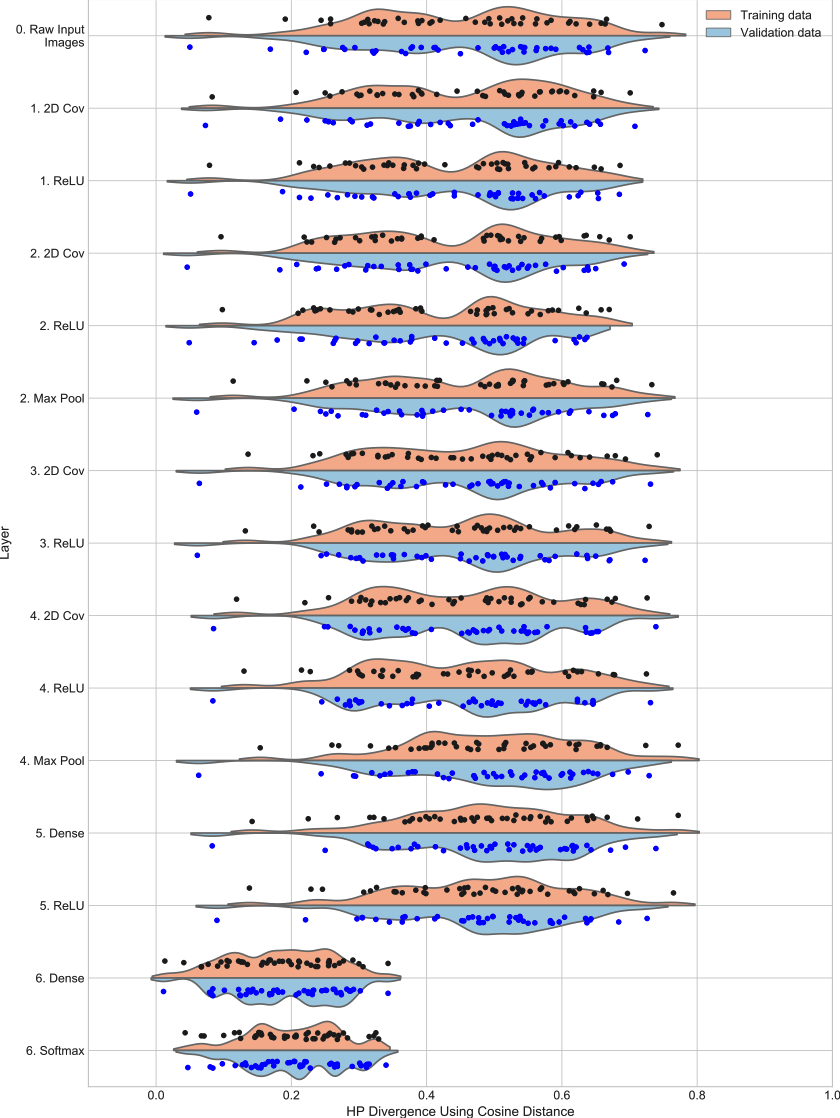}
  \caption{Untrained model}
\end{subfigure}
\begin{subfigure}[(B)]{0.49\linewidth}
  \includegraphics[width=\linewidth]{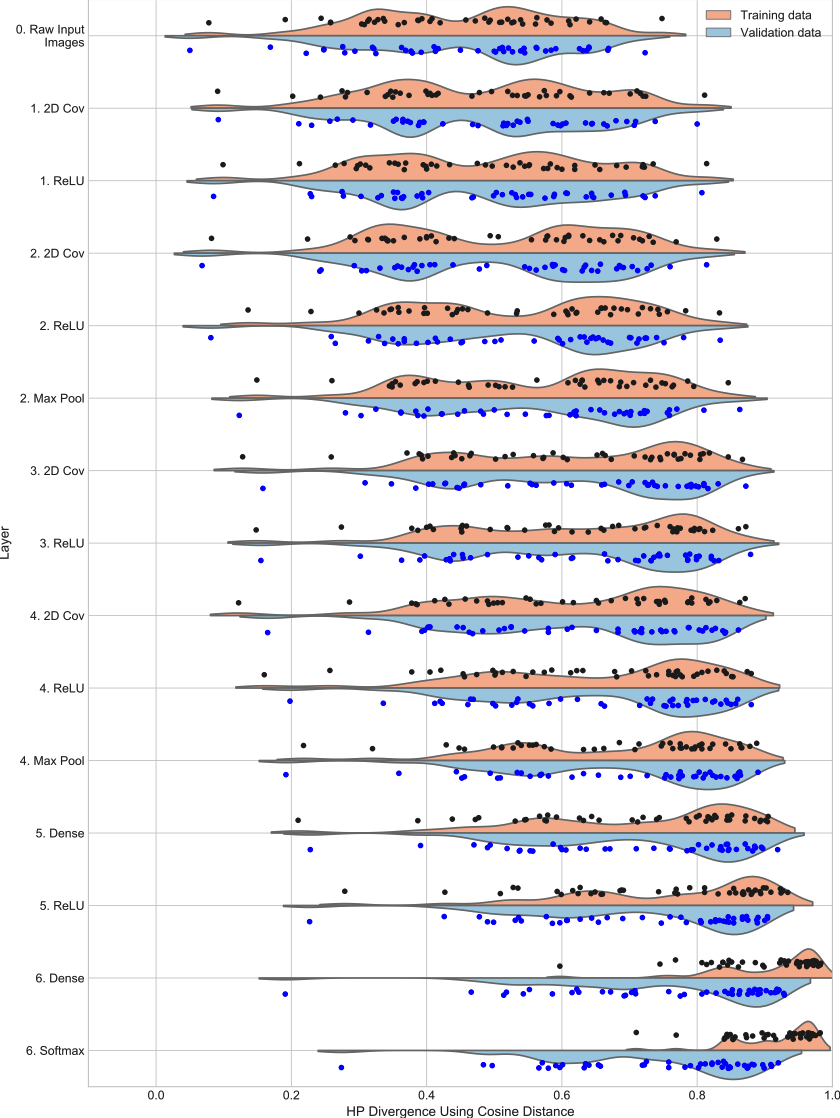}
  \caption{Trained model}
\end{subfigure}
\caption{$\mathcal{H}$ class-pair statistics at each layer for instance 2 of the model for CIFAR10 with
         true class labels. (a) shows results for the data for passing through the randomly
         initialized model  (epoch~0 state).  (b) shows the results for the data  passing through
         the fully trained  model  (stopping at peak validation set accuracy). (Note:~Cosine
         distance is used as the proximity measure.)}
\label{figure:B8}
\end{figure}

\begin{figure}[p!]
\centering
\begin{subfigure}[(A)]{0.49\linewidth}
  \includegraphics[width=\linewidth]{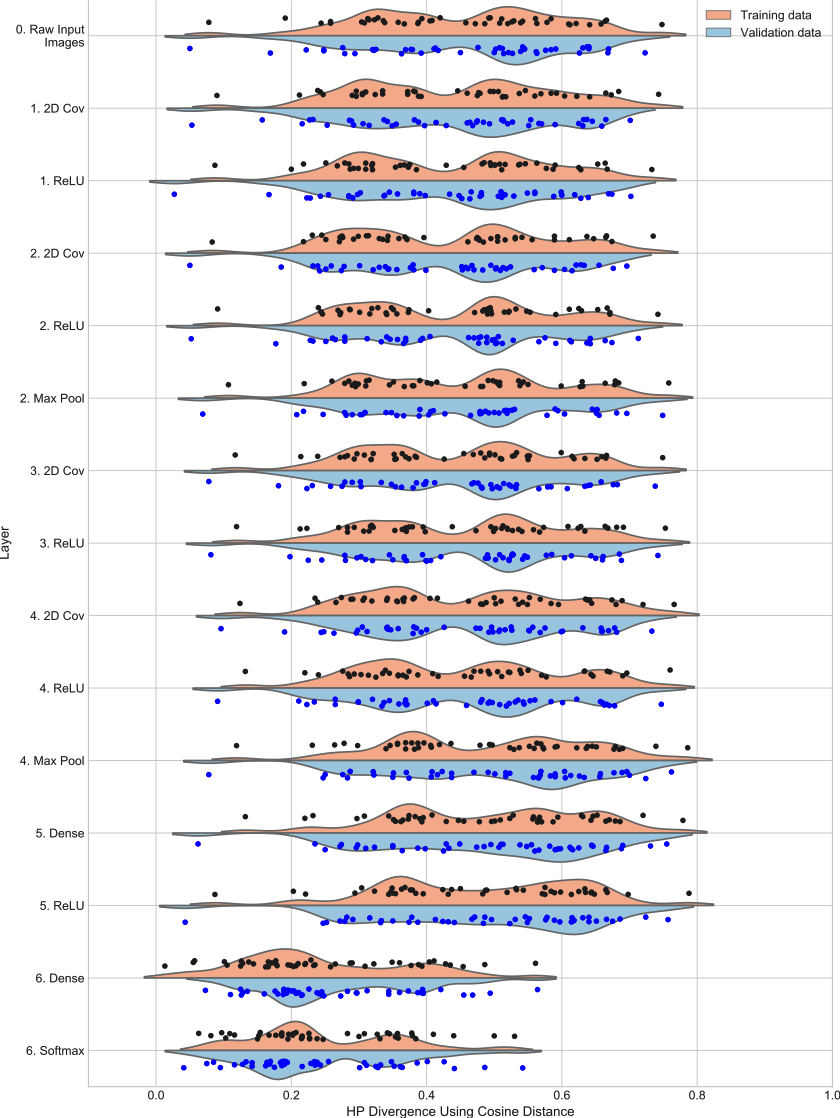}
  \caption{Untrained model}
\end{subfigure}
\begin{subfigure}[(B)]{0.49\linewidth}
  \includegraphics[width=\linewidth]{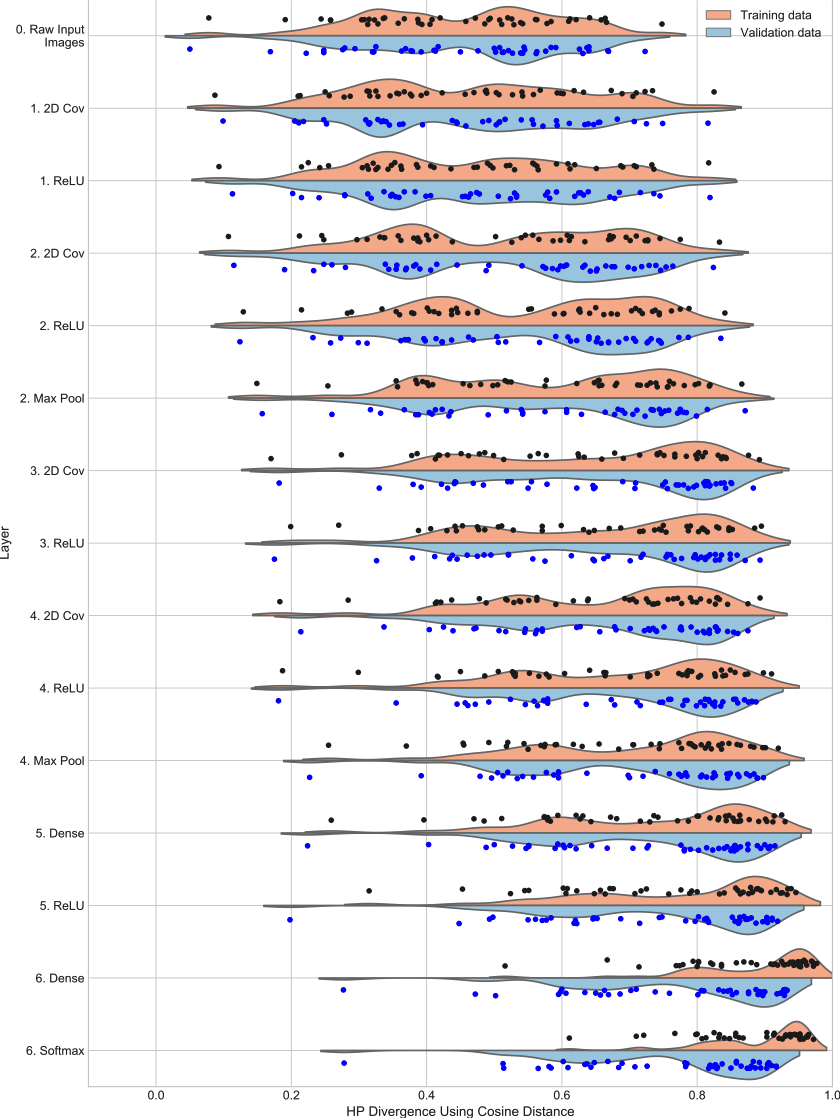}
  \caption{Trained model}
\end{subfigure}
\caption{$\mathcal{H}$ class-pair statistics at each layer for instance 3 of the model for CIFAR10 with
         true class labels. (a) shows results for the data for passing through the randomly
         initialized model  (epoch~0 state).  (b) shows the results for the data  passing through
         the fully trained  model  (stopping at peak validation set accuracy). (Note:~Cosine
         distance is used as the proximity measure.)}
\label{figure:B9}
\end{figure}

\begin{figure}[p!]
\centering
\begin{subfigure}[(A)]{0.49\linewidth}
  \includegraphics[width=\linewidth]{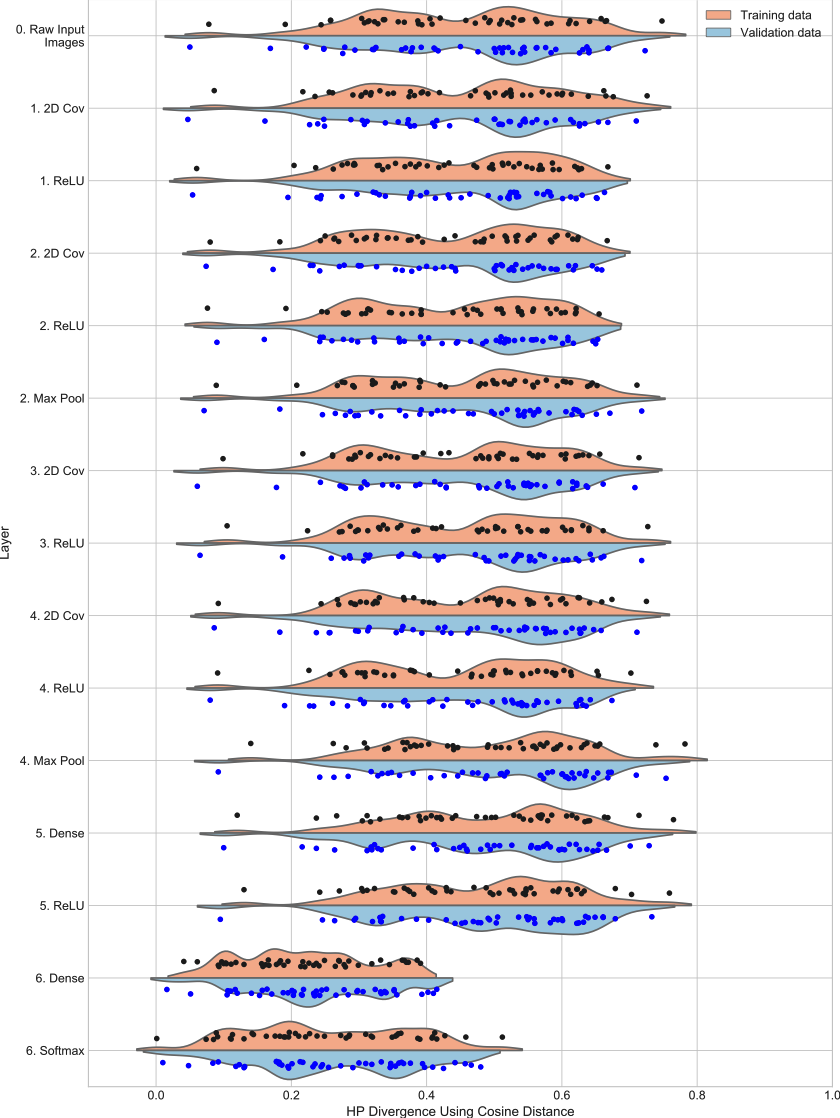}
  \caption{Untrained model}
\end{subfigure}
\begin{subfigure}[(B)]{0.49\linewidth}
  \includegraphics[width=\linewidth]{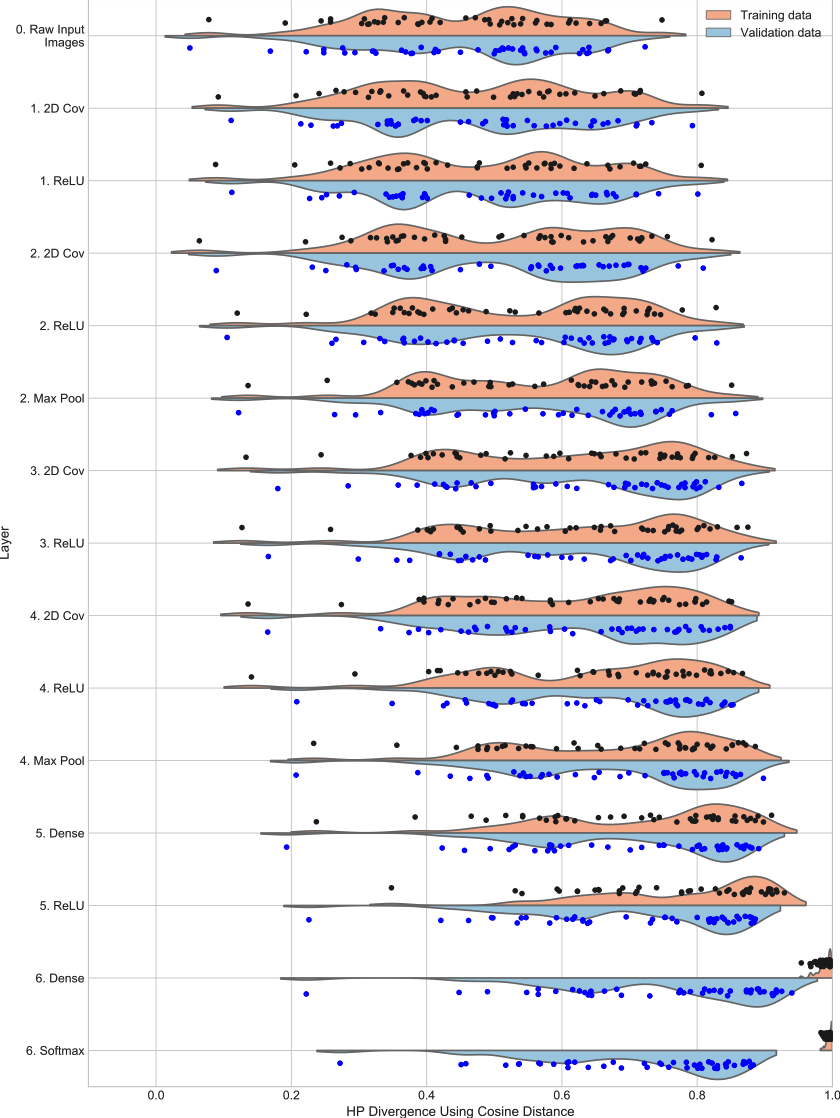}
  \caption{Trained model}
\end{subfigure}
\caption{$\mathcal{H}$ class-pair statistics at each layer for instance 4 of the model for CIFAR10 with
         true class labels. (a) shows results for the data for passing through the randomly
         initialized model  (epoch~0 state).  (b) shows the results for the data  passing through
         the fully trained  model  (stopping at peak validation set accuracy). (Note:~Cosine
         distance is used as the proximity measure.)}
\label{figure:B10}
\end{figure}

\begin{figure}[p!]
\centering
\begin{subfigure}[(A)]{0.49\linewidth}
  \includegraphics[width=\linewidth]{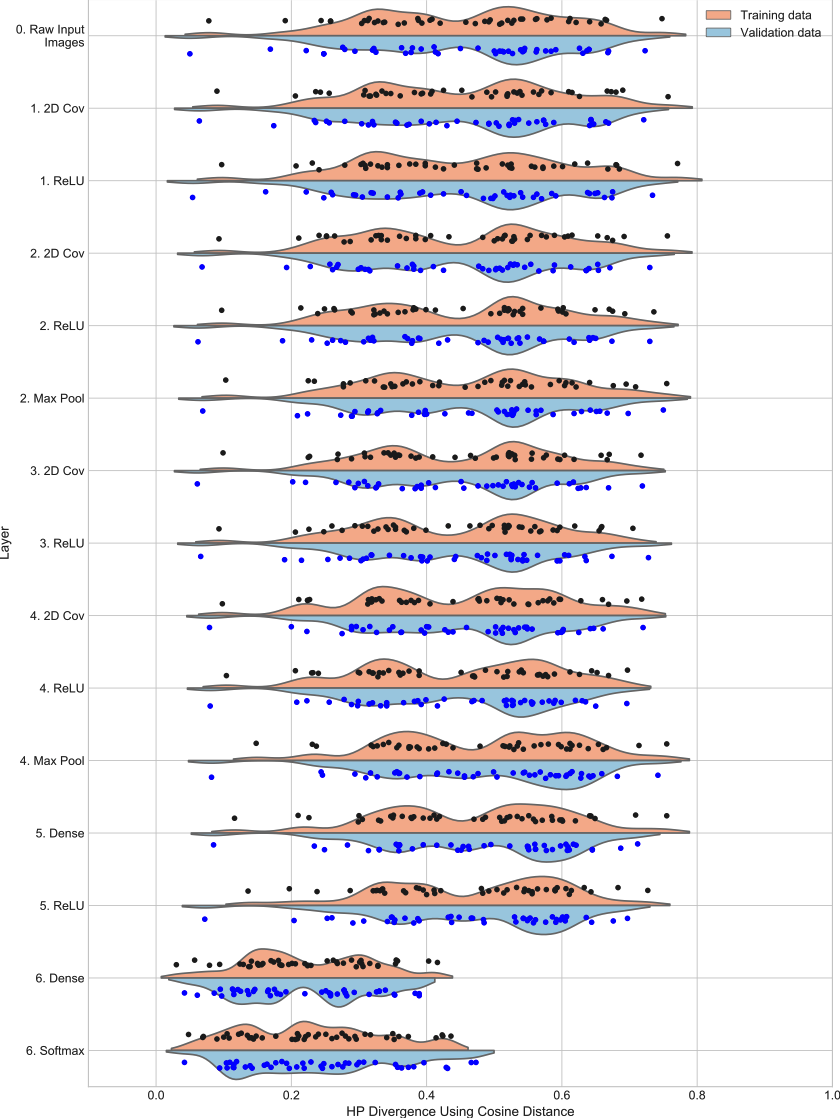}
  \caption{Untrained model}
\end{subfigure}
\begin{subfigure}[(B)]{0.49\linewidth}
  \includegraphics[width=\linewidth]{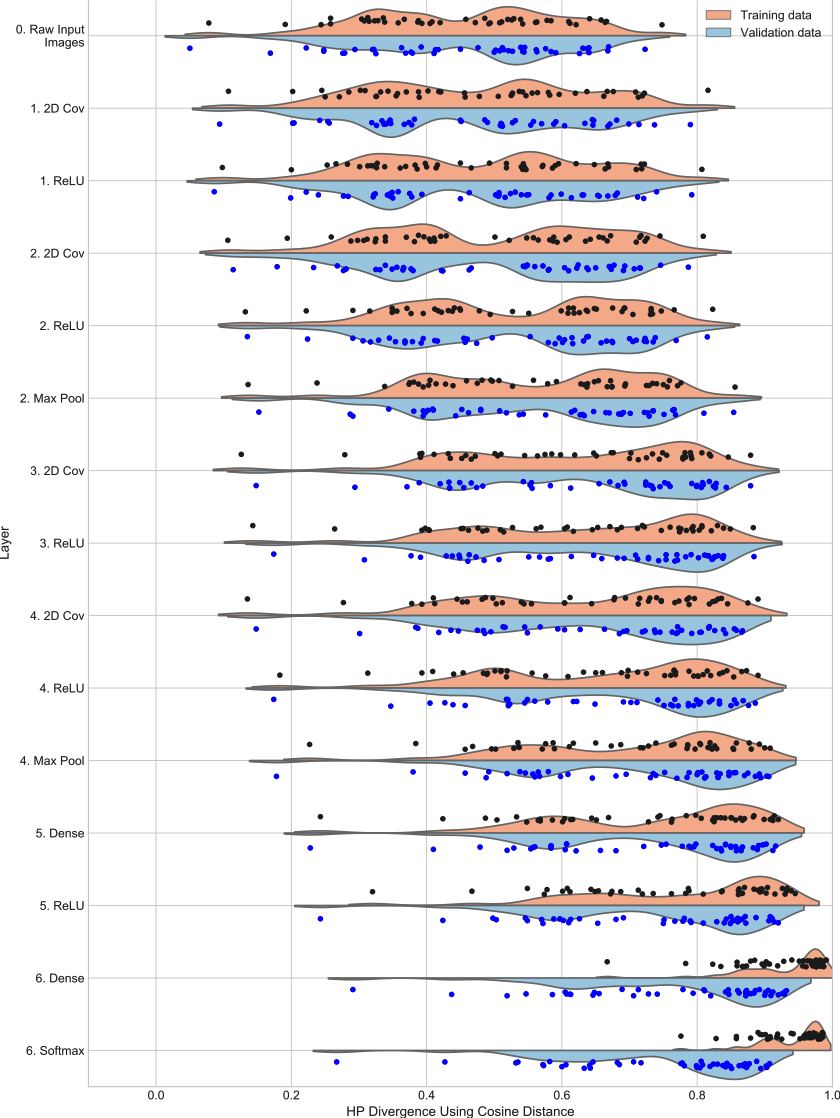}
  \caption{Trained model}
\end{subfigure}
\caption{$\mathcal{H}$ class-pair statistics at each layer for instance 5 of the model for CIFAR10 with
         true class labels. (a) shows results for the data for passing through the randomly
         initialized model  (epoch~0 state).  (b) shows the results for the data  passing through
         the fully trained  model  (stopping at peak validation set accuracy). (Note:~Cosine
         distance is used as the proximity measure.)}
\label{figure:B11}
\end{figure}

\begin{figure}[p!]
\centering
\begin{subfigure}[(A)]{0.49\linewidth}
  \includegraphics[width=\linewidth]{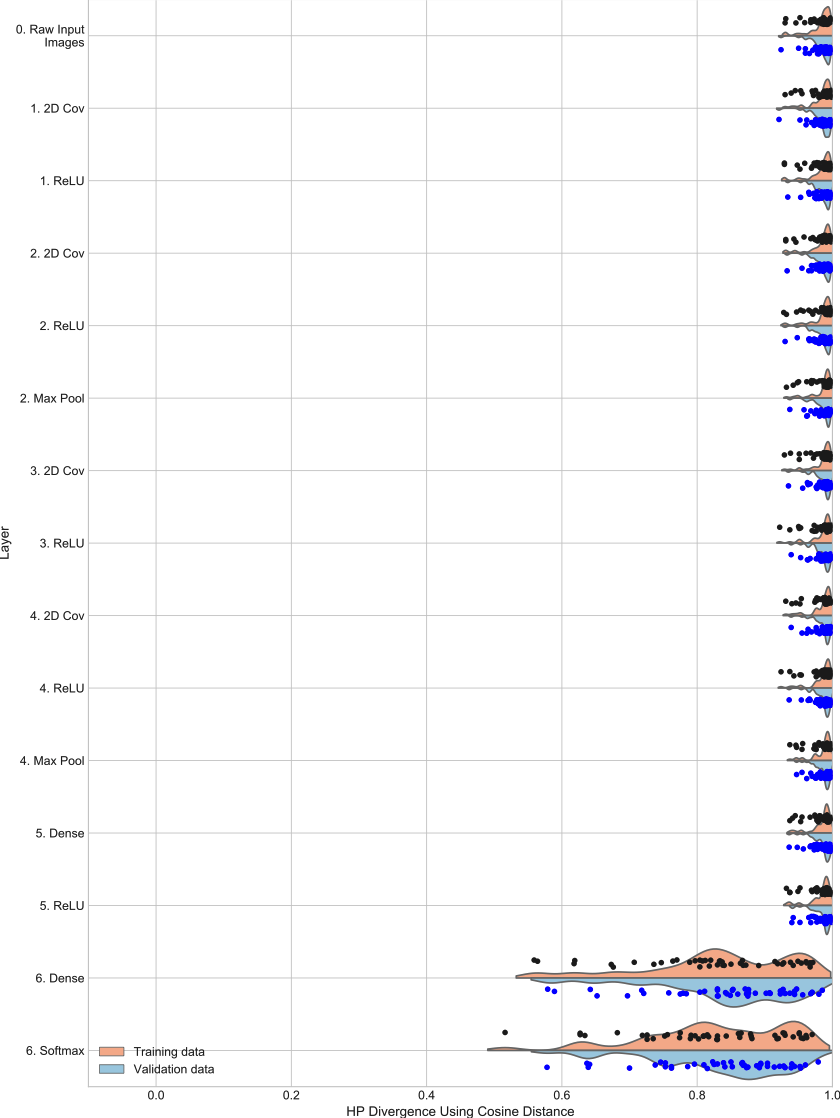}
  \caption{Untrained model}
\end{subfigure}
\begin{subfigure}[(B)]{0.49\linewidth}
  \includegraphics[width=\linewidth]{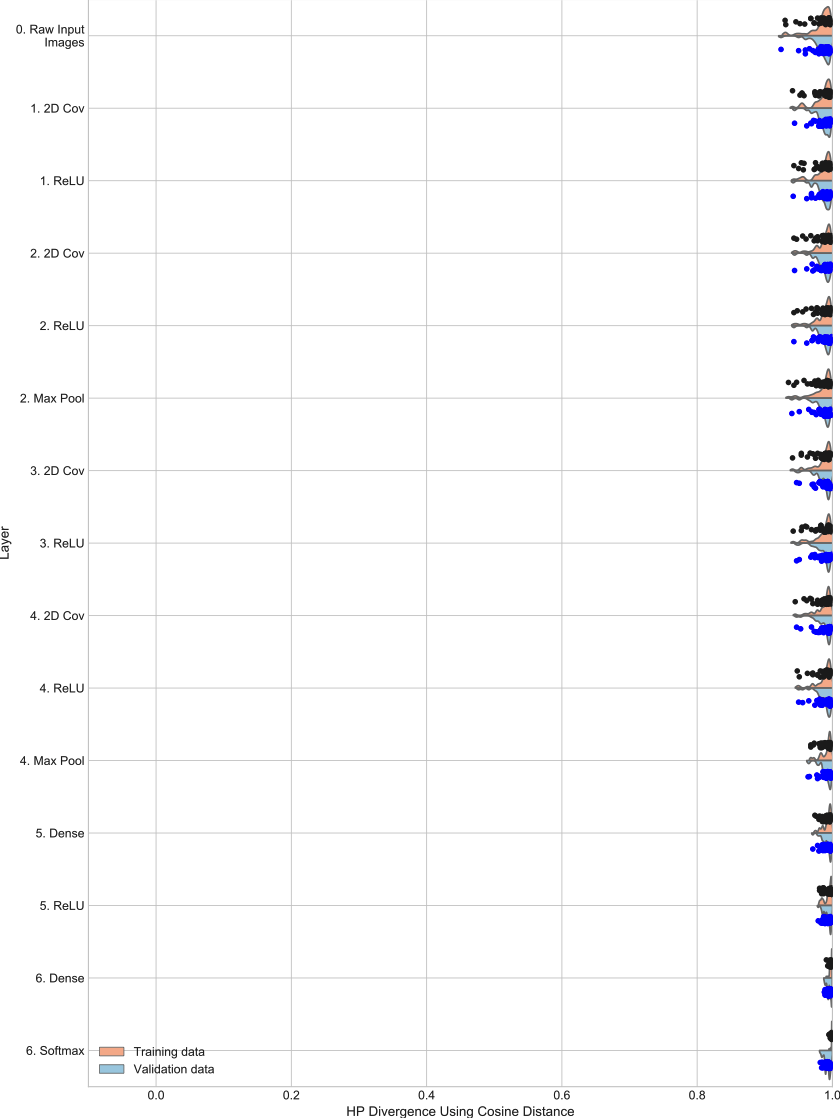}
  \caption{Trained model}
\end{subfigure}
\caption{$\mathcal{H}$ class-pair statistics at each layer for instance 1 of the model for MNIST with
         true class labels. (a) shows results for the data for passing through the randomly
         initialized model  (epoch~0 state).  (b) shows the results for the data  passing through
         the fully trained  model  (stopping at peak validation set accuracy). (Note:~Cosine
         distance is used as the proximity measure.)}
\label{figure:B12}
\end{figure}

\begin{figure}[p!]
\centering
\begin{subfigure}[(A)]{0.49\linewidth}
  \includegraphics[width=\linewidth]{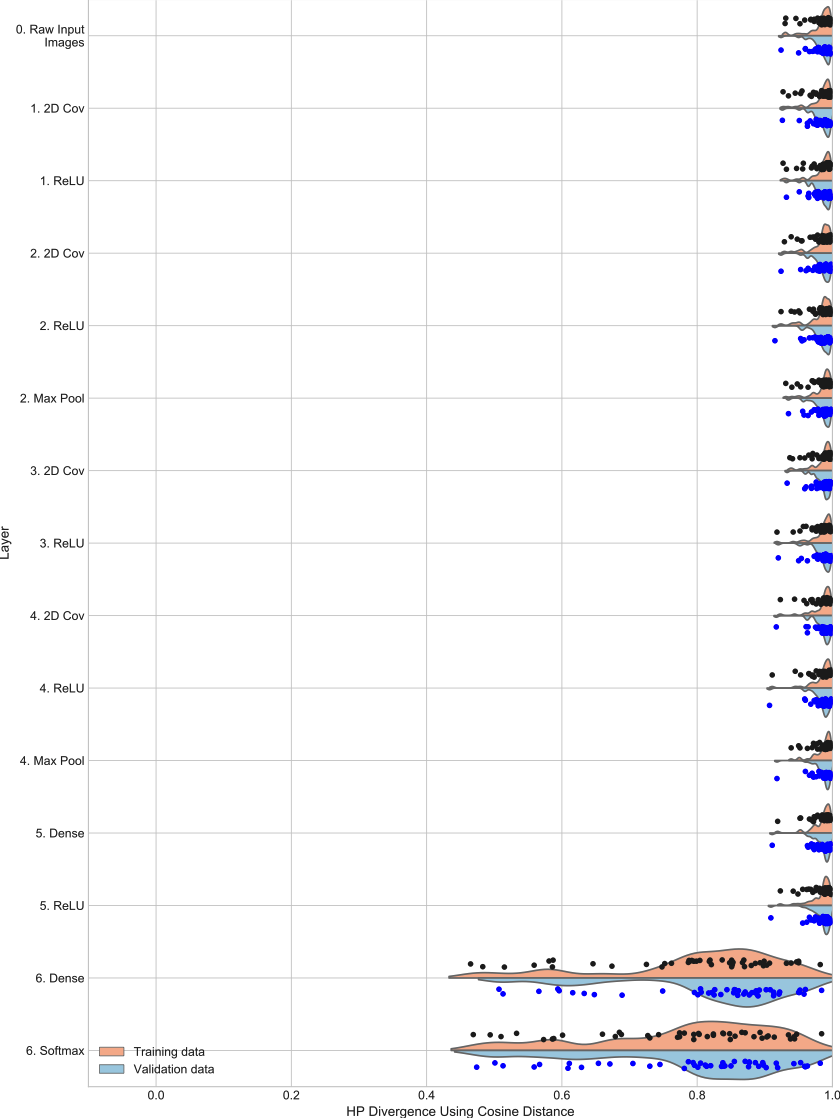}
  \caption{Untrained model}
\end{subfigure}
\begin{subfigure}[(B)]{0.49\linewidth}
  \includegraphics[width=\linewidth]{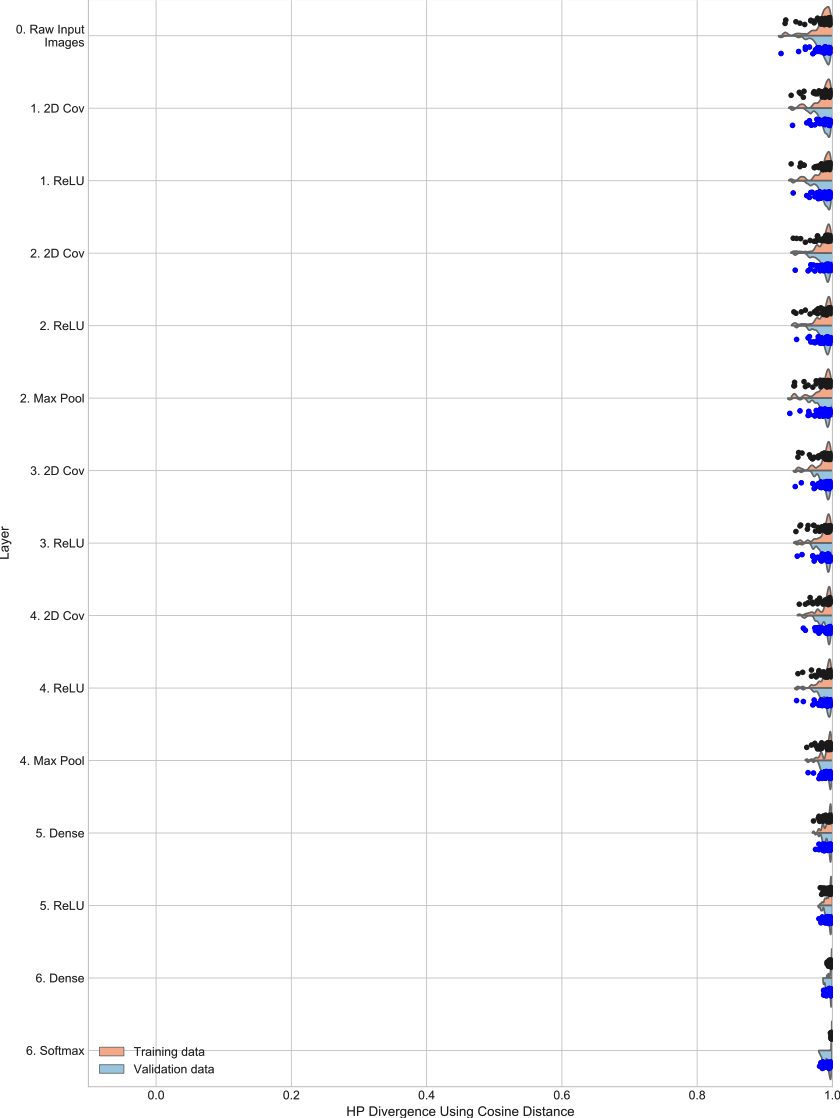}
  \caption{Trained model}
\end{subfigure}
\caption{$\mathcal{H}$ class-pair statistics at each layer for instance 2 of the model for MNIST with
         true class labels. (a) shows results for the data for passing through the randomly
         initialized model  (epoch~0 state).  (b) shows the results for the data  passing through
         the fully trained  model  (stopping at peak validation set accuracy). (Note:~Cosine
         distance is used as the proximity measure.)}
\label{figure:B13}
\end{figure}

\begin{figure}[p!]
\centering
\begin{subfigure}[(A)]{0.49\linewidth}
  \includegraphics[width=\linewidth]{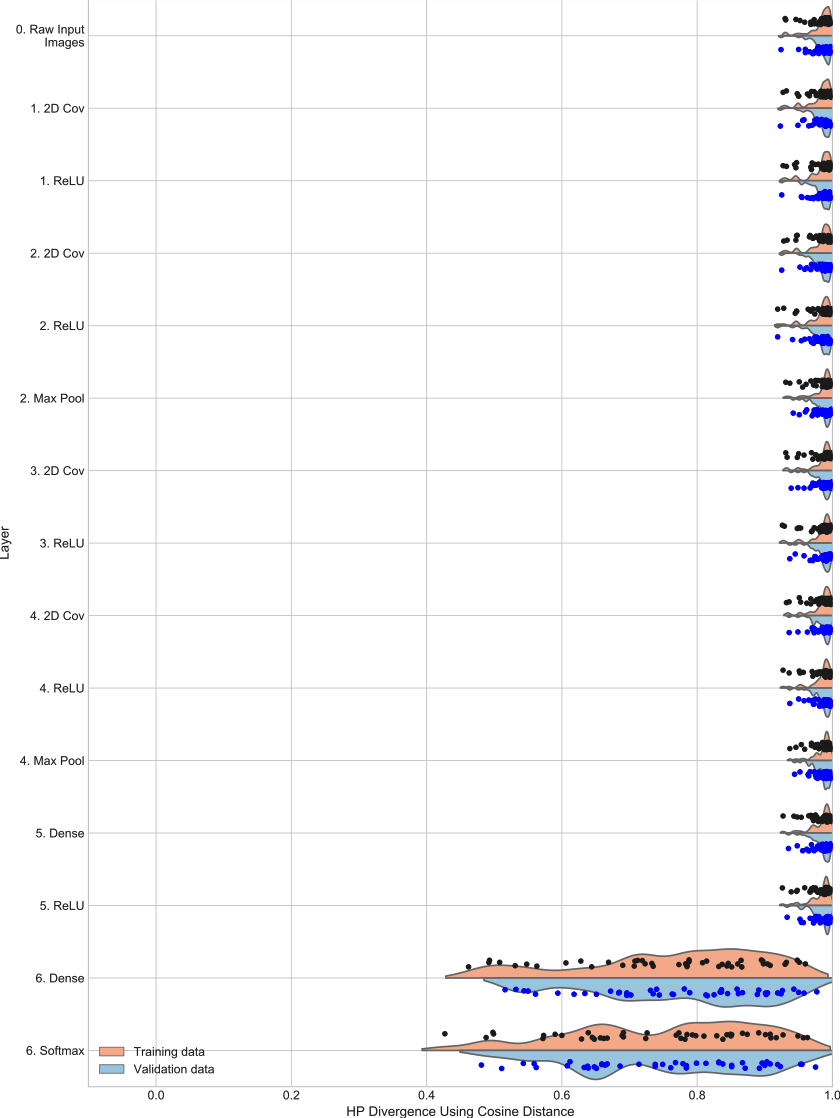}
  \caption{Untrained model}
\end{subfigure}
\begin{subfigure}[(B)]{0.49\linewidth}
  \includegraphics[width=\linewidth]{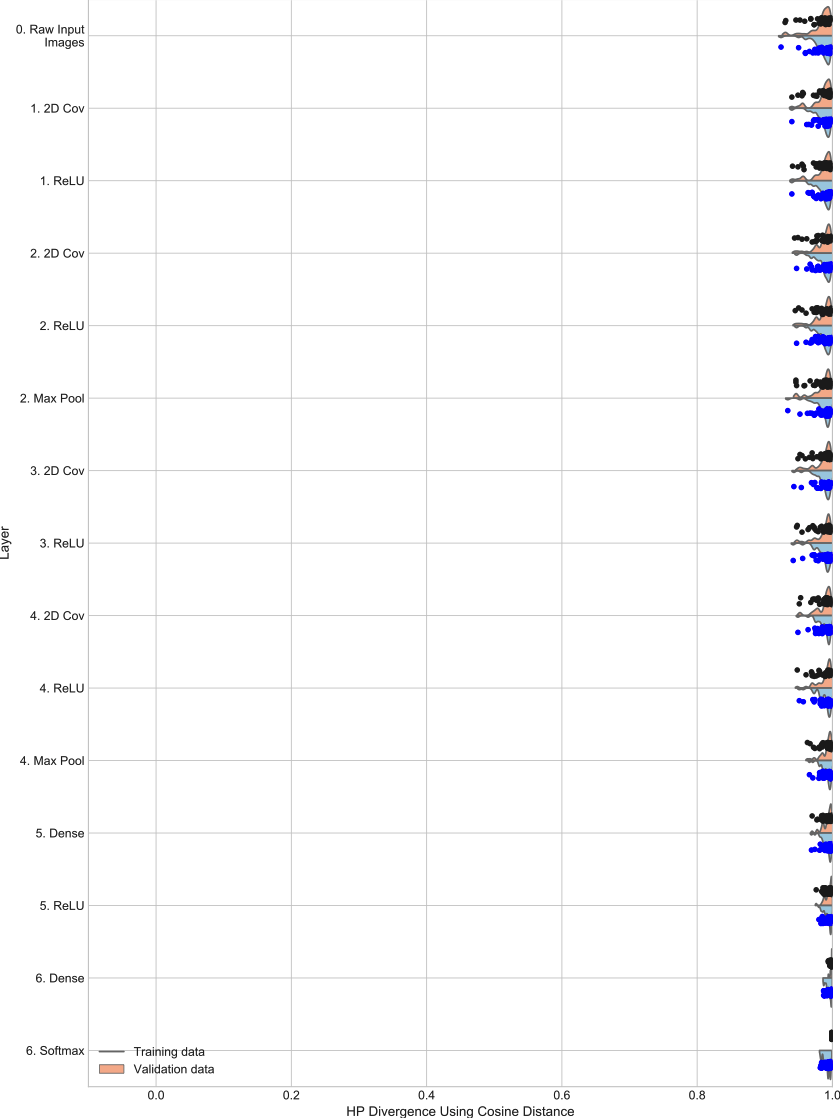}
  \caption{Trained model}
\end{subfigure}
\caption{$\mathcal{H}$ class-pair statistics at each layer for instance 3 of the model for MNIST with
         true class labels. (a) shows results for the data for passing through the randomly
         initialized model  (epoch~0 state).  (b) shows the results for the data  passing through
         the fully trained  model  (stopping at peak validation set accuracy). (Note:~Cosine
         distance is used as the proximity measure.)}
\label{figure:B14}
\end{figure}

\begin{figure}[p!]
\centering
\begin{subfigure}[(A)]{0.49\linewidth}
  \includegraphics[width=\linewidth]{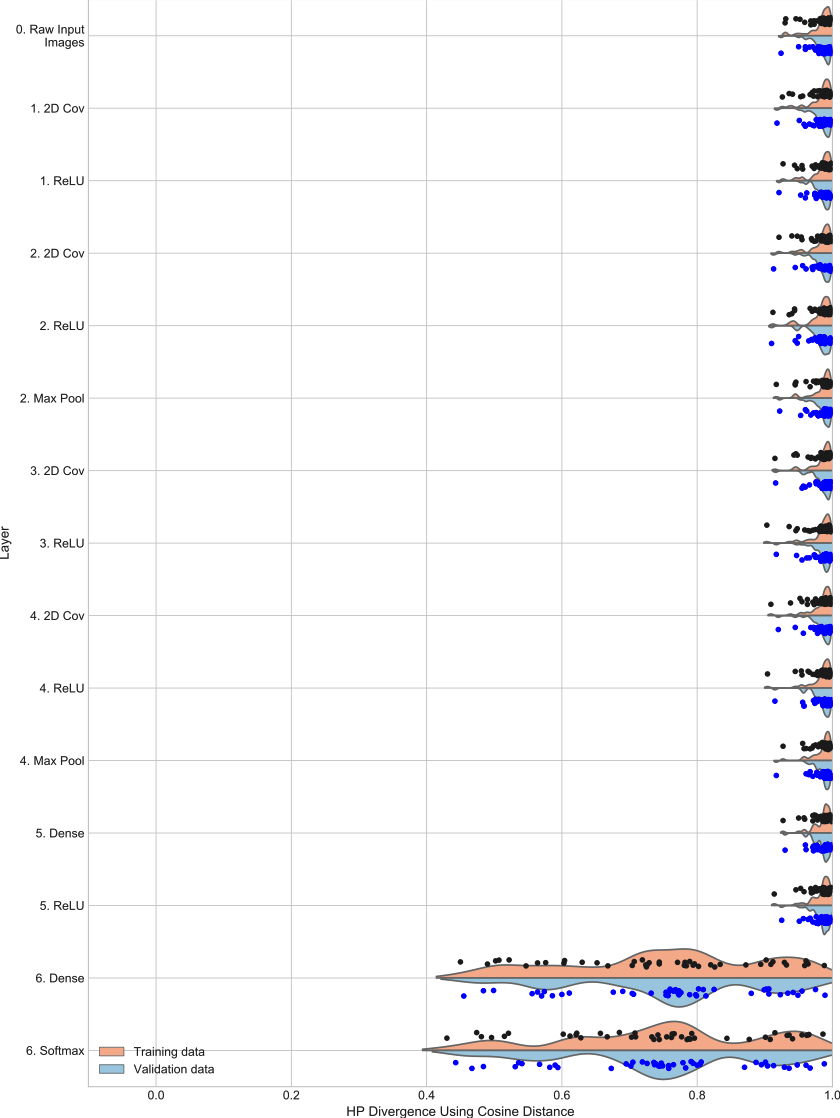}
  \caption{Untrained model}
\end{subfigure}
\begin{subfigure}[(B)]{0.49\linewidth}
  \includegraphics[width=\linewidth]{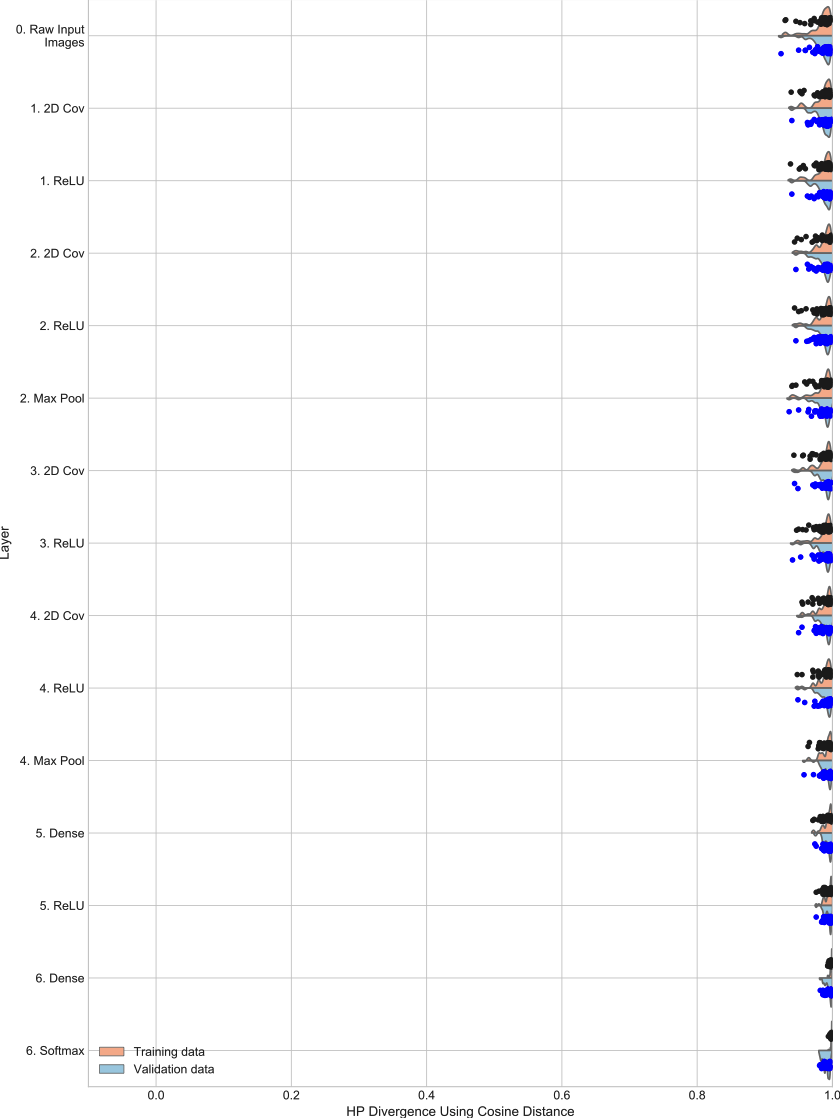}
  \caption{Trained model}
\end{subfigure}
\caption{$\mathcal{H}$ class-pair statistics at each layer for instance 4 of the model for MNIST with
         true class labels. (a) shows results for the data for passing through the randomly
         initialized model  (epoch~0 state).  (b) shows the results for the data  passing through
         the fully trained  model  (stopping at peak validation set accuracy). (Note:~Cosine
         distance is used as the proximity measure.)}
\label{figure:B15}
\end{figure}

\begin{figure}[p!]
\centering
\begin{subfigure}[(A)]{0.49\linewidth}
  \includegraphics[width=\linewidth]{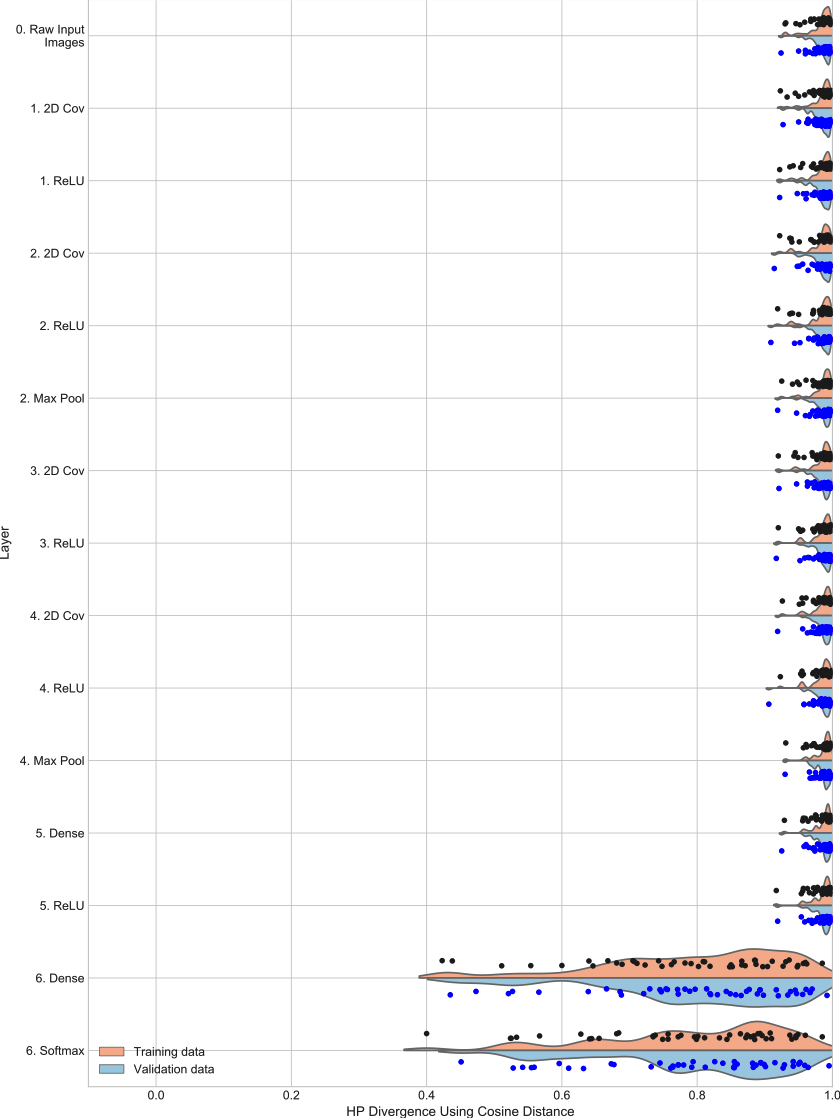}
  \caption{Untrained model}
\end{subfigure}
\begin{subfigure}[(B)]{0.49\linewidth}
  \includegraphics[width=\linewidth]{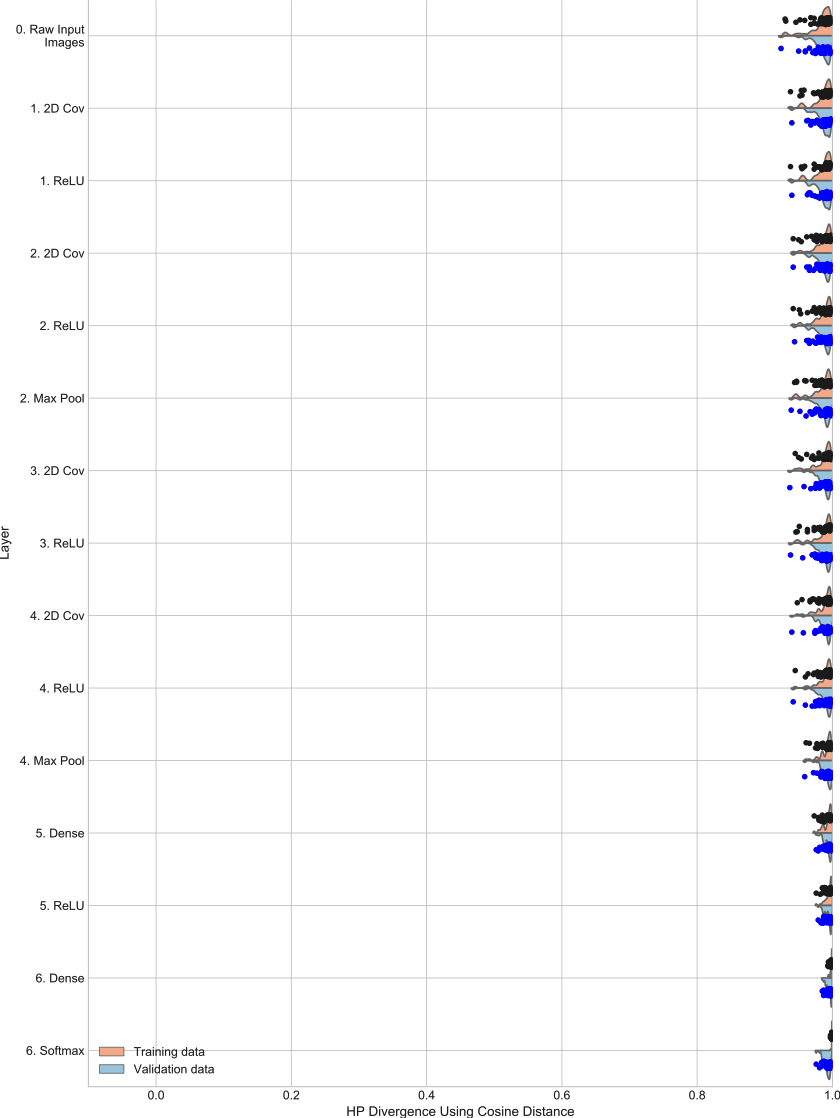}
  \caption{Trained model}
\end{subfigure}
\caption{$\mathcal{H}$ class-pair statistics at each layer for instance 5 of the model for MNIST with
         true class labels. (a) shows results for the data for passing through the randomly
         initialized model  (epoch~0 state).  (b) shows the results for the data  passing through
         the fully trained  model  (stopping at peak validation set accuracy). (Note:~Cosine
         distance is used as the proximity measure.)}
\label{figure:B16}
\end{figure}

Figures~\ref{figure:B2} through~\ref{figure:B6} present the class-wise HP statistics of the training
and validation samples of the CIFAR10 data with random labels as they pass through an associated
model. Each figure plots the results for one of the five training instances discussed in
Section~3 of the paper. The~(a)~subfigures show the data passing through the untrained models
and the~(b)~subfigures show the data passing through the trained version of the models. Similarly,
Figures~\ref{figure:B7} through~\ref{figure:B11} present plots for the~5 model instances trained on
the CIFAR10 data with true labels, and Figures~\ref{figure:B12} through~\ref{figure:B16} show the
plots for the~5 MNIST-trained models. As mentioned above, the cosine distance is used for all of
these cases\footnote{Note that that the results using cosine distance and the Euclidian distance
show similar trends.}.

Tables~\ref{table:B1Table} through~\ref{table:B9Table} present results of two-sample null hypothesis
test of the difference of means, where cosine distance is used in the HP divergence calculations.
The tables flag cases where the estimated $p$-values are $< 0.025$\footnote{Note that the tests were
performed using the random permutation algorithm with~50,000 Monte Carlo trials.}.
\begin{table}[p!]
\tiny
\centering
\caption{Two-sided permutation test of training data to detect change in  $\bar{\mathcal{H}}$ between
         layers (before training) with critical value $\alpha = 0.025$. \textcolor{red}{Red font}
         denote layer instances for which we reject $\mathbf{H_0}$, black font denotes layers for
         which we fail to reject $\mathbf{H_0}$ ~(Eq.~14 in paper).
         (Note:~Cosine distance is used as the proximity
         measure. 50,000 Monte Carlo trials were used to estimate the $p$-values. $\Delta{\bar{\mathcal{H}}} = \bar{\mathcal{H}}^{(t)}_{(k,0)} - \bar{\mathcal{H}}^{(t)}_{(k-1,0)}$))}
\label{table:B1Table}
\begin{tabular}{p{12mm}|p{12mm}|K{14mm}|K{14mm}|K{14mm}|K{14mm}|K{14mm}|K{14mm}}
\toprule
Input Space&Output Space& \multicolumn{2}{c}{CIFAR10 w Random}& \multicolumn{2}{c}{CIFAR10 w True}&
\multicolumn{2}{c}{MNIST w True} \\
& &$\Delta{\bar{\mathcal{H}}}$ & \textit{p}-values & $\Delta{\bar{\mathcal{H}}}$ & \textit{p}-values&
$\Delta{\bar{\mathcal{H}}}$ & \textit{p}-values \\
\midrule
0.Input&1.Conv&-.001; -.004; -.001; -.002; -.001&.855; .302; .923; .736; .838&-.004; .002; -.011;
-.002; .003&.901; .956; .725; .942; .932&.001; .000; -.001; -.000; -.000&.827; .978; .793; .991;
.994\\
\hline
1.Conv&1.ReLU&-.006; .005; -.003; .002; -.002&.158; .211; .619; .675; .719&.007; -.017; -.006;
-.007; -.001&.828; .567; .843; .806; .973&-.000; .001; .000; .001; .000&.958; .870; .901; .754;
.936\\
\hline
1.ReLU&2.Conv&-.001; -.002; .001; .001; -.003&.745; .725; .830; .893; .536&.005; -.001; -.000;
-.007; -.004&.871; .984; .991; .821; .893&.000; -.000; -.000; -.001; -.001&.962; .948; .977; .854;
.870\\
\hline
2.Conv&2.ReLU&-.004; .006; .001; .001; .000&.298; .148; .866; .924; .995&.002; -.014; .006; .001;
.002&.947; .642; .848; .973; .940&.000; -.000; -.001; -.002; .000&.940; .905; .816; .644; .942\\
\hline
2.ReLU&2.MaxPool&.005; -.004; .006; .000; .010&.129; .337; .283; .960; .071&.016; .036; .011; .018;
.012&.623; .236; .721; .537; .698&.001; .002; .003; .002; .001&.775; .658; .366; .522; .703\\
\hline
2.MaxPool&3.Conv&-.001; -.002; .001; -.008; -.000&.833; .627; .895; .220; .922&-.002; -.001; -.004;
-.002; -.012&.958; .981; .897; .951; .690&-.000; .001; -.001; -.000; -.001&.984; .865; .876; .959;
.836\\
\hline
3.Conv&3.ReLU&.000; .000; .003; -.002; .004&.899; .913; .519; .788; .405&.005; -.006; .008; .005;
-.006&.883; .846; .793; .876; .832&-.001; -.001; -.001; -.001; .001&.844; .843; .851; .768; .827\\
\hline
3.ReLU&4.Conv&.003; .004; .003; .002; -.001&.403; .363; .525; .730; .811&-.008; .005; .005; .001;
.004&.799; .863; .885; .974; .888&.001; .001; .000; .001; .001&.884; .783; .889; .872; .846\\
\hline
4.Conv&4.ReLU&.004; -.002; -.003; -.004; .003&.252; .692; .449; .411; .481&-.001; -.005; -.005;
-.009; -.003&.983; .852; .868; .771; .931&-.001; -.001; -.000; .000; -.001&.880; .778; .920; .972;
.862\\
\hline
4.ReLU&4.MaxPool&.002; -.005; .001; .002; -.001&.728; .280; .733; .660; .779&.041; .044; .034; .043;
.037&.187; .120; .281; .143; .218&.000; .001; .001; .002; .002&.894; .711; .800; .643; .595\\
\hline
4.MaxPool&5.Dense&-.001; .004; -.001; .001; -.007&.796; .431; .828; .860; .096&-.010; -.007; .000;
-.004; -.016&.734; .794; .997; .900; .594&-.001; -.001; -.002; -.001; -.001&.736; .742; .597; .828;
.743\\
\hline
5.Dense&5.ReLU&.003; .002; .001; -.001; .005&.583; .773; .811; .856; .153&-.013; -.007; -.011;
-.009; -.000&.673; .805; .730; .764; .993&-.001; -.002; -.001; -.001; -.001&.735; .575; .672; .664;
.721\\
\hline
5.ReLU&6.Dense&.006; .003; -.009; -.001; .001&.205; .571; .058; .846; .812&\textcolor{red}{-.243};
\textcolor{red}{-.297}; \textcolor{red}{-.231}; \textcolor{red}{-.261};
\textcolor{red}{-.244}&\textcolor{red}{.000}; \textcolor{red}{.000}; \textcolor{red}{.000};
\textcolor{red}{.000}; \textcolor{red}{.000}&\textcolor{red}{-.152}; \textcolor{red}{-.188};
\textcolor{red}{-.221}; \textcolor{red}{-.232}; \textcolor{red}{-.193}&\textcolor{red}{.000};
\textcolor{red}{.000}; \textcolor{red}{.000}; \textcolor{red}{.000}; \textcolor{red}{.000}\\
\hline
6.Dense&6.Softmax&.000; -.004; .001; -.001; -.002&.997; .401; .827; .745; .661&.002; .020; .004;
.035; .007&.942; .205; .885; .132; .726&-.003; -.015; -.011; -.001; -.001&.884; .594; .711; .963;
.966\\
\bottomrule
\end{tabular}
\end{table}
\begin{table}[p!]
\tiny
\centering
\caption{Differences between the trained and initialized $\boldsymbol{\mathcal{H}}$ class-pair statistics of a layer,
         and respective $p$-values for the
         corresponding one-sided permutation test ~(Eq.~15 in paper). \textcolor{red}{Red font}
         denotes layer instances for which we reject $\mathbf{H_0}$, and black font denotes layers for
         which we fail to reject $\mathbf{H_0}$. 
         (Note: ~Cosine distance is used as the proximity measure. 50,000 Monte Carlo trials to estimate the $p$-values.
         $\Delta{\bar{\mathcal{H}}} = \bar{\mathcal{H}}^{(t)}_{(k,T)} - \bar{\mathcal{H}}^{(t)}_{(k,0)}$)}
\label{table:B2Table}
\begin{tabular}{p{12mm}|K{16mm}|K{16mm}|K{16mm}|K{16mm}|K{16mm}|K{16mm}}
\toprule
Output Space& \multicolumn{2}{c|}{CIFAR10 w Random}& \multicolumn{2}{c|}{CIFAR10 w True}&
\multicolumn{2}{c}{MNIST w True} \\
 &$\Delta{\bar{\mathcal{H}}}$ & \textit{p}-values & $\Delta{\bar{\mathcal{H}}}$ & \textit{p}-values&
 $\Delta{\bar{\mathcal{H}}}$ & \textit{p}-values \\
\midrule
1.Conv&.003; .001; -.002; -.003; .003&.220; .394; .631; .674; .220&.020; .031; .022; .026;
.020&.269; .173; .254; .211; .277&.002; .002; .003; .002; .003&.288; .241; .160; .231; .222\\
\hline
1.ReLU&.007; -.004; .001; -.006; .006&.055; .807; .409; .853; .094&.018; .050; .035; .042;
.025&.297; .061; .147; .096; .222&.002; .002; .003; .002; .002&.275; .282; .183; .310; .227\\
\hline
2.Conv&\textcolor{red}{.017}; .001; .006; -.013; .006&\textcolor{red}{.000}; .406; .115; .989;
.098&.049; \textcolor{red}{.075}; \textcolor{red}{.076}; \textcolor{red}{.075}; .050&.076;
\textcolor{red}{.014}; \textcolor{red}{.016}; \textcolor{red}{.013}; .071&.002; .002; .004; .003;
.003&.246; .229; .135; .209; .166\\
\hline
2.ReLU&\textcolor{red}{.012}; -.007; .007; -.016; .006&\textcolor{red}{.000}; .945; .087; .997;
.116&\textcolor{red}{.081}; \textcolor{red}{.116}; \textcolor{red}{.106}; \textcolor{red}{.100};
\textcolor{red}{.082}&\textcolor{red}{.008}; \textcolor{red}{.000}; \textcolor{red}{.001};
\textcolor{red}{.001}; \textcolor{red}{.008}&.002; .003; .005; .004; .003&.271; .199; .095; .099;
.177\\
\hline
2.MaxPool&\textcolor{red}{.014}; .001; -.003; -.017; -.001&\textcolor{red}{.000}; .441; .726; .993;
.607&\textcolor{red}{.105}; \textcolor{red}{.108}; \textcolor{red}{.139}; \textcolor{red}{.111};
\textcolor{red}{.101}&\textcolor{red}{.001}; \textcolor{red}{.001}; \textcolor{red}{.000};
\textcolor{red}{.001}; \textcolor{red}{.001}&-.000; .000; -.000; .001; .001&.550; .473; .505; .436;
.393\\
\hline
3.Conv&\textcolor{red}{.021}; .001; -.001; -.005; .007&\textcolor{red}{.000}; .430; .585; .822;
.080&\textcolor{red}{.164}; \textcolor{red}{.154}; \textcolor{red}{.196}; \textcolor{red}{.158};
\textcolor{red}{.168}&\textcolor{red}{.000}; \textcolor{red}{.000}; \textcolor{red}{.000};
\textcolor{red}{.000}; \textcolor{red}{.000}&.001; .001; .002; .003; .003&.327; .343; .270; .213;
.203\\
\hline
3.ReLU&\textcolor{red}{.023}; .002; -.008; -.011; -.001&\textcolor{red}{.000}; .348; .966; .972;
.552&\textcolor{red}{.164}; \textcolor{red}{.169}; \textcolor{red}{.199}; \textcolor{red}{.162};
\textcolor{red}{.188}&\textcolor{red}{.000}; \textcolor{red}{.000}; \textcolor{red}{.000};
\textcolor{red}{.000}; \textcolor{red}{.000}&.002; .002; .003; .004; .002&.293; .269; .212; .155;
.249\\
\hline
4.Conv&\textcolor{red}{.010}; -.000; -.005; -.012; \textcolor{red}{.012}&\textcolor{red}{.003};
.506; .892; .984; \textcolor{red}{.002}&\textcolor{red}{.178}; \textcolor{red}{.164};
\textcolor{red}{.193}; \textcolor{red}{.154}; \textcolor{red}{.186}&\textcolor{red}{.000};
\textcolor{red}{.000}; \textcolor{red}{.000}; \textcolor{red}{.000}; \textcolor{red}{.000}&.002;
.002; .004; .005; .003&.245; .219; .105; .066; .168\\
\hline
4.ReLU&\textcolor{red}{.016}; \textcolor{red}{.015}; .008; -.002; -.001&\textcolor{red}{.000};
\textcolor{red}{.002}; .051; .633; .597&\textcolor{red}{.203}; \textcolor{red}{.200};
\textcolor{red}{.224}; \textcolor{red}{.194}; \textcolor{red}{.212}&\textcolor{red}{.000};
\textcolor{red}{.000}; \textcolor{red}{.000}; \textcolor{red}{.000}; \textcolor{red}{.000}&.003;
.004; .005; .005; .004&.149; .114; .061; .061; .074\\
\hline
4.MaxPool&\textcolor{red}{.014}; \textcolor{red}{.016}; .001; .003; -.006&\textcolor{red}{.002};
\textcolor{red}{.000}; .428; .255; .930&\textcolor{red}{.188}; \textcolor{red}{.183};
\textcolor{red}{.214}; \textcolor{red}{.185}; \textcolor{red}{.214}&\textcolor{red}{.000};
\textcolor{red}{.000}; \textcolor{red}{.000}; \textcolor{red}{.000};
\textcolor{red}{.000}&\textcolor{red}{.006}; \textcolor{red}{.005}; \textcolor{red}{.006};
\textcolor{red}{.006}; \textcolor{red}{.005}&\textcolor{red}{.012}; \textcolor{red}{.020};
\textcolor{red}{.004}; \textcolor{red}{.007}; \textcolor{red}{.024}\\
\hline
5.Dense&\textcolor{red}{.052}; \textcolor{red}{.051}; \textcolor{red}{.041}; \textcolor{red}{.033};
\textcolor{red}{.046}&\textcolor{red}{.000}; \textcolor{red}{.000}; \textcolor{red}{.000};
\textcolor{red}{.000}; \textcolor{red}{.000}&\textcolor{red}{.231}; \textcolor{red}{.227};
\textcolor{red}{.244}; \textcolor{red}{.227}; \textcolor{red}{.264}&\textcolor{red}{.000};
\textcolor{red}{.000}; \textcolor{red}{.000}; \textcolor{red}{.000};
\textcolor{red}{.000}&\textcolor{red}{.008}; \textcolor{red}{.008}; \textcolor{red}{.009};
\textcolor{red}{.008}; \textcolor{red}{.007}&\textcolor{red}{.000}; \textcolor{red}{.001};
\textcolor{red}{.000}; \textcolor{red}{.000}; \textcolor{red}{.001}\\
\hline
5.ReLU&\textcolor{red}{.362}; \textcolor{red}{.360}; \textcolor{red}{.350}; \textcolor{red}{.337};
\textcolor{red}{.375}&\textcolor{red}{.000}; \textcolor{red}{.000}; \textcolor{red}{.000};
\textcolor{red}{.000}; \textcolor{red}{.000}&\textcolor{red}{.286}; \textcolor{red}{.277};
\textcolor{red}{.293}; \textcolor{red}{.297}; \textcolor{red}{.309}&\textcolor{red}{.000};
\textcolor{red}{.000}; \textcolor{red}{.000}; \textcolor{red}{.000};
\textcolor{red}{.000}&\textcolor{red}{.011}; \textcolor{red}{.011}; \textcolor{red}{.012};
\textcolor{red}{.010}; \textcolor{red}{.010}&\textcolor{red}{.000}; \textcolor{red}{.000};
\textcolor{red}{.000}; \textcolor{red}{.000}; \textcolor{red}{.000}\\
\hline
6.Dense&\textcolor{red}{.995}; \textcolor{red}{1.000}; \textcolor{red}{1.003};
\textcolor{red}{.994}; \textcolor{red}{.995}&\textcolor{red}{.000}; \textcolor{red}{.000};
\textcolor{red}{.000}; \textcolor{red}{.000}; \textcolor{red}{.000}&\textcolor{red}{.698};
\textcolor{red}{.725}; \textcolor{red}{.637}; \textcolor{red}{.770};
\textcolor{red}{.706}&\textcolor{red}{.000}; \textcolor{red}{.000}; \textcolor{red}{.000};
\textcolor{red}{.000}; \textcolor{red}{.000}&\textcolor{red}{.167}; \textcolor{red}{.203};
\textcolor{red}{.238}; \textcolor{red}{.247}; \textcolor{red}{.207}&\textcolor{red}{.000};
\textcolor{red}{.000}; \textcolor{red}{.000}; \textcolor{red}{.000}; \textcolor{red}{.000}\\
\hline
6.Softmax&\textcolor{red}{.995}; \textcolor{red}{1.004}; \textcolor{red}{1.002};
\textcolor{red}{.996}; \textcolor{red}{.997}&\textcolor{red}{.000}; \textcolor{red}{.000};
\textcolor{red}{.000}; \textcolor{red}{.000}; \textcolor{red}{.000}&\textcolor{red}{.709};
\textcolor{red}{.714}; \textcolor{red}{.638}; \textcolor{red}{.741};
\textcolor{red}{.709}&\textcolor{red}{.000}; \textcolor{red}{.000}; \textcolor{red}{.000};
\textcolor{red}{.000}; \textcolor{red}{.000}&\textcolor{red}{.171}; \textcolor{red}{.219};
\textcolor{red}{.250}; \textcolor{red}{.249}; \textcolor{red}{.209}&\textcolor{red}{.000};
\textcolor{red}{.000}; \textcolor{red}{.000}; \textcolor{red}{.000}; \textcolor{red}{.000}\\
\bottomrule
\end{tabular}
\end{table}
%
\begin{table}[p!]
\tiny
\centering
\caption{Training data difference between input and output of each layer's $\boldsymbol{\mathcal{H}}$ class-pair statistics
         for the trained models, and respective
         $p$-values for the one sided permutation test~(Eq.~17 in paper).  \textcolor{red}{Red font}
         denotes layer instances for which we reject $\mathbf{H_0}$, and black font denotes layers for
         which we fail to reject $\mathbf{H_0}$. 
         (Note: ~Cosine distance is used as the proximity measure.  50,000
         Monte Carlo trials used to estimate the $p$-values. $\Delta{\bar{\mathcal{H}}} = \bar{\mathcal{H}}^{(t)}_{(k,T)} - \bar{\mathcal{H}}^{(t)}_{(k-1,T)}$)}
\label{table:B3Table}
\begin{tabular}{p{10mm}p{12mm}|K{16mm}|K{16mm}|K{16mm}|K{16mm}|K{16mm}|K{16mm}}
\toprule
Input Space &Output Space& \multicolumn{2}{c|}{CIFAR10 w Random}& \multicolumn{2}{c|}{CIFAR10 w
True}& \multicolumn{2}{c}{MNIST w True} \\
 & &$\Delta{\bar{\mathcal{H}}}$ & \textit{p}-values & $\Delta{\bar{\mathcal{H}}}$ & \textit{p}-values&
 $\Delta{\bar{\mathcal{H}}}$ & \textit{p}-values \\
\midrule
0.Input&1.Conv&.002; -.003; -.002; -.004; .002&.296; .794; .667; .767; .292&.016; .032; .011; .024;
.023&.312; .163; .366; .231; .247&.003; .002; .002; .002; .003&.214; .230; .237; .231; .220\\
\hline
1.Conv&1.ReLU&-.002; .001; .001; -.002; .001&.686; .408; .467; .603; .414&.004; .003; .007; .008;
.004&.457; .468; .422; .398; .459&-.000; .000; .000; .000; .000&.511; .487; .465; .469; .477\\
\hline
1.ReLU&2.Conv&.009; .003; .006; -.006; -.003&.028; .236; .154; .825; .731&.037; .024; .040; .027;
.021&.153; .249; .133; .221; .271&.000; .000; .000; .000; .000&.441; .467; .440; .450; .466\\
\hline
2.Conv&2.ReLU&-.008; -.002; .002; -.003; .000&.969; .685; .368; .667; .478&.034; .027; .036; .026;
.034&.169; .221; .161; .235; .169&.000; .000; -.000; .000; .000&.494; .501; .499; .491; .480\\
\hline
2.ReLU&2.MaxPool&.007; .004; -.004; -.001; .002&.050; .193; .818; .544; .361&.040; .028; .043; .028;
.031&.132; .216; .121; .205; .186&-.001; -.001; -.001; -.001; -.001&.669; .630; .669; .682; .617\\
\hline
2.MaxPool&3.Conv&.006; -.002; .002; .004; .008&.065; .683; .291; .257; .063&.057; .045; .053; .045;
.054&.060; .108; .081; .103; .063&.002; .002; .001; .002; .001&.289; .302; .307; .268; .359\\
\hline
3.Conv&3.ReLU&.002; .002; -.004; -.007; -.004&.290; .368; .826; .919; .745&.005; .009; .012; .009;
.014&.443; .399; .379; .404; .357&-.000; .000; -.000; -.000; .000&.548; .487; .498; .524; .463\\
\hline
3.ReLU&4.Conv&-.010; .002; .006; .001; \textcolor{red}{.012}&.994; .350; .103; .424;
\textcolor{red}{.016}&.006; .000; -.001; -.007; .002&.437; .501; .509; .581; .475&.001; .001; .002;
.002; .001&.375; .318; .279; .246; .315\\
\hline
4.Conv&4.ReLU&\textcolor{red}{.010}; \textcolor{red}{.013}; \textcolor{red}{.010}; .006;
-.010&\textcolor{red}{.011}; \textcolor{red}{.006}; \textcolor{red}{.017}; .118; .987&.024; .031;
.025; .031; .024&.250; .204; .240; .193; .263&.001; .000; .001; .000; .001&.398; .431; .405; .459;
.373\\
\hline
4.ReLU&4.MaxPool&-.000; -.005; -.006; .007; -.006&.504; .832; .885; .051; .903&.026; .028; .024;
.034; .039&.227; .217; .247; .162; .133&.003; .003; .003; .003; .002&.089; .116; .107; .095; .155\\
\hline
4.MaxPool&5.Dense&\textcolor{red}{.037}; \textcolor{red}{.040}; \textcolor{red}{.039};
\textcolor{red}{.031}; \textcolor{red}{.046}&\textcolor{red}{.000}; \textcolor{red}{.000};
\textcolor{red}{.000}; \textcolor{red}{.000}; \textcolor{red}{.000}&.032; .036; .030; .037;
.034&.170; .150; .184; .128; .153&.002; .002; .001; .001; .002&.176; .169; .224; .258; .157\\
\hline
5.Dense&5.ReLU&\textcolor{red}{.312}; \textcolor{red}{.310}; \textcolor{red}{.310};
\textcolor{red}{.303}; \textcolor{red}{.334}&\textcolor{red}{.000}; \textcolor{red}{.000};
\textcolor{red}{.000}; \textcolor{red}{.000}; \textcolor{red}{.000}&.043; .043; .038;
\textcolor{red}{.062}; .044&.091; .102; .124; \textcolor{red}{.022}; .085&.001; .002; .001; .001;
.002&.161; .065; .127; .163; .091\\
\hline
5.ReLU&6.Dense&\textcolor{red}{.640}; \textcolor{red}{.643}; \textcolor{red}{.645};
\textcolor{red}{.657}; \textcolor{red}{.620}&\textcolor{red}{.000}; \textcolor{red}{.000};
\textcolor{red}{.000}; \textcolor{red}{.000}; \textcolor{red}{.000}&\textcolor{red}{.169};
\textcolor{red}{.151}; \textcolor{red}{.113}; \textcolor{red}{.211};
\textcolor{red}{.153}&\textcolor{red}{.000}; \textcolor{red}{.000}; \textcolor{red}{.000};
\textcolor{red}{.000}; \textcolor{red}{.000}&\textcolor{red}{.004}; \textcolor{red}{.003};
\textcolor{red}{.004}; \textcolor{red}{.004}; \textcolor{red}{.004}&\textcolor{red}{.000};
\textcolor{red}{.000}; \textcolor{red}{.000}; \textcolor{red}{.000}; \textcolor{red}{.000}\\
\hline
6.Dense&6.Softmax&-.000; -.000; -.000; .000; -.000&.945; .809; .515; .523; .652&.013; .009; .005;
\textcolor{red}{.007}; .010&.102; .267; .398; \textcolor{red}{.000}; .206&\textcolor{red}{.001};
\textcolor{red}{.001}; \textcolor{red}{.001}; \textcolor{red}{.001};
\textcolor{red}{.001}&\textcolor{red}{.000}; \textcolor{red}{.000}; \textcolor{red}{.000};
\textcolor{red}{.016}; \textcolor{red}{.000}\\
\bottomrule
\end{tabular}
\end{table}
%
\begin{table}[p!]
\tiny
\centering
\caption{Training data difference between between multi-layer component input and output $\boldsymbol{\mathcal{H}}$ class-pair statistics
         for the trained models, and respective
         $p$-values for the one sided permutation test~(Eq.~17 in paper).  \textcolor{red}{Red font}
         denotes layer instances for which we reject $\mathbf{H_0}$, and black font denotes layers for
         which we fail to reject $\mathbf{H_0}$. 
         (Note: ~Cosine distance is used as the proximity measure.  50,000
         Monte Carlo trials used to estimate the $p$-values. $\Delta{\bar{\mathcal{H}}} = \bar{\mathcal{H}}^{(t)}_{(k_2,T)} - \bar{\mathcal{H}}^{(t)}_{(k_1,T)}$)}
\label{table:B4Table}
\begin{tabular}{p{10mm}p{12mm}|K{16mm}|K{16mm}|K{16mm}|K{16mm}|K{16mm}|K{16mm}}
\toprule
Input Space &Output Space& \multicolumn{2}{c|}{CIFAR10 w Random}& \multicolumn{2}{c|}{CIFAR10 w
True}& \multicolumn{2}{c}{MNIST w True} \\
 & &$\Delta{\bar{\mathcal{H}}}$ & \textit{p}-values & $\Delta{\bar{\mathcal{H}}}$ & \textit{p}-values&
 $\Delta{\bar{\mathcal{H}}}$ & \textit{p}-values \\
\midrule
0.Input&1.ReLU&.000; -.003; -.002; -.006; .004&.472; .739; .629; .838; .214&.020; .035; .018; .032;
.026&.267; .145; .289; .165; .209&.003; .003; .003; .003; .003&.224; .214; .209; .210; .199\\
\hline
1.ReLU&2.ReLU&.000; .001; .008; -.009; -.003&.457; .426; .091; .910; .702&\textcolor{red}{.071};
.052; \textcolor{red}{.077}; .053; .056&\textcolor{red}{.024}; .071; \textcolor{red}{.017}; .066;
.055&.000; .000; .000; .000; .000&.438; .466; .447; .438; .450\\
\hline
2.ReLU&2.MaxPool&.007; .004; -.004; -.001; .002&.049; .192; .819; .548; .353&.040; .028; .043; .028;
.031&.135; .217; .118; .208; .184&-.001; -.001; -.001; -.001; -.001&.672; .628; .666; .684; .617\\
\hline
2.MaxPool&3.ReLU&\textcolor{red}{.009}; -.000; -.002; -.003; .004&\textcolor{red}{.023}; .545; .669;
.695; .238&.062; .054; .064; .054; .068&.043; .068; .043; .064; .029&.001; .002; .001; .002;
.001&.322; .292; .313; .293; .329\\
\hline
3.ReLU&4.ReLU&.000; \textcolor{red}{.015}; \textcolor{red}{.016}; .007; .001&.491;
\textcolor{red}{.002}; \textcolor{red}{.001}; .078; .405&.030; .031; .024; .024; .026&.210; .202;
.252; .252; .241&.002; .002; .002; .002; .002&.284; .256; .210; .217; .218\\
\hline
4.ReLU&4.MaxPool&-.000; -.005; -.006; .007; -.006&.507; .830; .882; .054; .902&.026; .028; .024;
.034; .039&.224; .218; .248; .159; .133&.003; .003; .003; .003; .002&.087; .117; .106; .096; .154\\
\hline
4.MaxPool&5.ReLU&\textcolor{red}{.349}; \textcolor{red}{.350}; \textcolor{red}{.349};
\textcolor{red}{.334}; \textcolor{red}{.380}&\textcolor{red}{.000}; \textcolor{red}{.000};
\textcolor{red}{.000}; \textcolor{red}{.000}; \textcolor{red}{.000}&\textcolor{red}{.075};
\textcolor{red}{.079}; \textcolor{red}{.068}; \textcolor{red}{.100};
\textcolor{red}{.079}&\textcolor{red}{.011}; \textcolor{red}{.010}; \textcolor{red}{.019};
\textcolor{red}{.001}; \textcolor{red}{.008}&.003; \textcolor{red}{.003}; .003; .002;
\textcolor{red}{.003}&.034; \textcolor{red}{.011}; .037; .059; \textcolor{red}{.015}\\
\hline
5.ReLU&6.Softmax&\textcolor{red}{.639}; \textcolor{red}{.643}; \textcolor{red}{.645};
\textcolor{red}{.657}; \textcolor{red}{.620}&\textcolor{red}{.000}; \textcolor{red}{.000};
\textcolor{red}{.000}; \textcolor{red}{.000}; \textcolor{red}{.000}&\textcolor{red}{.182};
\textcolor{red}{.160}; \textcolor{red}{.118}; \textcolor{red}{.218};
\textcolor{red}{.163}&\textcolor{red}{.000}; \textcolor{red}{.000}; \textcolor{red}{.000};
\textcolor{red}{.000}; \textcolor{red}{.000}&\textcolor{red}{.005}; \textcolor{red}{.004};
\textcolor{red}{.005}; \textcolor{red}{.005}; \textcolor{red}{.004}&\textcolor{red}{.000};
\textcolor{red}{.000}; \textcolor{red}{.000}; \textcolor{red}{.000}; \textcolor{red}{.000}\\
\bottomrule
\end{tabular}
\end{table}
%
\begin{table}[p!]
\tiny
\centering
\caption{Training data difference between between multi-layer component input and output $\boldsymbol{\mathcal{H}}$ class-pair statistics
         for the trained models, and respective
         $p$-values for the one sided permutation test~(Eq.~17 in paper).  \textcolor{red}{Red font}
         denotes layer instances for which we reject $\mathbf{H_0}$, and black font denotes layers for
         which we fail to reject $\mathbf{H_0}$. 
         (Note: ~Cosine distance is used as the proximity measure.  50,000
         Monte Carlo trials used to estimate the $p$-values. $\Delta{\bar{\mathcal{H}}} = \bar{\mathcal{H}}^{(t)}_{(k_2,T)} - \bar{\mathcal{H}}^{(t)}_{(k_1,T)}$)}
\label{table:B5Table}
\begin{tabular}{p{10mm}p{12mm}|K{16mm}|K{16mm}|K{16mm}|K{16mm}|K{16mm}|K{16mm}}
\toprule
Input Space &Output Space& \multicolumn{2}{c|}{CIFAR10 w Random}& \multicolumn{2}{c|}{CIFAR10 w
True}& \multicolumn{2}{c}{MNIST w True} \\
 & &$\Delta{\bar{\mathcal{H}}}$ & \textit{p}-values & $\Delta{\bar{\mathcal{H}}}$ & \textit{p}-values&
 $\Delta{\bar{\mathcal{H}}}$ & \textit{p}-values \\
\midrule
0.Input&2.MaxPool&.008; .002; .002; -.015; .003&.047; .328; .371; .992; .271&\textcolor{red}{.131};
\textcolor{red}{.114}; \textcolor{red}{.139}; \textcolor{red}{.114};
\textcolor{red}{.113}&\textcolor{red}{.000}; \textcolor{red}{.000}; \textcolor{red}{.000};
\textcolor{red}{.000}; \textcolor{red}{.001}&.002; .002; .002; .002; .002&.316; .295; .298; .314;
.251\\
\hline
2.MaxPool&4.MaxPool&.009; \textcolor{red}{.010}; .008; .011; -.000&.031; \textcolor{red}{.006};
.046; .033; .537&\textcolor{red}{.119}; \textcolor{red}{.112}; \textcolor{red}{.113};
\textcolor{red}{.113}; \textcolor{red}{.133}&\textcolor{red}{.000}; \textcolor{red}{.001};
\textcolor{red}{.001}; \textcolor{red}{.001}; \textcolor{red}{.000}&\textcolor{red}{.006};
\textcolor{red}{.006}; \textcolor{red}{.006}; \textcolor{red}{.006};
\textcolor{red}{.006}&\textcolor{red}{.012}; \textcolor{red}{.013}; \textcolor{red}{.007};
\textcolor{red}{.006}; \textcolor{red}{.013}\\
\hline
4.MaxPool&5.ReLU&\textcolor{red}{.349}; \textcolor{red}{.350}; \textcolor{red}{.349};
\textcolor{red}{.334}; \textcolor{red}{.380}&\textcolor{red}{.000}; \textcolor{red}{.000};
\textcolor{red}{.000}; \textcolor{red}{.000}; \textcolor{red}{.000}&\textcolor{red}{.075};
\textcolor{red}{.079}; \textcolor{red}{.068}; \textcolor{red}{.100};
\textcolor{red}{.079}&\textcolor{red}{.012}; \textcolor{red}{.010}; \textcolor{red}{.019};
\textcolor{red}{.001}; \textcolor{red}{.009}&.003; \textcolor{red}{.003}; .003; .002;
\textcolor{red}{.003}&.033; \textcolor{red}{.011}; .037; .059; \textcolor{red}{.014}\\
\hline
5.ReLU&6.Softmax&\textcolor{red}{.639}; \textcolor{red}{.643}; \textcolor{red}{.645};
\textcolor{red}{.657}; \textcolor{red}{.620}&\textcolor{red}{.000}; \textcolor{red}{.000};
\textcolor{red}{.000}; \textcolor{red}{.000}; \textcolor{red}{.000}&\textcolor{red}{.182};
\textcolor{red}{.160}; \textcolor{red}{.118}; \textcolor{red}{.218};
\textcolor{red}{.163}&\textcolor{red}{.000}; \textcolor{red}{.000}; \textcolor{red}{.000};
\textcolor{red}{.000}; \textcolor{red}{.000}&\textcolor{red}{.005}; \textcolor{red}{.004};
\textcolor{red}{.005}; \textcolor{red}{.005}; \textcolor{red}{.004}&\textcolor{red}{.000};
\textcolor{red}{.000}; \textcolor{red}{.000}; \textcolor{red}{.000}; \textcolor{red}{.000}\\
\bottomrule
\end{tabular}
\end{table}
\begin{table}[p!]
\tiny
\centering
\caption{Validation data differences in mean of $\boldsymbol{\mathcal{H}}$ class-pair  statistics between the
         input and output representations of a layer, and respective one-sided
         permutation test~$p$-values.  \textcolor{red}{Red font} denotes layer
         instances for which we reject $\mathbf{H_0}$, and black font denotes layers for which we fail
         to reject $\mathbf{H_0}$~(Eq.~18 in paper). 
          (Note: Note: ~Cosine distance is used as the proximity measure.  50,000
         Monte Carlo trials used to estimate the $p$-values. 
         $\Delta{\bar{\mathcal{H}}} = \bar{\mathcal{H}}^{(v)}_{(k,T)} - \bar{\mathcal{H}}^{(v)}_{(k-1,T)}$)}

\label{table:B6Table}
\begin{tabular}{p{10mm}p{12mm}|K{16mm}|K{16mm}|K{16mm}|K{16mm}|K{16mm}|K{16mm}}
\toprule
Input Space &Output Space& \multicolumn{2}{c|}{CIFAR10 w Random}& \multicolumn{2}{c|}{CIFAR10 w
True}& \multicolumn{2}{c}{MNIST w True} \\
 & &$\Delta{\bar{\mathcal{H}}}$ & \textit{p}-values & $\Delta{\bar{\mathcal{H}}}$ & \textit{p}-values&
 $\Delta{\bar{\mathcal{H}}}$ & \textit{p}-values \\
\midrule
0.Input&1.Conv&-.002; .001; -.005; -.001; -.000&.669; .395; .886; .577; .524&.017; .032; .008; .022;
.017&.309; .168; .410; .245; .307&.003; .002; .002; .002; .002&.161; .206; .193; .194; .205\\
\hline
1.Conv&1.ReLU&.002; .000; .001; -.000; .001&.381; .490; .417; .517; .447&.006; -.001; .012; .009;
.004&.423; .512; .371; .396; .455&.000; .000; .000; .000; .000&.483; .477; .470; .475; .497\\
\hline
1.ReLU&2.Conv&.003; .003; -.002; .002; -.000&.287; .249; .640; .346; .501&.035; .026; .042; .025;
.023&.164; .227; .124; .238; .264&.000; .000; .000; .000; .000&.432; .435; .453; .478; .461\\
\hline
2.Conv&2.ReLU&-.003; .001; .001; -.000; -.002&.754; .432; .399; .503; .717&.030; .024; .029; .028;
.032&.200; .259; .229; .219; .188&-.000; .000; -.000; -.000; .000&.517; .499; .512; .505; .496\\
\hline
2.ReLU&2.MaxPool&.004; .003; -.001; -.006; .007&.219; .267; .578; .895; .050&.044; .030; .048; .030;
.038&.112; .201; .099; .196; .136&-.001; -.001; -.001; -.001; -.001&.635; .692; .667; .670; .665\\
\hline
2.MaxPool&3.Conv&-.012; .000; .005; .006; .001&.989; .462; .127; .090; .396&.064; .054; .060; .053;
.061&.040; .068; .057; .069; .045&.001; .002; .002; .002; .002&.302; .237; .275; .234; .266\\
\hline
3.Conv&3.ReLU&-.003; .002; .001; -.002; .001&.721; .311; .427; .650; .437&.003; .005; .009; .004;
.012&.466; .443; .405; .448; .370&-.000; -.000; -.000; -.000; .000&.513; .520; .529; .531; .479\\
\hline
3.ReLU&4.Conv&-.001; -.004; .000; -.002; -.001&.556; .826; .498; .706; .593&.006; .004; .004; -.002;
.003&.436; .452; .458; .523; .474&.000; .001; .001; .001; .001&.448; .337; .340; .344; .410\\
\hline
4.Conv&4.ReLU&.002; .002; .003; -.000; .003&.313; .308; .273; .525; .267&.024; .030; .023; .032;
.030&.245; .203; .252; .176; .211&.001; -.000; .000; .000; .001&.383; .509; .472; .427; .408\\
\hline
4.ReLU&4.MaxPool&-.001; -.001; .005; .002; .004&.598; .605; .180; .355; .204&.018; .024; .017; .028;
.031&.295; .253; .309; .204; .192&.002; .003; .003; .002; .002&.143; .104; .084; .132; .118\\
\hline
4.MaxPool&5.Dense&-.007; -.005; -.002; -.003; -.006&.931; .881; .624; .719; .932&.025; .029; .023;
.025; .020&.233; .204; .246; .231; .283&.001; .001; .001; .001; .002&.203; .156; .298; .178; .160\\
\hline
5.Dense&5.ReLU&\textcolor{red}{.017}; .003; -.000; .001; .001&\textcolor{red}{.000}; .166; .543;
.450; .417&.009; .009; .012; .003; .009&.398; .400; .357; .461; .390&.001; .001; .001; .001;
.001&.239; .201; .167; .271; .283\\
\hline
5.ReLU&6.Dense&-.007; -.001; -.013; -.007; .002&.938; .552; .996; .932; .371&.040; .039; .035; .041;
.035&.112; .125; .145; .106; .143&\textcolor{red}{.002}; .001; .002; .001;
.002&\textcolor{red}{.009}; .144; .046; .112; .057\\
\hline
6.Dense&6.Softmax&-.002; .003; \textcolor{red}{.015}; \textcolor{red}{.010}; -.008&.658; .274;
\textcolor{red}{.002}; \textcolor{red}{.020}; .946&-.026; -.012; -.019; -.038; -.025&.793; .644;
.727; .883; .787&-.003; -.002; -.003; -.004; -.003&.999; .990; .999; 1.000; 1.000\\
\bottomrule
\end{tabular}
\end{table}
\begin{table}[p!]
\tiny
\centering
\caption{Validation data differences in mean of $\boldsymbol{\mathcal{H}}$ class-pair  statistics between the
         input and output representations of multilayer layer compnents, and respective one-sided
         permutation test~$p$-values.  \textcolor{red}{Red font} denotes layer
         instances for which we reject $\mathbf{H_0}$, and black font denotes layers for which we fail
         to reject $\mathbf{H_0}$~(Eq.~18 in paper). 
          (Note: Note: ~Cosine distance is used as the proximity measure.  50,000
         Monte Carlo trials used to estimate the $p$-values. 
         $\Delta{\bar{\mathcal{H}}} = \bar{\mathcal{H}}^{(v)}_{(k_2,T)} - \bar{\mathcal{H}}^{(v)}_{(k_1,T)}$)}
\label{table:B7Table}
\begin{tabular}{p{10mm}p{12mm}|K{16mm}|K{16mm}|K{16mm}|K{16mm}|K{16mm}|K{16mm}}
\toprule
Input Space &Output Space& \multicolumn{2}{c|}{CIFAR10 w Random}& \multicolumn{2}{c|}{CIFAR10 w
True}& \multicolumn{2}{c}{MNIST w True} \\
 & &$\Delta{\bar{\mathcal{H}}}$ & \textit{p}-values & $\Delta{\bar{\mathcal{H}}}$ & \textit{p}-values&
 $\Delta{\bar{\mathcal{H}}}$ & \textit{p}-values \\
\midrule

0.Input&1.ReLU&-.001; .001; -.005; -.001; .000&.555; .381; .838; .593; .472&.023; .031; .020; .031;
.021&.249; .176; .283; .174; .271&.003; .002; .003; .003; .002&.156; .188; .174; .183; .203\\
\hline
1.ReLU&2.ReLU&-.000; .004; -.001; .002; -.002&.527; .203; .549; .344; .715&.065; .050; .071; .053;
.055&.036; .082; .025; .064; .063&.000; .000; .000; .000; .000&.451; .435; .467; .481; .460\\
\hline
2.ReLU&2.MaxPool&.004; .003; -.001; -.006; .007&.225; .272; .582; .892; .051&.044; .030; .048; .030;
.038&.111; .199; .100; .198; .136&-.001; -.001; -.001; -.001; -.001&.638; .689; .670; .670; .658\\
\hline
2.MaxPool&3.ReLU&-.015; .003; .006; .005; .002&.997; .267; .079; .137; .341&.067; .059; .069; .057;
\textcolor{red}{.074}&.034; .053; .034; .053; \textcolor{red}{.021}&.001; .002; .001; .002;
.002&.318; .248; .300; .264; .251\\
\hline
3.ReLU&4.ReLU&.002; -.002; .003; -.003; .002&.380; .668; .275; .749; .354&.030; .034; .028; .030;
.032&.198; .174; .223; .195; .186&.001; .001; .001; .001; .001&.331; .354; .319; .289; .323\\
\hline
4.ReLU&4.MaxPool&-.001; -.001; .005; .002; .004&.598; .603; .180; .349; .201&.018; .024; .017; .028;
.031&.296; .248; .315; .205; .192&.002; .003; .003; .002; .002&.146; .105; .084; .131; .118\\
\hline
4.MaxPool&5.ReLU&\textcolor{red}{.009}; -.001; -.002; -.002; -.005&\textcolor{red}{.016}; .627;
.655; .667; .877&.034; .038; .035; .028; .029&.166; .136; .149; .203; .200&.002; .002; .002; .002;
.002&.070; .035; .068; .069; .069\\
\hline
5.ReLU&6.Softmax&-.009; .002; .001; .003; -.006&.973; .297; .380; .258; .896&.014; .027; .016; .004;
.010&.329; .203; .306; .453; .377&-.000; -.001; -.001; -.002; -.002&.635; .894; .909; .990; .941\\
\bottomrule
\end{tabular}
\end{table}
\begin{table}[p!]
\tiny
\centering
\caption{Validation data differences in mean of $\boldsymbol{\mathcal{H}}$ class-pair  statistics between the
         input and output representations of multilayer layer compnents, and respective one-sided
         permutation test~$p$-values.  \textcolor{red}{Red font} denotes layer
         instances for which we reject $\mathbf{H_0}$, and black font denotes layers for which we fail
         to reject $\mathbf{H_0}$~(Eq.~18 in paper). 
          (Note: ~Cosine distance is used as the proximity measure.  50,000
         Monte Carlo trials used to estimate the $p$-values. 
         $\Delta{\bar{\mathcal{H}}} = \bar{\mathcal{H}}^{(v)}_{(k_2,T)} - \bar{\mathcal{H}}^{(v)}_{(k_1,T)}$)}
\label{table:B8Table}
\begin{tabular}{p{10mm}p{12mm}|K{16mm}|K{16mm}|K{16mm}|K{16mm}|K{16mm}|K{16mm}}
\toprule
Input Space &Output Space& \multicolumn{2}{c|}{CIFAR10 w Random}& \multicolumn{2}{c|}{CIFAR10 w
True}& \multicolumn{2}{c}{MNIST w True} \\
 & &$\Delta{\bar{\mathcal{H}}}$ & \textit{p}-values & $\Delta{\bar{\mathcal{H}}}$ & \textit{p}-values&
 $\Delta{\bar{\mathcal{H}}}$ & \textit{p}-values \\
\midrule
0.Input&2.MaxPool&.003; .008; -.006; -.005; .005&.309; .039; .907; .843; .119&\textcolor{red}{.132};
\textcolor{red}{.111}; \textcolor{red}{.138}; \textcolor{red}{.115};
\textcolor{red}{.114}&\textcolor{red}{.000}; \textcolor{red}{.001}; \textcolor{red}{.000};
\textcolor{red}{.000}; \textcolor{red}{.001}&.002; .002; .002; .002; .002&.215; .287; .279; .294;
.298\\
\hline
2.MaxPool&4.MaxPool&-.014; -.000; \textcolor{red}{.013}; .004; .008&.999; .538;
\textcolor{red}{.003}; .213; .049&\textcolor{red}{.115}; \textcolor{red}{.117};
\textcolor{red}{.113}; \textcolor{red}{.115}; \textcolor{red}{.137}&\textcolor{red}{.000};
\textcolor{red}{.001}; \textcolor{red}{.001}; \textcolor{red}{.001};
\textcolor{red}{.000}&\textcolor{red}{.004}; \textcolor{red}{.005}; \textcolor{red}{.005};
\textcolor{red}{.005}; \textcolor{red}{.005}&\textcolor{red}{.022}; \textcolor{red}{.010};
\textcolor{red}{.009}; \textcolor{red}{.009}; \textcolor{red}{.008}\\
\hline
4.MaxPool&5.ReLU&\textcolor{red}{.009}; -.001; -.002; -.002; -.005&\textcolor{red}{.016}; .624;
.657; .670; .874&.034; .038; .035; .028; .029&.162; .139; .149; .200; .201&.002; .002; .002; .002;
.002&.069; .036; .066; .069; .069\\
\hline
5.ReLU&6.Softmax&-.009; .002; .001; .003; -.006&.973; .300; .380; .264; .898&.014; .027; .016; .004;
.010&.333; .202; .307; .451; .381&-.000; -.001; -.001; -.002; -.002&.634; .891; .909; .989; .942\\
\bottomrule
\end{tabular}
\end{table}
\begin{table}[p!]
\tiny
\centering
\caption{Two-sided permutation test~(Eq.~20 in paper) comparing the differences in the mean change induced on the
         training and validation statistics ( $\boldsymbol{\Delta{\mathcal{H}}}^{(t)}_{(k,k-1)}$ and $\boldsymbol{\Delta{\mathcal{H}}}^{(v)}_{(k,k-1)}$  ). \textcolor{red}{Red font}
         denotes layer instances for which we reject $\mathbf{H_0}$, and black font denotes layers for
         which we fail to reject $\mathbf{H_0}$. 
         (Note: ~Cosine distance is used as the proximity measure.  50,000
         Monte Carlo trials used to estimate the $p$-values. 
         $\Delta\mu = \overline{\Delta{\mathcal{H}}}^{\:(t)}_{(k, k-1)} -  \overline{\Delta{\mathcal{H}}}^{\:(v)}_{(k,k-1)} $)}

\label{table:B9Table}
\begin{tabular}{p{10mm}p{12mm}|K{16mm}|K{16mm}|K{16mm}|K{16mm}|K{16mm}|K{16mm}}
\toprule
Input Space &Output Space& \multicolumn{2}{c|}{CIFAR10 w Random}& \multicolumn{2}{c|}{CIFAR10 w
True}& \multicolumn{2}{c}{MNIST w True} \\
  & &$\Delta\mu$ & \textit{p}-values &
  $\Delta\mu$ & \textit{p}-values&
  $\Delta\mu$ & \textit{p}-values \\
\midrule
0.Input&1.Conv&.005; -.005; .003; -.003; .003&.201; .253; .467; .408; .472&-.000; -.000; .004; .001;
.006&.961; .982; .694; .868; .442&-.000; .000; -.000; .000; .000&.914; .906; .948; .979; .776\\
\hline
1.Conv&1.ReLU&-.004; .001; -.000; -.001; .000&.046; .741; .833; .455; .805&-.002; .004; -.005;
-.000; -.000&.346; .133; .085; .835; .956&-.000; -.000; .000; .000; .000&.362; 1.000; .628; .881;
.252\\
\hline
1.ReLU&2.Conv&.006; .000; .008; -.008; -.003&.244; .995; .089; .089; .560&.002; -.002; -.002; .002;
-.001&.848; .774; .862; .821; .834&.000; -.000; .000; .000; .000&.911; .748; .772; .613; 1.000\\
\hline
2.Conv&2.ReLU&-.005; -.003; .001; -.003; .003&.128; .282; .825; .434; .382&.004; .004; .008; -.002;
.002&.293; .293; .059; .574; .506&.000; .000; .000; .000; .000&.347; 1.000; .652; .436; .446\\
\hline
2.ReLU&2.MaxPool&.003; .001; -.004; .005; -.006&.436; .739; .332; .198; .160&-.004; -.002; -.004;
-.002; -.007&.354; .575; .428; .654; .110&-.000; .000; -.000; -.000; .000&.483; .748; .831; .646;
.841\\
\hline
2.MaxPool&3.Conv&\textcolor{red}{.018}; -.002; -.003; -.002; .007&\textcolor{red}{.000}; .541; .515;
.605; .146&-.007; -.009; -.007; -.008; -.007&.446; .242; .393; .295; .404&.000; -.000; -.000; -.000;
-.001&.430; .651; .927; .865; .319\\
\hline
3.Conv&3.ReLU&.006; -.001; -.005; -.005; -.005&.085; .844; .120; .081; .208&.002; .004; .003; .005;
.001&.215; .040; .197; .027; .613&-.000; .000; .000; .000; .000&.422; .610; .604; .943; .649\\
\hline
3.ReLU&4.Conv&-.009; .006; .006; .003; \textcolor{red}{.013}&.051; .171; .167; .474;
\textcolor{red}{.008}&-.001; -.004; -.005; -.005; -.000&.888; .346; .412; .303; .944&.001; .000;
.001; .001; .001&.394; .630; .366; .122; .202\\
\hline
4.Conv&4.ReLU&.008; .011; .007; .006; \textcolor{red}{-.013}&.082; .033; .181; .275;
\textcolor{red}{.008}&.000; .001; .002; -.001; -.006&.967; .875; .652; .908; .232&-.000; .000; .000;
-.000; .000&.965; .419; .420; .802; .568\\
\hline
4.ReLU&4.MaxPool&.001; -.003; -.011; .006; -.010&.822; .446; .033; .221; .055&.008; .004; .007;
.007; .008&.050; .375; .057; .162; .049&.001; .000; -.000; .001; -.000&.384; 1.000; 1.000; .508;
.740\\
\hline
4.MaxPool&5.Dense&\textcolor{red}{.044}; \textcolor{red}{.044}; \textcolor{red}{.041};
\textcolor{red}{.033}; \textcolor{red}{.052}&\textcolor{red}{.000}; \textcolor{red}{.000};
\textcolor{red}{.000}; \textcolor{red}{.000}; \textcolor{red}{.000}&\textcolor{red}{.007}; .007;
.007; \textcolor{red}{.013}; \textcolor{red}{.014}&\textcolor{red}{.013}; .038; .046;
\textcolor{red}{.004}; \textcolor{red}{.000}&.000; .000; .001; -.000; .000&.686; .936; .342; .548;
.875\\
\hline
5.Dense&5.ReLU&\textcolor{red}{.296}; \textcolor{red}{.307}; \textcolor{red}{.311};
\textcolor{red}{.302}; \textcolor{red}{.333}&\textcolor{red}{.000}; \textcolor{red}{.000};
\textcolor{red}{.000}; \textcolor{red}{.000}; \textcolor{red}{.000}&\textcolor{red}{.034};
\textcolor{red}{.034}; \textcolor{red}{.026}; \textcolor{red}{.059};
\textcolor{red}{.035}&\textcolor{red}{.000}; \textcolor{red}{.000}; \textcolor{red}{.000};
\textcolor{red}{.000}; \textcolor{red}{.000}&.000; .001; .000; .001; \textcolor{red}{.001}&.386;
.069; .650; .162; \textcolor{red}{.015}\\
\hline
5.ReLU&6.Dense&\textcolor{red}{.647}; \textcolor{red}{.644}; \textcolor{red}{.658};
\textcolor{red}{.664}; \textcolor{red}{.619}&\textcolor{red}{.000}; \textcolor{red}{.000};
\textcolor{red}{.000}; \textcolor{red}{.000}; \textcolor{red}{.000}&\textcolor{red}{.129};
\textcolor{red}{.112}; \textcolor{red}{.079}; \textcolor{red}{.169};
\textcolor{red}{.118}&\textcolor{red}{.000}; \textcolor{red}{.000}; \textcolor{red}{.000};
\textcolor{red}{.000}; \textcolor{red}{.000}&.002; \textcolor{red}{.002}; \textcolor{red}{.002};
\textcolor{red}{.003}; \textcolor{red}{.002}&.041; \textcolor{red}{.000}; \textcolor{red}{.004};
\textcolor{red}{.000}; \textcolor{red}{.006}\\
\hline
6.Dense&6.Softmax&.002; -.003; \textcolor{red}{-.015}; -.010; .008&.664; .481;
\textcolor{red}{.004}; .033; .129&\textcolor{red}{.039}; \textcolor{red}{.021};
\textcolor{red}{.023}; \textcolor{red}{.045}; \textcolor{red}{.035}&\textcolor{red}{.000};
\textcolor{red}{.000}; \textcolor{red}{.000}; \textcolor{red}{.000};
\textcolor{red}{.000}&\textcolor{red}{.004}; \textcolor{red}{.003}; \textcolor{red}{.004};
\textcolor{red}{.004}; \textcolor{red}{.004}&\textcolor{red}{.000}; \textcolor{red}{.000};
\textcolor{red}{.000}; \textcolor{red}{.000}; \textcolor{red}{.000}\\
\bottomrule
\end{tabular}
\end{table}